%% file: main.tex
\documentclass{article}

\PassOptionsToPackage{numbers, compress}{natbib}

    \usepackage[final]{neurips_data_2024}

\usepackage[utf8]{inputenc} 
\usepackage[T1]{fontenc}    
\usepackage{hyperref}       
\usepackage{url}            
\usepackage{booktabs}       
\usepackage{amsfonts}       
\usepackage{nicefrac}       
\usepackage{microtype}      
\usepackage{xcolor}         
\usepackage{titletoc}       

\input{utils/general_utils}
\input{utils/includes}
\input{utils/math_utils}

\definecolor{bluegray}{rgb}{0.4, 0.6, 0.8}
\definecolor{ceruleanblue}{rgb}{0.16, 0.32, 0.75}

\hypersetup{
colorlinks=true,
citecolor=ceruleanblue,
linkcolor=ceruleanblue,
urlcolor=black
}

\newcommand{\ours}{\textsc{UnlearnCanvas}}

\title{{\ours}: A Stylized Image Dataset for Enhanced Machine Unlearning Evaluation  in Diffusion Models 
}

\author{%
  Yihua Zhang$^{1}$,
  Chongyu Fan$^{1}$,  
  Yimeng Zhang$^{1}$, 
  Yuguang Yao$^{1}$, 
  Jinghan Jia$^{1}$, 
  Jiancheng Liu$^{1}$, 
  \\
  \textbf{Gaoyuan Zhang}$^{2}$,
  \textbf{Gaowen Liu}$^{3}$,
  \textbf{Ramana Kompella}$^{3}$,
  \textbf{Xiaoming Liu}$^{1}$, 
  \textbf{Sijia Liu}$^{1,2}$
  \\
  \textsuperscript{1}Michigan State University, \textsuperscript{2}IBM Research, \textsuperscript{3}Cisco Research
}

\begin{document}

\maketitle

\input{sections/abstract}

\input{sections/introduction}

\input{sections/problem_statement}

\input{sections/method}

\input{sections/experiment}

\input{sections/conclusion}

\input{sections/acknowledgement}

\bibliography{refs}
\bibliographystyle{IEEEtranN}

\input{sections/checklist}
\input{sections/appendix}

\end{document}

%% file: utils/general_utils.tex
\usepackage{pifont}
\newcommand{\cmark}{\textcolor{green}{\ding{51}}}
\newcommand{\xmark}{\textcolor{red}{\ding{55}}}

\usepackage{color, colortbl}
\definecolor{Gray}{gray}{0.93}
\definecolor{Orange}{rgb}{1,0.5,0}
\definecolor{DGray}{gray}{0.83}
\definecolor{LightCyan}{rgb}{0.88,1,1}

\definecolor{WarnREd}{rgb}{1,0.4,0.4}
\definecolor{WarnOrange}{rgb}{1,0.682,0.502}
\definecolor{WarnPink}{rgb}{0.9176, 0.7215, 0.7215}
\definecolor{GoodGreen}{rgb}{0.5019, 0.9215, 0.6039}

\newcommand{\warnorange}[1]{\sethlcolor{WarnOrange}\hl{#1}}
\newcommand{\warn}[1]{\sethlcolor{WarnPink}\hl{#1}}
\newcommand{\congrat}[1]{\sethlcolor{GoodGreen}\hl{#1}}

\newcommand{\warncellorange}{\cellcolor{WarnOrange}}

\usepackage[T1]{fontenc}

\definecolor{styleblue}{HTML}{504099}
\definecolor{mypurple}{HTML}{9391ff}

%% file: utils/includes.tex
\usepackage{wrapfig}

\usepackage{multirow,mathtools }

\usepackage{pifont}
\usepackage{color, colortbl}

\usepackage{blindtext}
\usepackage{lipsum}

\usepackage{multirow}
\usepackage{listings}

\usepackage{bbm}

\usepackage [english]{babel}
\usepackage [autostyle, english = american]{csquotes}
\usepackage[nottoc]{tocbibind}

\usepackage{pifont}
\usepackage{url}
\usepackage[most]{tcolorbox}

\usepackage{lipsum}
\usepackage{soul}
\usepackage{xcolor}
\usepackage{wrapfig}
\usepackage{multirow,mathtools } 

\usepackage{adjustbox}
\MakeOuterQuote{"}

\usepackage{microtype}
\usepackage{graphicx}
\usepackage{subfigure}
\usepackage{booktabs} 

\usepackage{hyperref}

\usepackage{amsmath}

\usepackage[capitalize,noabbrev]{cleveref}

\usepackage{tablefootnote}

%% file: utils/math_utils.tex
\usepackage{amsmath,amsfonts,bm}

\def\eqref#1{(\ref{#1})}

\def\1{\bm{1}}

\DeclareMathAlphabet{\mathsfit}{\encodingdefault}{\sfdefault}{m}{sl}
\SetMathAlphabet{\mathsfit}{bold}{\encodingdefault}{\sfdefault}{bx}{n}

\newcommand{\btheta}{{\boldsymbol{\theta}}}

\newcommand{\bx}{\mathbf{x}}

%% file: sections/abstract.tex
\begin{abstract}
The technological advancements in diffusion models (DMs) have demonstrated unprecedented capabilities in text-to-image generation and are widely used in diverse applications. However, they have also raised significant societal concerns, such as the generation of harmful content and copyright disputes.
Machine unlearning (MU) has emerged as a promising solution, capable of removing undesired generative capabilities from DMs. However, existing MU evaluation systems present several key challenges that can result in incomplete and inaccurate assessments.
To address these issues, we propose {\ours}, a comprehensive high-resolution stylized image dataset that facilitates the evaluation of the unlearning of artistic styles and associated objects. This dataset enables the establishment of a standardized, automated evaluation framework with 7 quantitative metrics assessing various aspects of the unlearning performance for DMs. Through extensive experiments, we benchmark 9 state-of-the-art MU methods for DMs, revealing novel insights into their strengths, weaknesses, and underlying mechanisms. Additionally, we explore challenging unlearning scenarios for DMs to evaluate worst-case performance against adversarial prompts, the unlearning of finer-scale concepts, and sequential unlearning.
We hope that this study can pave the way for developing more effective, accurate, and robust DM unlearning methods, ensuring safer and more ethical applications of DMs in the future.
The dataset, benchmark, and codes are publicly available at \url{https://unlearn-canvas.netlify.app/}.

\end{abstract}

%% file: sections/introduction.tex
\section{Introduction}
\label{sec: intro}
The recent technological breakthroughs in text-to-image generation, driven by diffusion models (\textbf{DMs}), have shown an unprecedented capability to produce high-resolution, high-quality images across a diverse range of subjects \cite{rombach2022high, midjourney, kang2023scaling, tao2023galip, tao2022df, li2019controllable, reed2016generative, zhou2022learning, Kawar_2023_CVPR, zhang2022motiondiffuse}.
These models have become widely accessible to the public and are applied across various sectors, including advertising, the creative arts \cite{fotor}, and forensic sketching \cite{lablab2023forensic}.
One reason for DMs being able to generate a broad spectrum of content is their reliance on diverse internet-sourced data \cite{schuhmann2022laion}.
However, this inclusiveness comes with risks, such as harmful generation \cite{schramowski2023safe}, copyright issues \cite{author2023midjourney}, and biases or stereotypes \cite{vinker2023concept, turk2023ai}.

\begin{figure}
\centering
\begin{tabular}{cc}
     \hspace*{-4mm} \includegraphics[width=.45\linewidth,height=!]{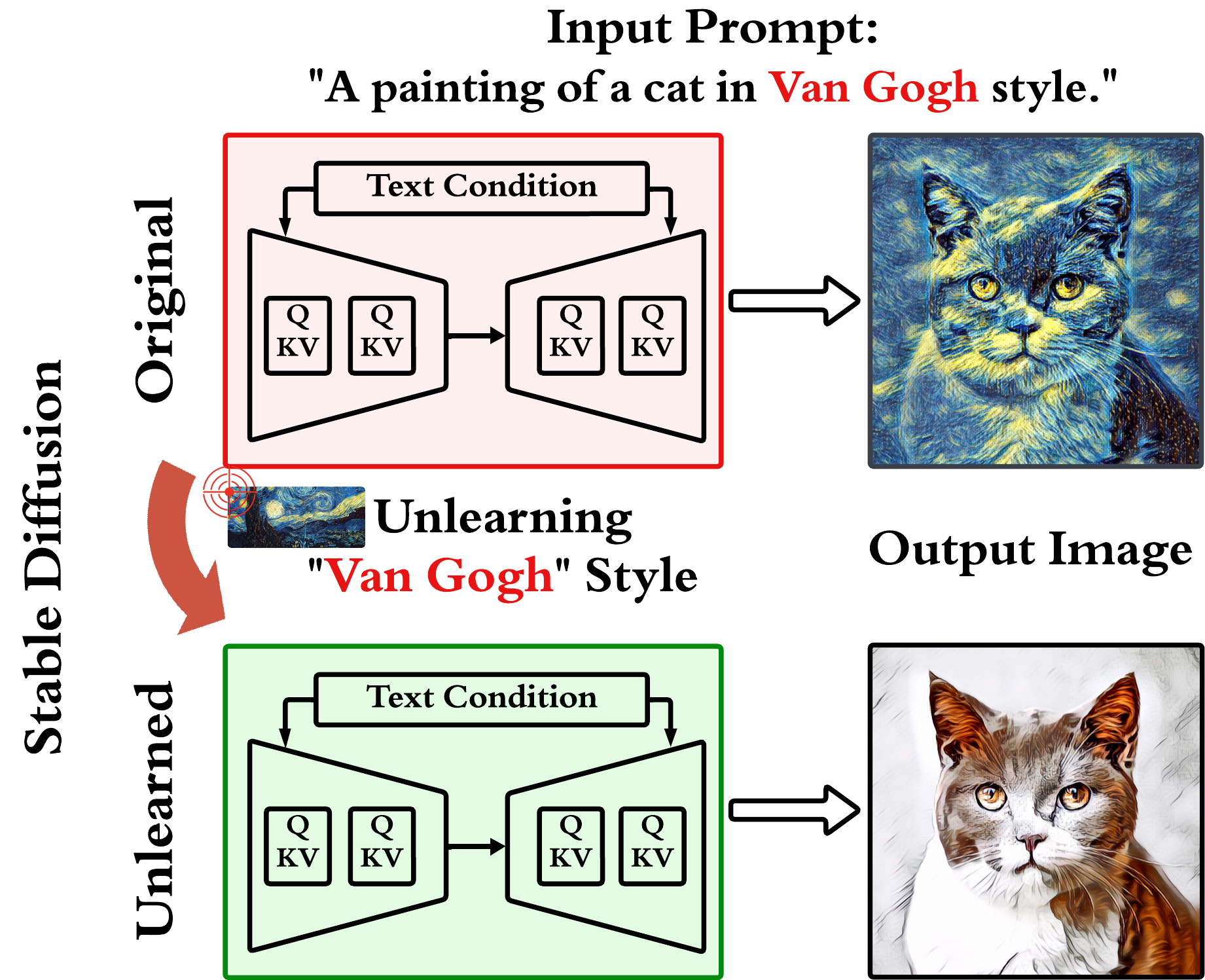} & \hspace*{2mm} \includegraphics[width=.47\linewidth,height=!]{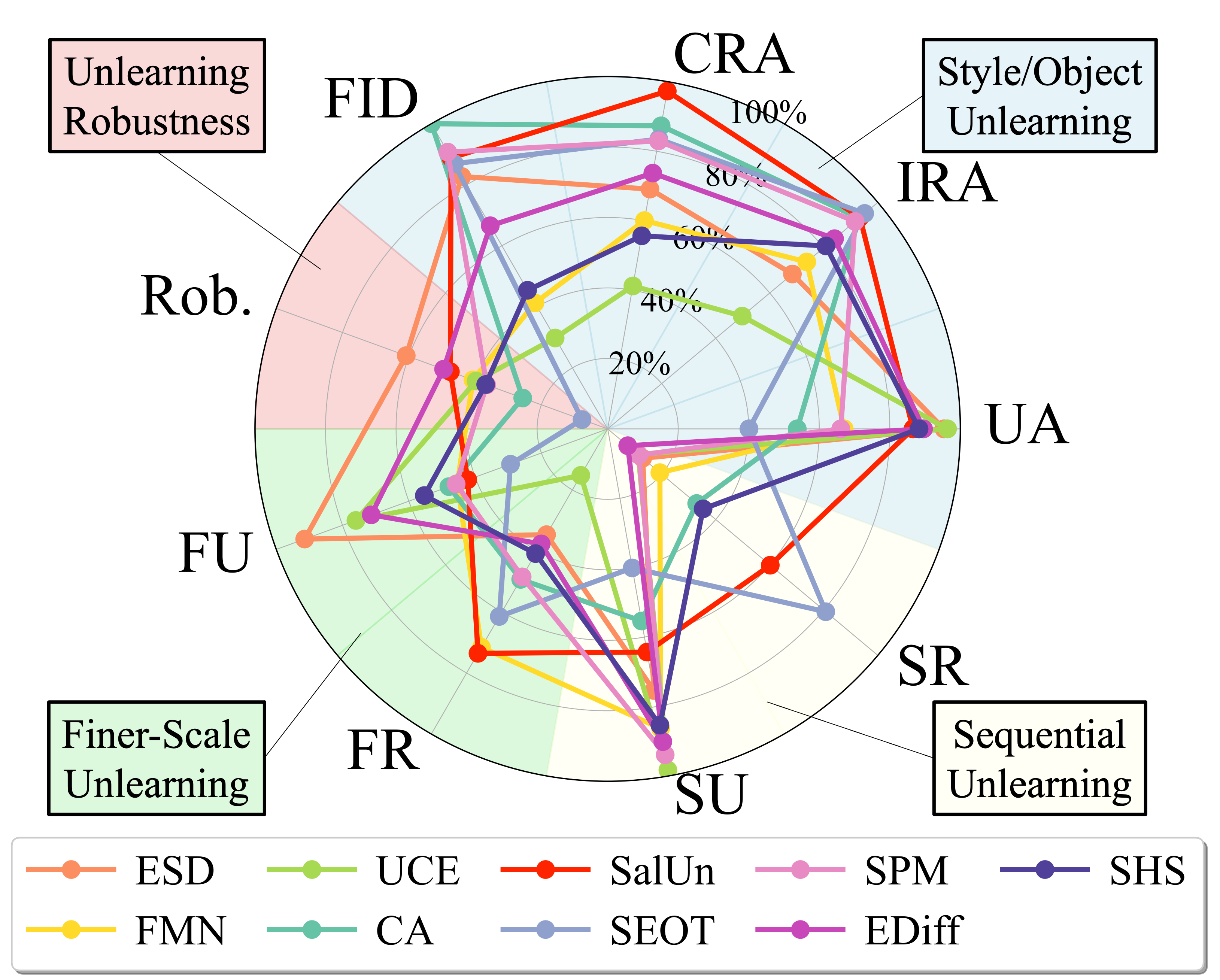} \\
     \hspace*{-2mm}(a) & \hspace*{0mm}  (b)
\end{tabular}
\vspace*{-1em}
\caption{(a) An illustration of MU for DMs. (b) Overview of experiment settings and benchmark results. This benchmark focuses on three categories of quantitative metrics: the unlearning effectiveness (UA, Rob., FU, SU); the retainability of innocent knowledge (IRA, CRA, FR, SR); and the image generation quality (FID). Results are normalized to $0\%\sim100\%$ per metric. No single method excels across all metrics. See a summary of these metrics in Tab.\,\ref{tab: metrics_summary} and more results in Sec.\,\ref{sec: mu_experiment_results}.}
\label{fig: problems_illustration}
\vspace*{-2em}
\end{figure}

To alleviate the negative social impacts associated with DMs, machine unlearning (\textbf{MU}) techniques are catching increasing attention and have been studied in the field of text-to-image generation via DMs \cite{tsai2023ring, dong2024towards, hong2024all, zhao2024separable, huang2023receler, gandikota2023erasing, gandikota2023unified, kumari2023ablating, lyu2023one, fan2023salun, zhang2023forget, heng2023selective, li2024get, wu2024erasediff, wu2024scissorhands, zhang2024defensive}.  
MU for DMs, also referred to as \textit{DM unlearning} \cite{fan2023salun}, aims to prevent the model from generating images when conditioned on an undesired concept (typically specified by a text prompt to be forgotten).
Therefore, the problem of DM unlearning is also known as \textit{concept erasing} \cite{tsai2023ring, dong2024towards, hong2024all, zhao2024separable, huang2023receler, gandikota2023erasing, gandikota2023unified, kumari2023ablating, lyu2023one}.
Based on the types of unlearning requests, the targeted concept to be erased can be diverse, including not-safe-for-work (NSFW) prompts \cite{schramowski2023safe, gandikota2023erasing}, concrete object entities \cite{tsai2023ring, dong2024towards, hong2024all, zhao2024separable, huang2023receler, gandikota2023erasing, gandikota2023unified, kumari2023ablating, lyu2023one, fan2023salun, zhang2023forget, heng2023selective, li2024get, wu2024erasediff, wu2024scissorhands, zhang2024defensive}, and copyrighted information like artistic styles \cite{gandikota2023erasing, gandikota2023unified, kumari2023ablating, lyu2023one, zhang2023forget, heng2023selective, li2024get, wu2024scissorhands}.
In Fig.\,\ref{fig: problems_illustration} (a), we provide an illustration of DM unlearning by demonstrating the removal of an undesired artistic style from DM-generated images.
Additionally, we refer readers to Sec.\,\ref{sec: preliminary} for more related work on DM unlearning.

\begin{wrapfigure}{r}{0.5\linewidth}
    \vspace*{-1em}
    \includegraphics[width=\linewidth]{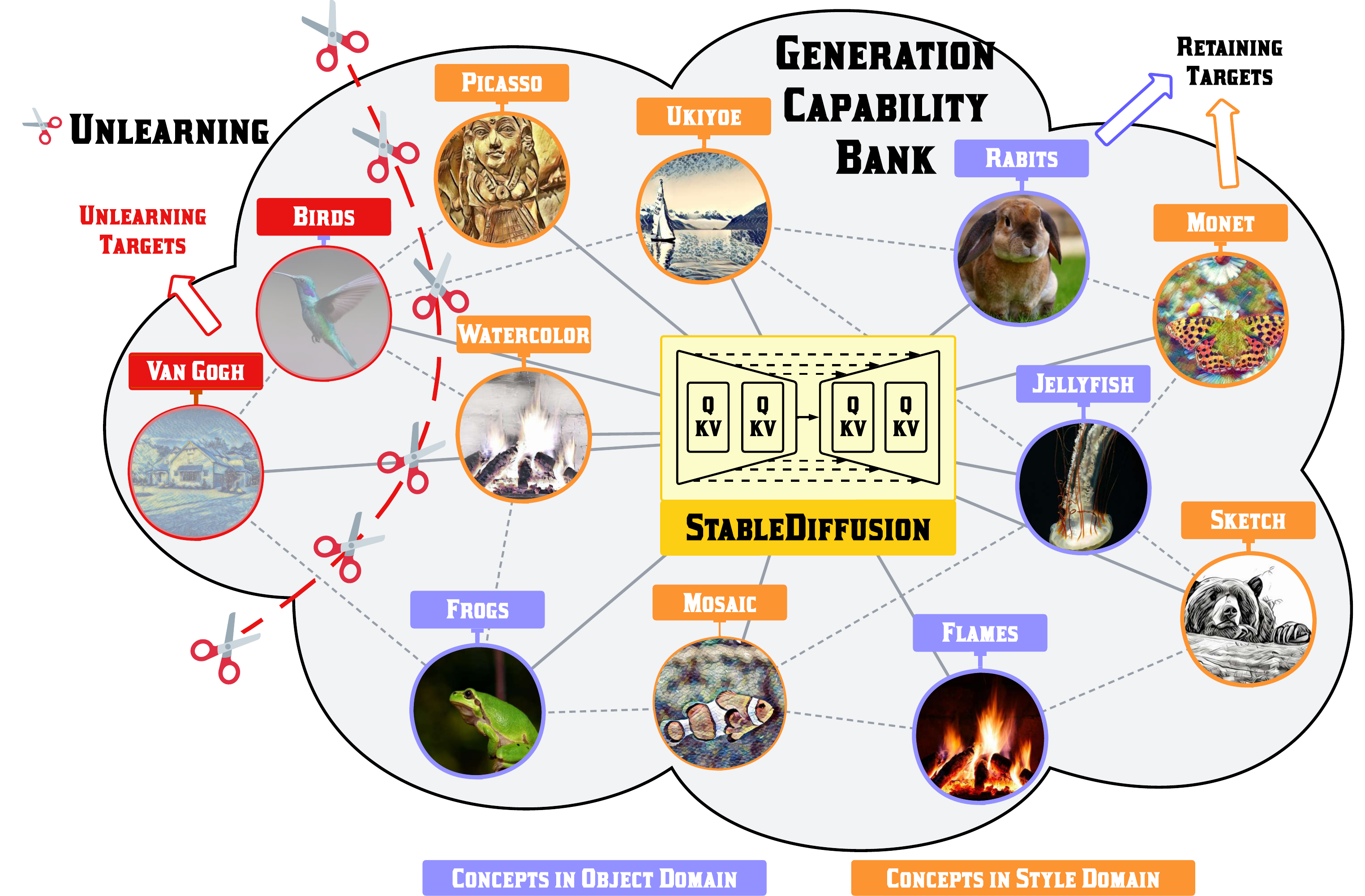}
    \vspace*{-1em}
    \caption{An illustration of machine unlearning using {\ours}. Concepts in the knowledge bank are categorized into different domains (style and object) and serve as potential unlearning targets. When one concept is unlearned, the rest concepts in both the same and different domains are required to be retained.}
    \label{fig: unlearn_ours}
    \vspace*{-1em}
\end{wrapfigure}
Compared to MU for DMs, the unlearning studies in computer vision have primarily focused on image classification models \cite{cao2015towards,golatkar2020eternal, becker2022evaluating, thudi2022unrolling, jia2023model, chen2023boundary, warnecke2021machine,fan2024challenging,ginart2019making,neel2021descent,ullah2021machine,sekhari2021remember, bourtoule2021machine, nguyen2022survey, izzo2021approximate}. Given an image classification dataset, benchmarking unlearning performance for discriminative models is simpler, because MU typically involves either class-wise forgetting \cite{graves2021amnesiac, golatkar2020eternal} or data-wise forgetting \cite{fan2023salun, warnecke2021machine}. The former aims to eliminate the influence of a specific image class, while the latter removes the influence of specific data points from the entire training set. In contrast, DM unlearning is often applied to scenarios involving the removal of higher-level and more abstract concepts,  rather than specific training data points, which can also be difficult to localize within the DM training set.
The primary method for assessing the performance of DM unlearning is using the I2P (inappropriate image prompts) dataset \cite{schramowski2023safe}. This dataset mainly focuses on the safety assessment of \textit{unlearned DMs} (\textit{i.e.}, DMs post-unlearning), designed to erase specific harmful concepts like `nudity' and `violence'. Although I2P specifies which concepts to unlearn, evaluating the unlearning effectiveness and generation retainability of unlearned DMs under I2P remains challenging. This difficulty arises primarily from the lack of ground-truth data or objective criteria to precisely define safety versus non-safety. Furthermore, I2P cannot evaluate the performance of DM unlearning in other scenarios, such as object unlearning \cite{dong2024towards, hong2024all, zhao2024separable, huang2023receler, gandikota2023erasing, gandikota2023unified, kumari2023ablating, lyu2023one, fan2023salun, zhang2023forget, heng2023selective, li2024get, wu2024erasediff, wu2024scissorhands, zhang2024defensive} and style unlearning \cite{gandikota2023erasing, gandikota2023unified, kumari2023ablating, lyu2023one, fan2023salun, zhang2023forget, heng2023selective, li2024get}.

To enhance the assessment of machine unlearning in DMs and establish a standardized evaluation framework, we propose the development of a new benchmark dataset, referred to as \textbf{\ours}. Unlike I2P, {\ours} is designed to evaluate the unlearning of artistic painting styles along with associated image objects (see an illustration in Fig.\,\ref{fig: unlearn_ours}). It also provides ground-truth data annotations to precisely define the criteria for the effectiveness of unlearning and the retained model utility post-unlearning (see a result overview in Fig.\,\ref{fig: problems_illustration} (b)). We summarize \textbf{our contributions} below. 

$\bullet$ We conduct a systematic review of existing MU methods
for DMs and identified three unresolved challenges in their
evaluations. We further provide an in-depth analysis of why a standardized benchmark for DM unlearning evaluation is crucial.

{$\bullet$ We propose {\ours}, a large-scale, high-resolution stylized image dataset with diverse styles and objects, to address current challenges in DM unlearning evaluation. Its dual supervision of styles and objects ensures stylistic consistency, enabling a comprehensive, standardized evaluation approach to precisely characterize unlearning efficacy, generation utility preservation, and efficiency.
}

$\bullet$ We benchmark 9 state-of-the-art DM unlearning methods using \textbf{\ours}, covering standard assessments and examining more challenging scenarios, such as adversarial robustness, grouped object-style unlearning, and sequential unlearning. This comprehensive evaluation provides previously unknown insights into their strengths and weaknesses.

%% file: sections/problem_statement.tex
\section{Related Work, Problem Statement, and Open Challenges}
\label{sec: preliminary}

\paragraph{Related work.}
MU (machine unlearning) was originally developed to {mitigate} the (potentially detrimental) influence of specific data points in a pretrained ML model, without necessitating a complete retraining of the model after removing these unlearning data  \cite{cao2015towards,golatkar2020eternal, becker2022evaluating, thudi2022unrolling, jia2023model, chen2023boundary, warnecke2021machine,fan2024challenging,ginart2019making,neel2021descent,ullah2021machine,sekhari2021remember, bourtoule2021machine, nguyen2022survey, izzo2021approximate,graves2021amnesiac,guo2019certified,xu2023machine,jia2024soul}.
The significance of MU has emerged with the purpose of data privacy protection, in response to regulations such as `the right to be forgotten' \cite{hoofnagle2019european}. However, it has rapidly gained recognition for its role in promoting security, safety, and trustworthiness of ML models.
Examples include defense against backdoor attacks \cite{jia2023model,liu2022backdoor}, fairness enhancement  \cite{chen2023fast,oesterling2023fair}, and controllable federated learning \cite{halimi2022federated,jin2023forgettable}.
More recently, DM unlearning has proven crucial in text-to-image generation to prevent the production of harmful, private, or illegal content \cite{gandikota2023erasing, gandikota2023unified, kumari2023ablating,lyu2023one, fan2023salun, zhang2023forget, heng2023selective,li2024get,wu2024erasediff,wu2024scissorhands}.  For example, DMs have demonstrated a susceptibility to generating NSFW (not-safe-for-work) images when conditioned on inappropriate text prompts (\textit{e.g.}, `nudity' and `violence') \cite{schramowski2023safe,zhang2023generate,chin2023prompting4debugging}. 
Similarly, MU has also found applications in preventing the generation of copyrighted information, such as the misuse of artistic painting styles \cite{gandikota2023erasing, zhang2023forget}.
In brief, {DM unlearning can be regarded as a model edit operation, aimed at preventing DMs from producing undesired or inappropriate images.}


As unlearning approaches continue to evolve, the need for \textit{benchmarking} their performance becomes increasingly critical. Standard datasets like CIFAR-10 \cite{Krizhevsky2009learning} and ImageNet \cite{deng2009imagenet} have long been used to evaluate unlearning performance when forgetting a specific subset of data points \cite{fan2023salun, warnecke2021machine} or a targeted image class \cite{graves2021amnesiac, golatkar2020eternal}. Additionally, a recent benchmark specialized for unlearning in face recognition has been developed \cite{choi2023towards}.
However, crafting effective benchmarks for generative models introduces additional challenges. In the language domain, recent initiatives have specifically aimed to evaluate the efficacy of MU for large language models (LLMs). Notable examples include TOFU \cite{maini2024tofu}, which explores the unlearning of fictitious elements, and WMDP \cite{li2024wmdp}, aimed at removing hazardous knowledge from LLMs. In contrast, to the best of our knowledge, there is currently no effective benchmark dataset designed for DM unlearning in text-to-image generation. The predominant dataset, I2P \cite{schramowski2023safe}, targets the removal of harmful concepts for safe image generation. However, I2P only covers a specialized unlearning scenario and lacks precise evaluation criteria to support a comprehensive, quantitative, and precise assessment of unlearning performance.
This gap inspires us to develop a new benchmark dataset for DM unlearning.

\input{tables/method_overview}

 \textbf{Problem statement.}
Let us denote a pretrained DM by $\boldsymbol{\theta}_\mathrm{o}$, capable of synthesizing high-quality images conditioned on a provided text prompt  $c$ (\textit{e.g.}, `A painting of a cat in Van Gogh style'). When there is a request to adapt the model $\boldsymbol{\theta}_\mathrm{o}$ to prevent generating images in a specific concept, such as the `Van Gogh' painting style, the MU  problem arises. Here, the `Van Gogh' style becomes the specified \textit{unlearning target}, also known as the \textit{erasing concept} \cite{gandikota2023erasing}. See  Fig.\,\ref{fig: problems_illustration}-(a) for a visual representation of this unlearning problem.
In this work, we specify the unlearning target as artistic styles and/or object classes to encompass the settings of style unlearning and object unlearning as described in the literature \cite{gandikota2023erasing, gandikota2023unified, kumari2023ablating, lyu2023one, fan2023salun,zhang2023forget, heng2023selective, li2024get, wu2024erasediff, wu2024scissorhands}.

\textbf{Motivation: Unresolved challenges in MU evaluation.}
When building a rigorous \textit{quantitative} evaluation framework for DM unlearning, we identified three urgent challenges \textit{(C1)}-\textit{(C3)}
after carefully examining existing studies as shown in  \textbf{Tab.\,\ref{tab: mu_comparison}}.

\textit{(C1) The absence of a consensus on a diverse unlearning target test repository.}
As shown in Tab.\,\ref{tab: mu_comparison}, assessments of DM unlearning in various studies, in terms of both {effectiveness} (\textit{i.e.}, concept erasure performance) and {retainability} (\textit{i.e.}, preserved generation quality under non-forgotten concepts),   typically use {manually-selected} unlearning targets from a \textit{limited pool} and focus on \textit{object-centric} evaluation. Precise evaluation for style unlearning and grouped object-style unlearning is lacking.
Although the existing dataset \textsc{WikiArt} \cite{phillips2011wiki} features a collection of real-world artworks by various artists, it does \textit{not} apply to object unlearning.

\textit{(C2) The lack of a systematic study on  `retainability' of DMs post-unlearning.} 
As highlighted in the `retainability' column of Tab.\,\ref{tab: mu_comparison}, there is a notable deficiency in the quantitative assessment of retainability for DMs post-unlearning. Retainability is crucial for measuring the potential side effects of a MU method. 
For instance, when unlearning an artistic style, it is essential to assess the DM's ability to generate images in other styles and all objects. This makes the \textsc{WikiArt} dataset less suitable for benchmarking DM unlearning, as it lacks object labels needed to characterize the side effects of style unlearning on object recognition.

\textit{(C3) The precision challenge in evaluating DM-generated images.} Artistic styles are inherently complex to define and differentiate precisely, complicating the quantitative evaluation of forgetting and retaining performance in style unlearning. For example, the \textsc{WikiArt} dataset was recently used in \cite{moayeri2024rethinking} to train a classifier for detecting copyright infringement, merely achieving an accuracy of 72.80\%. The difficulty in precisely recognizing artistic styles in \textsc{WikiArt} may further hamper the evaluation of DMs' post-unlearning generation. \textbf{Tab.\,\ref{tab: classifier_preliminary}} illustrates the challenge of accurately recognizing the styles of images generated by DMs, even when finetuned on \textsc{WikiArt}. This is evidenced by the test-time classification accuracy using the style classifier, ViT-L/16 \cite{dosovitskiy2020image}, finetuned on \textsc{WikiArt}. As we can see, only about half of the images are correctly classified into their corresponding style classes, \textit{e.g.}, 56.7\% accuracy for classification on generated images using finetuned stable diffusion (SD) v2.0 \cite{rombach2022high}, which is significantly lower than that on test images from the original \textsc{WikiArt} (85.4\%).

The above challenges \textit{(C1)}-\textit{(C3)} drive us to develop the {\ours} dataset as a solution for the systematic and comprehensive evaluation of DM unlearning in Sec.\,\ref{sec: dataset_proposal}.

%% file: tables/method_overview.tex
\begin{wraptable}{r}{0.55\linewidth}
\centering
\vspace*{-2.0em}
\caption{
Overview of existing MU evaluations for DMs, which covers two unlearning scenarios: object unlearning and style unlearning. Each case is demonstrated by the number of targeted unlearning concepts (target \#), qualitative evaluation (Qual.) via generated image visualization, and quantitative evaluation (Quant.) of unlearning effectiveness (UE) and retainability (RT).
}
\label{tab: mu_comparison}
\resizebox{\linewidth}{!}{%
\begin{tabular}{c|cccc|cccc}
\toprule[1pt]
\midrule
\multirow{3}{*}{\begin{tabular}[c]{@{}c@{}}Related\\ Work\end{tabular}} & \multicolumn{4}{c|}{Object Unlearning} & \multicolumn{4}{c}{Style Unlearning} \\
& \multirow{2}{*}{\begin{tabular}[c]{@{}c@{}}Target \# \end{tabular}} & \multicolumn{2}{c}{Quant.} & \multirow{2}{*}{Qual.} & \multirow{2}{*}{\begin{tabular}[c]{@{}c@{}}Target \# \end{tabular}} & \multicolumn{2}{c}{Quant.} & \multirow{2}{*}{Qual.} \\
&  & UE & RT &  &  & UE & RT & \\
\midrule

\rowcolor{Gray}
ESD \cite{gandikota2023erasing} & 11 & {\cmark} & {\cmark} & {\cmark} & 5 & {\xmark} & \xmark & {\cmark} \\

CA \cite{vinker2023concept} & 4 & \cmark & \xmark & {\cmark} & 4 & \xmark & \xmark  & {\cmark} \\

\rowcolor{Gray}
FMN \cite{zhang2023forget} & 7 & \cmark & \xmark  & {\cmark} & 1 & \xmark & \xmark  & {\cmark} \\

UCE \cite{gandikota2023unified} & 11 & \cmark & \xmark  & {\cmark} & 7 & \cmark & \xmark  & {\cmark} \\

\rowcolor{Gray}
SA \cite{heng2023selective} & 31 & \cmark & \xmark  & {\cmark} & 1 & \xmark & \xmark  & {\cmark} \\

SalUn \cite{fan2023salun} & 11 & \cmark & \cmark  & {\cmark} & 0 & \xmark & \xmark  & {\xmark} \\

\rowcolor{Gray}
SPM \cite{lyu2023one} & 24 & \cmark & \cmark  & {\cmark} & 5 & \xmark & \xmark  & {\cmark} \\

SEOT \cite{li2024get} & 22 & \cmark & \cmark  & {\cmark} & 2 & \xmark & \xmark  & {\cmark} \\

\rowcolor{Gray}
EDiff \cite{wu2024erasediff} & 2 & \cmark & \cmark  & {\cmark} & 0 & \xmark & \xmark  & {\xmark} \\

SHS \cite{wu2024scissorhands} & 1 & \cmark & \xmark  & {\cmark} & 0 & \xmark & \xmark  & {\xmark} \\
\midrule
\bottomrule[1pt]
\end{tabular}%
}
\vspace*{-1.5em}
\end{wraptable}

%% file: sections/method.tex
\section{Our Proposal: \textsc{\ours} Dataset}
\label{sec: dataset_proposal}

\begin{wrapfigure}{r}{.6\linewidth}
    \vspace*{-2.6em}
    \centering    
    \includegraphics[width=\linewidth]{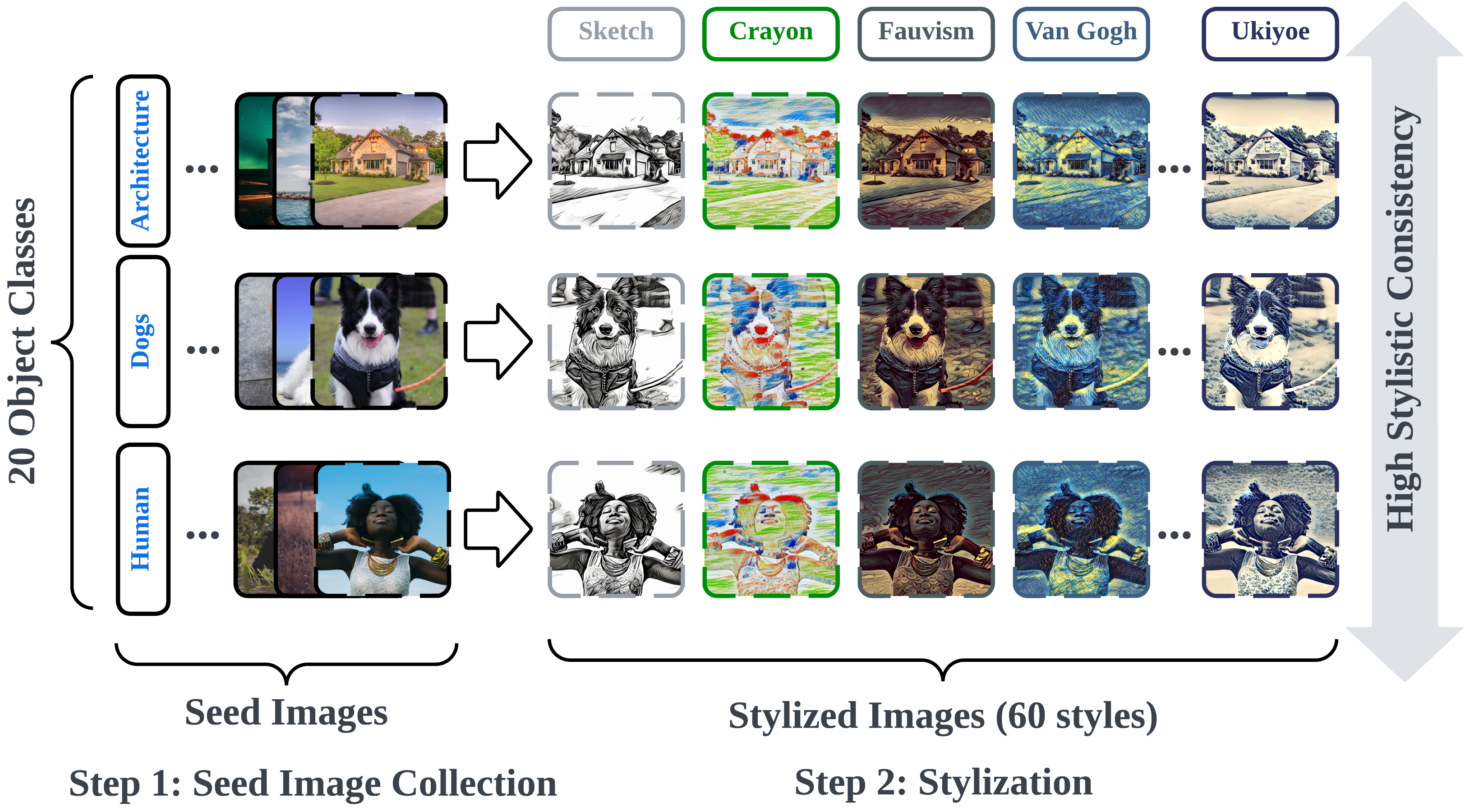}
    \vspace*{-6mm}
    \caption{Illustration of curating {\ours}.
    }
    \label{fig: dataset_overview}
    \vspace*{-6mm}
\end{wrapfigure}

\textbf{Construction of {\ours}.}
As motivated in Sec.\,\ref{sec: preliminary}, {\ours} is created for ease of MU evaluation in DMs.  Its construction involves two main steps: seed image collection and subsequent image stylization; see \textbf{Fig.\,\ref{fig: dataset_overview}} for a schematic overview. More information about the dataset is provided in Appx.\,\ref{app: dataset_overview}.

\textit{Seed image collection.} {\ours} includes images in $60$ unique artistic styles across $20$ distinct objects. {The style categories are built upon a set of high-resolution \textit{seed images}, given by real-world photos from Pexels \cite{pexels}. There are $20$ seed images collected for each of the $20$ object classes. See step 1 in Fig.\,\ref{fig: dataset_overview}.}

\textit{Image stylization.} 
After collecting the seed images, the next step is the \textit{stylization} process. During this phase, each seed image is transformed into 60 predetermined artistic styles, ensuring high stylistic consistency while preserving the original content details. This stylization process is carried out using services provided by Fotor \cite{fotor}. Once all seed images have been stylized in all 60 styles, the dataset is constructed with super-high-resolution images and organized in a hierarchical structure that balances both style and object categories. See step 2 in Fig.\,\ref{fig: dataset_overview}.

\textbf{Advantages of {\ours}.}
Next, we will elaborate on the advantages \textit{(A1)}-\textit{(A3)} of \textbf{\ours} to address the MU evaluation challenges \textit{(C1)}-\textit{(C3)} discussed in Sec.\,\ref{sec: preliminary}.

\textit{(A1) Style-object dual supervision enables a rich unlearning target bank for comprehensive evaluation}. 
To address \textit{(C1)}, {\ours} offers a diverse range of unlearning targets by annotating each image with both style and object labels. We categorized the unlearning targets into two main domains: a style domain comprising 60 styles, and an object domain with 20 object classes. This setup allows us to investigate the impact of style unlearning on the fidelity of object generation and vice versa.

\begin{wrapfigure}{r}{.3\linewidth}
    \centering
    \vspace*{-6mm}
    \includegraphics[width=\linewidth]{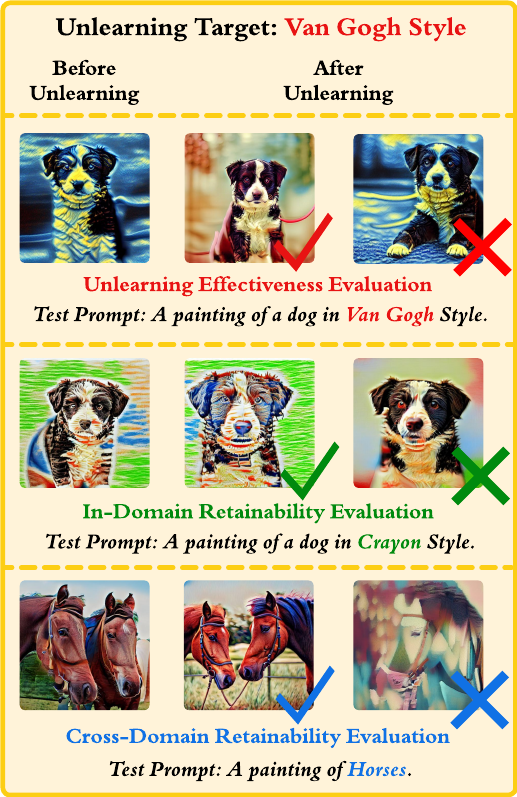}
    \vspace*{-1.5em}    
    \caption{{Illustration of in-domain and cross-domain retainability evaluation,  with the Van Gogh style as the unlearning target. \ding{51} and \ding{55} indicate satisfactory and undesired results post unlearning.}
    }
    \label{fig: unlearn_cases}
    \vspace*{-6mm}
\end{wrapfigure}

\textit{(A2) Enabling both `in-domain' and `cross-domain' retainability analyses for MU evaluation.} 
To address \textit{(C2)}, {\ours} facilitates us to assess DMs' retainability post-unlearning from two perspectives: \textit{in-domain retainability}, which evaluates the DM's generation performance within the same domain as the unlearning target, and \textit{cross-domain retainability}, which assesses the model's generation ability under conditions from a different domain. For example, as depicted in \textbf{Fig.\,\ref{fig: unlearn_cases}}, we can define the `domain' as either `artistic style' or `object class'. If the unlearning target is the `Van Gogh' style, then generating images in all styles except `Van Gogh' within the same object class (\textit{e.g.}, `Dog' in Fig.\,\ref{fig: unlearn_cases}) represents the in-domain case. Conversely, cross-domain retainability refers to generating images of different object classes (\textit{e.g.}, `Horses' in Fig.\,\ref{fig: unlearn_cases}), corresponding to an object domain different from the unlearning target within the style domain.

\textit{(A3) High stylistic consistency ensures precise style definitions and enables accurate quantitative evaluations.} 
In {\ours}, images labeled with the same style exhibit high intra-style consistency and distinct inter-style differences. This facilitates the distinction between different styles, effectively addressing challenge \textit{(C2)}. In a similar study to that presented in Tab.\,\ref{tab: classifier_preliminary}, we observe that the styles within {\ours} can be readily classified using ViT-Large \cite{dosovitskiy2020image}, achieving an accuracy of 98\% for images generated by DM finetuned on {\ours}. For detailed results, please refer to Tab.\,\ref{tab: classifier_ours}.

The above advantages \textit{(A1)}-\textit{(A3)} make {\ours} uniquely suitable for assessing the performance of DM unlearning. For instance, {\ours} includes a greater number of high-resolution images (15M) compared to the existing dataset \textsc{WikiArt} (2M). Moreover, {\ours} surpasses \textsc{WikiArt} in terms of both intra-style coherence and inter-style distinctiveness. See Appx.\,\ref{app: ours_wikiart} for more detailed comparisons.

%% file: sections/experiment.tex
\textbf{Evaluation pipeline via {\ours}.}
We next introduce how {\ours} can be used to benchmark the performance of DM unlearning and the corresponding evaluation metrics. \textbf{Fig.\,\ref{fig: mu_pipeline}} illustrates the proposed evaluation pipeline with a designated unlearning target. It comprises four phases (\textit{I}-\textit{IV}) to evaluate unlearning effectiveness, retainability, generation quality, and efficiency.

\textit{Phase I: Testbed preparation (Fig.\,\ref{fig: mu_pipeline} (a)).} 
Prior to unlearning, we commence by finetuning a specific DM, given by SD (StableDiffusion) v1.5, on {\ours} for text-to-image generation. We also train a vision transformer  (ViT-Large) \cite{dosovitskiy2020image} for the purposes of style and object classification after unlearning.
We ensure that both the SD model and classifiers attain high performance in image generation and classification across all styles and objects. See detailed testbed preparation in Appx.\,\ref{app: exp_setting}.

\textit{Phase II: Machine unlearning (Fig.\,\ref{fig: mu_pipeline} (b)).} 
Next, we utilize a given MU method to update the DM acquired in Phase I, aiming to eliminate a designated unlearning target (e.g., Van Gogh style). This results in the unlearned DM (denoted as $\btheta_\mathrm{u}$), which avoids generating images in response to Van Gogh style-specific prompts.

\textit{Phase III: Answer set generation (Fig.\,\ref{fig: mu_pipeline} (c)).}
Further, we utilize the unlearned model $\btheta_\mathrm{u}$ to generate a set of images conditioned on both the unlearning-related prompts and other benign prompts. To comprehensively evaluate unlearning performance, three types of answer sets are generated.
\textit{(a)} To assess unlearning effectiveness, images are generated in response to prompts that involve the unlearning target (\textit{e.g.}, any prompt like `An image in Van Gogh style' for the case in Fig.\,\ref{fig: mu_pipeline}).
\textit{(b)} To assess in-domain retainability (\textit{i.e.}, the unlearned model's ability to retain generation quality within the same domain), images are generated with prompts in the same domain as the unlearning target (\textit{e.g.}, style-related prompts for style unlearning). These prompts cover all other styles provided in {\ours} except the one to be unlearned. \textit{(c)} To measure cross-domain retainability, images are generated with prompts in other domains. For instance, in the case of style unlearning as shown in Fig.\,\ref{fig: mu_pipeline}, the object domain is considered separate from the style domain. Prompts related to objects are used to evaluate the DM's cross-domain generation capability after style unlearning.

\textit{Phase IV: MU performance assessment (Fig.\,\ref{fig: mu_pipeline} (d)).}  After Phase III, the answer set undergoes style/object classification for unlearning performance assessment. This classification results in three quantitative metrics \ding{172}-\ding{174}.
\ding{172} Unlearning accuracy (\textbf{UA}): This represents the proportion of images generated by the unlearned DM using the unlearning target-related prompt that is \textit{not} correctly classified into the corresponding class. A \textit{higher UA} indicates better unlearning performance in preventing image generation from the unlearning target-related prompts.
\ding{173} In-domain retain accuracy (\textbf{IRA}): This is the classification accuracy of images generated by the unlearned DM using innocent prompts (not relevant to the unlearning target) within the same domain.
\ding{174} Cross-domain retain accuracy (\textbf{CRA}): Similar to IRA, this is the classification accuracy of images generated by the unlearned DM using innocent prompts in different domains.
In addition to the accuracy metrics, we also use the \textbf{FID} score (\ding{175}) to evaluate the distribution-wise generation quality of the unlearned DM. Furthermore, we monitor the {efficiency} of the unlearning process, considering factors such as \textbf{run-time} (\ding{176}), \textbf{storage} space requirements (\ding{177}), and \textbf{memory} costs (\ding{178}).
\definecolor{metrics}{HTML}{0f71e5}
\begin{figure}[t]
    \vspace*{-1em}
    \centering
    \includegraphics[width=\linewidth]{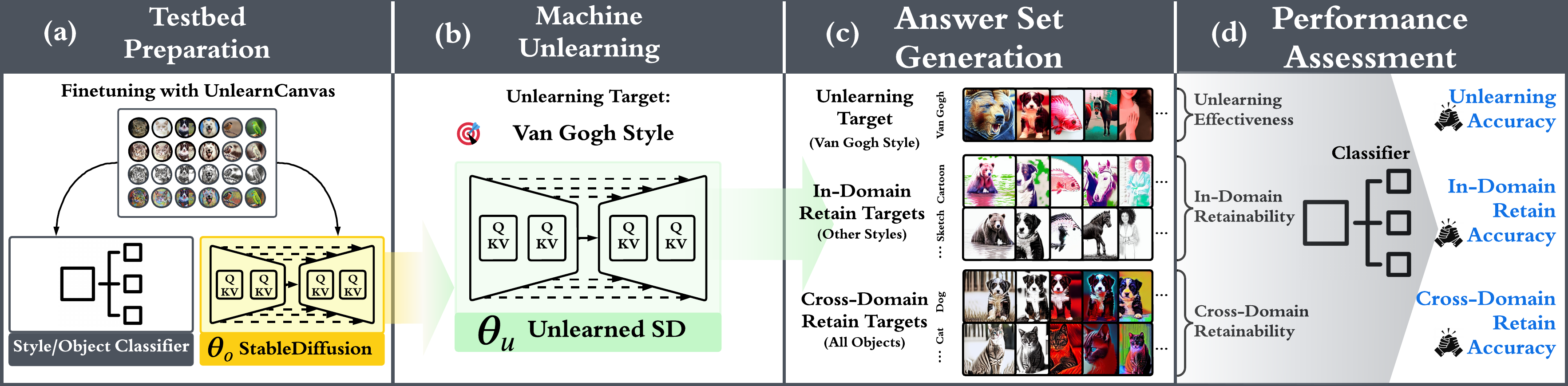}
    \vspace*{-1.7em}
    \caption{An illustration of the evaluation pipeline proposed in this work using {\ours} when unlearning a specific target concept `Van Gogh Style'.
    Unlearning performances (including the unlearning effectiveness and retainability) are quantitatively assessed  (marked in \textcolor{metrics}{blue}) to accurately reflect the unlearning performance portrait. The unlearning target of the pipeline could traverse all the styles and objects to achieve a   comprehensive evaluation.}
    \label{fig: mu_pipeline}
    \vspace*{-1em}
\end{figure}

\textbf{DM unlearning methods to be benchmarked.} 
In this work, we assess  \textbf{9} most recently proposed MU methods for DMs, including  \textbf{ESD} \cite{gandikota2023erasing}, \textbf{FMN} \cite{zhang2023forget}, \textbf{UCE} \cite{gandikota2023unified}, \textbf{CA} \cite{kumari2023ablating}, \textbf{SalUn} \cite{fan2023salun},  \textbf{SEOT} \cite{li2024get}, \textbf{SPM} \cite{lyu2023one}, \textbf{EDiff} \cite{wu2024erasediff}, and \textbf{SHS} \cite{wu2024scissorhands}. We remark that SA \cite{heng2023selective}, a recently proposed method, is excluded from our evaluation due to its excessive computational resource requirements and time consumption. Unless specified otherwise, SD v1.5 is the model used for unlearning.
Detailed training settings for each method can be found in Appx.\,\ref{app: exp_setting}.

\section{Experiment Results}
\label{sec: mu_experiment_results}

\begin{wrapfigure}{r}{0.5\linewidth}
    \vspace*{-1.4em}
    \includegraphics[width=\linewidth]{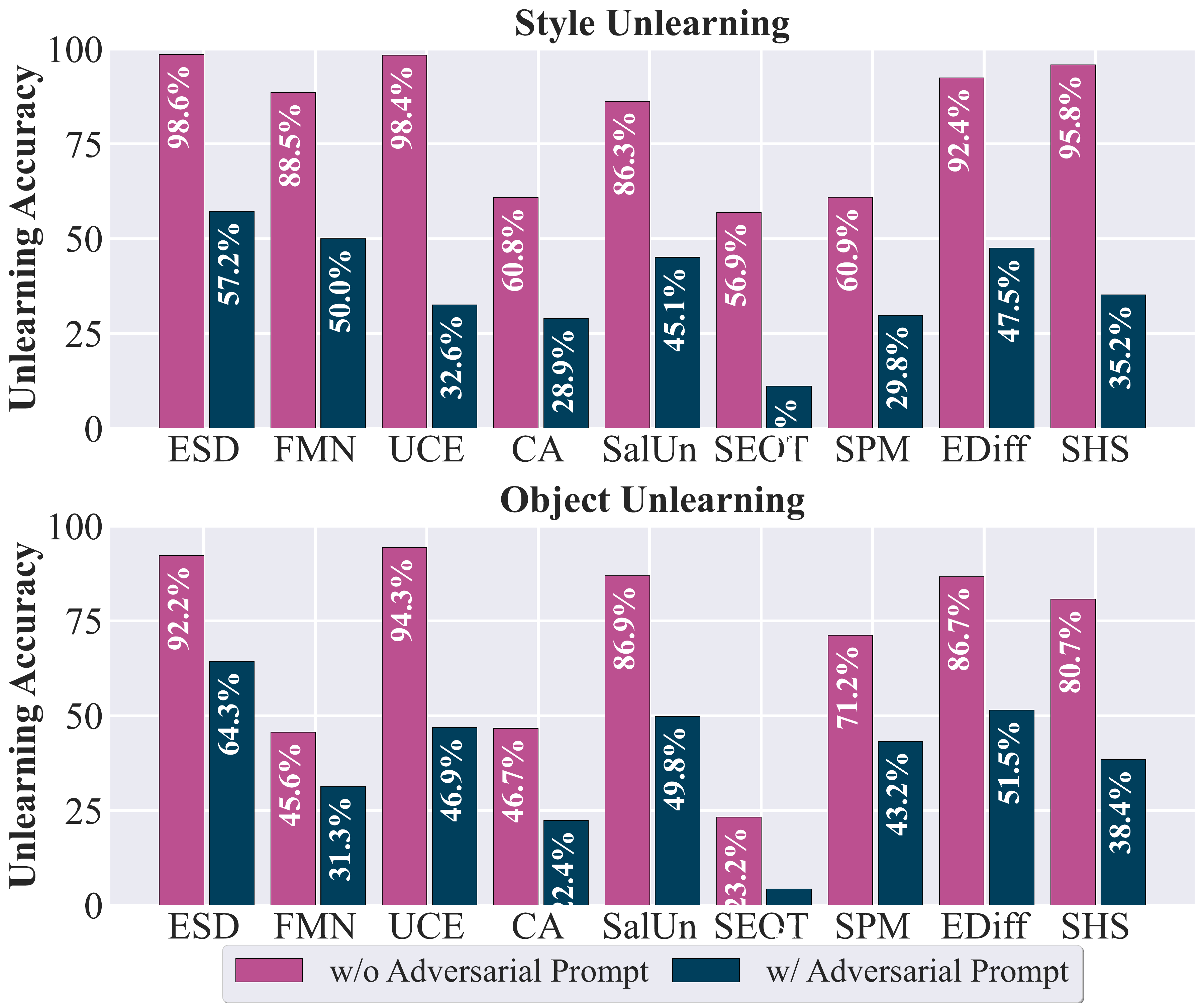}
    \vspace*{-1.7em}
    \caption{UA of DM unlearning against adversarial prompts \cite{zhang2023generate}. Unlearned models in Tab.\,\ref{tab: unlearn_overview} are used as victim models to generate adversarial prompts.}
    \label{fig: adv_prompt}
    \vspace*{-3.9mm}
\end{wrapfigure}

In this section, our benchmarking results are divided into two main parts.
In Sec.\,\ref{sec: exp_basic}, we comprehensively evaluate the existing DM unlearning methods on the established tasks of style and object unlearning. Extensive studies following the proposed evaluation pipeline (Fig.\,\ref{fig: mu_pipeline}) reveal that focusing solely on unlearning effectiveness can lead to a biased perspective for DM unlearning if retainability is not concurrently assessed (Tab.\,\ref{tab: unlearn_overview} \& Fig.\,\ref{fig: unlearn_radar}). We also show that preserving CRA (cross-domain retainability) is more challenging than IRA (in-domain retainability), highlighting a gap in the current literature (Fig.\,\ref{fig: unlearning_dissection}). 
Furthermore, we find that different DM unlearning methods exhibit distinct unlearning mechanisms by examining their unlearning directions (Fig.\,\ref{fig: mu_direction}).
In Sec.\,\ref{sec: exp_more_challenging}, we introduce more challenging unlearning scenarios. First, we assess the robustness of current DM unlearning methods against adversarial prompts crafted based on (Fig.\,\ref{fig: adv_prompt}). Additionally, we examine the performance when facing unlearning targets with finer granularity, formed by style-object concept combinations (Tab.\,\ref{tab: unlearn_combination}).
Further, we leverage {\ours} to provide insights into DM unlearning in a sequential unlearning fashion (Tab.\,\ref{tab: continue_unlearning}).

\subsection{Benchmarking Current DM Unlearning Methods for Style and Object Unlearning}
\label{sec: exp_basic}

\paragraph{Overall performance of DM unlearning.} 
In \textbf{Tab.\,\ref{tab: unlearn_overview}}, we provide an overview of the performance of existing unlearning methods, using the evaluation metrics associated with {\ours}.

\begin{table}[t]
\vspace*{-1em}
\centering
\caption{Performance overview of different DM unlearning methods evaluated on {\ours}. The performance metrics include UA (unlearning accuracy), IRA, CRA, and FID. The symbols $\uparrow$ or $\downarrow$ indicate whether a higher or lower value represents better performance. Results are averaged over all the style and object unlearning cases. The best performance for each metric is highlighted in \congrat{green}, while significantly underperforming results are marked in \warn{red}, indicating areas needing improvement for existing DM unlearning methods.
}
\label{tab: unlearn_overview}
\resizebox{\textwidth}{!}{%
\begin{tabular}{c|ccc|ccc|c|ccc}
\toprule[1pt]
\midrule
\multirow{3}{*}{\textbf{Method}} &
\multicolumn{7}{c|}{\textbf{Effectiveness}} &
\multicolumn{3}{c}{\textbf{Efficiency}} \\
&
\multicolumn{3}{c|}{\textbf{Style Unlearning}} 
& \multicolumn{3}{c|}{\textbf{Object Unlearning}} 
& \multirow{2}{*}{\textbf{FID} ($\downarrow$)} 
& \textbf{Time} 
& \textbf{Memory} 
& \textbf{Storage} \\
& \textbf{UA} ($\uparrow$) & \textbf{IRA} ($\uparrow$) & \textbf{CRA} ($\uparrow$) & \textbf{UA} ($\uparrow$) & \textbf{IRA} ($\uparrow$) & \textbf{CRA} ($\uparrow$) &  &  (s) ($\downarrow$)  &  (GB) ($\downarrow$) & (GB) ($\downarrow$) \\
\midrule
ESD \cite{gandikota2023erasing}  & \congrat{${98.58}\%$} & $80.97\%$ & $93.96\%$ & $92.15\%$ & \warn{${55.78\%}$} & \warn{${44.23\%}$} & $65.55$ & \warn{6163} & 17.8 & 4.3 \\
FMN \cite{zhang2023forget}  & $88.48\%$ & \warn{${56.77\%}$} & \warn{${46.60\%}$} & \warn{${45.64\%}$} & $90.63\%$ & $73.46\%$ & \warn{${131.37}$} & $350$ & 17.9 & 4.2    \\
UCE \cite{gandikota2023unified}  & ${98.40}\%$ & \warn{${60.22\%}$} & \warn{${47.71\%}$} & \congrat{${94.31}\%$} & \warn{${39.35\%}$} & \warn{${34.67\%}$} & \warn{${182.01}$} & 434 & \congrat{${5.1}$} & ${1.7}$ \\
CA \cite{kumari2023ablating}   & \warn{${60.82\%}$} & \congrat{${96.01}\%$} & $92.70\%$ & \warn{${46.67\%}$} & $90.11\%$ & $81.97\%$ & \congrat{${54.21}$} & 734 & 10.1  & 4.2 \\
SalUn \cite{fan2023salun} & $86.26\%$ & $90.39\%$ & $95.08\%$ & $86.91\%$ & \congrat{${96.35}\%$} & \congrat{${99.59}\%$} &  $61.05$ & 667 & \warn{30.8} & 4.0 \\
SEOT \cite{li2024get} & \warn{${56.90\%}$} & $94.68\%$ & $84.31\%$ & \warn{${23.25\%}$} & $95.57\%$ & $82.71\%$ & 62.38 & \congrat{${95}$} & 7.34 & \congrat{0.0}\\
SPM \cite{lyu2023one} & \warn{${60.94\%}$} & $92.39\%$ & $84.33\%$ & $71.25\%$ & $90.79\%$ & $81.65\%$ & 59.79 & \warn{29700} & 6.9 & \congrat{0.0}\\
EDiff \cite{wu2024erasediff} & $92.42\%$ & ${73.91\%}$ & \congrat{${98.93}\%$} & $86.67\%$ & $94.03\%$ & \warn{${48.48\%}$} & 81.42 & 1567 & 27.8 & 4.0\\
SHS \cite{wu2024scissorhands} & $95.84\%$ & $80.42\%$ & \warn{${43.27\%}$} & $80.73\%$ & $81.15\%$ & \warn{${67.99\%}$} & \warn{119.34} & 1223 & \warn{31.2} & 4.0\\
\midrule
\bottomrule[1pt]
\end{tabular}%
}
\vspace*{-1.5em}
\end{table}

\underline{First}, as seen in {Tab.\,\ref{tab: unlearn_overview}}, retainability is essential for a comprehensive assessment of DM unlearning.
For example, in the scenario of style unlearning, ESD and UCE achieve similar UA (both around $98\%$), but their retainability (IRA and CRA) differs significantly, with one over $80\%$ and the other around $60\%$. Therefore, relying solely on UA can provide a skewed view of the performance of DM unlearning.
\underline{Second}, we observe that CRA is harder to retain than IRA. 
For example, UCE exhibits a significant gap between its IRA and CRA, both in style ($60.22\%$ vs. $47.71\%$) and object ($39.35\%$ vs. $34.67\%$) unlearning scenarios. Similar patterns are observed with other methods like FMN, CA, SEOT, SPM, and SHS. This pattern suggests that while MU methods are somewhat effective at preserving concepts within the same domain, they face greater challenges in maintaining performance on unlearning target-unrelated concepts across domains. This issue has been overlooked in previous studies due to the absence of a multi-label dataset like {\ours} for systematic unlearning analyses.
\underline{Third}, we find that a single unlearning method can perform differently across various domains, and no single method excels in all aspects. For example,  FMN achieves a UA of 88.48\% in style unlearning but a much lower UA of 45.64\% in object unlearning. 
A similar phenomenon can be observed with the unlearning methods SEOT, SPM, and SHS, which exhibit significantly large performance gaps between style and object unlearning.
Furthermore, each method exhibits unique strengths and weaknesses. For example, while ESD  typically shows improvement in UA, it may lag in terms of IRA and CRA, particularly in the case of object unlearning.
Some methods, such as SEOT and SPM, achieve extremely high storage efficiency, requiring no additional storage, but suffer from inferior UA (below 80\%). 
These observations underscore the challenges of DM unlearning and highlight the need for further advancements in the field.

\begin{figure*}[htb]
    \centering
    \includegraphics[width=\linewidth]{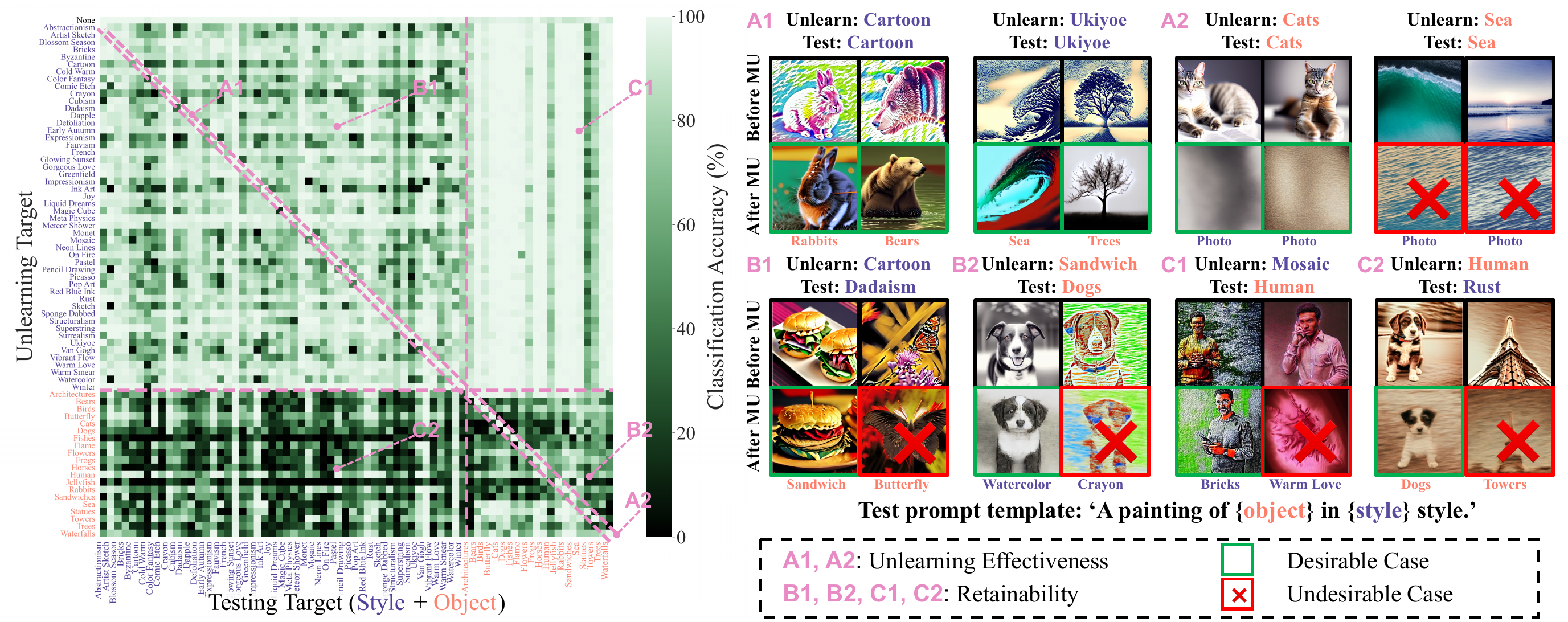}
    \vspace*{-2em}
    \caption{\textit{Left}: Heatmap visualization of the unlearning accuracy and retainability of ESD  on {\ours}. The $x$-axis shows the tested concepts for image generation using the unlearned model, while the $y$-axis indicates the unlearning target. Concept types are distinguished by color: \textcolor{styleblue}{styles} in blue and \textcolor{orange}{objects} in orange.  The figure is divided into regions representing corresponding evaluation metrics and unlearning scopes ($A$ for UA, $B$ for IRA, $C$ for CRA; `1' for style unlearning, `2' for object unlearning). 
    Higher values in lighter colors denote better performance. The first row serves as a reference for comparison before unlearning. 
    Zooming into the figure is recommended for detailed observation.
    \textit{Right}: Representative cases illustrating each region with images generated before and after unlearning a specific concept.
    }
    \vspace*{-1em}
    \label{fig: unlearning_dissection}
\end{figure*}

\paragraph{A closer look into retainability: A case study on ESD.} We next perform an in-depth analysis of DMs' retainability after unlearning, focusing on the most popular ESD method \cite{gandikota2023erasing}. In \textbf{Fig.\,\ref{fig: unlearning_dissection}  (left)}, we present a heatmap illustrating the per-style/object unlearning accuracy (diagonal) and retain accuracy (off-diagonal) for ESD. The heatmap segments are labeled to indicate different regions of interest (ROIs) in style and object unlearning related to our evaluation metrics: \textbf{A1}/\textbf{A2} for UA, \textbf{B1}/\textbf{B2} for IRA, and \textbf{C1}/\textbf{C2} for CRA. Key observations are highlighted below.

\underline{First},  we observe distinct behaviors of ESD in unlearning styles compared to objects. For example, ROI C2 appears darker than C1, suggesting a reduced ability to retain styles during object unlearning. Similar patterns are seen between B2 and B1. For a clearer explanation, \textbf{Fig.\,\ref{fig: unlearning_dissection} (right)} provides image generation examples of the DM before and after unlearning within different ROIs.
\underline{Second}, We observe that style/object unlearning is relatively easier compared to retaining the generation performance of unlearned DMs conditioned on unlearning-unrelated prompts. This is evident in the contrast between ROIs A1/A2 and ROIs B1/B2 or C1/C2. One exception is the case of unlearning the object `Sea', which exhibits relatively low UA in A2. This may be attributed to post-unlearning image generation still resembling sea waves, as shown by the image examples in Fig.\,\ref{fig: unlearning_dissection} (right) for A2.
\underline{Third}, we observe an inherent trade-off between unlearning effectiveness and retainability. For example, while SalUn exhibits more consistent performance across different scenarios than ESD, it does not achieve the same level of UA, as indicated by the lighter diagonal values in Fig.\,\ref{fig: salun_heatmap}. For the heatmap performance of other unlearning methods, please refer to Appx.\,\ref{app: additional_exp}.

\begin{figure}[htb]
    \centering
    \begin{tabular}[c]{ccc}
    \hspace*{-1em} \includegraphics[width=.3\linewidth]{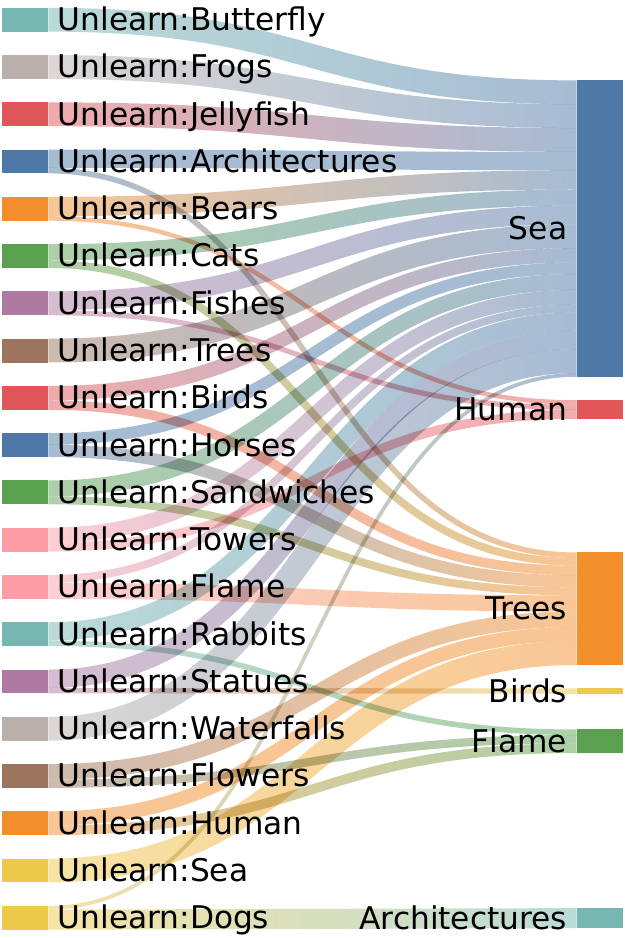}
    &\hspace*{0em} \includegraphics[width=.3\linewidth]{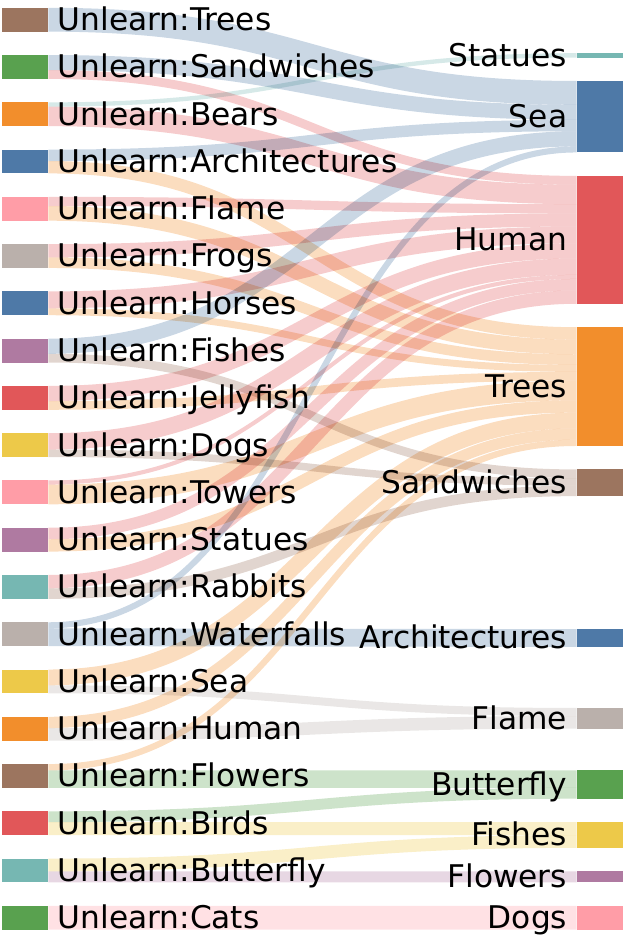} 
    & \hspace*{0em} \includegraphics[width=.3\linewidth]{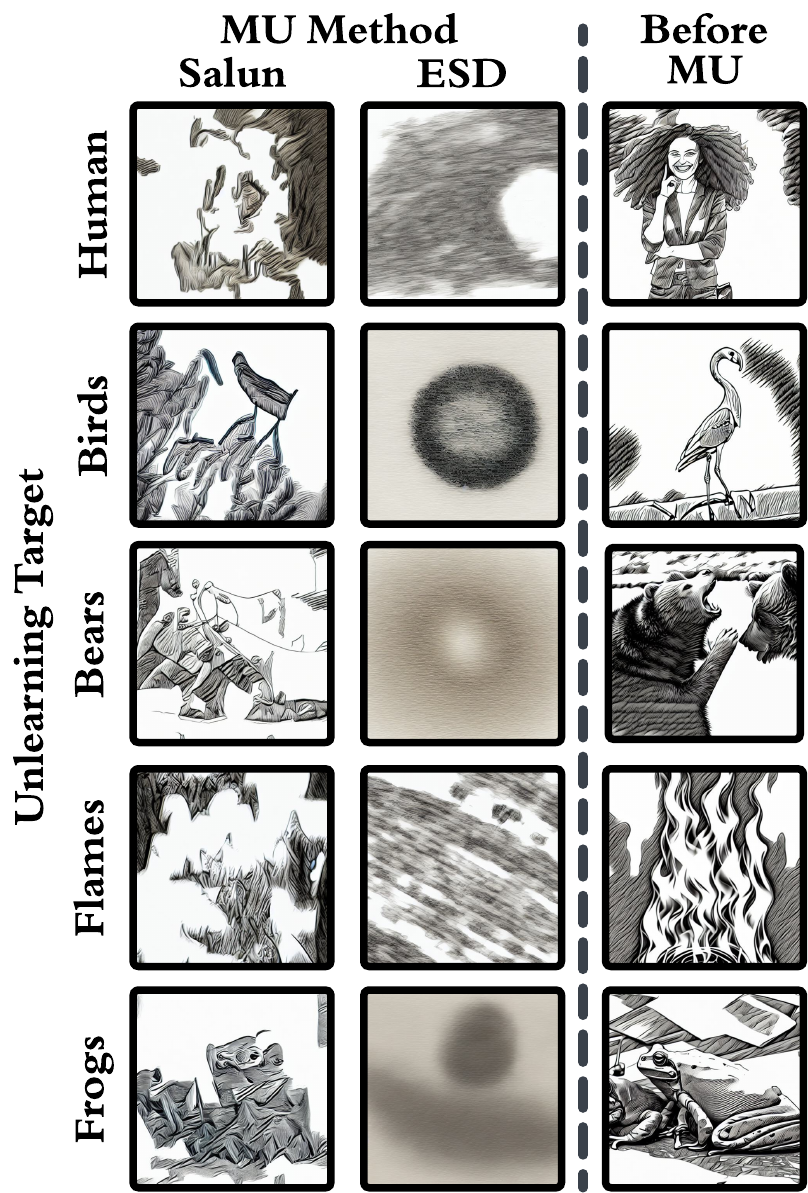} 
    \\
    \hspace*{-1em} (a) ESD
    & \hspace*{0em} (b) SalUn 
    & \hspace*{0em} (c) Illustration 
    \vspace*{-.5em}
    \end{tabular}
    \caption{
    Visualization of the unlearning directions of (a) ESD and (b) SalUn. This figure illustrates the conceptual shift of the generated images of an unlearned model conditioned on the unlearning target.  
    Images generated by the post-unlearning models are classified and used to understand this shift. 
    Edges leading from the object in the left column to the right signify that images generated conditioned on unlearning targets are instead classified as the shifted concepts after unlearning. 
    This reveals the primary unlearning direction for each unlearning method. The most dominant unlearning direction for an object is visualized. 
    Figure (c) provides visualizations of generated images using the prompt template `A painting of \{\textit{object}\} in Sketch style.' with \textit{object} being each unlearning target.
    }
    \vspace*{-1em}
    \label{fig: mu_direction}
\end{figure}

\paragraph{Understanding unlearning method's behavior via unlearning directions.}
As noted earlier, different unlearning methods display distinct unlearning behaviors. To gain insights into the underlying reasons for these differences, \textbf{Fig.\,\ref{fig: mu_direction}} (a) and (b) visualize the `unlearning directions' for ESD and SalUn, respectively. These unlearning directions are determined by connecting the unlearning target with the predicted label of the generated image from the unlearned DM conditioned on the unlearning target.
As shown in Fig.\,\ref{fig: mu_direction} (a), ESD demonstrates a focused shift in image generation after object unlearning, with a predominant transition towards generating images labeled by `Sea' and `Trees'.
This behavior arises from ESD's optimization process,  designed to steer the generation of the DM away from a predefined concept.
Consequently, images generated by the ESD-induced unlearned model consistently lack clearly identifiable objects, resembling waves and trees, which leads to their classification into the `Sea' and `Trees' classes; see Fig.\,\ref{fig: mu_direction} (c) for examples of generated images.
In contrast, SalUn exhibits a more diverse range of unlearning directions, shifting images to 11 different objects. This diversity results from SalUn's requirement to replace the unlearning target with a random concept. As shown in Fig.\,\ref{fig: mu_direction} (c), images generated by SalUn post-object unlearning still maintain some object contours (different from the original unlearning target) and better retain style information compared to ESD.

\subsection{Benchmarking Current DM Unlearning Methods in More Challenging Scenarios}
\label{sec: exp_more_challenging}

\paragraph{Unlearning robustness against adversarial prompts.} Recent studies \cite{zhang2023generate,chin2023prompting4debugging} have highlighted the vulnerabilities of DM unlearning to adversarial prompts, such as jailbreak attacks. In \textbf{Fig.\,\ref{fig: adv_prompt}}, we use the state-of-the-art attacking method, UnlearnDiffAtk \cite{zhang2023generate}, to craft adversarial prompts based on {\ours} and evaluate the robustness of unlearned DMs shown in Tab.\,\ref{tab: unlearn_overview}. See Appx.\,\ref{app: exp_setting_adv_prompt} for detailed setup.
As we can see, all the DM unlearning methods experience a significant drop in UA, falling below $60\%$ when confronted with adversarial prompts. Notably, methods like UCE, SEOT, and SHS exhibit particularly steep declines. Moreover, a high UA under normal conditions does not necessarily imply robustness to adversarial attacks. For instance, although UCE achieves a UA over 90\% in normal settings, its performance plummets to below 50\% against adversarial prompts. This stark contrast highlights the importance of the worst-case evaluation for MU methods.

\begin{wraptable}{r}{.5\linewidth}
\vspace*{-1.5em}
\centering
\caption{Performance of unlearning style-object combinations. The assessment includes UA and retainability in three contexts: SC (style consistency), OC (object consistency), and UP (unrelated prompting).
The best result in each metric is highlighted in \congrat{green}.
}
\label{tab: unlearn_combination}
\resizebox{\linewidth}{!}{%
\begin{tabular}{c|c|ccc}
\toprule[1pt]
\midrule
{\textbf{Method}}
&
{\textbf{UA}}
& 
\textbf{SC} & \textbf{OC} & \textbf{UP} \\
\midrule
ESD \cite{gandikota2023erasing} & \congrat{$91.42\%$} & {${4.88\%}$} & {${14.72\%}$} & $84.38\%$ \\
FMN \cite{zhang2023forget} & {${45.37\%}$} & \congrat{$68.73\%$} & $62.74\%$ & $83.25\%$ \\
UCE \cite{gandikota2023unified} & $75.97\%$ & {${4.53\%}$} & {${5.72\%}$} & {${35.42\%}$} \\
CA \cite{kumari2023ablating} & {${47.92\%}$} & {${10.08\%}$} & {${56.35\%}$} & $81.54\%$\\
SalUn \cite{fan2023salun} & {${42.21\%}$} & $62.45\%$ & \congrat{$70.93\%$} & \congrat{$87.28\%$} \\
SEOT \cite{li2024get} & {${29.32\%}$} & {${45.31\%}$} & {${53.64\%}$} & $85.45\%$ \\
SPM \cite{lyu2023one} & {${45.72\%}$} & {${41.34\%}$} & {${36.32\%}$} & $67.82\%$ \\
EDiff \cite{wu2024erasediff} & $71.33\%$ & {${35.23\%}$} & {${26.32\%}$} & {${51.52\%}$} \\
SHS \cite{wu2024scissorhands} & {${55.32\%}$} & {${14.34\%}$} & {${24.32\%}$} & $83.95\%$ \\
\midrule
\bottomrule[1pt]
\end{tabular}%
}
\vspace*{-0.8em}
\end{wraptable}

\paragraph{Unlearning at a finer scale: Style-object combinations as unlearning targets.} 
Next, we consider the fine granularity of the unlearning target, defined by a style-object combination, such as ``An image of dogs in Van Gogh style''. This unlearning challenge requires the unlearned model to avoid affecting image generation for dogs in  non-Van Gogh styles and Van Gogh-style images with non-dog objects.
See Appx.\,\ref{app: exp_setting_combination} for detailed setup.
Here, besides UA, the retainability performance is assessed in three contexts. (a) {Style consistency} (\textbf{SC}): retainability under prompts with different objects but in the same style, \textit{e.g.}, ``An image of \textit{cats} in Van Gogh style''.
(b) Object consistency (\textbf{OC}): retainability under prompts featuring the same object in different styles, \textit{e.g.}, ``An image of dogs in \textit{Picasso} style''. 
(c) Unrelated prompting (\textbf{UP}): retainability for all other prompts. \textbf{Tab.\,\ref{tab: unlearn_overview}} presents the performance of unlearning  style-object combinations.
Style-object combinations are more challenging to unlearn than individual objects or styles, as evidenced by a significant drop in UA—over 20\% lower compared to values in Tab.\,\ref{tab: unlearn_overview}. Retainability drops to below 20\% for top-performing methods like ESD and UCE, originally highlighted for their efficacy. This is presumably due to ESD's underlying unlearning mechanism, which requires only a single prompt, resulting in a poor ability to precisely define the unlearning scope.

\paragraph{Evaluation in sequential unlearning.} 
Furthermore, we evaluate the unlearning performance in sequential unlearning (SU) \cite{jang2022knowledge,chen2023unlearn}, where unlearning requests arrive sequentially. This parallels the continual learning (CL) task, which requires models to not only unlearn new targets effectively but also maintain the unlearning of previous targets while retaining all other knowledge. Here, we consider unlearning 6 styles sequentially and the results are presented in \textbf{Tab.\,\ref{tab: continue_unlearning}}. We remark that the method SEOT does not support sequential unlearning in its original implementation and thus is not included in Tab.\,\ref{tab: continue_unlearning}.
Our findings reveal significant insights.
(1) \textit{Degraded retainability}: Sequential unlearning requests generally degrade retainability across all methods, with RA values frequently dropping below the average levels previously seen in Tab.\,\ref{tab: unlearn_overview}. Here RA is given by the average of IRA and CRA.
(2) \textit{Unlearning rebound effect}: Knowledge previously unlearned can be inadvertently reactivated by new unlearning requests. This is evidenced by decreasing UA values for earlier objectives as more unlearning tasks are introduced. This suggests that residual knowledge remains within the model and can be reactivated, aligning with findings from Fig.\,\ref{fig: adv_prompt}. This indicates the unlearned models by some MU methods do not essentially lose the generation ability of the unlearning target.
(3) \textit{Catastrophic retaining failure}: RA significantly drops at a certain request, exemplified by a sudden decrease in RA of UCE from $81.42\%$ to $29.38\%$ after the second request, $\mathcal{T}_2$. This indicates that the seemingly acceptable side effects generated by some unlearning methods will drastically modify the knowledge representations when accumulated. This experiment illuminates the complex dynamics of knowledge removal and retention within DMs and highlights the potential pitfalls of existing unlearning methods when faced with sequential unlearning tasks. The observation of the `unlearning rebound effect' and `catastrophic retaining failure' particularly emphasizes the need for a more nuanced understanding of how knowledge is managed within DMs.

\paragraph{Visualizations.} We provide plenty of visualizations, illustrations, and qualitative results. These visualizations are intended to deepen the understanding of the effects of different MU methods and clearly illustrate the challenges identified in previous sections. Specifically, in Fig.\,\ref{fig: visual_style_unlearning}, we provide abundant generation examples of all the 9 methods benchmarked in this work in a case study of unlearning the `Cartoon' style. Both the successful and failure cases are demonstrated in the context of unlearning effectiveness, in-domain retainability, and cross-domain retainability. Besides, in Fig.\,\ref{fig: visual_adv_prompt}, we provide visualizations for the effect of adversarial prompts enabled by UnlearnDiffAtk\,\cite{zhang2023generate}.

%% file: sections/conclusion.tex
\section{Conclusion, Discussion, and Limitation}
\label{sec: conclusion}

In this paper, we systematically reviewed existing MU (machine unlearning) methods on DMs (diffusion models) and identified key challenges in their evaluation systems that could lead to incomplete, inaccurate, and biased assessments. In response, we propose {\ours}, a high-resolution stylized image dataset designed to facilitate comprehensive evaluation of MU methods. We also introduce novel systematic evaluation metrics.
By benchmarking nine state-of-the-art MU methods, we reveal novel insights into their strengths and weaknesses and also deepen the understanding of their underlying mechanisms.
Our analysis of three challenging unlearning tasks highlights significant shortcomings of current methods, including a lack of robustness, difficulties in fine-scale unlearning, and issues arising from sequential unlearning tasks. These findings also point to meaningful future research directions aimed at developing satisfactory and practical MU methods for real-world applications. Furthermore, we recognize limitations in our benchmark, such as its focus on specific Stable Diffusion models and the text-to-image task in DMs. We hope this work serves as a foundation for future research to broaden MU evaluations across a wider range of model architectures and tasks in DMs.

%% file: sections/acknowledgement.tex
\section*{Acknowledgement}

The work of Y. Zhang, C. Fan, Y. Zhang, Y. Yao, J. Jia, J. Liu, and S. Liu was supported by the Cisco Research Award, the ARO Award W911NF2310343, the National Science Foundation (NSF) CISE Core Program Award IIS-2207052, and the NSF CAREER Award IIS-2338068, and the Amazon Research Award for AI in Information Security. 

%% file: sections/checklist.tex
\section*{Checklist}

\begin{enumerate}

\item For all authors...
\begin{enumerate}
  \item Do the main claims made in the abstract and introduction accurately reflect the paper's contributions and scope?
    \answerYes{The introduction (Sec.\,\ref{sec: intro}) clearly specifies the scope of this paper and the major contributions.}
  \item Did you describe the limitations of your work?
    \answerYes{The limitations are discussed in Sec.\,\ref{sec: conclusion}}
  \item Did you discuss any potential negative societal impacts of your work?
    \answerNo{This work aims to improve the evaluation of machine unlearning technique, which will bring many positive societal impacts, such as mitigating copyright infringement, avoiding sexual or violent content generation, and mitigating the biases and stereotypes of the diffusion models. Given the technical focus of this work, there are no specific negative societal consequences directly stemming from it that need to be highlighted here.}
  \item Have you read the ethics review guidelines and ensured that your paper conforms to them?
    \answerYes{}
\end{enumerate}

\item If you are including theoretical results...
\begin{enumerate}
  \item Did you state the full set of assumptions of all theoretical results?
    \answerNA{There are no theoretical results in this paper.}
	\item Did you include complete proofs of all theoretical results?
    \answerNA{There are no theoretical results in this paper.}
\end{enumerate}

\item If you ran experiments (e.g. for benchmarks)...
\begin{enumerate}
  \item Did you include the code, data, and instructions needed to reproduce the main experimental results (either in the supplemental material or as a URL)?
    \answerYes{There are detailed instructions of the codes in the github repo to facilitate reproducing the results. The detailed experiment settings are also disclosed in Appx.\,\ref{app: exp_setting}.}
  \item Did you specify all the training details (e.g., data splits, hyperparameters, how they were chosen)?
    \answerYes{Detailed training settings are disclosed in Sec.\,\ref{sec: mu_experiment_results} and Appx.\,\ref{app: additional_exp}.}
	\item Did you report error bars (e.g., with respect to the random seed after running experiments multiple times)?
    \answerNo{The results reported in this work were averaged over dozens of individual unlearning cases, and the variance confirmed to be below 1\%. We believe this indicates negligible variability. Given the large scale of the experiments, error bars were omitted to enhance readability and maintain focus on the key findings.}
	\item Did you include the total amount of compute and the type of resources used (e.g., type of GPUs, internal cluster, or cloud provider)?
    \answerYes{The compute resources are disclosed in Appx.\,\ref{app: exp_setting}.}
\end{enumerate}

\item If you are using existing assets (e.g., code, data, models) or curating/releasing new assets...
\begin{enumerate}
  \item If your work uses existing assets, did you cite the creators?
    \answerYes{The seed images used to curate {\ours} belong to the existing assets, and their sources are properly cited in Sec.\,\ref{sec: dataset_proposal}.}
  \item Did you mention the license of the assets?
    \answerYes{}
  \item Did you include any new assets either in the supplemental material or as a URL?
    \answerYes{This paper includes a new dataset {\ours} and the link is released both in the abstract as well as in Appx.\,\ref{app: dataset_overview}.}
  \item Did you discuss whether and how consent was obtained from people whose data you're using/curating?
    \answerYes{The data being used is collected from an open-source photography website and this is mentioned in Sec.\,\ref{sec: dataset_proposal}.}
  \item Did you discuss whether the data you are using/curating contains personally identifiable information or offensive content?
    \answerYes{The data does not contain any personally identifiable information or offensive content, as discussed in Appx.\,\ref{app: dataset_overview}.}
\end{enumerate}

\item If you used crowdsourcing or conducted research with human subjects...
\begin{enumerate}
  \item Did you include the full text of instructions given to participants and screenshots, if applicable?
    \answerNA{No human subjects are involved.}
  \item Did you describe any potential participant risks, with links to Institutional Review Board (IRB) approvals, if applicable?
    \answerNA{No human subjects are involved.}
  \item Did you include the estimated hourly wage paid to participants and the total amount spent on participant compensation?
    \answerNA{No human subjects are involved.}
\end{enumerate}

\end{enumerate}

%% file: sections/appendix.tex
\clearpage\newpage
\onecolumn
\section*{\Large{Appendix}}
\setcounter{section}{0}
\setcounter{figure}{0}
\setcounter{table}{0}
\makeatletter 
\renewcommand{\thesection}{\Alph{section}}
\renewcommand{\theHsection}{\Alph{section}}
\renewcommand{\thefigure}{A\arabic{figure}} 
\renewcommand{\theHfigure}{A\arabic{figure}} 
\renewcommand{\thetable}{A\arabic{table}}
\renewcommand{\theHtable}{A\arabic{table}}
\makeatother

\renewcommand{\thetable}{A\arabic{table}}
\setcounter{mylemma}{0}
\renewcommand{\themylemma}{A\arabic{mylemma}}
\setcounter{equation}{0}
\renewcommand{\theequation}{A\arabic{equation}}

\begin{center}
{
\large
\textbf{Table of Contents}
}    
\end{center}

\startcontents
\printcontents{ }{1}{}
\clearpage
\newpage

\section{{\ours} Dataset Details}
\label{app: dataset_overview}

In this section, we provide a detailed description of the dataset following `datasheet for datasets' \cite{gebru2021datasheets}.

\subsection{Motivation} This dataset is collected to enable a precise, comprehensive, and automated quantitative evaluation framework for MU (machine unlearning) methods in DMs (diffusion models). The current evaluation plans used in the existing literature have exposed several weaknesses, which may lead to incomplete, inaccurate or even biased results, see a more detailed discussion in Sec.\,\ref{sec: preliminary}. To the best of our knowledge, there are no datasets specifically designed to meet the assessing requirements of DM unlearning. Therefore, {\ours} is designed, collected, and made to fill in this gap.

\subsection{Composition: Styles and Object Classes} There are 60 predetermined artistic styles provided by Fotor\,\cite{fotor}. The images in the same artistic style all share high stylistic consistency, which enables a high-precision style classifier to be trained on them. In Fig.\,\ref{fig: dataset_style_details}, we list some examples of the images in each style to illustrate these styles. There are 20 distinct object classes in {\ours}. In Fig.\,\ref{fig: dataset_object_details}, we list some examples of the images in each object to illustrate these classes.

\subsection{Collection and Labeling Process} The construction of {\ours} involves two main steps: seed image collection and subsequent image stylization; see Fig.\,\ref{fig: dataset_overview} for an illustration. For seed image collection, a set of high-resolution real-world photos are collected from Pexels\,\cite{pexels}, providing open-sourced photographs. There are 20 seed images collected for each of the 20 object classes; see Fig.\,\ref{fig: dataset_object_details}. After collecting the seed images, we stylize each and every seed image into all 60 predetermined artistic styles; see Fig.\,\ref{fig: dataset_style_details} with Fotor\,\cite{fotor}. After the stylization of all the images, the dataset is structured in a hierarchical manner, and each image is labeled with both its style and object classes. In order to support text-to-image training, each image is annotated with the prompt `An image of \textit{object} in \textit{style} style.'.

\subsection{Uses} {\ours} can be used to evaluate MU methods in different unlearning scenarios. Please see Sec.\,\ref{sec: mu_experiment_results} for more details. In addition, we stress that {\ours} can be used for more real-world tasks than unlearning, and we provide an example of how {\ours} can be used to systematically evaluate another generative task, style transfer, in Appx.\,\ref{app: broader}. We provide very detailed instructions with codes in the \href{https://github.com/OPTML-Group/UnlearnCanvas}{GitHub code repository}.

\subsection{Distribution} {\ours} is an open-sourced dataset and is based on the existing open-sourced data \cite{pexels}. We make access to the dataset public under the MIT license. We remark that no personally identifiable information or offensive content is included in this dataset. The dataset can be accessed either through \href{https://drive.google.com/drive/folders/1sJBLhng3I2fD3dbEzxx_Jvm1bMBcMQH7?usp=sharing}{Google Drive} and \href{https://huggingface.co/datasets/OPTML-Group/UnlearnCanvas}{HuggingFace}. 
More resources on the dataset, such as the introduction video and the benchmark, can be found in the official \textit{\href{https://openreview.net/attachment?id=VH1vxapUTs&name=supplementary_material}{project webpage}}.

\subsection{Maintenance} The dataset will be maintained by the lead author Yihua Zhang. If needed, the email for contacting is zhan1908@msu.edu. The dataset may be updated if needed (with the inclusion of more seed images, more artistic styles, and more objects). The updates will be ad-hoc and will not be periodical. Each time the dataset is updated, the updates will be reflected in the same \href{https://github.com/OPTML-Group/UnlearnCanvas}{GitHub code repository}.

\subsection{Author Statements} The collector and the lead author of this dataset, Yihua Zhang, bears full responsibility for any violation of rights that may arise from the collection of the data included in this research.

\section{Reproducibility Statement and Detailed Experiment Settings}
\label{app: exp_setting}

In this section, we provide detailed instructions on the reproduction of our results in Sec.\,\ref{sec: mu_experiment_results}, including the settings of training, the implementation details of the tested machine unlearning methods, and the evaluation details in each unlearning scenario. We denote in Sec.\,\ref{sec: mu_experiment_results}, 50 out of the 60 styles proposed in {\ours} are studied and evaluated, with the other 10 styles being intentionally reserved for future explorations, such as generalizations.

\subsection{Finetuning Style and Object Classifier with {\ours}} 
\label{app: exp_setting_classifier_training}

Style and object classifiers need to be trained as part of the testbed proposed in our evaluation pipeline (Fig.\,\ref{fig: mu_pipeline}). Here, we adopted a   ViT-L/16 model \cite{dosovitskiy2020image} pretrained on ImageNet and finetune it on {\ours}. {\ours} are split into the train set and test set with a ratio of $9:1$. After hyper-parameter tuning, the classifiers are trained with Adam optimizer at a learning rate of $0.01$ for 10 epochs. 

\subsection{Finetuning Stable Diffusion with {\ours}}
\label{app: exp_setting_sd_finetuning}

The other part of the testbed is a diffusion model capable of generating high quality images in all the styles associated with all the objects encompassed in {\ours} in order to guarantee a trustworthy and unbiased evaluation. 

\paragraph{Training settings.} Practically, we finetune the pretrained Stable Diffusion (SD) v1.5 on {\ours} for 20k steps with a learning rate of $1e-6$. Unless otherwise stated, we strictly follow the training configurations used in Stable Diffusion \cite{rombach2022high}. For each image in {\ours}, we annotate the data with text prompt \texttt{An image of \{\textit{object}\} in \{\textit{style}\}}, where the \textit{object} and \textit{style} are the corresponding object and style label. For seed images, the \textit{style} label we use is `photo' style. We use the training scripts provided by Diffuser official tutorial (\url{https://huggingface.co/docs/diffusers/v0.13.0/en/training/text2image}) 
and the pretraining model card is \texttt{runwayml/stable-diffusion-v1-5}. During training, the checkpoints will be saved every 1000 steps.

\paragraph{Evaluation.} To evaluate the quality of the saved checkpoints and selcect the best one for unlearning study, the checkpoints are first used to generate an image set with the same prompt as training (\texttt{An image of \{\textit{object}\} in \{\textit{style}\}}) by traversing all the possible style and object labels. Each prompt are used to generate $5$ images with different random seed. Each image are sampled with 100 steps with a guidance coeeficient of $9$. The image set for each checkpoint are fed into the style and object classifier trained in Appx.\,\ref{app: exp_setting_classifier_training}. The model with the highest average performance on all the styles and objects are selected as the testbed for MU study. The classification performance are also used as a reference for later IRA/CRA comparison, which are disclosed in the first row of Fig.\,\ref{fig: unlearning_dissection} (left).

\paragraph{Computing resource.} In this work, we employ $40\times$ NVIDIA RTX A6000 GPUs to conduct all the model training, unlearning, image generation, and evaluations. When we finetuned the StableDiffusion on {\ours}, $8\times$ GPUs were used for parallel computing. Other experiments were all carried out in a single-GPU environment. Around 60,000 GPU hours in total were spent to complete all the experiments.

\subsection{Implementation of DM Unlearning Methods Studied in This Work}

In this work, we inspected a series of stateful MU methods for DMs. For each method, we use their publicly released source codes as code bases, which are listed below:
\begin{itemize}
    \item ESD \cite{gandikota2023erasing}: \url{https://github.com/rohitgandikota/erasing}
    \item CA \cite{kumari2023ablating}: \url{https://github.com/nupurkmr9/concept-ablation}
    \item UCE \cite{gandikota2023unified}: \url{https://github.com/rohitgandikota/unified-concept-editing}
    \item FMN \cite{zhang2023forget}: \url{https://github.com/SHI-Labs/Forget-Me-Not}
    \item SalUn: \cite{fan2023salun}: \url{https://github.com/OPTML-Group/Unlearn-Saliency}
    \item SEOT: \cite{li2024get}: \url{https://github.com/sen-mao/SuppressEOT}
    \item SPM: \cite{lyu2023one}: \url{https://github.com/Con6924/SPM}
    \item EDiff: \cite{wu2024erasediff}: \url{https://github.com/JingWu321/EraseDiff}
    \item SHS: \cite{wu2024scissorhands}: \url{https://github.com/JingWu321/Scissorhands}
\end{itemize}

In particular, we adopt the following training settings to adapt the methods to our dataset:

\begin{itemize}
    \item ESD: Based on the suggestions from the authors, ESD is used to only finetune the cross attention-related model weights (ESD-x). Other settings strict follow the ones used in the paper.
    \item CA: In order to ablate concepts using CA, we first use ChatGPT to generate a list of simple prompts for each concept (including the styles and the objects), namely anchor prompts. Each anchor prompt is a simple one-sentence description of the unlearning target.
    \item UCE: This method requires a guided concept (prompt) for each unlearning concept. For style unlearning, we use the prompt \texttt{An image in \{\textit{style*}\}} as the guided concept, where \textit{\texttt{style*}} represents the next style in {\ours} in alphabetical order. Similarly, \texttt{An image of \textit{object*}} is used for object unlearning, where \textit{\texttt{object*}} is the next object in {\ours} in alphabetical order.
    \item FMN: This method requires the images associated with the unlearning target. For simplicity and best performance, we randomly select 20 images associated with the unlearning concept. For the first stage of FMN, we run text inversion for 500 steps with a learning rate of $1e-4$, and for the second step, we used the inversed text to unlearn the cross attention layers of the model for $100$ steps. The hyper-parameters of learning rate, maximum steps, and tunable parameters (cross-attention or non-cross-attention) are carefully tuned with grid search.
    \item SalUn: This method involves two steps, the mask finding (weight saliancy analysis) and the model unlearning. For mask finding, we tuned the mask ratio, while for unlearning, we tune the hyper-parameter learning rate and unlearning intensity. All the parameters are tuned with grid search. For both steps, the mask or model is trained with $10$ epochs.
    \item SEOT: To generate unlearned images, we use the prompt \texttt{An \{\textit{object*}\} image in \{\textit{style*}\}}. We then suppress either {\texttt{object*}} or {\texttt{style*}} individually. Other settings strict follow the ones used in the paper.
    \item SPM: Following the hyperparameters provided by the authors, we trained and obtained Pre-tuned SPMs for all {\texttt{object*}} and {\texttt{style*}}. During image generation, we combine the pre-tuned SPMs with the DM. By calculating the association between words in the prompt and the target word, we determine whether to allow the specified word to preserve, and generate the corresponding image.
    \item EDiff: Based on the authors' suggestions, EDiff is used to finetune only the cross-attention-related model weights (EraseDiff-x). During the unlearning process, we adjusted the hyperparameters, specifically the learning rate and the number of unlearning epochs. The model is trained with $5$ epochs.
    \item SHS: SHS consists of two stages: trimming and repairing. During the trimming stage, certain weights are re-initialized. The repairing stage then restores the model's utility. Throughout the unlearning process, we finetuned the hyperparameters, focusing on the learning rate and the number of unlearning epochs. Ultimately, we selected $2$ epochs as the optimal number.
\end{itemize}

\subsection{Evaluation Details of the Adversarial Prompt Generation for Unlearning Robustness}
\label{app: exp_setting_adv_prompt}

In Sec.\,\ref{sec: mu_experiment_results}, we evaluated the robustness of different MU methods against adversarial prompts. Here, we use the state-of-the-art method, UnlearnDiffAtk\,\cite{zhang2023generate} to generate adversarial prompts. We set the prepended prompt perturbations by $N=5$ tokens for both style and object unlearning. Following the original attack setting in UnlearnDiffAtk \cite{zhang2023generate}, to optimize the adversarial perturbations, we sample 50 diffusion time steps and perform PGD running for 40 iterations with a learning rate of 0.01 at each step. Prior to projection onto the discrete text space, we utilize the AdamW optimizer.

\subsection{Experiment Details of the Style-Object Combination Unlearning}
\label{app: exp_setting_combination}

\paragraph{Unlearning targets.} In Sec.\,\ref{sec: mu_experiment_results}, we evaluate the capability of different MU methods on performing unlearning at a finer scale, and we use the style-object combinations as unlearning targets for evaluation. Ideally, the {\ours} dataset can generate 1200 ($60\times20$) style-object combinations. In this work, we randomly select 50 of these combinations for evaluation and we list these combinations below. For each method, the same hyper-parameters are used for each MU method as the ones for style and object unlearning in Tab.\,\ref{tab: unlearn_overview}. The unlearning targets include:
\begin{itemize}
\item `An image of Architectures in Abstractionism style.'
\item `An image of Bears in Artist Sketch style.'
\item `An image of Birds in Blossom Season style.'
\item `An image of Butterfly in Bricks style.'
\item `An image of Cats in Byzantine style.'
\item `An image of Dogs in Cartoon style.'
\item `An image of Fishes in Cold Warm style.'
\item `An image of Flame in Color Fantasy style.'
\item `An image of Flowers in Comic Etch style.'
\item `An image of Frogs in Crayon style.'
\item `An image of Horses in Cubism style.'
\item `An image of Human in Dadaism style.'
\item `An image of Jellyfish in Dapple style.'
\item `An image of Rabbits in Defoliation style.'
\item `An image of Sandwiches in Early Autumn style.'
\item `An image of Sea in Expressionism style.'
\item `An image of Statues in Fauvism style.'
\item `An image of Towers in French style.'
\item `An image of Trees in Glowing Sunset style.'
\item `An image of Waterfalls in Gorgeous Love style.'
\item `An image of Architectures in Greenfield style.'
\item `An image of Bears in Impressionism style.'
\item `An image of Birds in Ink Art style.'
\item `An image of Butterfly in Joy style.'
\item `An image of Cats in Liquid Dreams style.'
\item `An image of Dogs in Magic Cube style.'
\item `An image of Fishes in Meta Physics style.'
\item `An image of Flame in Meteor Shower style.'
\item `An image of Flowers in Monet style.'
\item `An image of Frogs in Mosaic style.'
\item `An image of Horses in Neon Lines style.'
\item `An image of Human in On Fire style.'
\item `An image of Jellyfish in Pastel style.'
\item `An image of Rabbits in Pencil Drawing style.'
\item `An image of Sandwiches in Picasso style.'
\item `An image of Sea in Pop Art style.'
\item `An image of Statues in Red Blue Ink style.'
\item `An image of Towers in Rust style.'
\item `An image of Waterfalls in Sketch style.'
\item `An image of Architectures in Sponge Dabbed style.'
\item `An image of Bears in Structuralism style.'
\item `An image of Birds in Superstring style.'
\item `An image of Butterfly in Surrealism style.'
\item `An image of Cats in Ukiyoe style.'
\item `An image of Dogs in Van Gogh style.'
\item `An image of Fishes in Vibrant Flow style.'
\item `An image of Flame in Warm Love style.'
\item `An image of Flowers in Warm Smear style.'
\item `An image of Frogs in Watercolor style.'
\item `An image of Horses in Winter style.'
\end{itemize}

\paragraph{Evaluation.} The evaluation of the style-object combination unlearning concerns four quantitative metrics, one for unlearning effectiveness and three for retainability. Before the evaluation, an answer set will be generated exactly following the same procedure introduced in Sec.\,\ref{sec: dataset_proposal} and Fig.\,\ref{fig: mu_pipeline} after unlearning each target. First, the UA (unlearning accuracy) will be evaluated for each answer set, which stands for the ratio of images generated by the target prompt that are neither classified into the target object nor the target style class. A high UA denotes a better ability to successfully unlearn the target combination. Second, the retainability of generation associated with those prompts close to the unlearning target prompt will be evaluated. These prompts can be divided into two groups, the ones sharing the same style but not the object class and the ones sharing the object but not the style class. The classification accuracy of the former corresponds to the retainability of the style, \textit{i.e.,} style consistency (SC), while the latter one denotes the object consistency (OC). These two quantitative metrics evaluate how well the unlearning method precisely define the unlearning scope and retain the generation ability of those close but innocent concept. Thirdly, the retainability of the rest unrelated prompts (UP) are evaluated, which is the last quantitative evaluation metric. The results reported in Tab.\,\ref{tab: unlearn_combination} are averaged over all the unlearning cases shown above.

\subsection{Experiment Details of Sequential Unlearning}
\label{app: exp_setting_continue}

In Sec.\,\ref{sec: mu_experiment_results}, we also evaluated the MU methods with the task of sequential unlearning (SU), where the efficacy of MU methods in handling multiple sequential unlearning requests $\{\mathcal{T}_i\}$ are evaluated. This requires models not only to unlearn new targets effectively but also to maintain the unlearning of previous targets, while retaining all other knowledge. In the experiments, 6 styles are randomly selected as the unlearning targets and excluded from the RA evaluation. The UA of all the already unlearned target will be assessed each time a new request is accomplished. The selected 6 styles include:
\begin{itemize}
    \item Abstractionism
    \item Byzantine 
    \item Cartoon 
    \item Cold Warm
    \item Ukiyoe
    \item Van Gogh
\end{itemize}

After each unlearning request, the unlearning effectiveness and retainability are evaluated. Specifically, the unlearning accuracy of all the unlearning targets in the previous requests are evaluated to evaluate how the unlearning effect lasts when new unlearning requests arrive. In the meantime, the retainability of all the other concepts that are not selected as unlearning targets are evaluated, and to ease the presentation, the retain accuracy of all the concepts (styles and objects) are averaged and reported.

\subsection{Metrics Summary in All Unlearning Settings}

Besides the {\ours} dataset, the various quantitative evaluation metrics proposed in this work are part of the major contributions to a comprehensive and precise evaluation for DM unlearning methods. As there are various unlearning scenarios studied in this work, we provide a summary of these metrics in Tab.\,\ref{tab: metrics_summary}, including their abbreviations, descriptions, and related tables or figures. 

\begin{table}[thb]
    \centering
    \caption{A summary of the quantitative metrics used in this work, including their abbreviations, meanings and where they are used.}
    \resizebox{\textwidth}{!}{%
    \begin{tabular}{c|lc}
    \toprule[1pt]
    \midrule
    \textbf{Metrics} & \textbf{Description} & \textbf{Usages (Table \& Figure)} \\
    \midrule
    \multicolumn{3}{c}{Style/Object Unlearning} \\
    \midrule
    UA & Unlearning accuracy & Fig.\,\ref{fig: problems_illustration}, Tab.\,\ref{tab: mu_comparison} \\
    IRA & In-domain unlearning accuracy & Fig.\,\ref{fig: problems_illustration}, Tab.\,\ref{tab: mu_comparison}  \\
    CRA & Cross-domain unlearning accuracy & Fig.\,\ref{fig: problems_illustration}, Tab.\,\ref{tab: mu_comparison} \\
    \midrule
    \multicolumn{3}{c}{Unlearning Robustness against Adversarial Prompts} \\
    \midrule
    Rob. & Unlearning robustness, unlearning accuracy in the presence of adversarial prompts & Fig.\,\ref{fig: problems_illustration} \\
    \midrule
    \multicolumn{3}{c}{Style-Object Combination Finer-Scale Unlearning} \\
    \midrule
    FU/UA & Unlearning accuracy in finer-scale unlearning & Fig.\,\ref{fig: problems_illustration}, Tab.\,\ref{tab: unlearn_combination}  \\
    SC & Retainability evaluation of style consistency & Tab.\,\ref{tab: unlearn_combination} \\
    OC & Retainability evaluation of object consistency & Tab.\,\ref{tab: unlearn_combination} \\
    UP & Retainability evaluation of unrelated prompts & Tab.\,\ref{tab: unlearn_combination} \\
    FR & Retainability evaluation in finer-scale unlearning, averaged by SC, OC, and UP & Fig.\,\ref{fig: problems_illustration}  \\
    \midrule
    \multicolumn{3}{c}{Sequential or Continual Unlearning} \\
    \midrule
    SU or CU & Unlearning accuracy in the context of sequential unlearning & Fig.\,\ref{fig: problems_illustration}, Tab.\,\ref{tab: continue_unlearning}\\
    SR or CR & Retainability in the context of sequential unlearning & Fig.\,\ref{fig: problems_illustration}, Tab.\,\ref{tab: continue_unlearning} \\
    \midrule
    \bottomrule[1pt]
    \end{tabular}%
    }
    \label{tab: metrics_summary}
\end{table}

\newpage
\clearpage
\section{A Detailed Comparison between {\ours} and \textsc{WikiArt}}
\label{app: ours_wikiart}

\begin{figure}[htb]
\centering
\includegraphics[width=\linewidth]{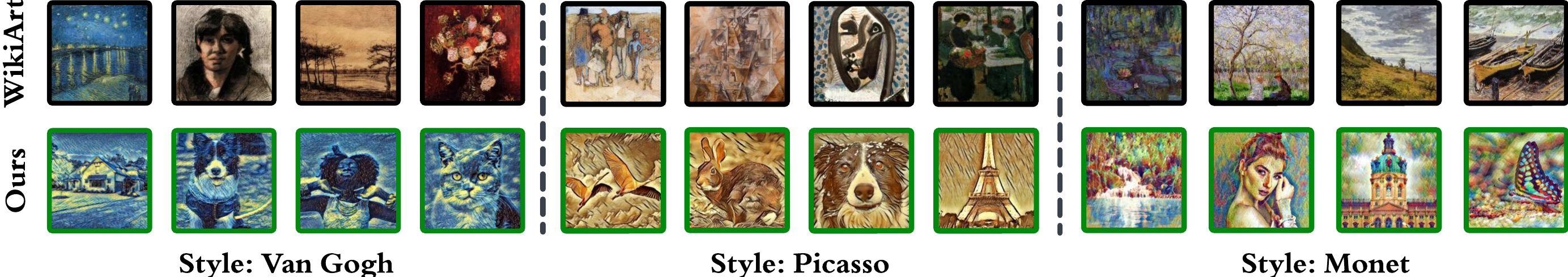}
    \caption{Image examples with the same style label from \textsc{WikiArt} \cite{phillips2011wiki} and {\ours}. Images of the same artistic style in {\ours} exhibit high stylistic consistency compared to \textsc{WikiArt}.
    }
    \label{fig: wikiart_examples}
\end{figure}

\paragraph{{\ours} vs.  \textsc{WikiArt}.}
To the best of our knowledge, \textsc{WikiArt} \cite{nichol2016painter}  is the most relevant baseline dataset to ours. In \textbf{Tab.\,\ref{tab: dataset_comparison}}, we provide a direct comparison of the key attributes of these two datasets. {\ours} differs from \textsc{WikiArt} in the following aspects.

\textbf{First}, {{\ours} includes a greater number of high-resolution images ($15$M) compared to \textsc{WikiArt} ($2$M), a factor that may enhance the training of state-of-the-art DMs.}

\begin{table}[htb]
\centering
\caption{Comparison with \textsc{WikiArt}, the most relevant dataset containing stylized images to ours. {\ours} stands out notably from \textsc{WikiArt} due to its characteristics of being supervised, balanced, and maintaining high stylistic cohesiveness.
}
\label{tab: dataset_comparison}
\resizebox{0.7\linewidth}{!}{%
\begin{tabular}{c|cccc}
\toprule[1pt]
\midrule
\textbf{Dataset} 
& \begin{tabular}[c]{@{}c@{}}\textbf{Resolution}\\ \textbf{(Pixels/Image)}\end{tabular}
& \begin{tabular}[c]{@{}c@{}}\textbf{Style-wise}\\ \textbf{Supervised}\end{tabular}  &  \begin{tabular}[c]{@{}c@{}}\textbf{High Stylistic}\\ \textbf{Consistency}\end{tabular}
& \begin{tabular}[c]{@{}c@{}}\textbf{Class-wise}\\ \textbf{Balanced}\end{tabular} 
\\
\midrule
\textsc{WikiArt} \cite{nichol2016painter} 
& $\sim 2M$  
& {\xmark} 
& {\xmark}
& \begin{tabular}[c]{@{}c@{}}Style-wise {\xmark}\\ Object-wise {\xmark}\end{tabular} 
\\
\midrule
{\ours} 
& $\sim 15M$ 
& {\cmark}
& {\cmark}
& \begin{tabular}[c]{@{}c@{}}Style-wise {\cmark}\\ Object-wise {\cmark}\end{tabular} 
\\
\midrule
\bottomrule[1pt]
\end{tabular}
}
\end{table}

\textbf{Second}, {\ours} surpasses \textsc{WikiArt} in terms of both intra-style coherence and inter-style distinctiveness, as illustrated in \textbf{Fig.\,\ref{fig: wikiart_examples}}, where images labeled with `Van Gogh Style' from both datasets are compared. In {\ours}, the images exhibit high stylistic consistency, while \textsc{WikiArt} lacks the necessary clarity for precise assessment. This will hamper the MU evaluation as discussed in the challenges \textbf{(C2)} and \textbf{(C3)}. This benefits can also be reflected by the training performance using {\ours} and \textsc{WikiArt}. The results are reported in Tab.\,\ref{tab: classifier_preliminary} and Tab.\,\ref{tab: classifier_ours}, respectively. As we can see, the classifier is much more easily trained on {\ours}, justifying the higher discernible features within each style in {\ours}.

\begin{table}[htb]
\centering
\caption{
Art style reproduction quality using SD v1.5 and SD v2.0 finetuned on \textsc{WikiArt}. Images are generated with the prompt ``A painting in {\textit{artist}} style'', where \textit{artist} refers to those included in \textsc{WikiArt}. The test accuracy on DM-generated images and original \textsc{WikiArt} test images is reported using the style classifier finetuned from the pretrained ViT-L/16 on \textsc{WikiArt}.
}
\label{tab: classifier_preliminary}
\resizebox{.5\linewidth}{!}{%
\begin{tabular}{c|cc|c}
\toprule[1pt]
\midrule
 \begin{tabular}[c]{@{}c@{}}Image \\ Source \end{tabular} &
  \begin{tabular}[c]{@{}c@{}}Images by\\ SD v1.5\end{tabular} &
  \begin{tabular}[c]{@{}c@{}}Images by\\ SD v2.0\end{tabular} & 
   \begin{tabular}[c]{@{}c@{}}\textsc{WikiArt}\\ Test Set\end{tabular} \\ \midrule
  Accuracy &
  41.2\% &
  56.7\% &
  85.4\% \\
  \midrule
\bottomrule[1pt]
\end{tabular}%
}
\end{table}

\begin{table}
\caption{Style classification results of a ViT-Large \cite{dosovitskiy2020image} as a style classifier trained on  {\ours}. After convergence, the classifier is tested on the test set and the image set generated by SD v1.5 finetuned on {\ours}.
}
\label{tab: classifier_ours}
\centering
\resizebox{0.8\linewidth}{!}{%
\begin{tabular}{cccc}
\toprule[1pt]
\midrule
 &
  \begin{tabular}[c]{@{}c@{}}{\ours}\\ Train Set\end{tabular} &
  \begin{tabular}[c]{@{}c@{}}{\ours}\\ Test Set\end{tabular} &
  \begin{tabular}[c]{@{}c@{}}Images by SD v1.5 \\ tuned on {\ours}\end{tabular} \\ \midrule
Accuracy &
  $100.0\%$ &
  $99.9\%$ &
  $98.8\%$
\\ \midrule
\bottomrule[1pt]
\end{tabular}
}
\end{table}
\textbf{Third}, the images in {\ours} are style-wise supervised. For each seed image, a stylized counterpart can be find in each style class. This is beneficial for tasks other than unlearning for text-to-image task, such as image editing, image stylization, and style transfer, which can provide a ground truth image for precise and robust evaluation. This will be detailed in Appx.\,\ref{app: broader}.

\section{Additional Experiment Results for  Unlearning Evaluation}
\label{app: additional_exp}

\subsection{ Visualization of the Style and Object Unlearning Performance} 

\begin{figure}[htb]
\centering
\includegraphics[width=.5\linewidth]{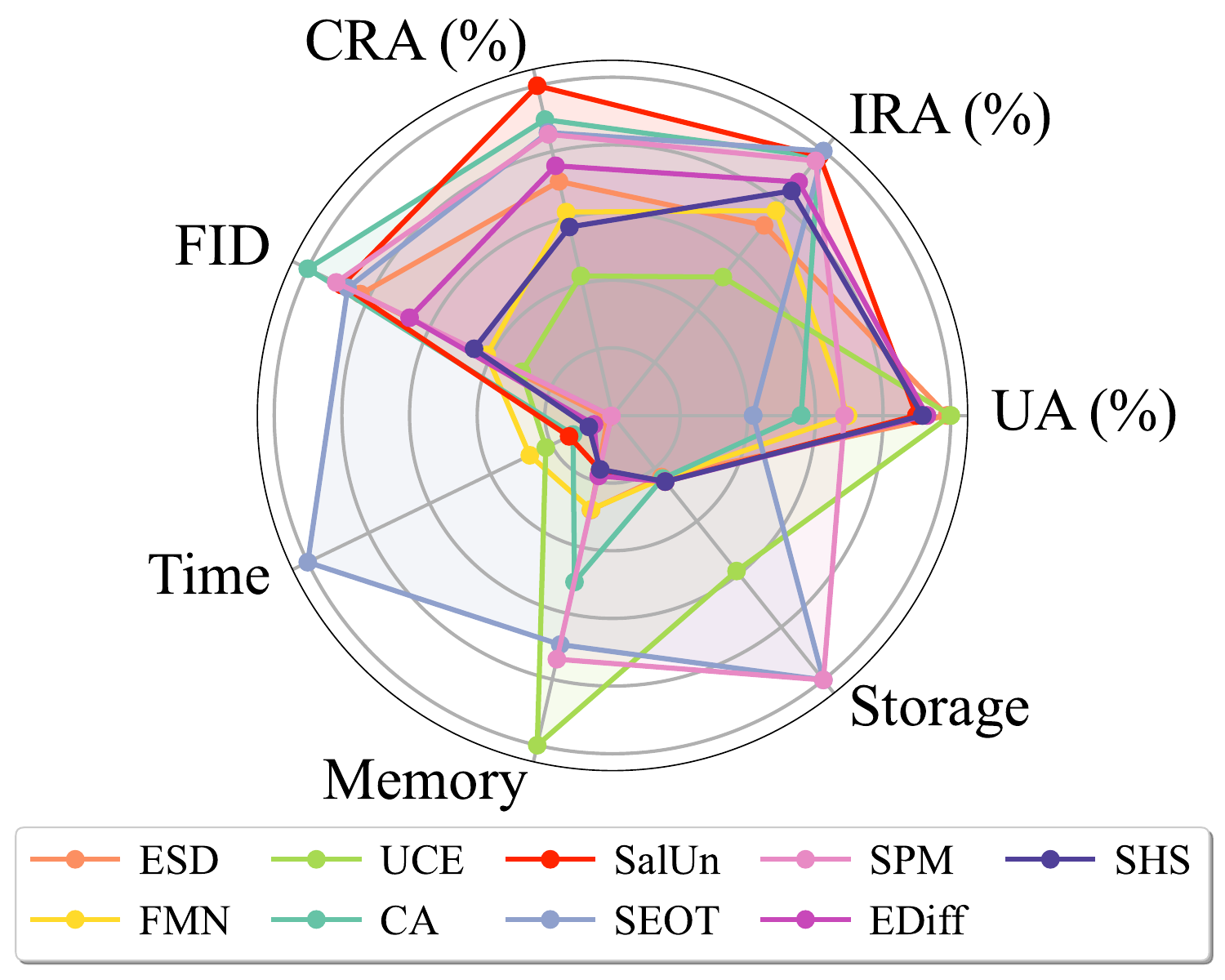}
\caption{Performance visualization for various unlearning methods as summarized in Table \ref{tab: unlearn_overview}. For UA, IRA, and CRA, the results are averaged over the style and object unlearning scenarios. Other metrics undertake the inverse operation as a smaller values represent better performance. Results are normalized to $0\%\sim100\%$ per metric.}
\label{fig: unlearn_radar}
\end{figure}

To make a more direct comparison among different MU methods reported in Tab.\,\ref{tab: mu_comparison}, the results are visualized in the radar chart Fig.\,\ref{fig: unlearn_radar}. This figure illustrates that no method dominates across all assessment dimensions. This underscores the complexity of unlearning in generative models and the need for further improvement.

\subsection{A Fine-Grained Comparison per Unlearning Target: ESD \textit{vs.} \textsc{Salun}} 

Following the analysis of ESD in Fig.\,\ref{fig: unlearning_dissection}, we next turn our focus to a comparative analysis with \textsc{Salun}, a method that demonstrated a better balance between unlearning and retaining according to Tab.\,\ref{tab: unlearn_overview}. A similar accuracy heatmap for \textsc{Salun} is presented in Fig.\,\ref{fig: salun_heatmap}. Compared to ESD, \textsc{Salun} exhibits more consistent performance across various unlearning scenarios, as indicated by the more uniform color distribution in the heatmap. This also suggests enhanced retainability. However, it is noticeable that \textsc{Salun} does not reach the same level of UA (Unlearning Accuracy) as ESD, as evidenced by the darker diagonal values in Fig.\,\ref{fig: salun_heatmap}. This observation reinforces the existence of a trade-off between unlearning effectiveness and retainability in the visual generative MU task, a phenomenon paralleled in other tasks such as classification.

\subsection{Unlearning Heatmaps of More Unlearning Methods} 
\begin{figure}[htb]
\centering
\includegraphics[width=.5\linewidth]{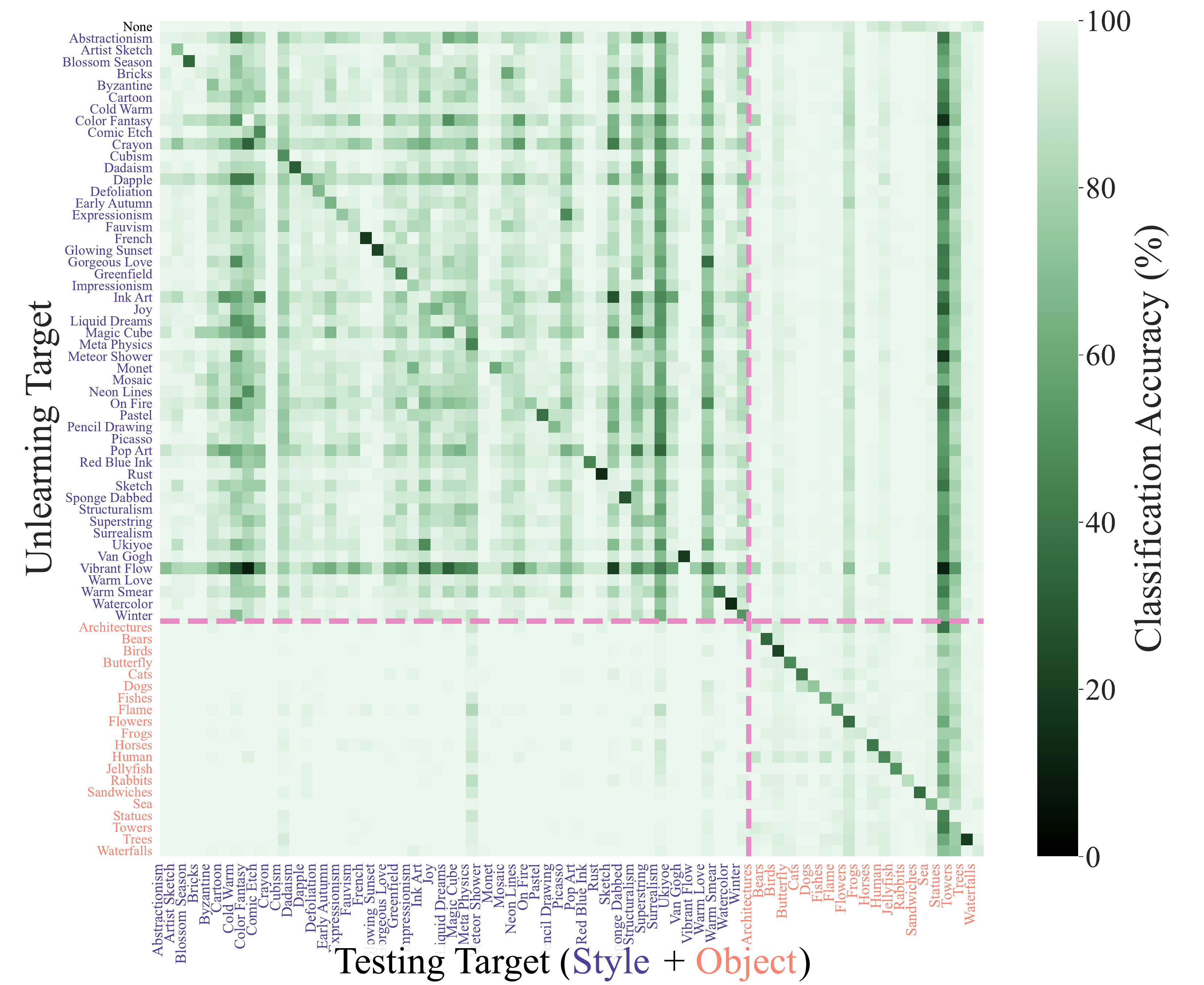}
    \caption{Heatmap visualization of SalUn. The plot setting is identical to Figure \ref{fig: unlearning_dissection}. }
    \label{fig: salun_heatmap}
\end{figure}

\begin{figure}[htb]
\centering
\includegraphics[width=.5\linewidth]{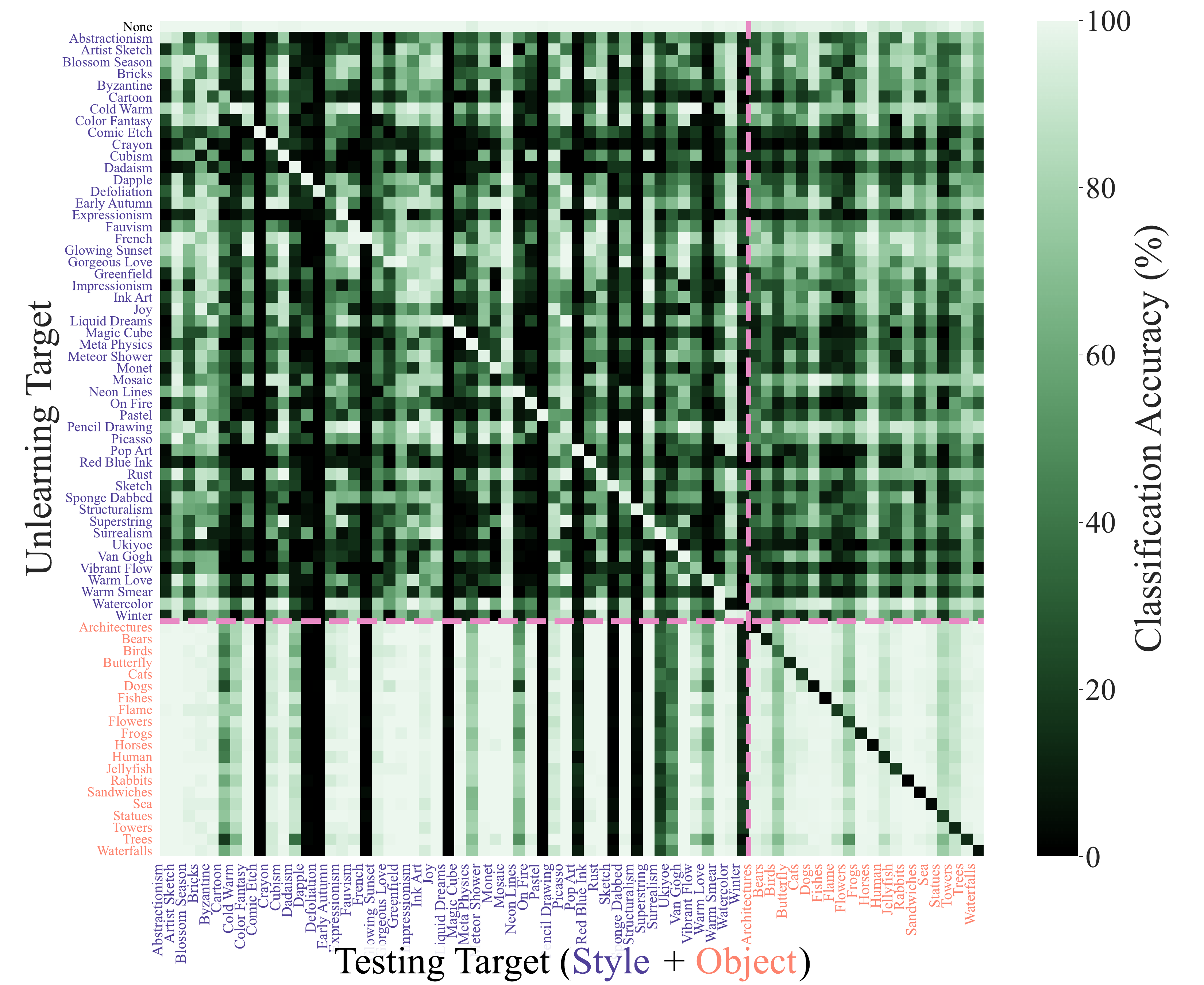}
    \caption{Heatmap visualization of FMN. The plot setting is identical to Figure \ref{fig: unlearning_dissection}. }
    \label{fig: fmn_heatmap}
\end{figure}

\begin{figure}[htb]
\centering
\includegraphics[width=.5\linewidth]{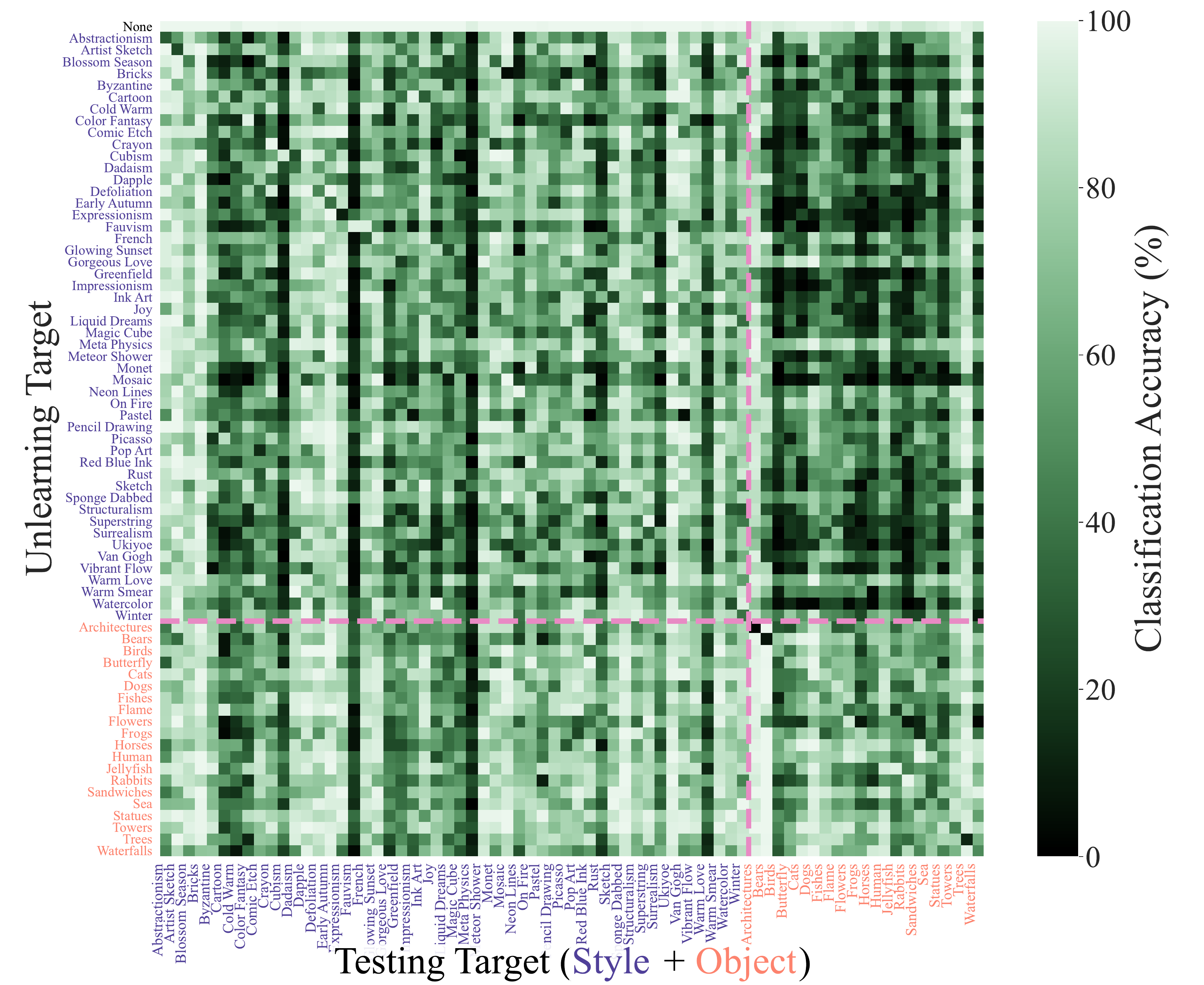}
    \caption{Heatmap visualization of SHS. The plot setting is identical to Figure \ref{fig: unlearning_dissection}. }
    \label{fig: shs_heatmap}
\end{figure}

In Fig.\,\ref{fig: fmn_heatmap}$\sim$Fig.\,\ref{fig: shs_heatmap}, we provide more unlearning heatmap visualizations in the same format as Fig.\,\ref{fig: unlearning_dissection} and Fig.\,\ref{fig: salun_heatmap} in order to provide a more detailed unlearning performance dissection for all the DM unlearning methods studied in this work. 

\begin{figure}[htb]
\centering
\includegraphics[width=.5\linewidth]{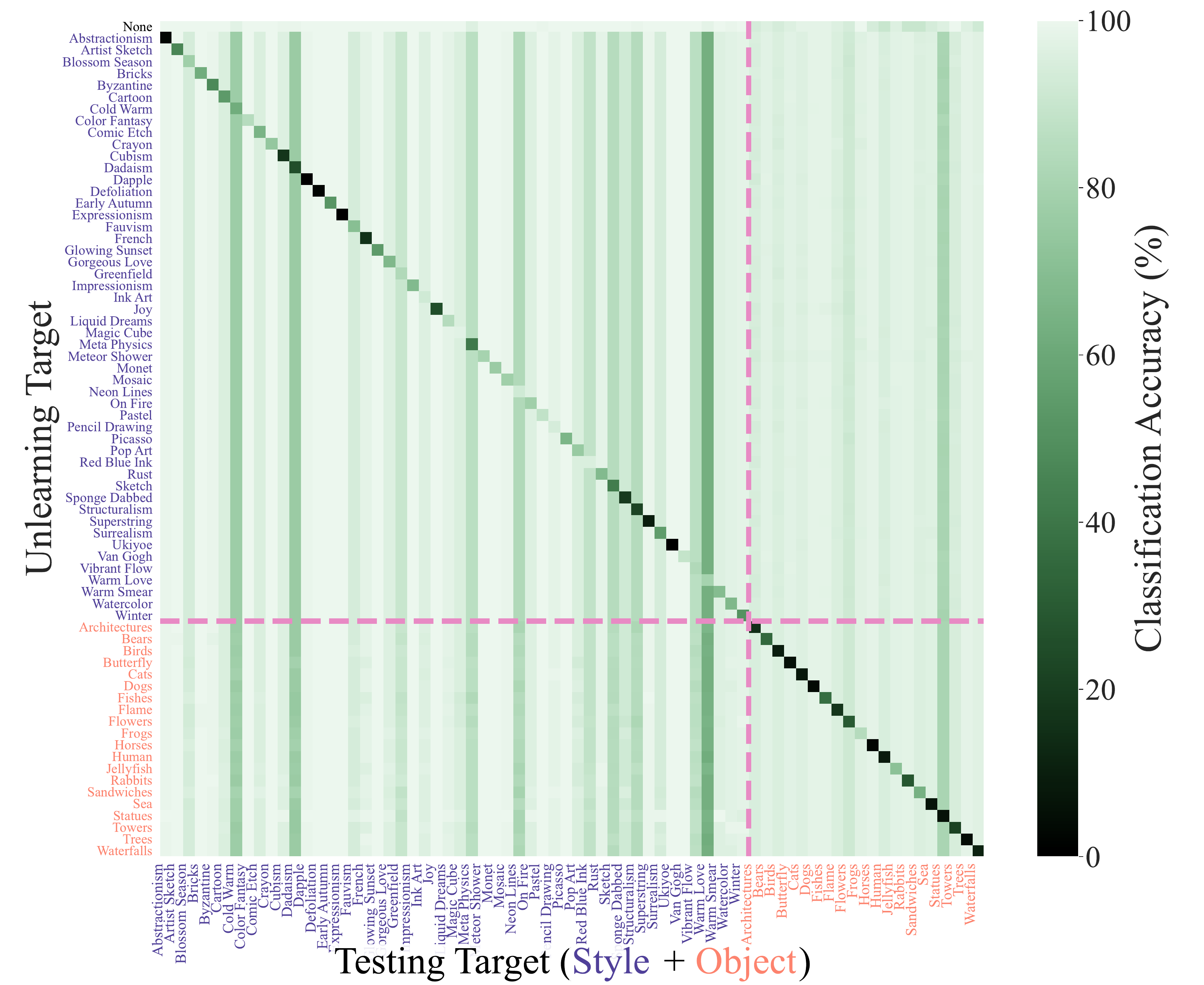}
    \caption{Heatmap visualization of SEOT. The plot setting is identical to Figure \ref{fig: unlearning_dissection}. }
    \label{fig: seot_heatmap}
\end{figure}

\begin{figure}[htb]
\centering
\includegraphics[width=.5\linewidth]{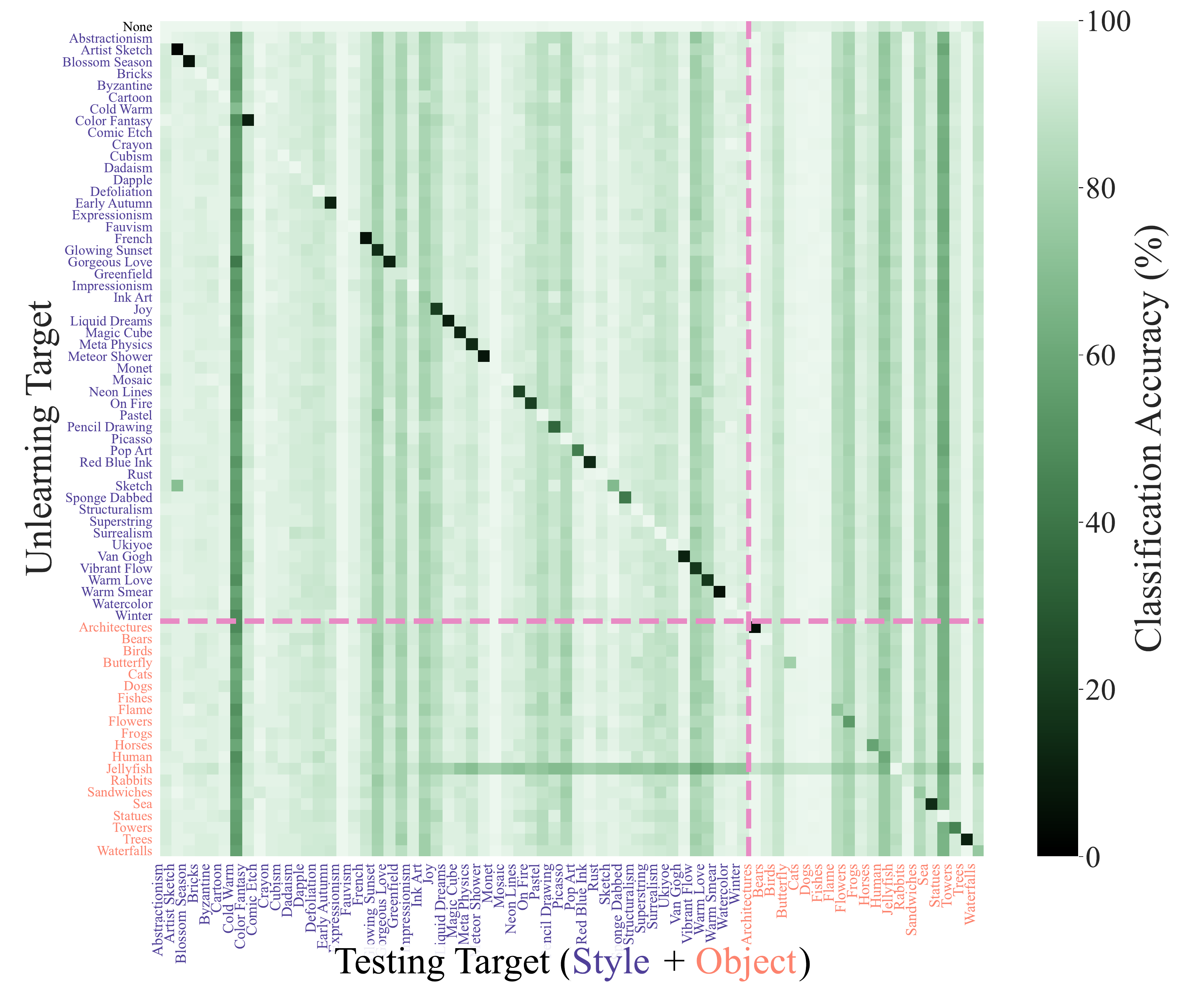}
    \caption{Heatmap visualization of SPM. The plot setting is identical to Figure \ref{fig: unlearning_dissection}. }
    \label{fig: spm_heatmap}
\end{figure}

\begin{figure}[htb]
\centering
\includegraphics[width=.5\linewidth]{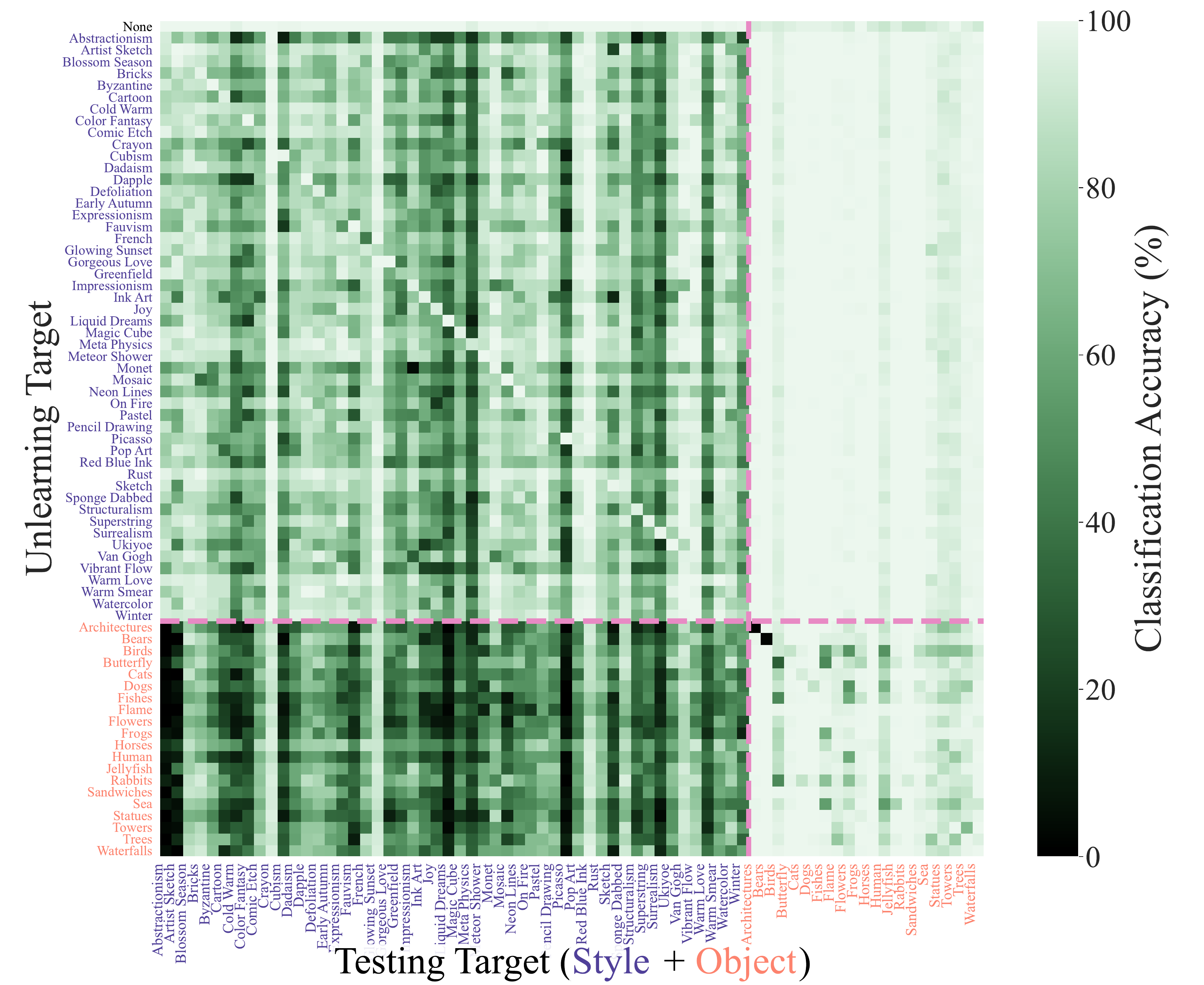}
    \caption{Heatmap visualization of Ediff. The plot setting is identical to Figure \ref{fig: unlearning_dissection}. }
    \label{fig: ediff_heatmap}
\end{figure}

\clearpage
\newpage

\section{Visualizations}
\label{app: visualization}

In this section, we aim to provide plenty of visualizations, illustrations, and qualitative results of the dataset and the quantitative results shown in Sec.\,\ref{sec: mu_experiment_results} and Appx.\,\ref{app: additional_exp}. These visualizations are intended to deepen the understanding of the effects of different MU methods and clearly illustrate the challenges identified in previous sections. We hope these visual aids will enable readers to more effectively grasp the nuances of MU methods and their implications for DMs (Diffusion Models).

\begin{table}
\centering
\caption{Performance comparison of different DM unlearning methods in the sequential unlearning setting. Each column represents a new unlearning request, denoted by $\mathcal{T}_i$, where $\mathcal{T}_1$ is the oldest. Each row represents the UA for a specific unlearning objective or the retaining accuracy (RA), given by the average of IRA and CRA. Results indicating \warnorange{\textit{unlearning rebound}} effect are highlighted in \warnorange{orange}, and those signifying \warn{\textit{catastrophic retaining failure}} are marked in \warn{red}.}

\label{tab: continue_unlearning}
\resizebox{\linewidth}{!}{%
\begin{tabular}{lc|cccccc | lc|cccccc}
\toprule[1pt]
\midrule
\multicolumn{8}{c|}{\textbf{Method: ESD}} & \multicolumn{8}{c}{\textbf{Method: FMN}} \\
\midrule
\multicolumn{2}{l|}{\multirow{2}{*}{\textbf{Metrics}}} 
& $\mathcal{T}_1$ & $\mathcal{T}_1\sim\mathcal{T}_2$ & $\mathcal{T}_1\sim\mathcal{T}_3$ & $\mathcal{T}_1\sim\mathcal{T}_4$ & $\mathcal{T}_1\sim\mathcal{T}_5$ & $\mathcal{T}_1\sim\mathcal{T}_6$ 

& \multicolumn{2}{l|}{\multirow{2}{*}{\textbf{Metrics}}} 
& $\mathcal{T}_1$ & $\mathcal{T}_1\sim\mathcal{T}_2$ & $\mathcal{T}_1\sim\mathcal{T}_3$ & $\mathcal{T}_1\sim\mathcal{T}_4$ & $\mathcal{T}_1\sim\mathcal{T}_5$ & $\mathcal{T}_1\sim\mathcal{T}_6$ 
\\ 
& & \multicolumn{6}{c|}{-------------------------- \textbf{Unlearning Request} --------------------------}
& & & \multicolumn{6}{c}{-------------------------- \textbf{Unlearning Request} --------------------------}
\\
\midrule

\multirow{6}{*}{\textbf{UA}} & $\mathcal{T}_1$ & \cellcolor{WarnOrange}$100\%$ & \cellcolor{WarnOrange}$99\%$ & \cellcolor{WarnOrange}$95\%$ & \cellcolor{WarnOrange}$87\%$ & \cellcolor{WarnOrange}$81\%$ & \cellcolor{WarnOrange}$75\%$ & 

\multirow{6}{*}{\textbf{UA}} & $\mathcal{T}_1$ & $88\%$ & $99\%$ & $99\%$ & $98\%$ & $99\%$ & $99\%$ 
\\

& $\mathcal{T}_2$ & - & $100\%$ & \cellcolor{WarnOrange}$100\%$ & \cellcolor{WarnOrange}$96\%$ & \cellcolor{WarnOrange}$87\%$ & \cellcolor{WarnOrange}$79\%$ & 

& $\mathcal{T}_2$ & - & $95\%$ & $99\%$ & $99\%$ & $98\%$ & $99\%$ \\

& $\mathcal{T}_3$ & - & - & $100\%$ & $98\%$ & $99\%$ & $98\%$ 

& & $\mathcal{T}_3$ & - & - & $97\%$ & $98\%$ & $99\%$ & $99\%$ 
\\

& $\mathcal{T}_4$ & - & - & - & $100\%$ & $100\%$ & $99\%$ 

& & $\mathcal{T}_4$ & - & - & - & $99\%$ & $99\%$ & $99\%$ 
\\

& $\mathcal{T}_5$ & - & - & - & - & $100\%$ & $99\%$ 

& & $\mathcal{T}_5$ & - & - & - & - & $99\%$ & $99\%$ 
\\

& $\mathcal{T}_6$ & - & - & - & - & - & $100\%$ 
& & $\mathcal{T}_6$ & - & - & - & - & - & $100\%$ 
\\
\midrule

\multicolumn{2}{l|}{\textbf{RA}} &$77.46\%$ & $52.94\%$ & $35.99\%$ & $24.86\%$ & $18.69\%$ & $12.95\%$ 
& \multicolumn{2}{l|}{\textbf{RA}} &\warn{$82.39\%$} & \warn{$14.56\%$} & $13.34\%$ & $10.42\%$ & $9.83\%$ & $8.76\%$ 
\\
\midrule

\multicolumn{8}{c|}{\textbf{Method: UCE}} & \multicolumn{8}{c}{\textbf{Method: CA}} \\
\midrule

\multicolumn{2}{c|}{\multirow{2}{*}{\textbf{Metrics}}} 
& $\mathcal{T}_1$ & $\mathcal{T}_1\sim\mathcal{T}_2$ & $\mathcal{T}_1\sim\mathcal{T}_3$ & $\mathcal{T}_1\sim\mathcal{T}_4$ & $\mathcal{T}_1\sim\mathcal{T}_5$ & $\mathcal{T}_1\sim\mathcal{T}_6$ 

& \multicolumn{2}{c|}{\multirow{2}{*}{\textbf{Metrics}}} 
& $\mathcal{T}_1$ & $\mathcal{T}_1\sim\mathcal{T}_2$ & $\mathcal{T}_1\sim\mathcal{T}_3$ & $\mathcal{T}_1\sim\mathcal{T}_4$ & $\mathcal{T}_1\sim\mathcal{T}_5$ & $\mathcal{T}_1\sim\mathcal{T}_6$ 
\\ 

& & \multicolumn{6}{c}{-------------------------- \textbf{Unlearning Request} --------------------------} 

& & & \multicolumn{6}{c}{-------------------------- \textbf{Unlearning Request} --------------------------} 
\\
\midrule

\multirow{6}{*}{\textbf{UA}} 
& $\mathcal{T}_1$ & $93\%$ & $95\%$ & $98\%$ & $96\%$ & $97\%$ & $98\%$ 

& \multirow{6}{*}{\textbf{UA}} 
& $\mathcal{T}_1$ & \cellcolor{WarnOrange}$58\%$ & \cellcolor{WarnOrange}$55\%$ & \cellcolor{WarnOrange}$59\%$ & \cellcolor{WarnOrange}$45\%$ & \cellcolor{WarnOrange}$44\%$ & \cellcolor{WarnOrange}$40\%$ 

\\

& $\mathcal{T}_2$ & - & $97\%$ & $98\%$ & $98\%$ & $98\%$ & $95\%$ 

& & $\mathcal{T}_2$ & - & \cellcolor{WarnOrange}$76\%$ & \cellcolor{WarnOrange}$58\%$ & \cellcolor{WarnOrange}$51\%$ & \cellcolor{WarnOrange}$47\%$ & \cellcolor{WarnOrange}$44\%$ 
\\

& $\mathcal{T}_3$ & - & - & $95\%$ & $97\%$ & $98\%$ & $99\%$ 

& & $\mathcal{T}_3$ & - & - & \cellcolor{WarnOrange}$45\%$ & \cellcolor{WarnOrange}$41\%$ & \cellcolor{WarnOrange}$40\%$ & \cellcolor{WarnOrange}$37\%$ 
\\

& $\mathcal{T}_4$ & - & - & - & $98\%$ & $98\%$ & $98\%$ 

& & $\mathcal{T}_4$ & - & - & - & \cellcolor{WarnOrange}$71\%$ & \cellcolor{WarnOrange}$70\%$ & \cellcolor{WarnOrange}$60\%$ 
\\

& $\mathcal{T}_5$ & - & - & - & - & $97\%$ & $99\%$ 

& & $\mathcal{T}_5$ & - & - & - & - & \cellcolor{WarnOrange}$69\%$ & \cellcolor{WarnOrange}$51\%$ 
\\

& $\mathcal{T}_6$ & - & - & - & - & - & $99\%$ 

& & $\mathcal{T}_6$ & - & - & - & - & - & $57\%$ 
\\

\midrule

\multicolumn{2}{l|}{\textbf{RA}} & \warn{$81.42\%$} & \warn{$29.38\%$} & $18.72\%$ & $15.34\%$ & $13.32\%$ & $11.31\%$

& \multicolumn{2}{l|}{\textbf{RA}} &$97.24\%$ & $93.39\%$ & $84.46\%$ & $79.32\%$ & $71.40\%$ & $60.53\%$
\\
\midrule

\multicolumn{8}{c}{\textbf{Method: SalUn}} 
& \multicolumn{8}{c}{\textbf{Method: SPM}}
\\
\midrule

\multicolumn{2}{c|}{\multirow{2}{*}{\textbf{Metrics}}} 
& $\mathcal{T}_1$ & $\mathcal{T}_1\sim\mathcal{T}_2$ & $\mathcal{T}_1\sim\mathcal{T}_3$ & $\mathcal{T}_1\sim\mathcal{T}_4$ & $\mathcal{T}_1\sim\mathcal{T}_5$ & $\mathcal{T}_1\sim\mathcal{T}_6$ 

& \multicolumn{2}{c|}{\multirow{2}{*}{\textbf{Metrics}}} 
& $\mathcal{T}_1$ & $\mathcal{T}_1\sim\mathcal{T}_2$ & $\mathcal{T}_1\sim\mathcal{T}_3$ & $\mathcal{T}_1\sim\mathcal{T}_4$ & $\mathcal{T}_1\sim\mathcal{T}_5$ & $\mathcal{T}_1\sim\mathcal{T}_6$ 
\\

& & \multicolumn{6}{c}{-------------------------- \textbf{Unlearning Request} --------------------------} 

& & & \multicolumn{6}{c}{-------------------------- \textbf{Unlearning Request} --------------------------} 
\\
\midrule

\multirow{6}{*}{\textbf{UA}}
& $\mathcal{T}_1$ & \cellcolor{WarnOrange}$84\%$ & \cellcolor{WarnOrange}$79\%$ & \cellcolor{WarnOrange}$78\%$ & \cellcolor{WarnOrange}$65\%$ & \cellcolor{WarnOrange}$67\%$ & \cellcolor{WarnOrange}$64\%$ 

& \multirow{6}{*}{\textbf{UA}}
& $\mathcal{T}_1$ & \cellcolor{WarnOrange}$55\%$ & \cellcolor{WarnOrange}$59\%$ & \cellcolor{WarnOrange}$50\%$ & \cellcolor{WarnOrange}$49\%$ & \cellcolor{WarnOrange}$47\%$ & \cellcolor{WarnOrange}$48\%$ 
\\

& $\mathcal{T}_2$ & - & \cellcolor{WarnOrange}$81.42\%$ & \cellcolor{WarnOrange}$75\%$ & \cellcolor{WarnOrange}$72\%$ & \cellcolor{WarnOrange}$69\%$ & \cellcolor{WarnOrange}$61\%$ 

& & $\mathcal{T}_2$ & - & $62\%$ & $59\%$ & $58\%$ & $60\%$ & $63\%$ 
\\

& $\mathcal{T}_3$ & - & - & $90\%$ & $85\%$ & $84\%$ & $87\%$ 

& & $\mathcal{T}_3$ & - & - & $42\%$ & $39\%$ & $40\%$ & $41\%$ 
\\

& $\mathcal{T}_4$ & - & - & - & $84\%$ & $86\%$ & $81\%$ 

& & $\mathcal{T}_4$ & - & - & - & $57\%$ & $59\%$ & $60\%$ 
\\

& $\mathcal{T}_5$ & - & - & - & - & $79\%$ & $81\%$ 

& & $\mathcal{T}_5$ & - & - & - & - & $51\%$ & $51\%$ 
\\

& $\mathcal{T}_6$ & - & - & - & - & - & $89\%$ 

& & $\mathcal{T}_6$ & - & - & - & - & - & $43\%$ 
\\
\midrule

\multicolumn{2}{c|}{\textbf{RA}} & $85.43\%$ & $80.32\%$ & $71.42\%$ & $65.41\%$ & $63.24\%$ & $60.19\%$ 

& \multicolumn{2}{c|}{\textbf{RA}} & $72.39\%$ & $70.42\%$ & $67.89\%$ & $60.45\%$ & $55.32\%$ & $51.12\%$ 
\\
\midrule

\multicolumn{8}{c}{\textbf{Method: EDiff}} 

& \multicolumn{8}{c}{\textbf{Method: SHS}} 
\\
\midrule

\multicolumn{2}{c|}{\multirow{2}{*}{\textbf{Metrics}}} 
& $\mathcal{T}_1$ & $\mathcal{T}_1\sim\mathcal{T}_2$ & $\mathcal{T}_1\sim\mathcal{T}_3$ & $\mathcal{T}_1\sim\mathcal{T}_4$ & $\mathcal{T}_1\sim\mathcal{T}_5$ & $\mathcal{T}_1\sim\mathcal{T}_6$ 

& \multicolumn{2}{c|}{\multirow{2}{*}{\textbf{Metrics}}} 
& $\mathcal{T}_1$ & $\mathcal{T}_1\sim\mathcal{T}_2$ & $\mathcal{T}_1\sim\mathcal{T}_3$ & $\mathcal{T}_1\sim\mathcal{T}_4$ & $\mathcal{T}_1\sim\mathcal{T}_5$ & $\mathcal{T}_1\sim\mathcal{T}_6$ 
\\ 

& & \multicolumn{6}{c}{-------------------------- \textbf{Unlearning Request} --------------------------} 

& & & \multicolumn{6}{c}{-------------------------- \textbf{Unlearning Request} --------------------------} 
\\
\midrule

\multirow{6}{*}{\textbf{UA}} & $\mathcal{T}_1$ & {\warncellorange}$97\%$ & {\warncellorange}$93\%$ & {\warncellorange}$91\%$ & {\warncellorange}$93\%$ & {\warncellorange}$85\%$ & {\warncellorange}$90\%$ 

& \multirow{6}{*}{\textbf{UA}} & $\mathcal{T}_1$ & $81\%$ & $73\%$ & $74\%$ & $93\%$ & $94\%$ & $97\%$ 
\\

& $\mathcal{T}_2$ & - & {\warncellorange}$92\%$ & {\warncellorange}$89\%$ & {\warncellorange}$93\%$ & {\warncellorange}$91\%$ & {\warncellorange}$87\%$ 

& & $\mathcal{T}_2$ & - & $69\%$ & $61\%$ & $89\%$ & $94\%$ & $97\%$
\\

& $\mathcal{T}_3$ & - & - & {\warncellorange}$96\%$ & {\warncellorange}$93\%$ & {\warncellorange}$90\%$ & {\warncellorange}$84\%$ 

& & $\mathcal{T}_3$ & - & - & $75\%$ & $92\%$ & $96\%$ & $90\%$
\\

& $\mathcal{T}_4$ & - & - & - & $91\%$ & $92\%$ & $90.22\%$ 

& & $\mathcal{T}_4$ & - & - & - & $91\%$ & $95\%$ & $97\%$
\\

& $\mathcal{T}_5$ & - & - & - & - & $99\%$ & $97\%$ 

& & $\mathcal{T}_5$ & - & - & - & - & $92\%$ & $96\%$
\\

& $\mathcal{T}_6$ & - & - & - & - & - & $94\%$ 

& & $\mathcal{T}_6$ & - & - & - & - & - & $94\%$
\\
\midrule

\multicolumn{2}{l|}{\textbf{RA}} & $92.34\%$ & \warn{$89.37\%$} & \warn{$14.35\%$} & $12.31\%$ & $12.82\%$ & $7.42\%$ 

& \multicolumn{2}{l|}{\textbf{RA}} & $88.41\%$ & $84.32\%$ & $73.98\%$ & \warn{$69.19\%$} & \warn{$10.76\%$} & $10.11\%$ 
\\
\midrule
\bottomrule[1pt]
\end{tabular}%
}
\end{table}

\subsection{Illustrations of the Styles and Objects in {\ours}}

\begin{figure}[h]
    \centering
    \begin{tabular}{cccccccccc}
         \hspace*{-4.1mm} \includegraphics[width=.08\linewidth,height=!]{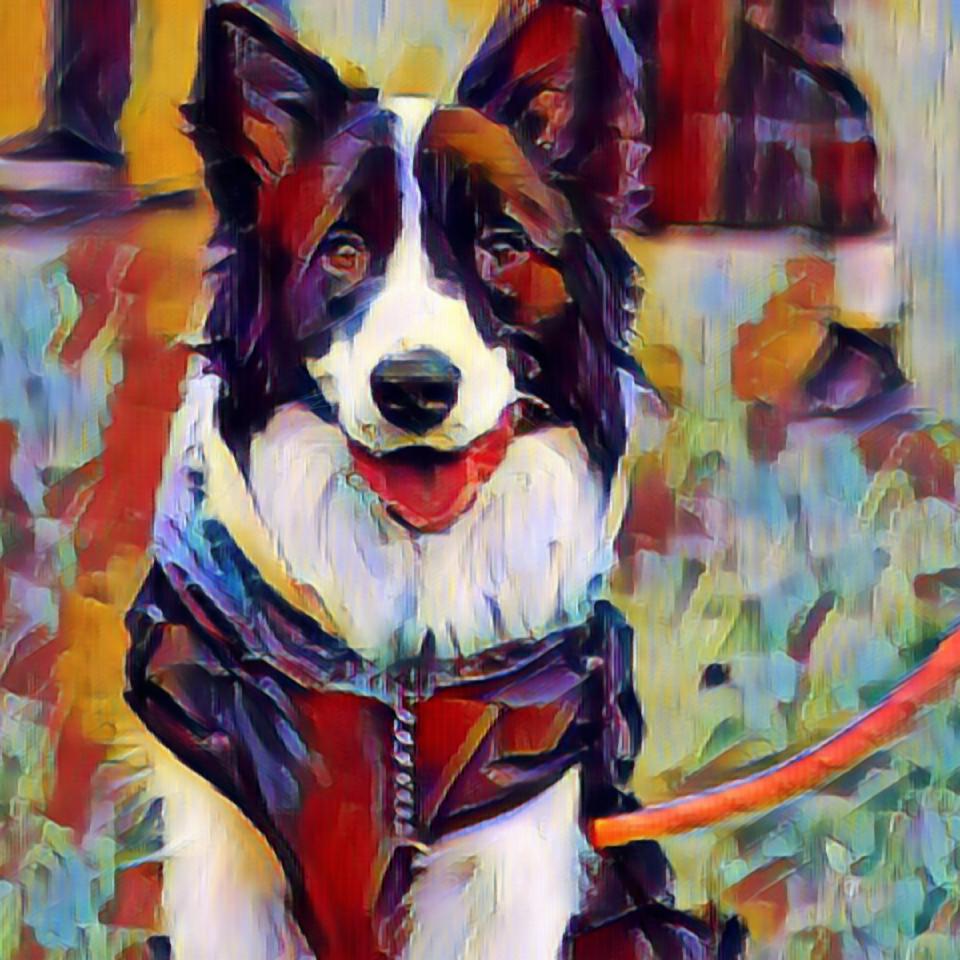} 
         & \hspace*{-4mm} \includegraphics[width=.08\linewidth,height=!]{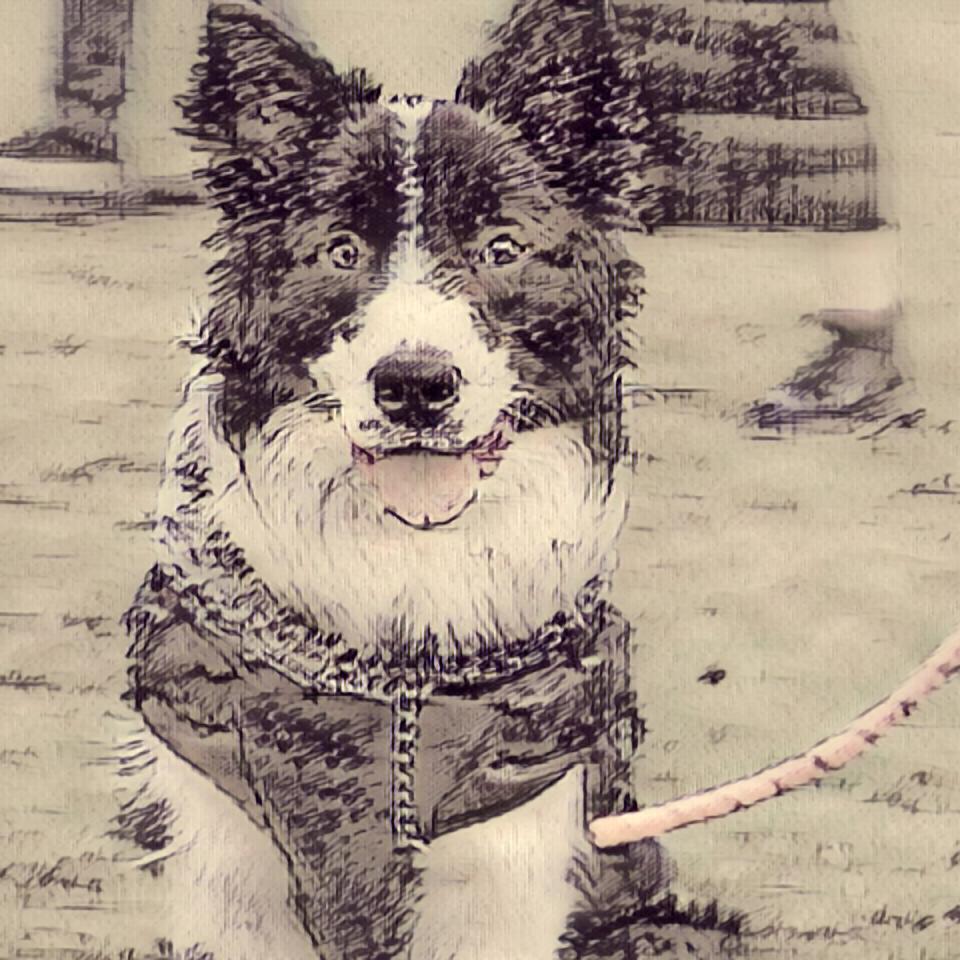}
         & \hspace*{-4mm} \includegraphics[width=.08\linewidth,height=!]{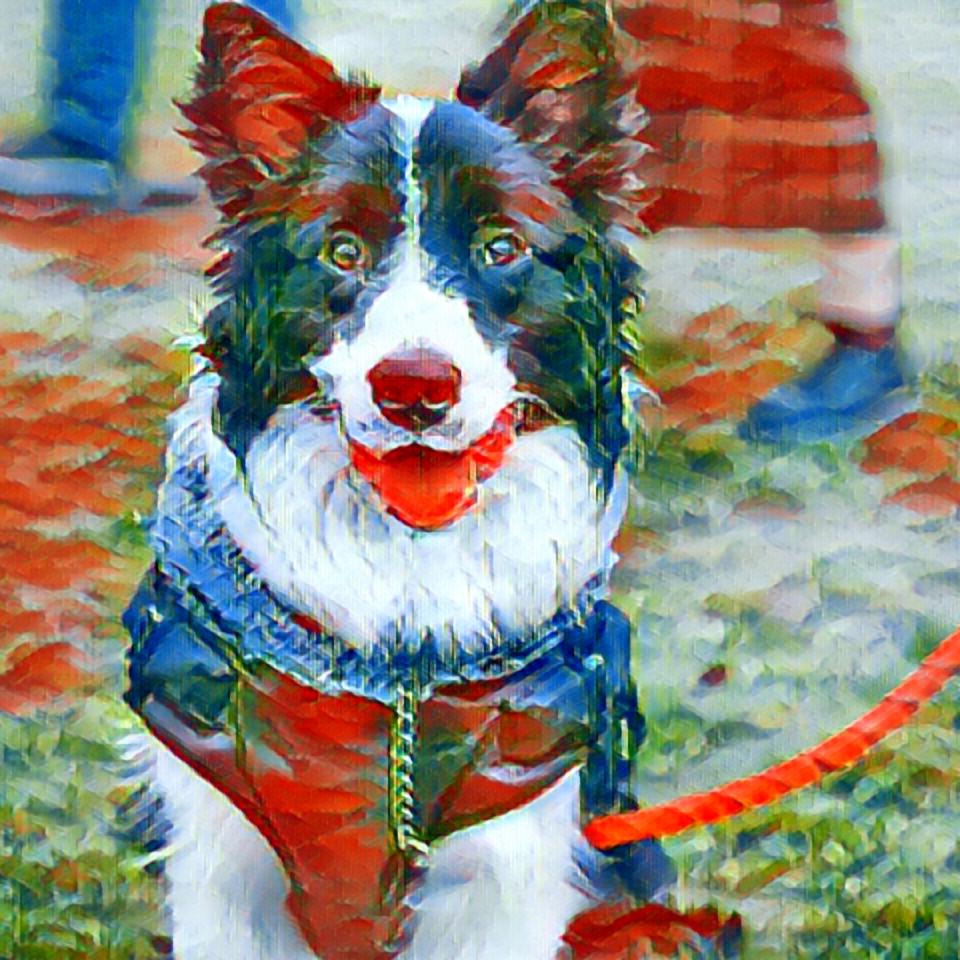}
         & \hspace*{-4mm} \includegraphics[width=.08\linewidth,height=!]{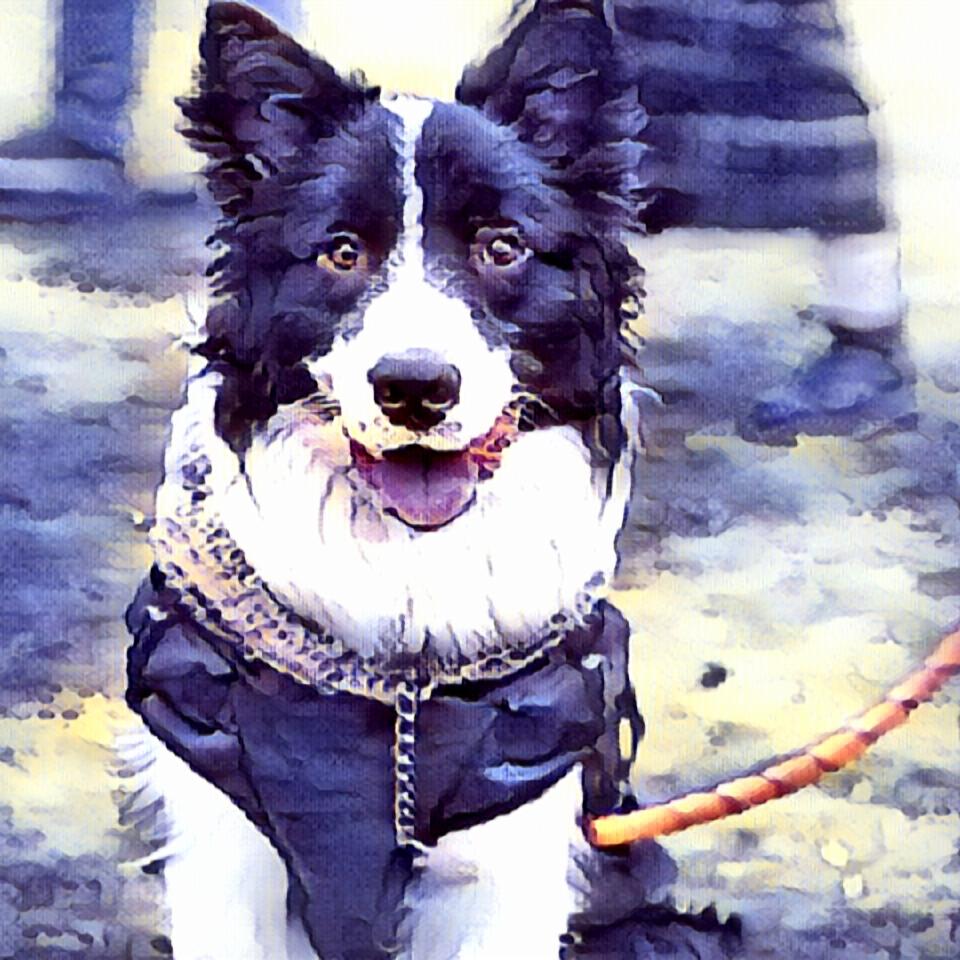}
         & \hspace*{-4mm} \includegraphics[width=.08\linewidth,height=!]{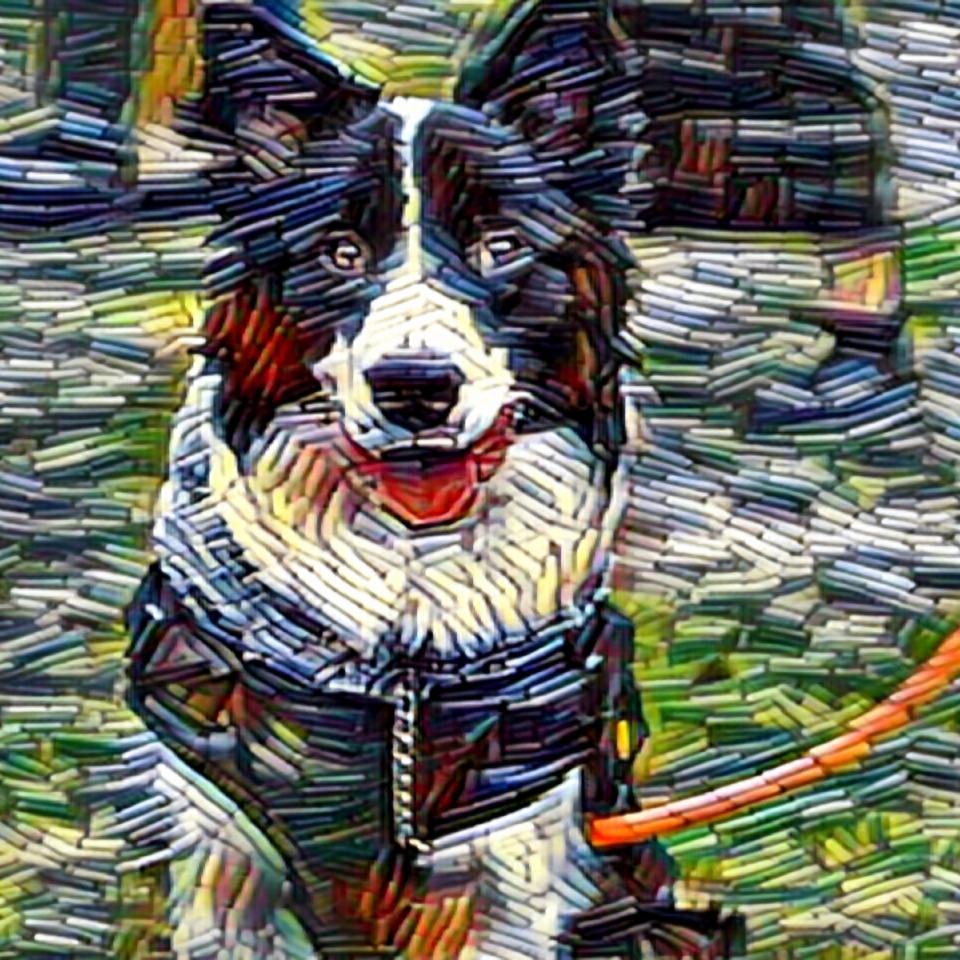}
         & \hspace*{-4mm} \includegraphics[width=.08\linewidth,height=!]{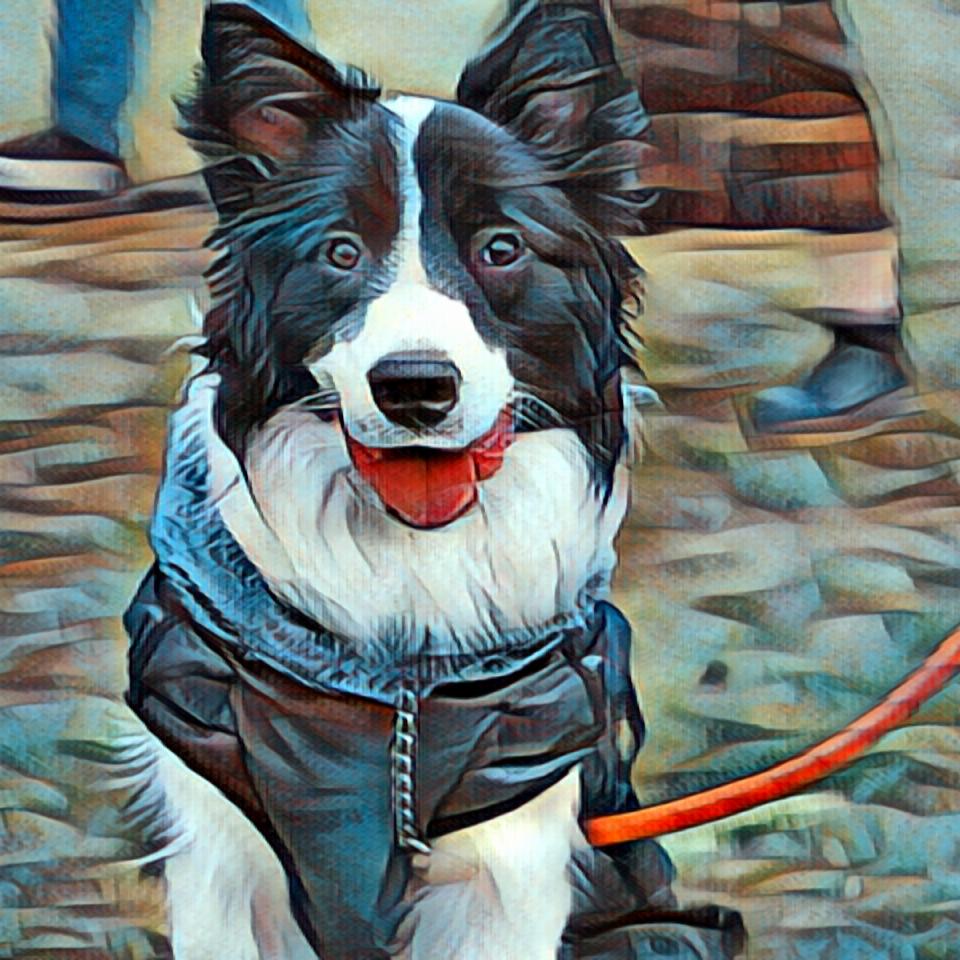} 
         & \hspace*{-4mm} \includegraphics[width=.08\linewidth,height=!]{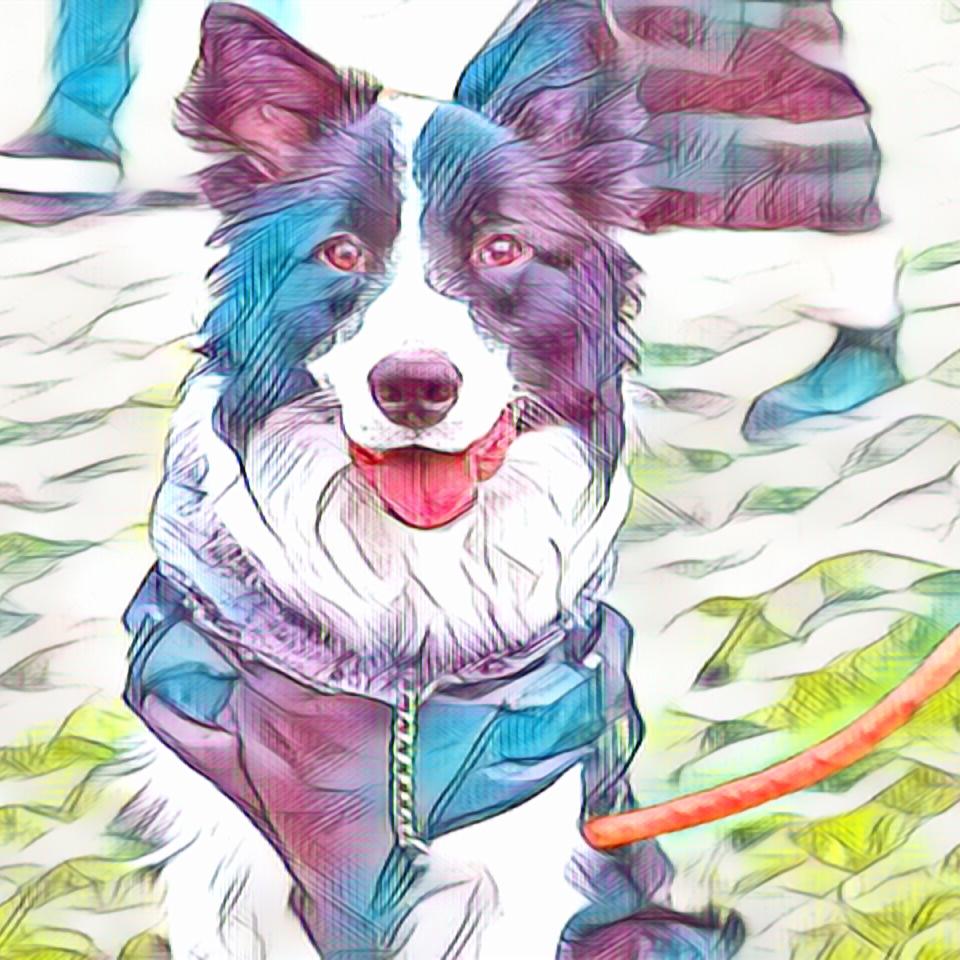} 
        & \hspace*{-4mm} \includegraphics[width=.08\linewidth,height=!]{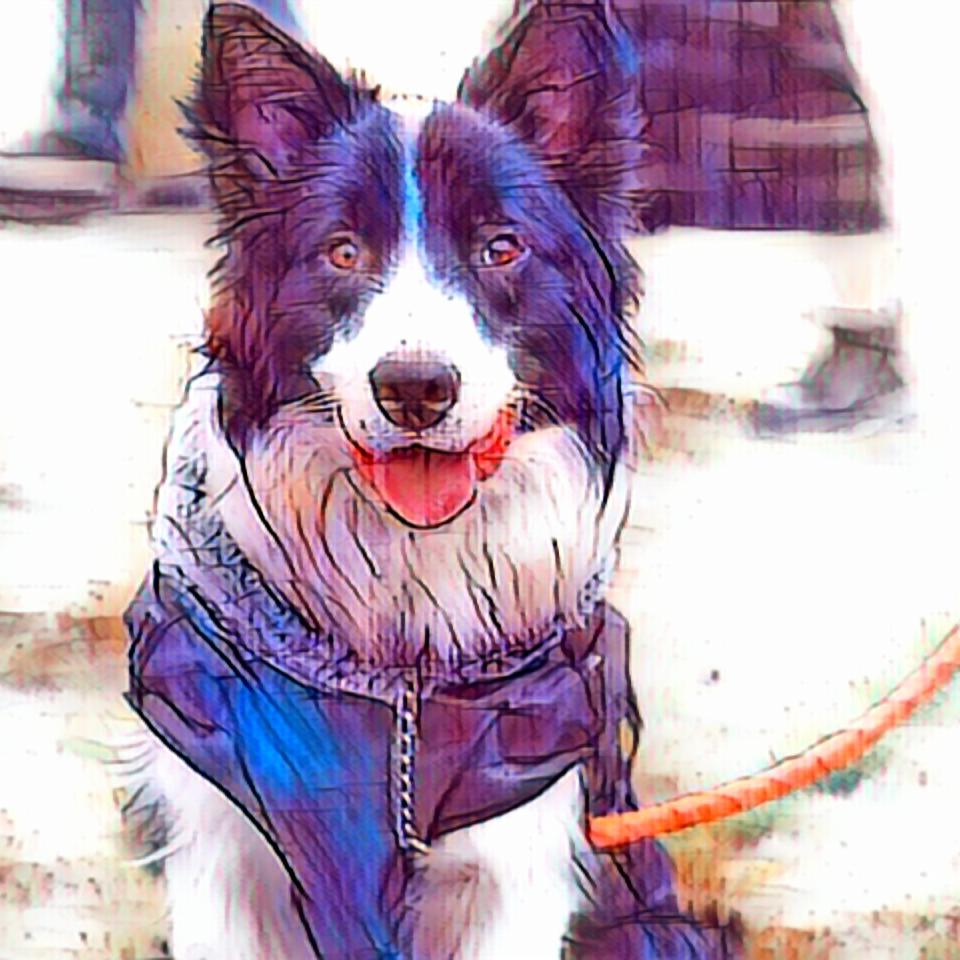} 
        & \hspace*{-4mm} \includegraphics[width=.08\linewidth,height=!]{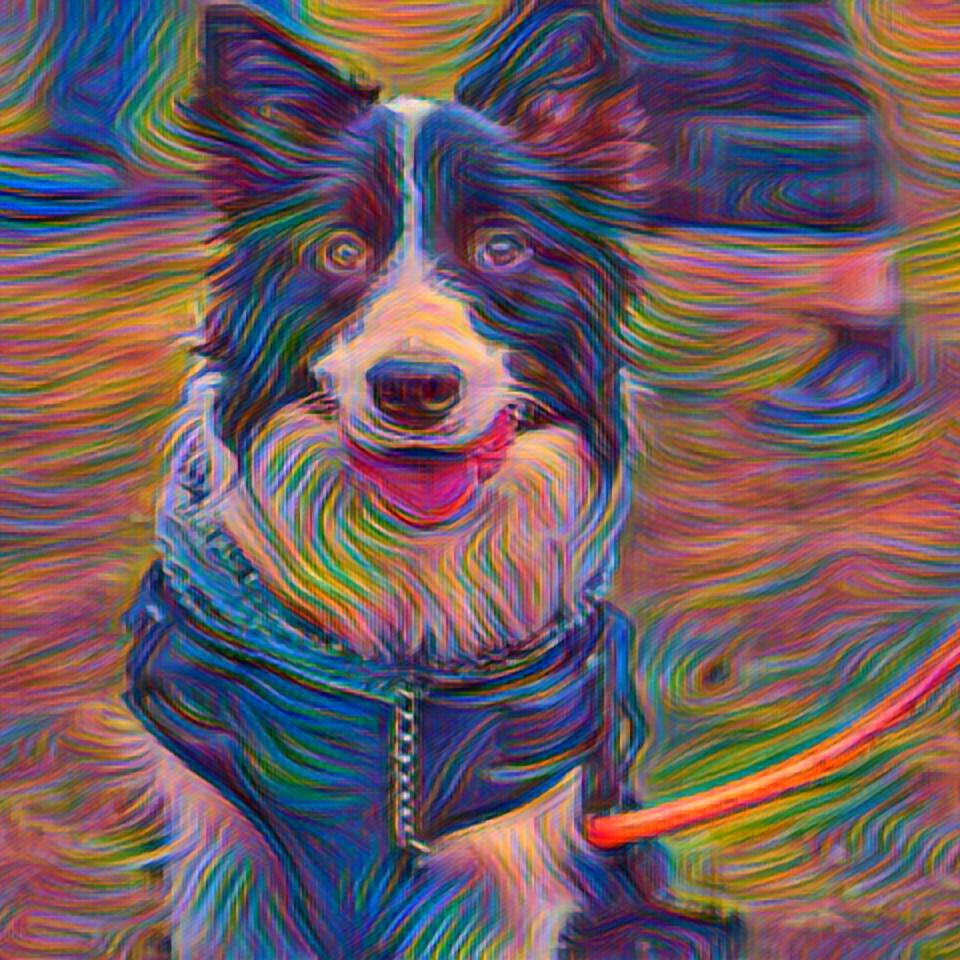} 
        & \hspace*{-4mm} \includegraphics[width=.08\linewidth,height=!]{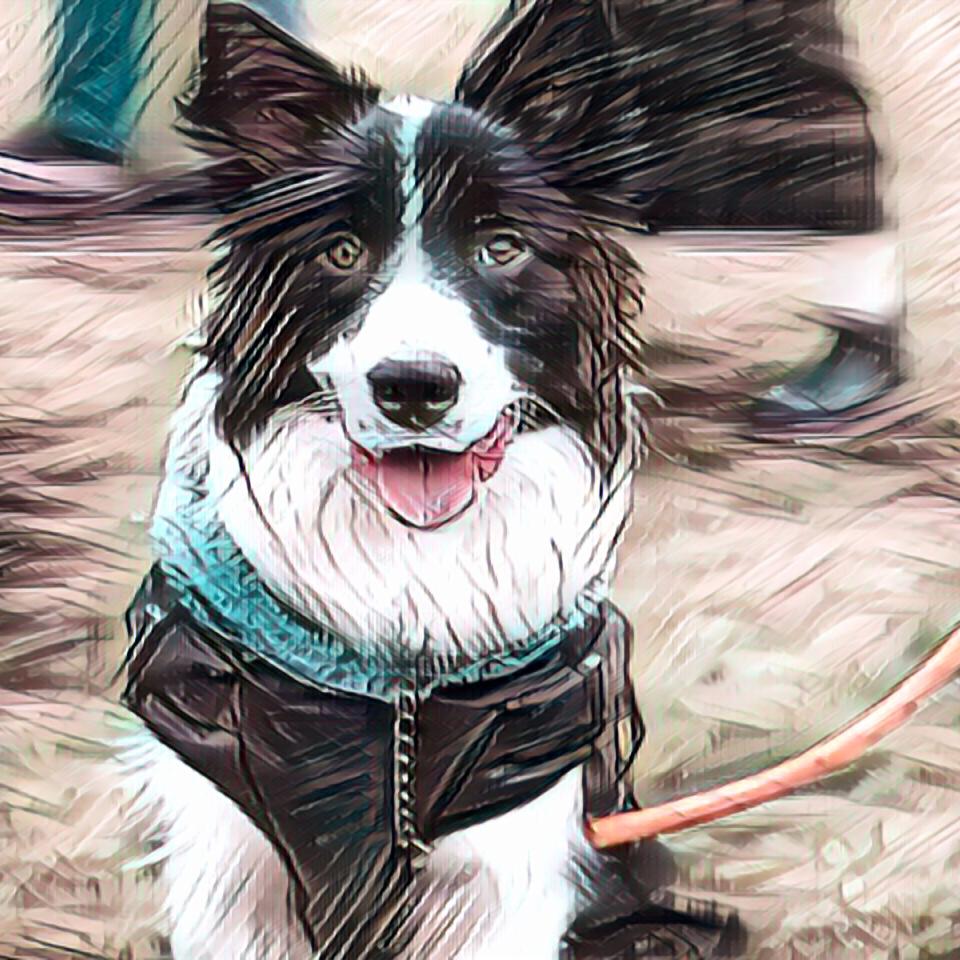} 
        \\
        \hspace*{-4.1mm} (1) 
        & \hspace*{-4mm} (2) 
        & \hspace*{-4mm} (3) 
        & \hspace*{-4mm} (4) 
        & \hspace*{-4mm} (5)  
        & \hspace*{-4mm} (6) 
        & \hspace*{-4mm} (7) 
        & \hspace*{-4mm} (8) 
        & \hspace*{-4mm} (9) 
        & \hspace*{-4mm} (10) 
        \\
        \hspace*{-4.1mm} \includegraphics[width=.08\linewidth,height=!]{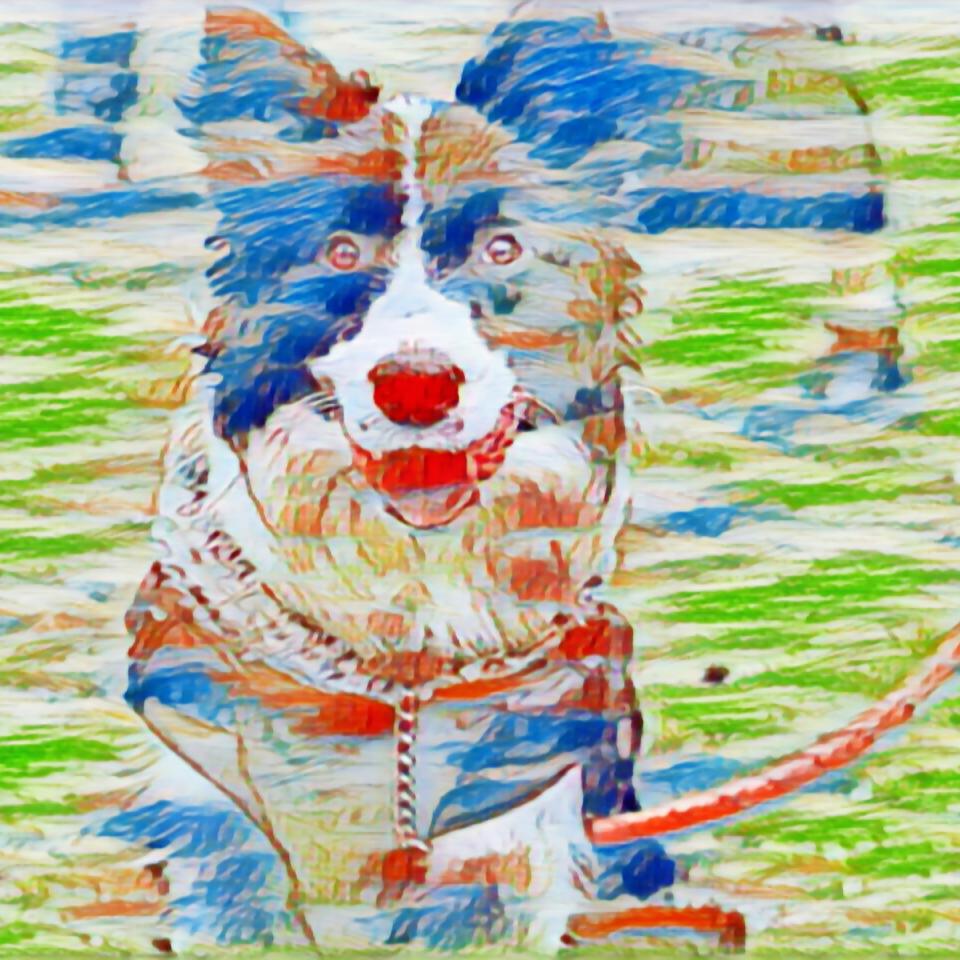} & \hspace*{-4mm} \includegraphics[width=.08\linewidth,height=!]{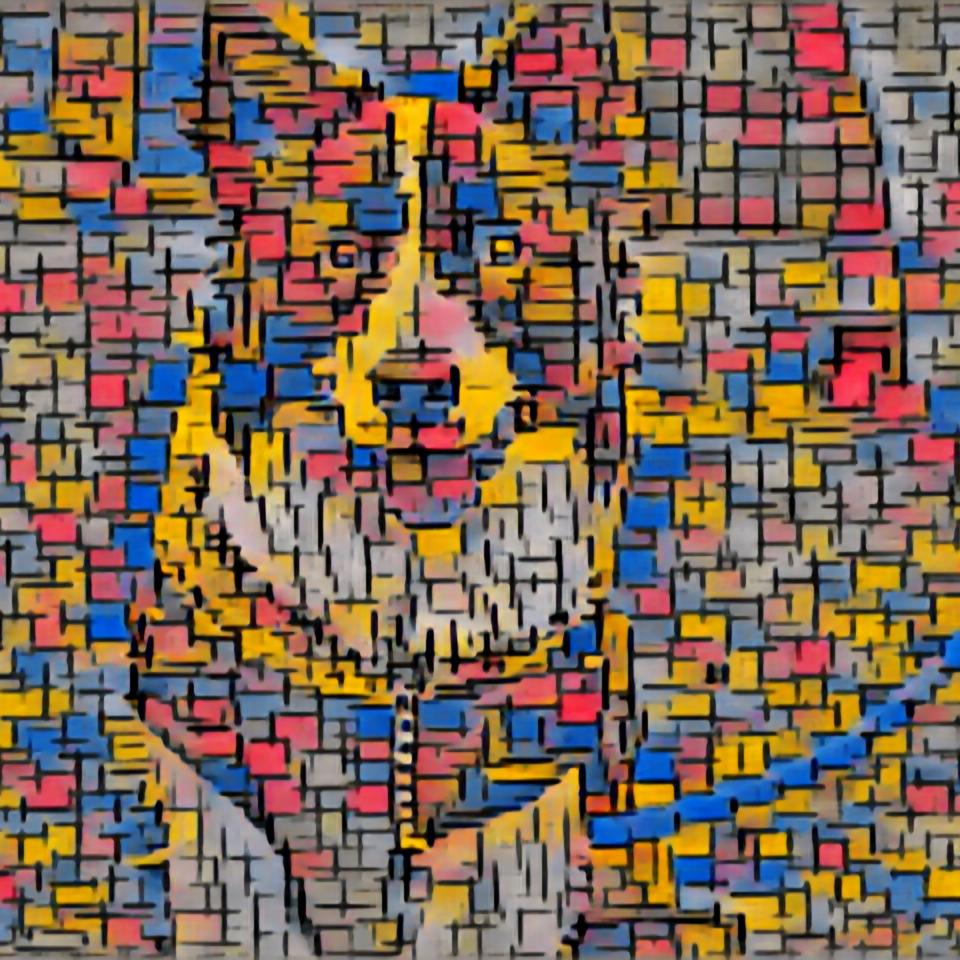} 
        & \hspace*{-4mm} \includegraphics[width=.08\linewidth,height=!]{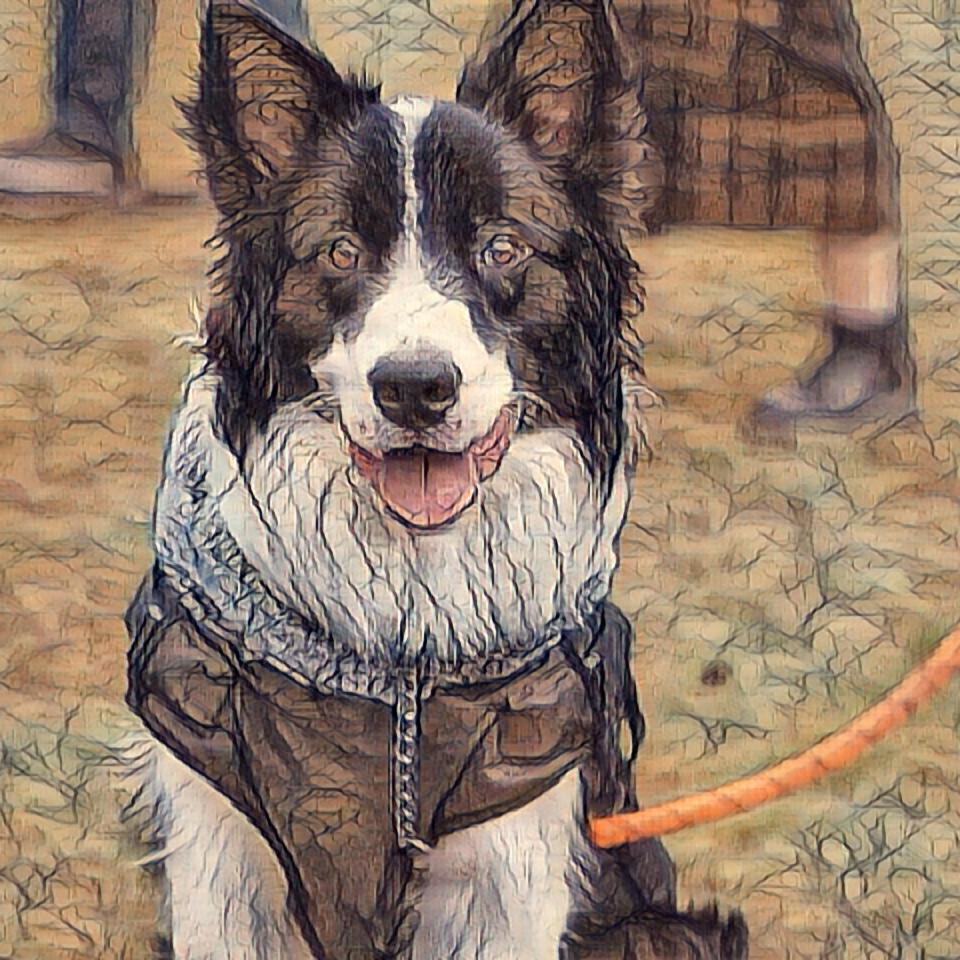} 
        & \hspace*{-4mm} \includegraphics[width=.08\linewidth,height=!]{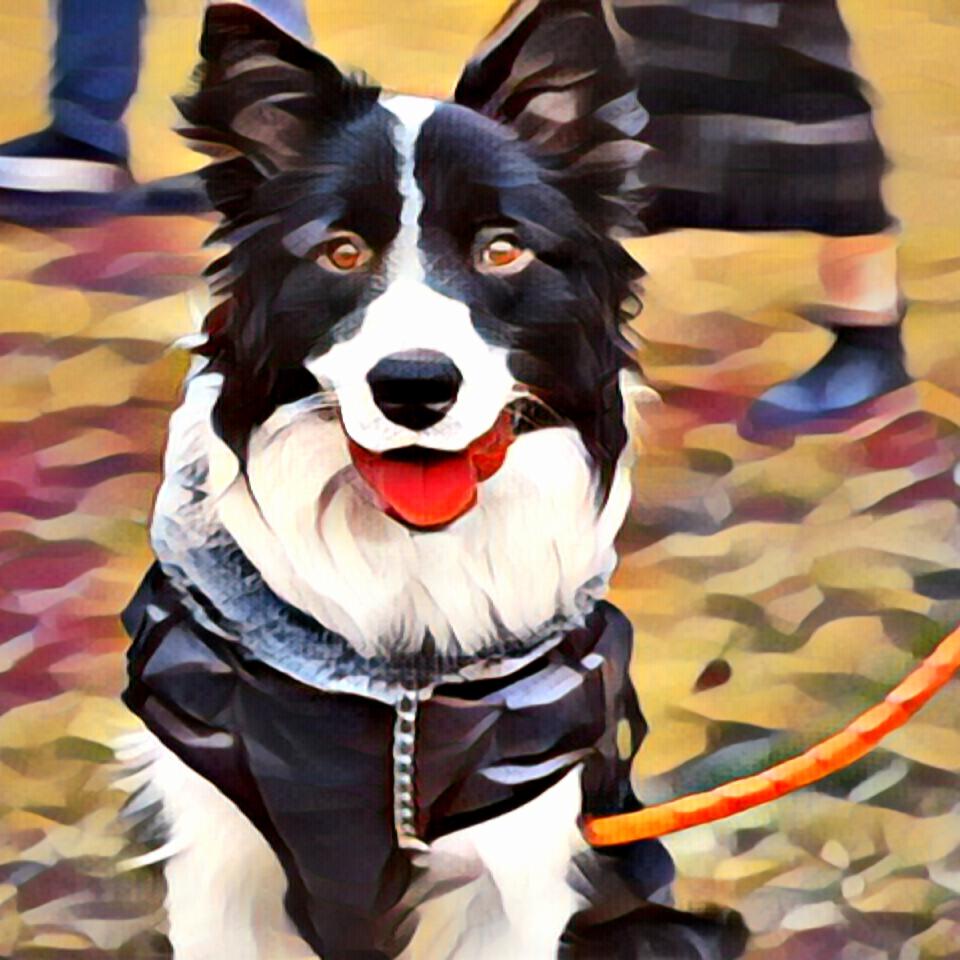} 
        & \hspace*{-4mm} \includegraphics[width=.08\linewidth,height=!]{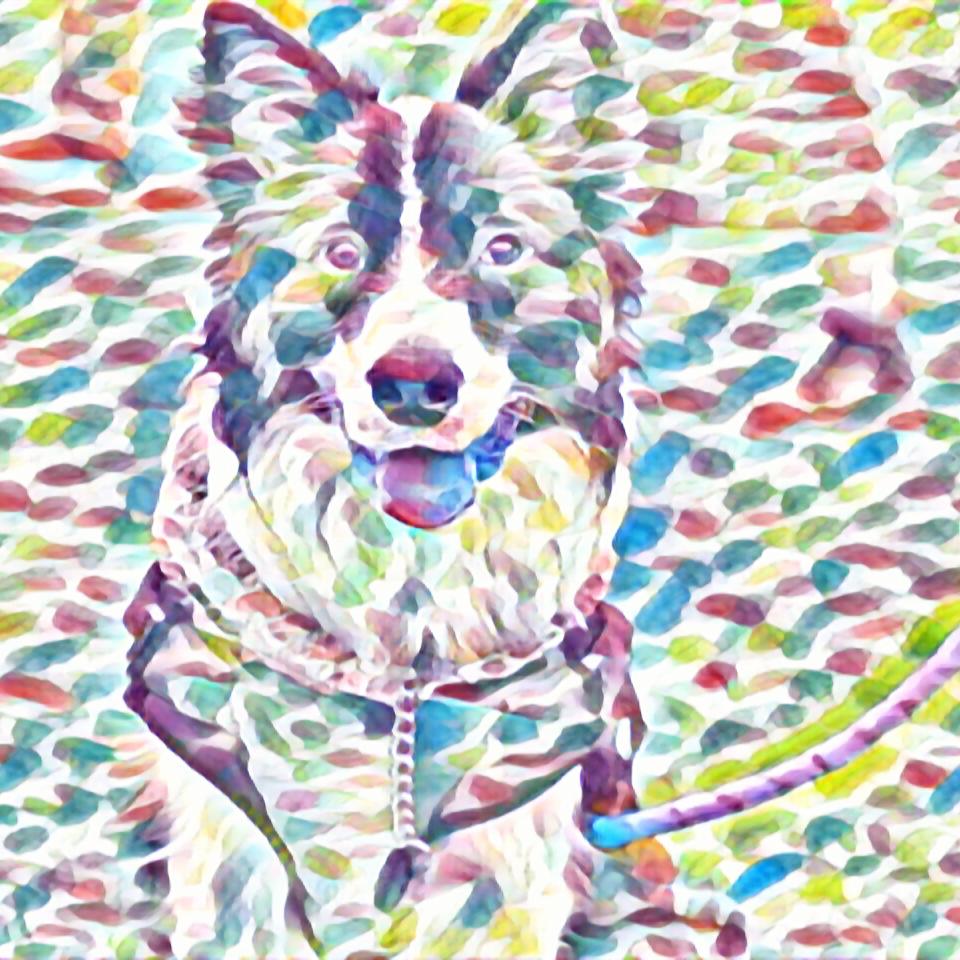} 
        & \hspace*{-4mm} \includegraphics[width=.08\linewidth,height=!]{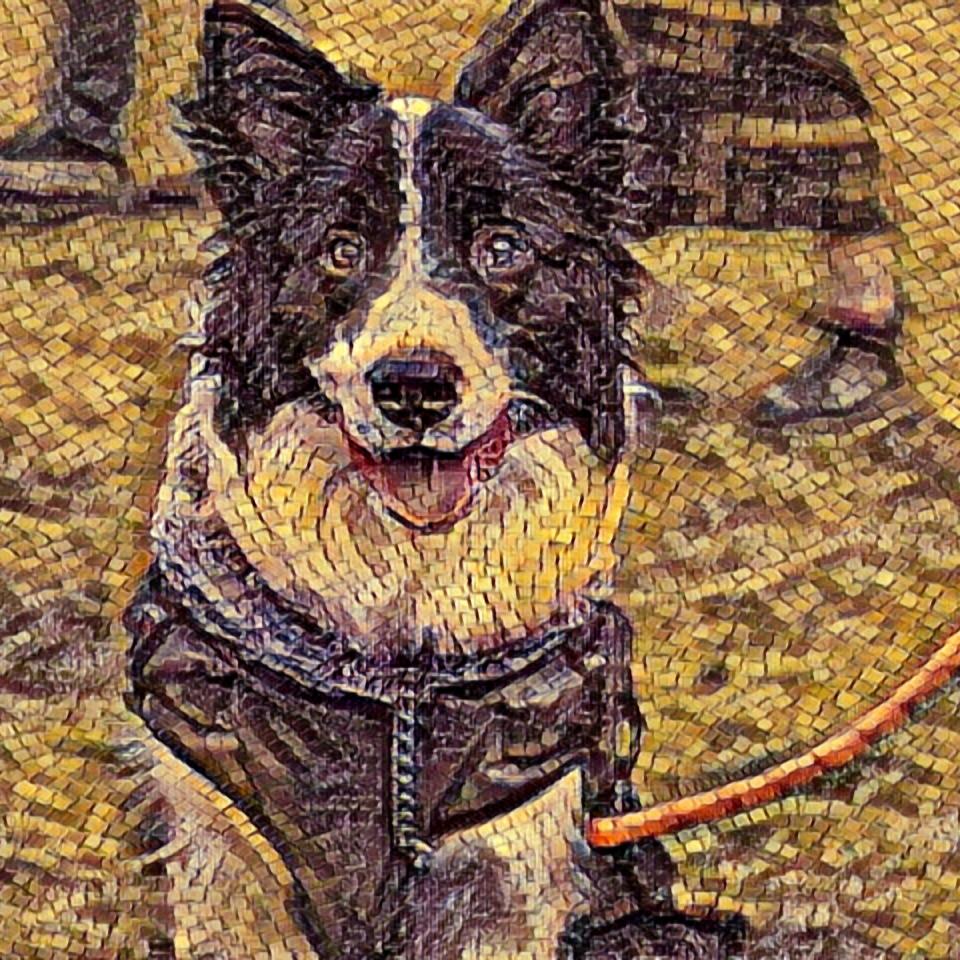} & \hspace*{-4mm} \includegraphics[width=.08\linewidth,height=!]{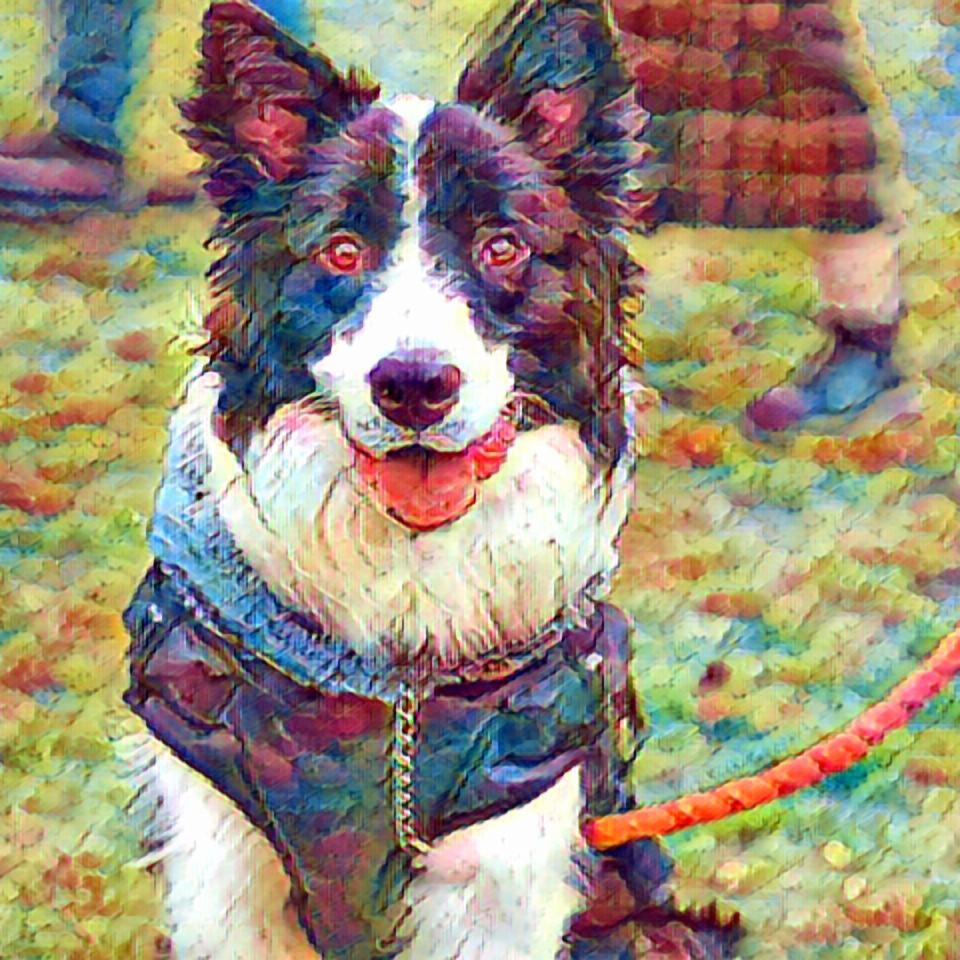} 
        & \hspace*{-4mm} \includegraphics[width=.08\linewidth,height=!]{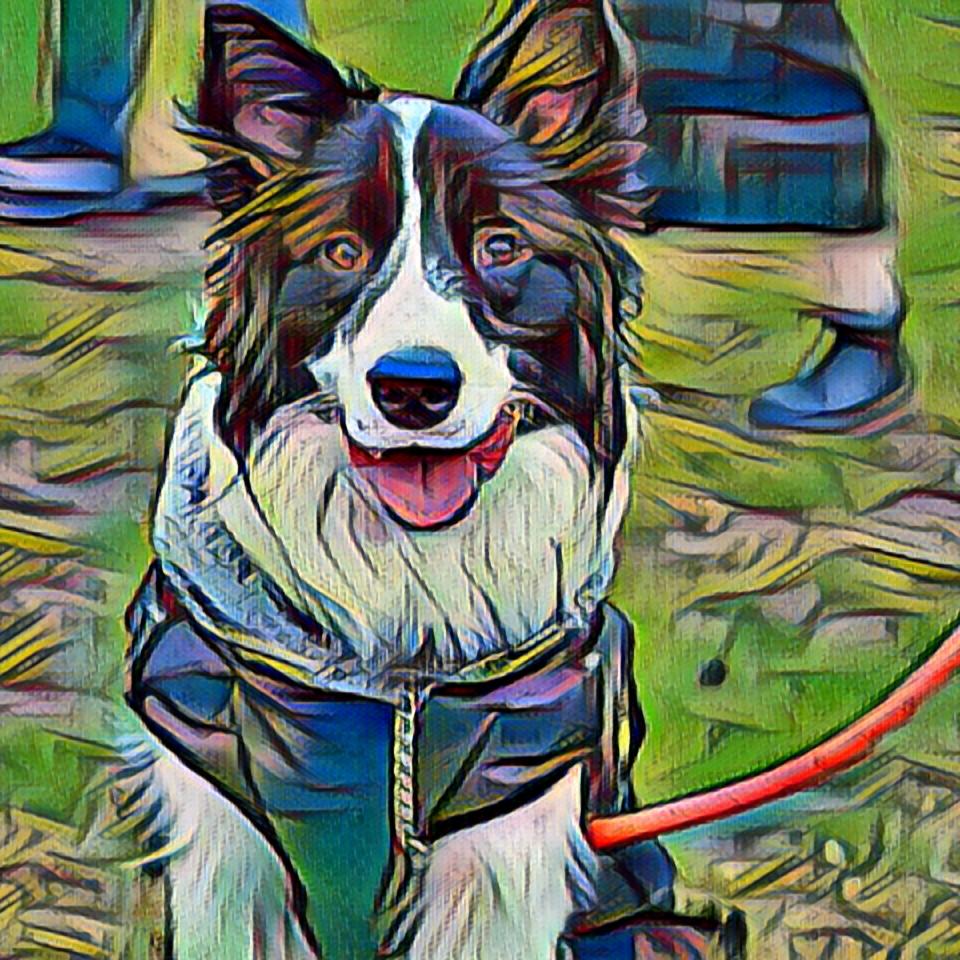} 
        & \hspace*{-4mm} \includegraphics[width=.08\linewidth,height=!]{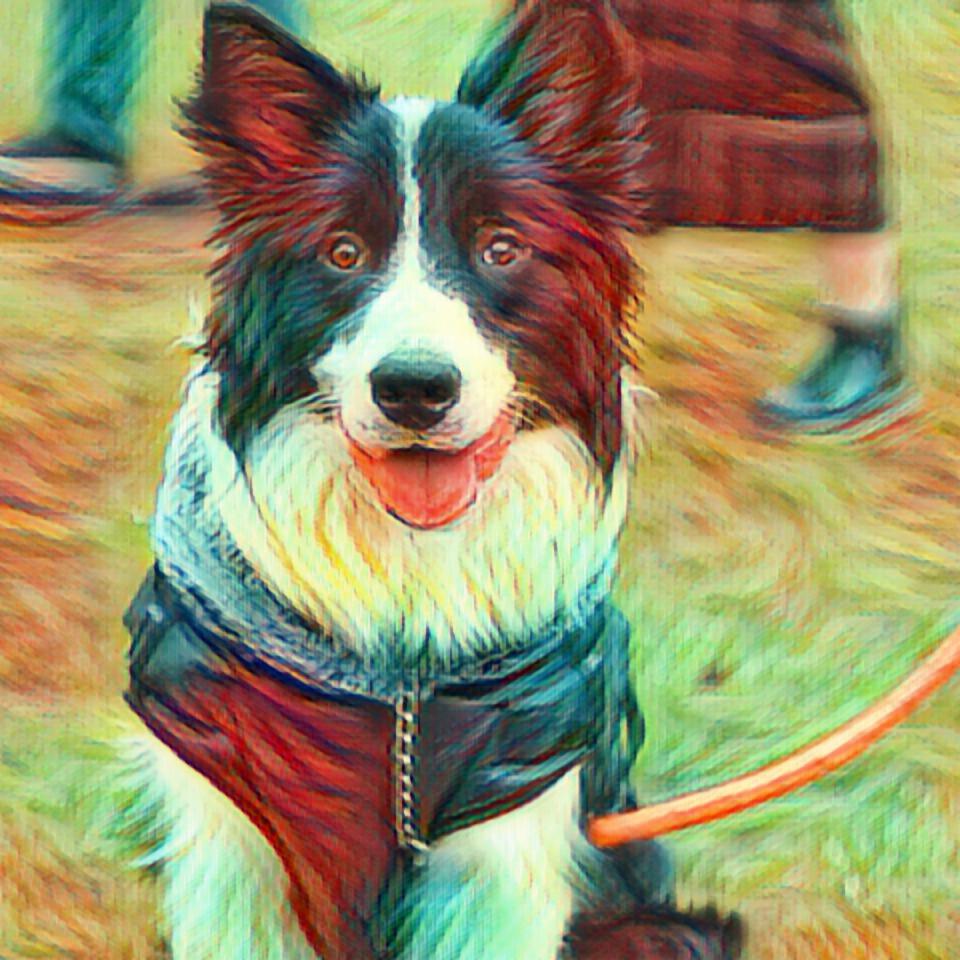} 
        & \hspace*{-4mm} \includegraphics[width=.08\linewidth,height=!]{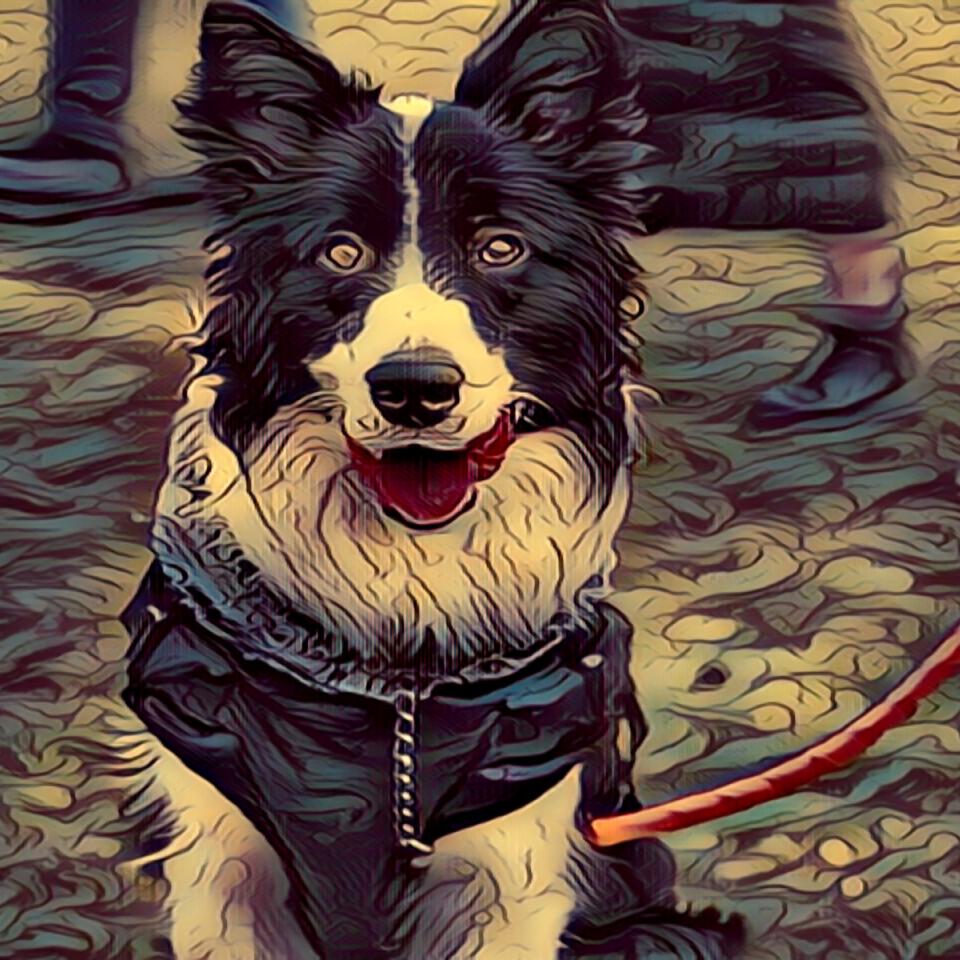} 
        \\
        \hspace*{-4.1mm} (11) 
        & \hspace*{-4mm} (12) 
        & \hspace*{-4mm} (13) 
        & \hspace*{-4mm} (14) 
        & \hspace*{-4mm} (15) 
        & \hspace*{-4mm} (16) 
        & \hspace*{-4mm} (17) 
        & \hspace*{-4mm} (18) 
        & \hspace*{-4mm} (19) 
        & \hspace*{-4mm} (20) 
        \\
        \hspace*{-4.1mm} \includegraphics[width=.08\linewidth,height=!]{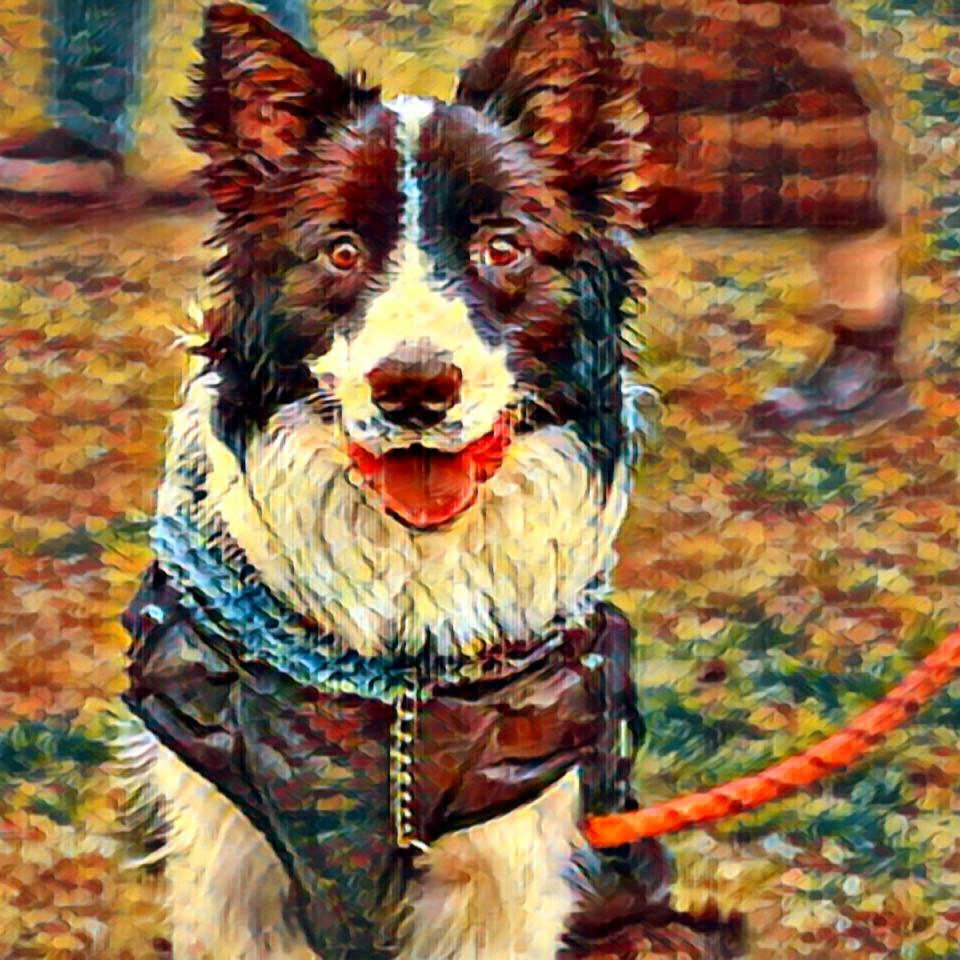} & \hspace*{-4mm} \includegraphics[width=.08\linewidth,height=!]{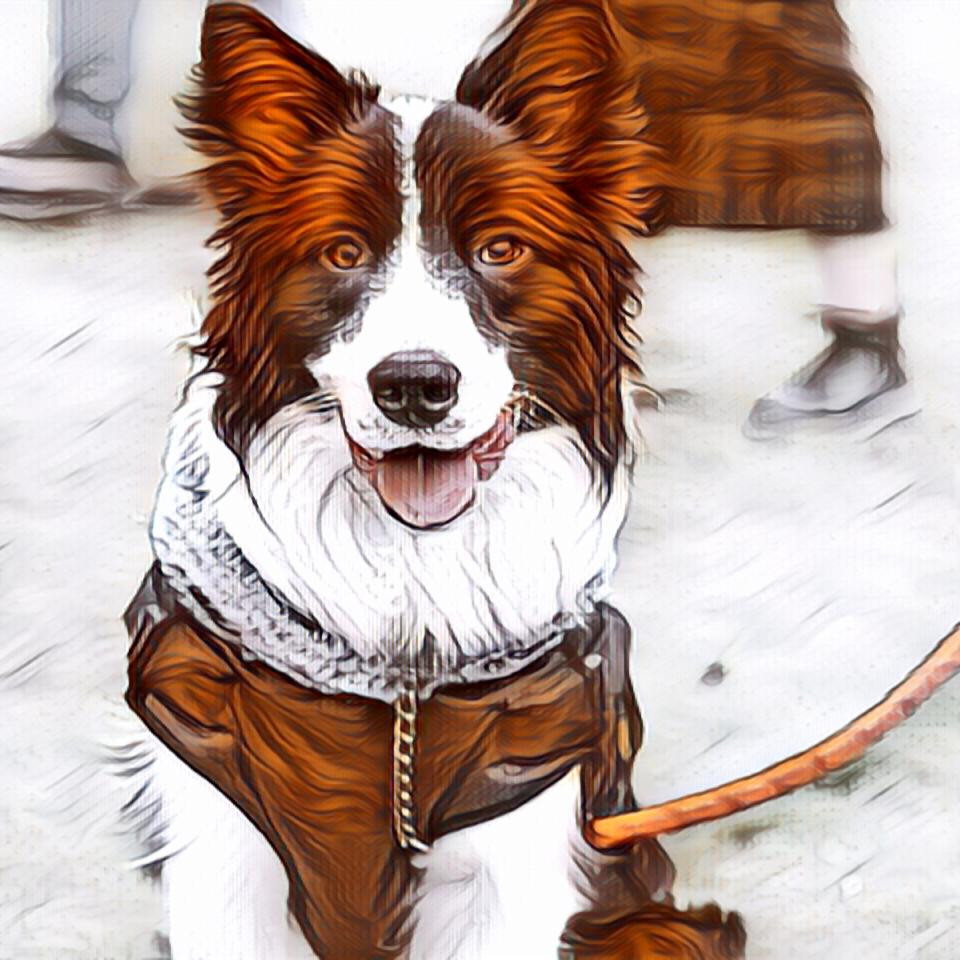} 
        & \hspace*{-4mm} \includegraphics[width=.08\linewidth,height=!]{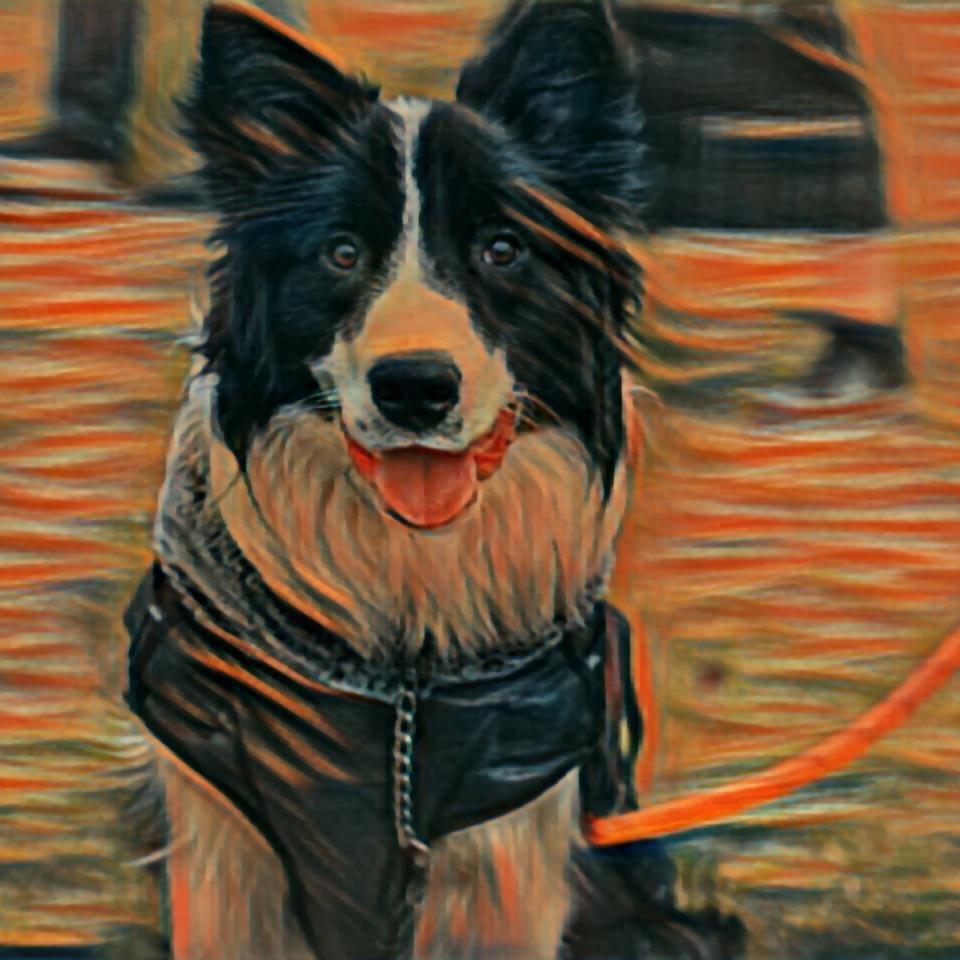} 
        & \hspace*{-4mm} \includegraphics[width=.08\linewidth,height=!]{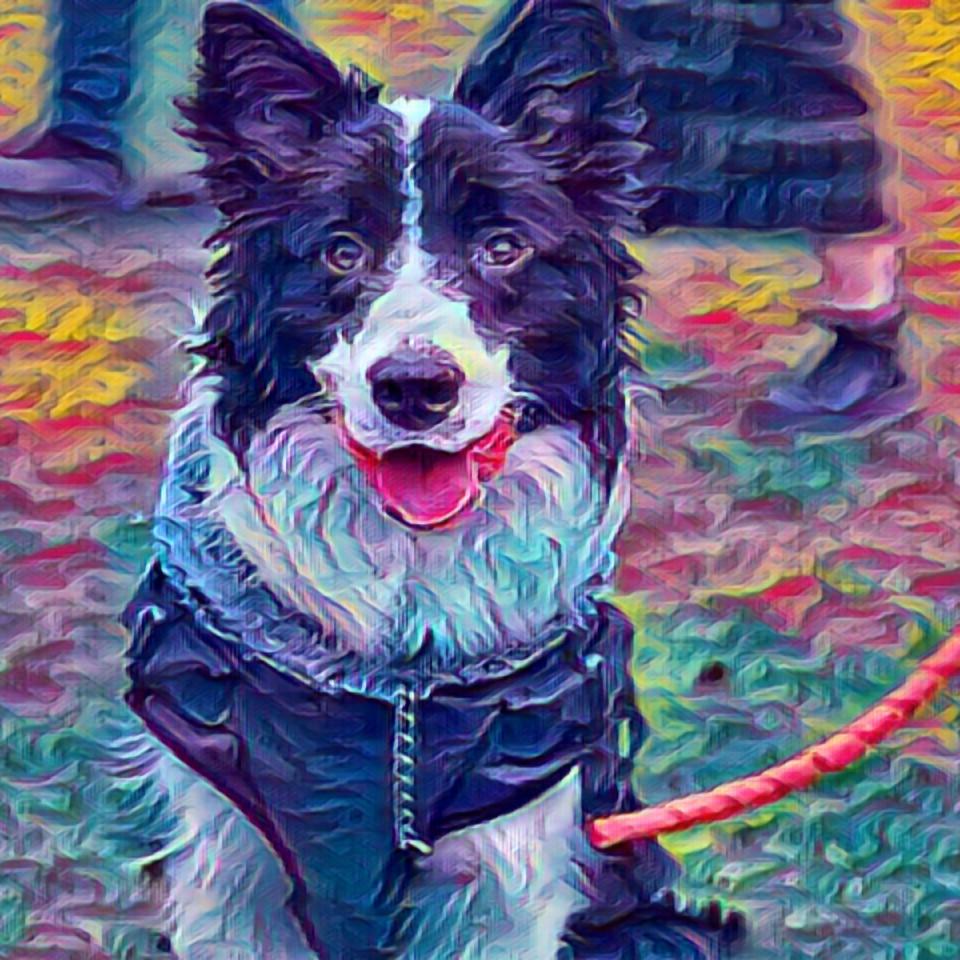} 
        & \hspace*{-4mm} \includegraphics[width=.08\linewidth,height=!]{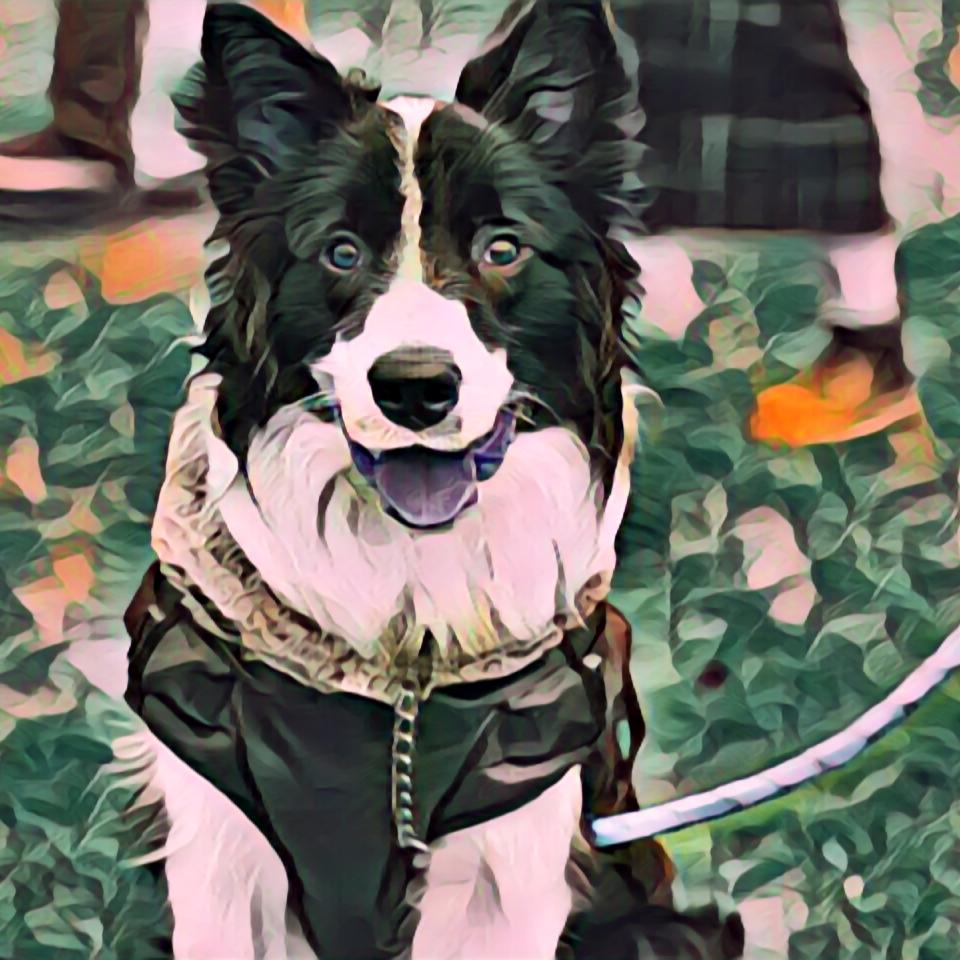} 
        & \hspace*{-4mm} \includegraphics[width=.08\linewidth,height=!]{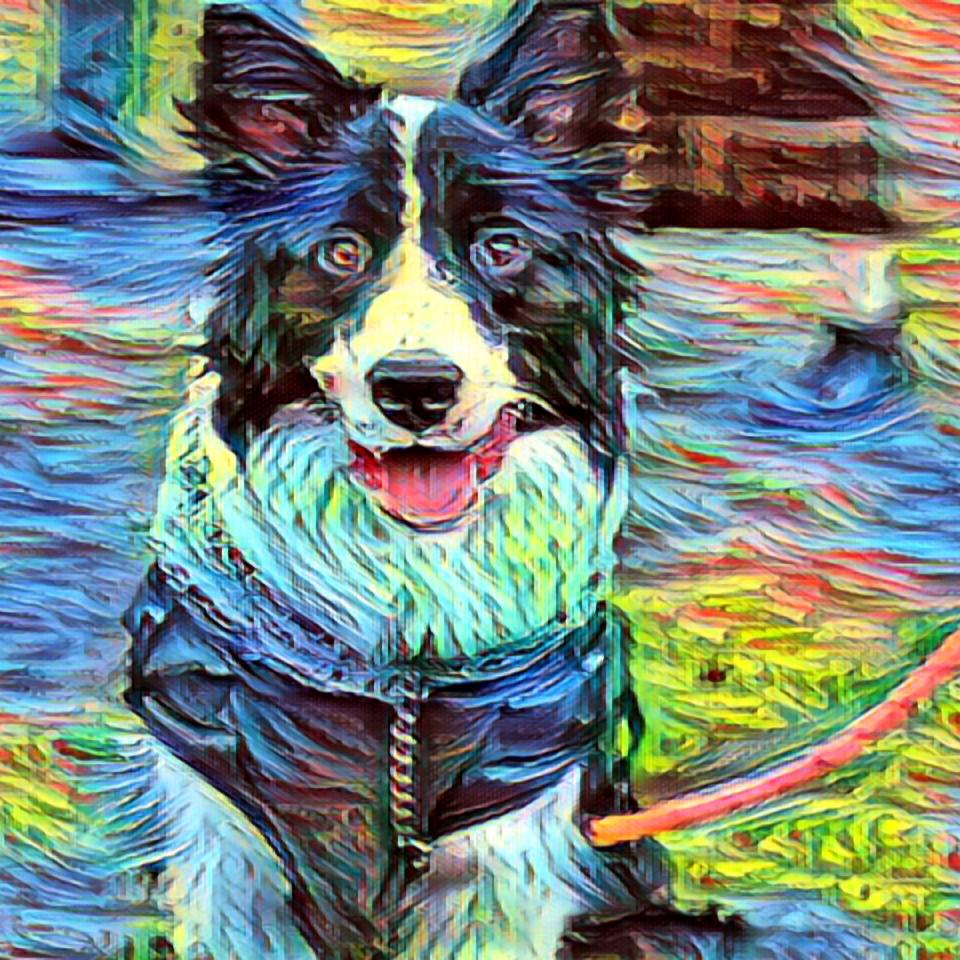} 
        & \hspace*{-4mm} \includegraphics[width=.08\linewidth,height=!]{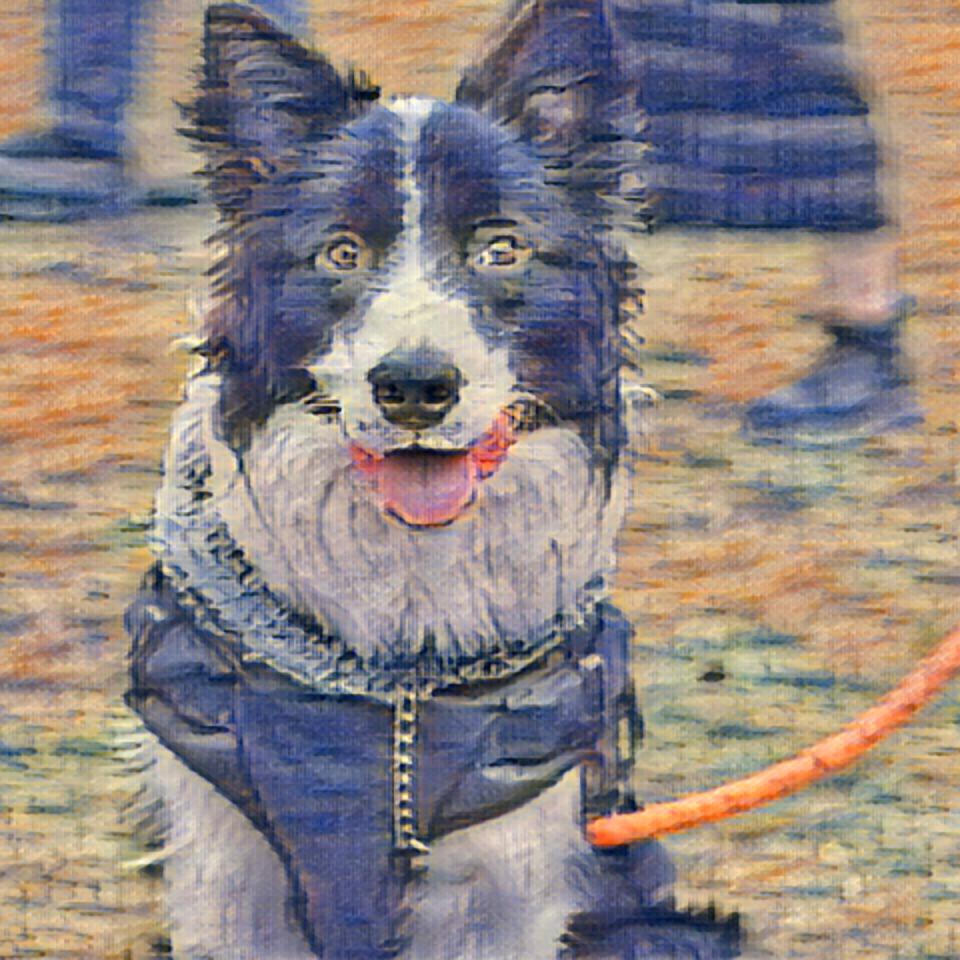} 
        & \hspace*{-4mm} \includegraphics[width=.08\linewidth,height=!]{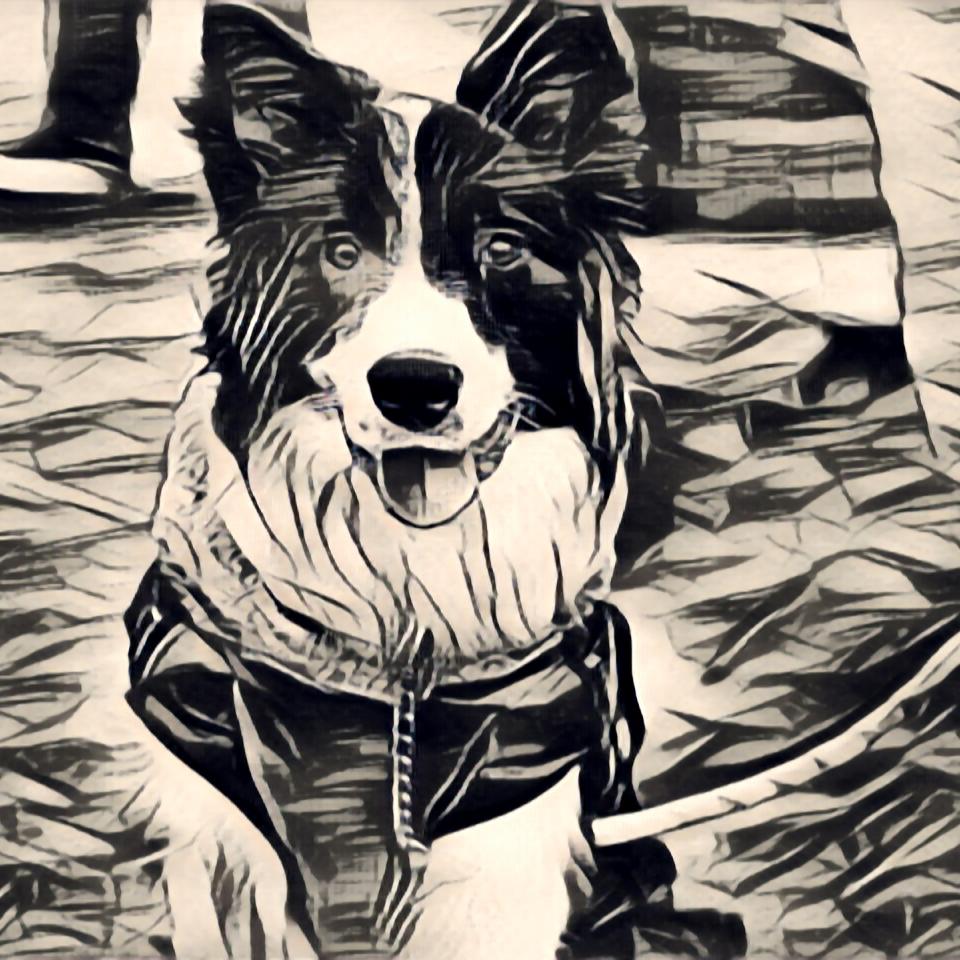} 
        & \hspace*{-4mm} \includegraphics[width=.08\linewidth,height=!]{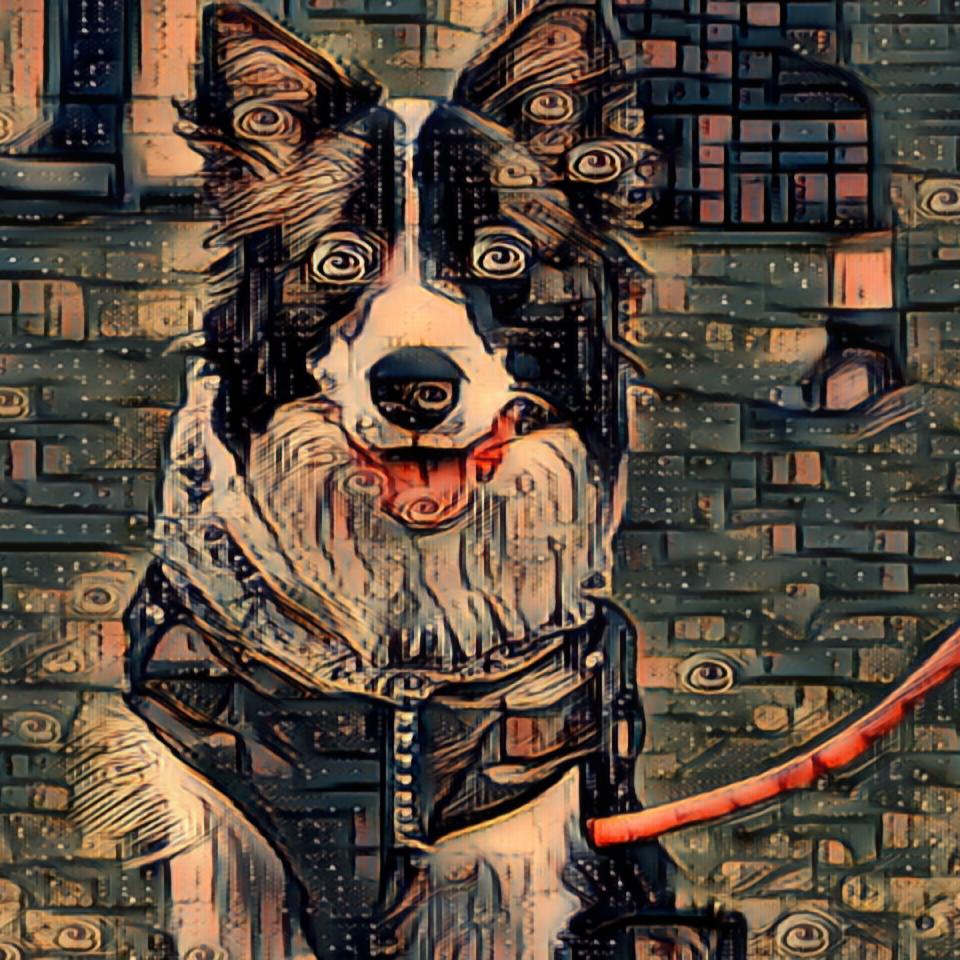} 
        & \hspace*{-4mm} \includegraphics[width=.08\linewidth,height=!]{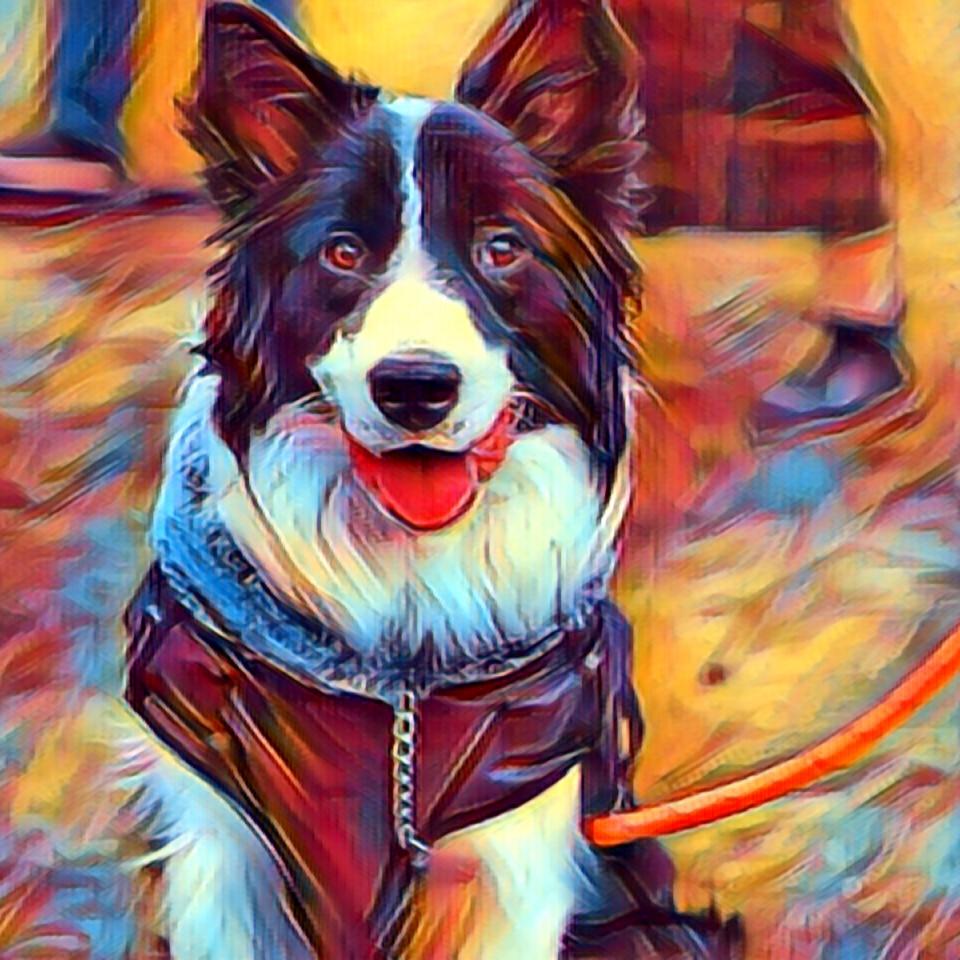} 
        \\
        \hspace*{-4.1mm} (21) 
        & \hspace*{-4mm} (22) 
        & \hspace*{-4mm} (23) 
        & \hspace*{-4mm} (24) 
        & \hspace*{-4mm} (25) 
        & \hspace*{-4mm} (26) 
        & \hspace*{-4mm} (27) 
        & \hspace*{-4mm} (28) 
        & \hspace*{-4mm} (29) 
        & \hspace*{-4mm} (30) 
        \\
        \hspace*{-4.1mm} \includegraphics[width=.08\linewidth,height=!]{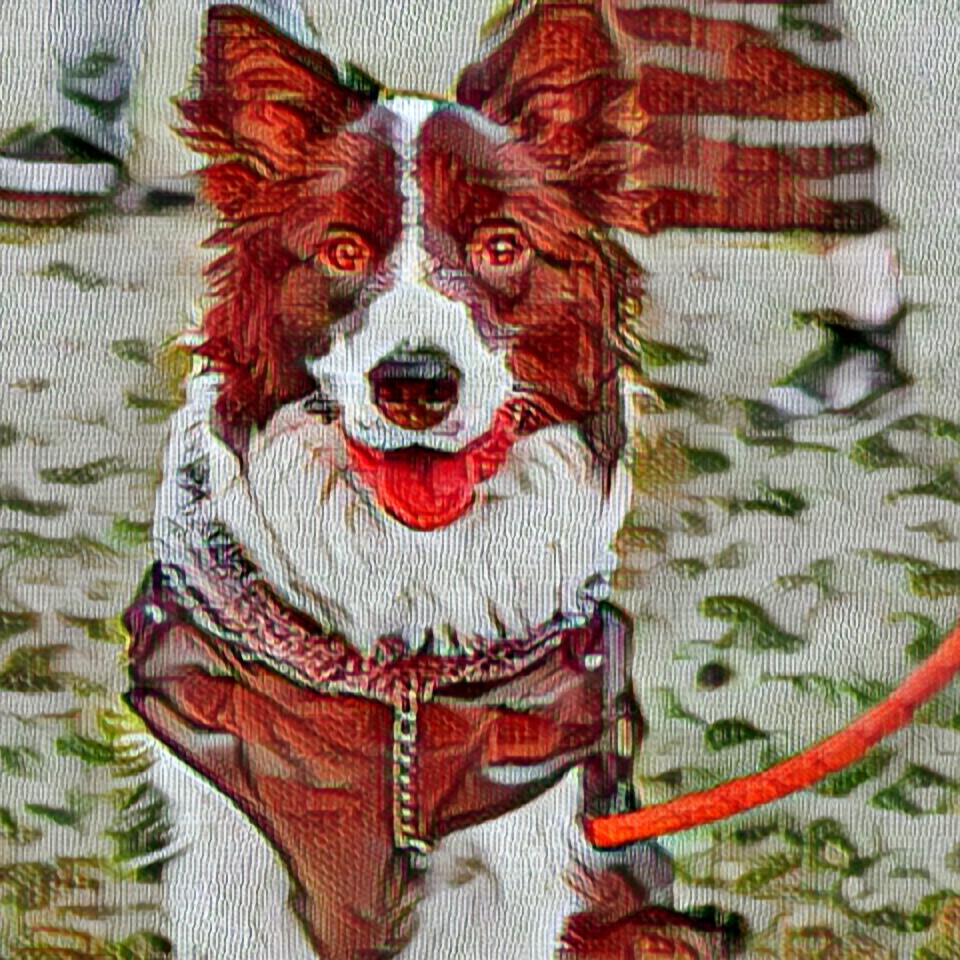} 
        & \hspace*{-4mm} \includegraphics[width=.08\linewidth,height=!]{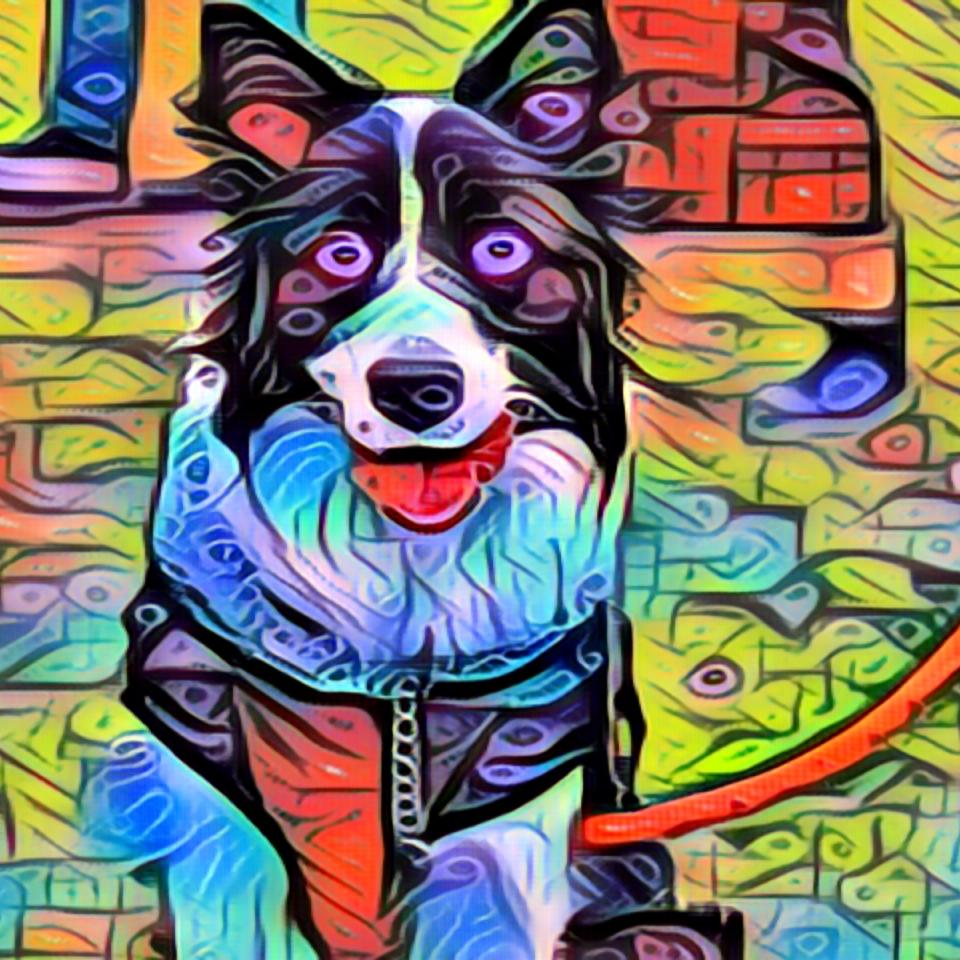} 
        & \hspace*{-4mm} \includegraphics[width=.08\linewidth,height=!]{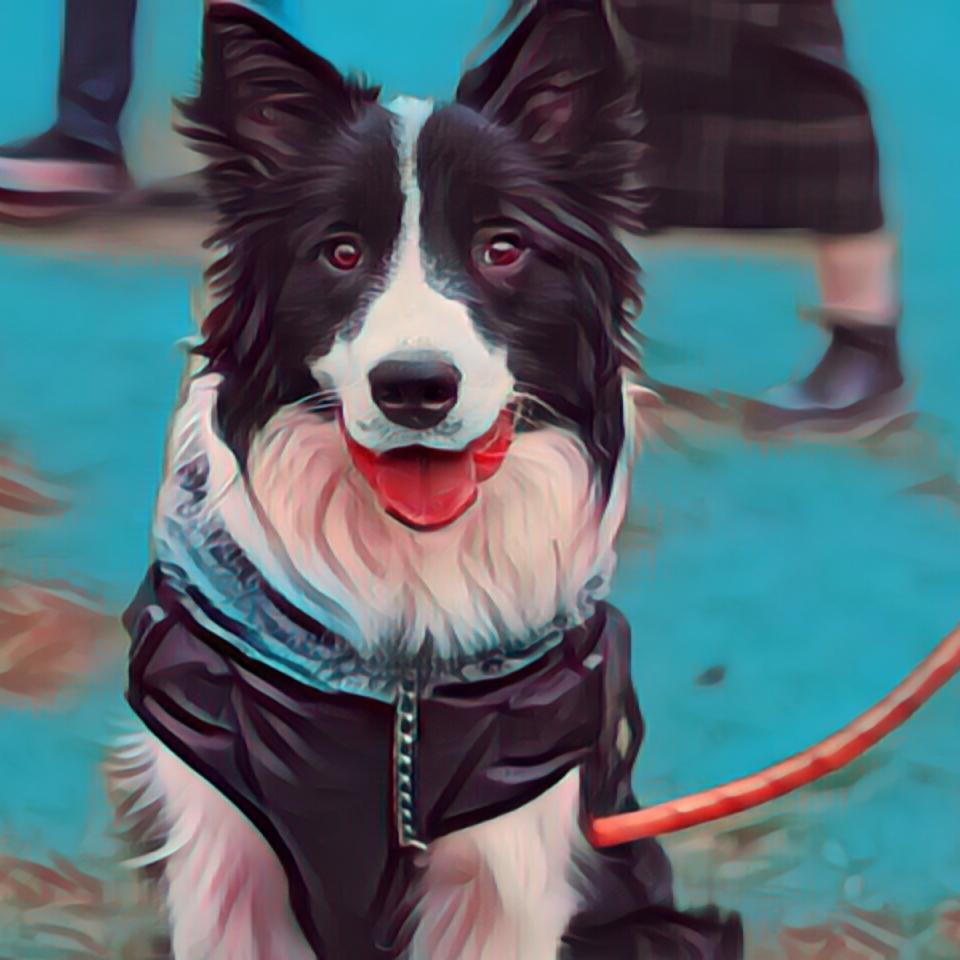} 
        & \hspace*{-4mm} \includegraphics[width=.08\linewidth,height=!]{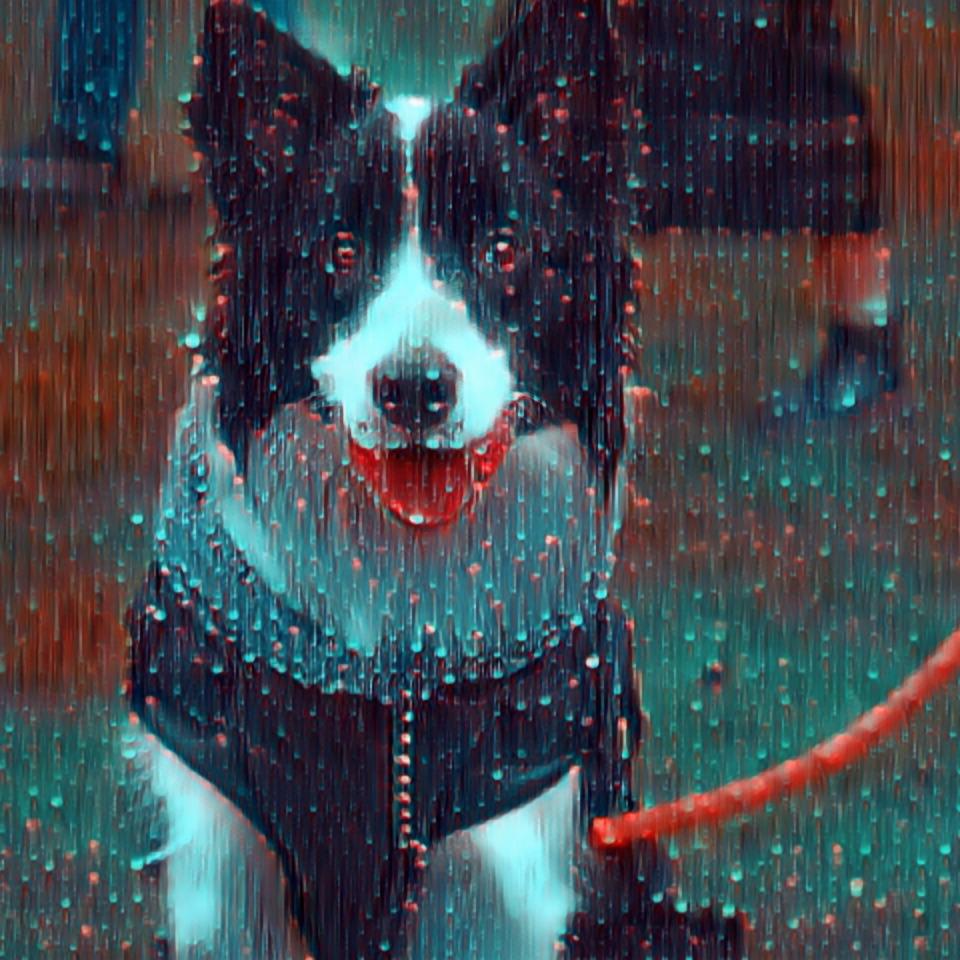} 
        & \hspace*{-4mm} \includegraphics[width=.08\linewidth,height=!]{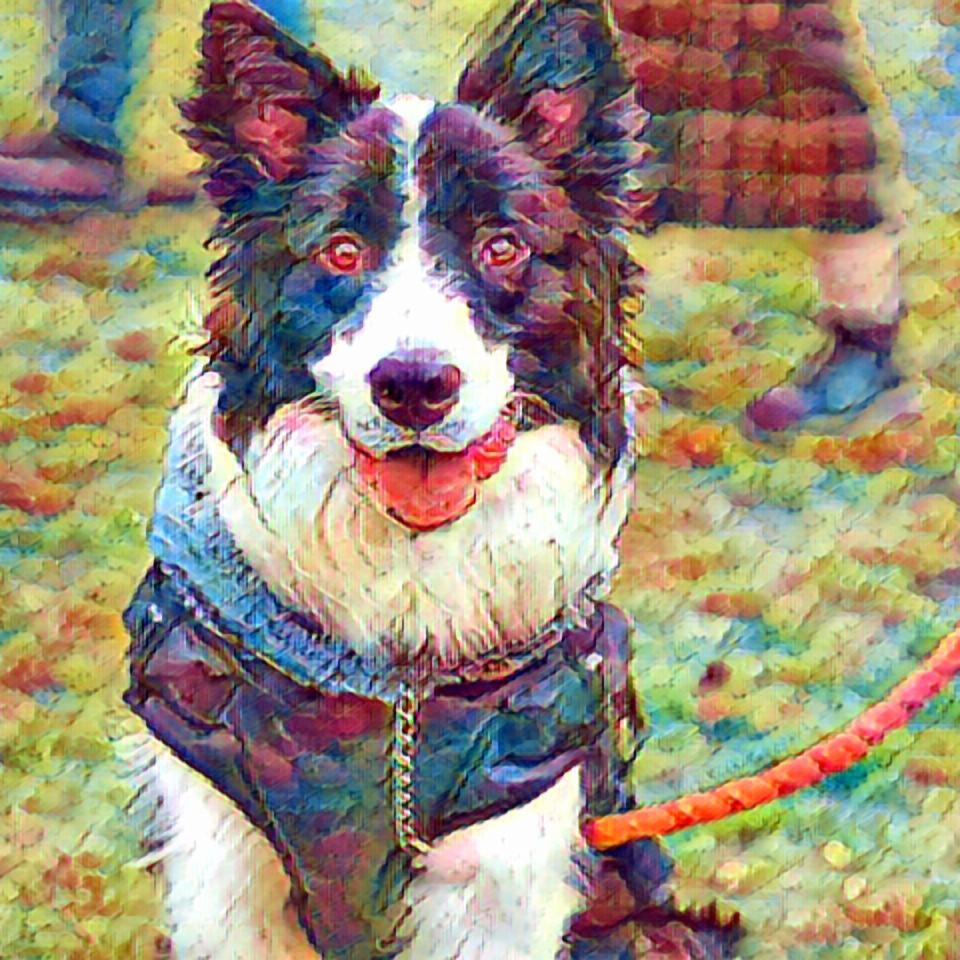} 
        & \hspace*{-4mm} \includegraphics[width=.08\linewidth,height=!]{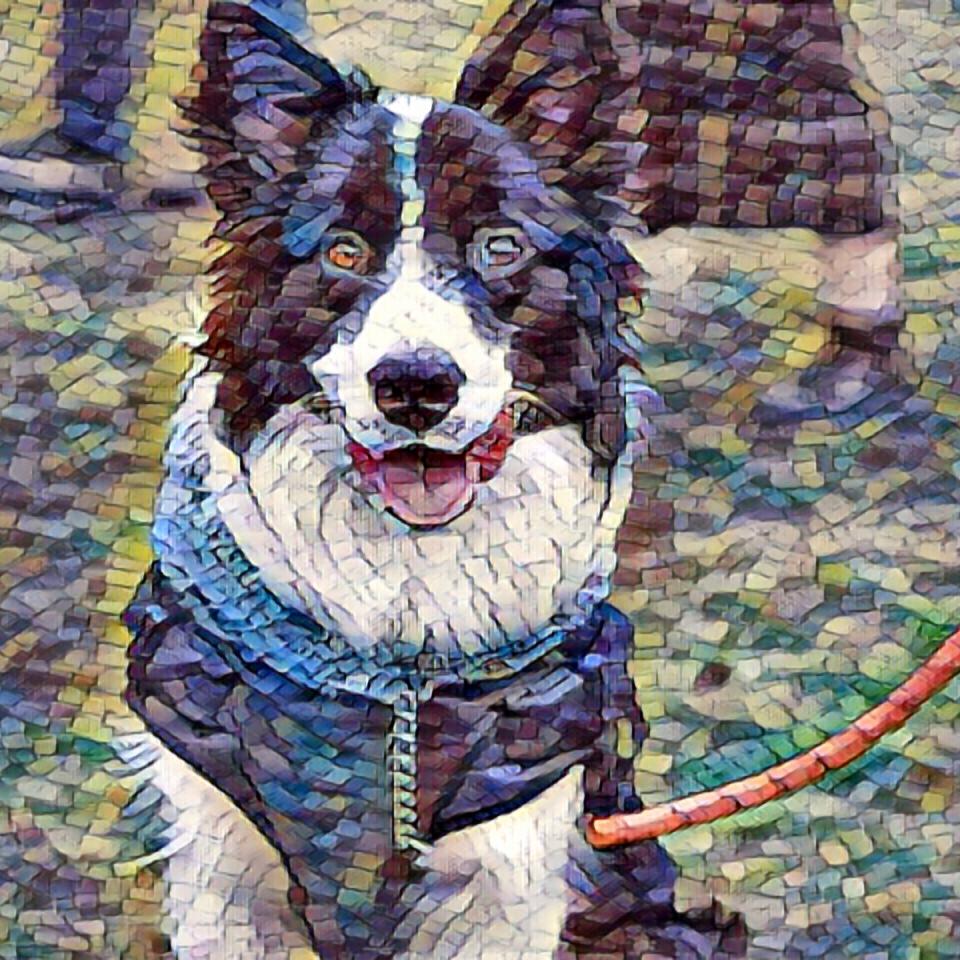} 
        & \hspace*{-4mm} \includegraphics[width=.08\linewidth,height=!]{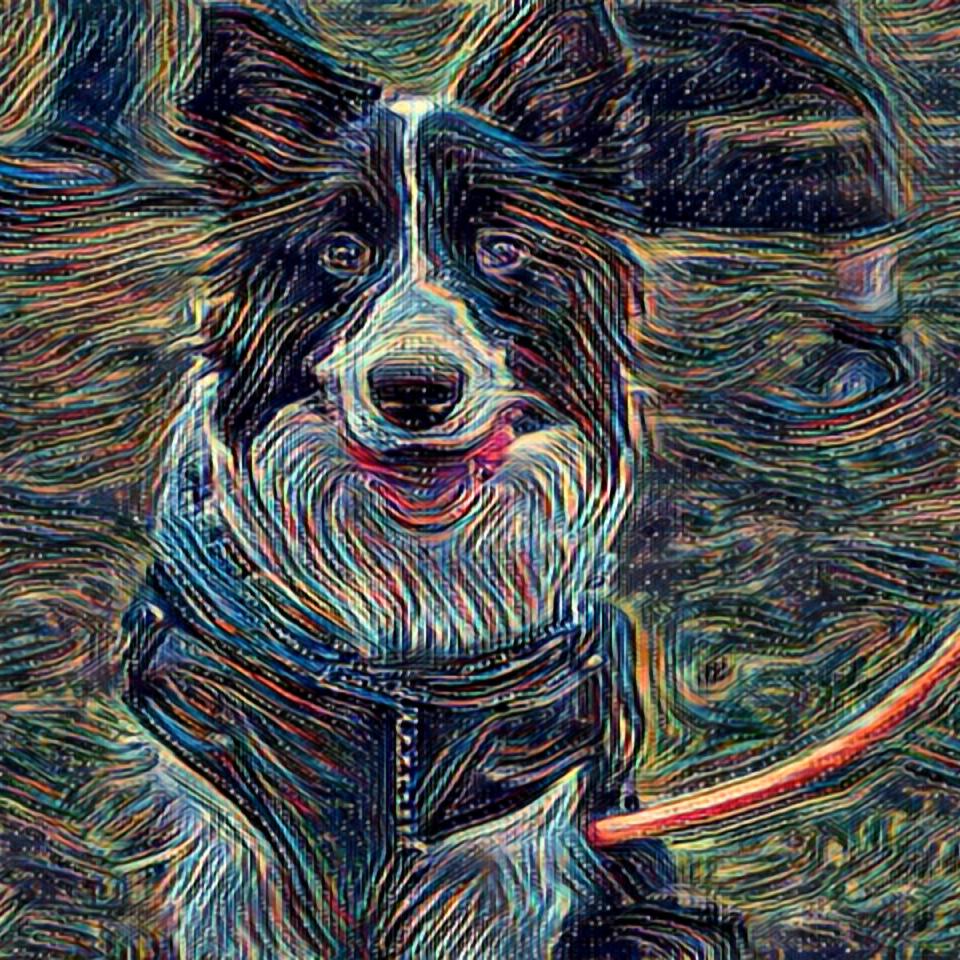} 
        & \hspace*{-4mm} \includegraphics[width=.08\linewidth,height=!]{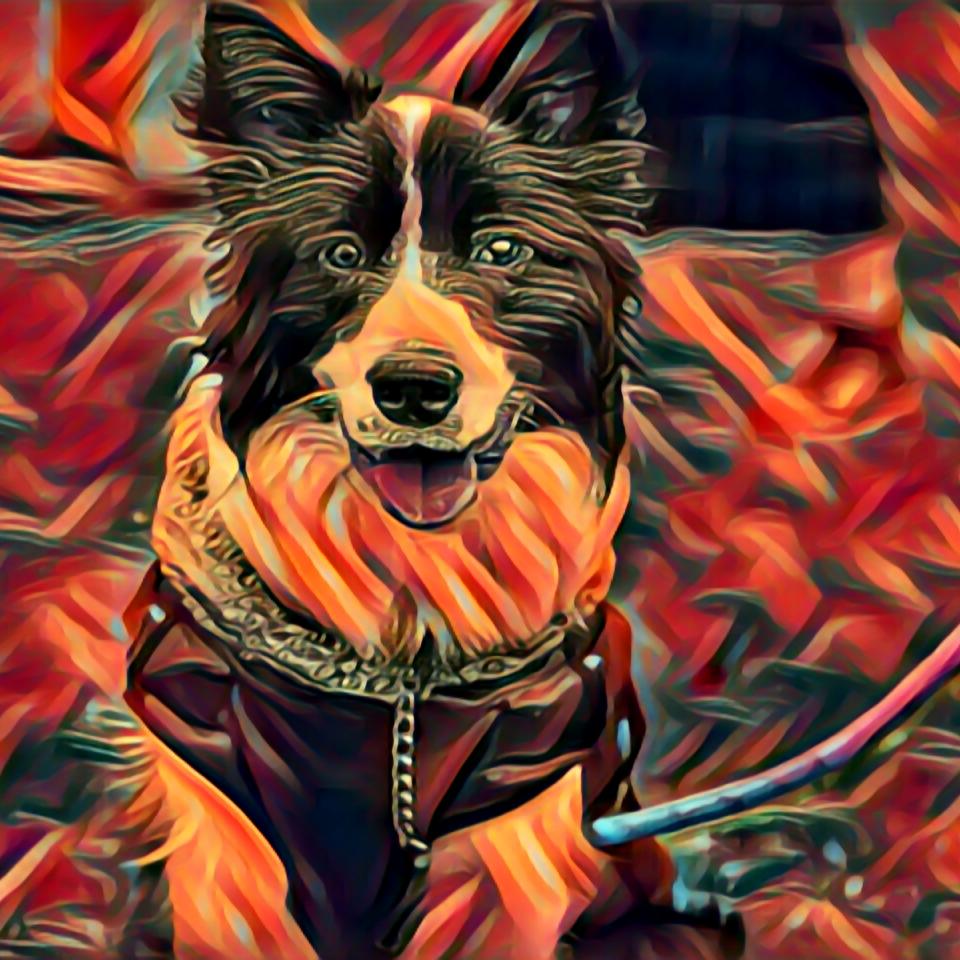} 
        & \hspace*{-4mm} \includegraphics[width=.08\linewidth,height=!]{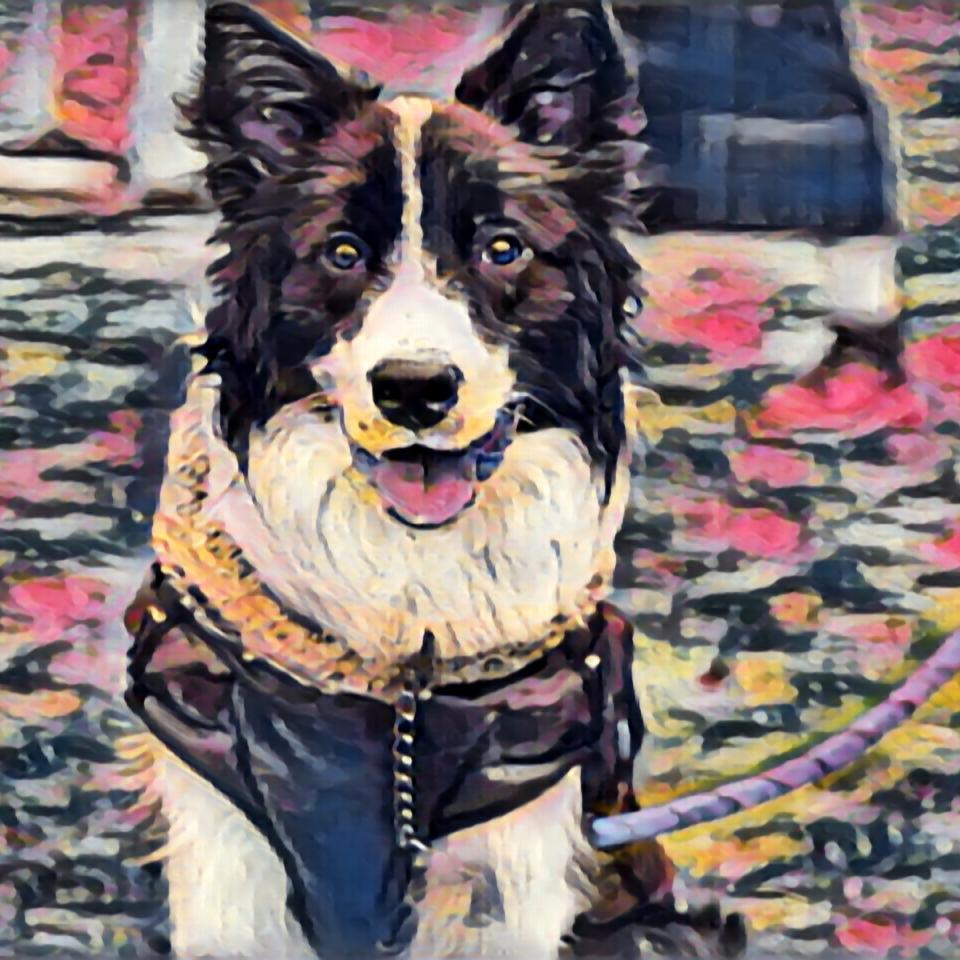} 
        & \hspace*{-4mm} \includegraphics[width=.08\linewidth,height=!]{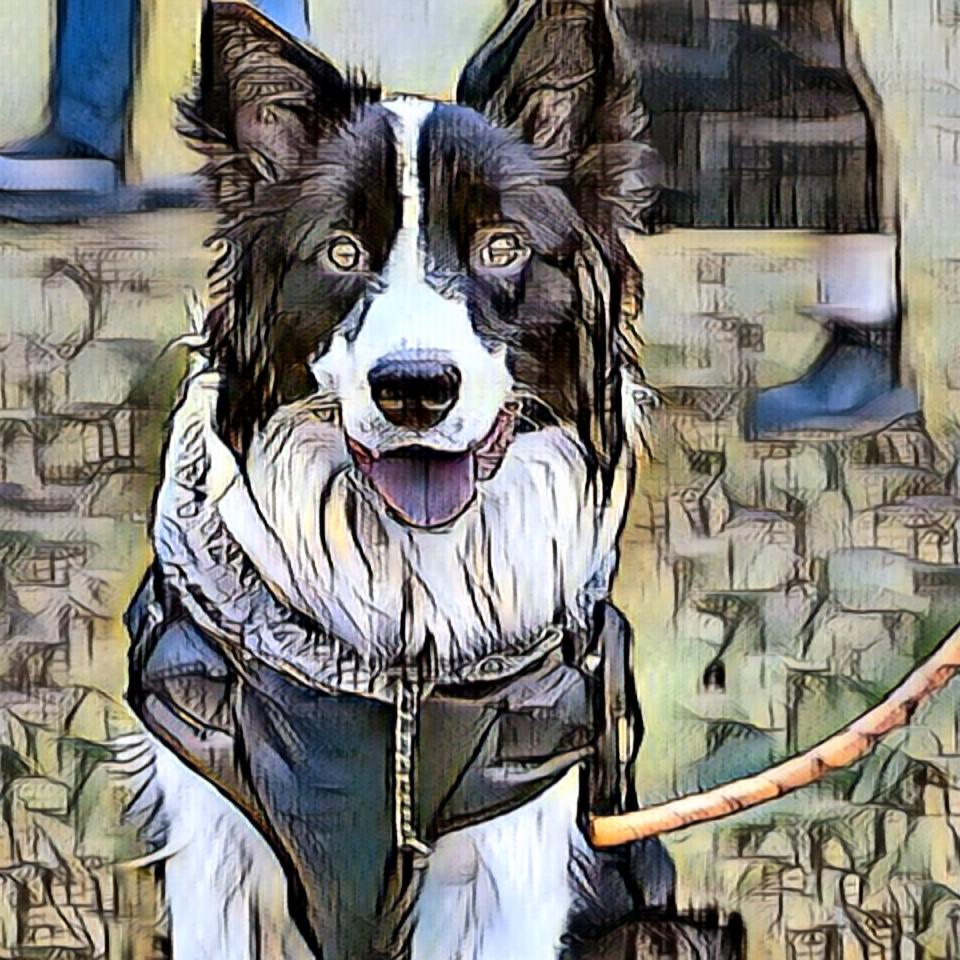} 
        \\
        \hspace*{-4.1mm} (31) 
        & \hspace*{-4mm} (32) 
        & \hspace*{-4mm} (33) 
        & \hspace*{-4mm} (34) 
        & \hspace*{-4mm} (35) 
        & \hspace*{-4mm} (36) 
        & \hspace*{-4mm} (37) 
        & \hspace*{-4mm} (38) 
        & \hspace*{-4mm} (39) 
        & \hspace*{-4mm} (40) 
        \\
        \hspace*{-4.1mm} \includegraphics[width=.08\linewidth,height=!]{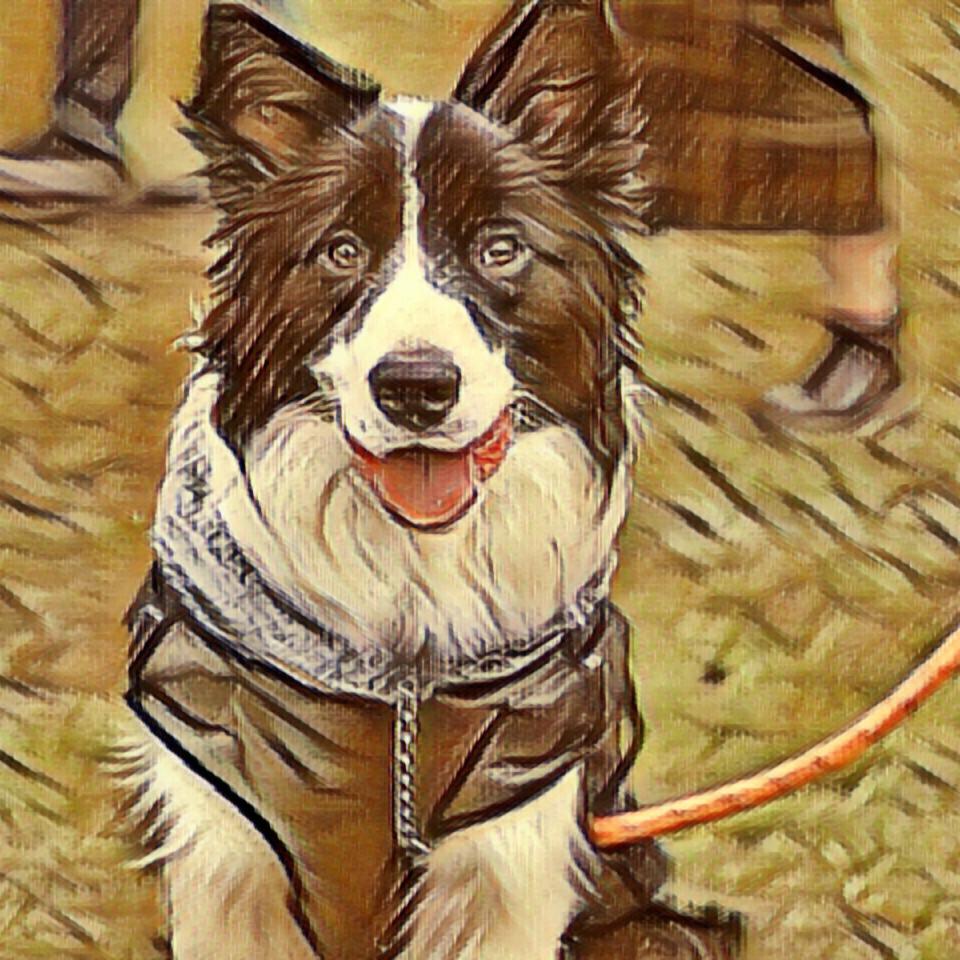} & \hspace*{-4mm} \includegraphics[width=.08\linewidth,height=!]{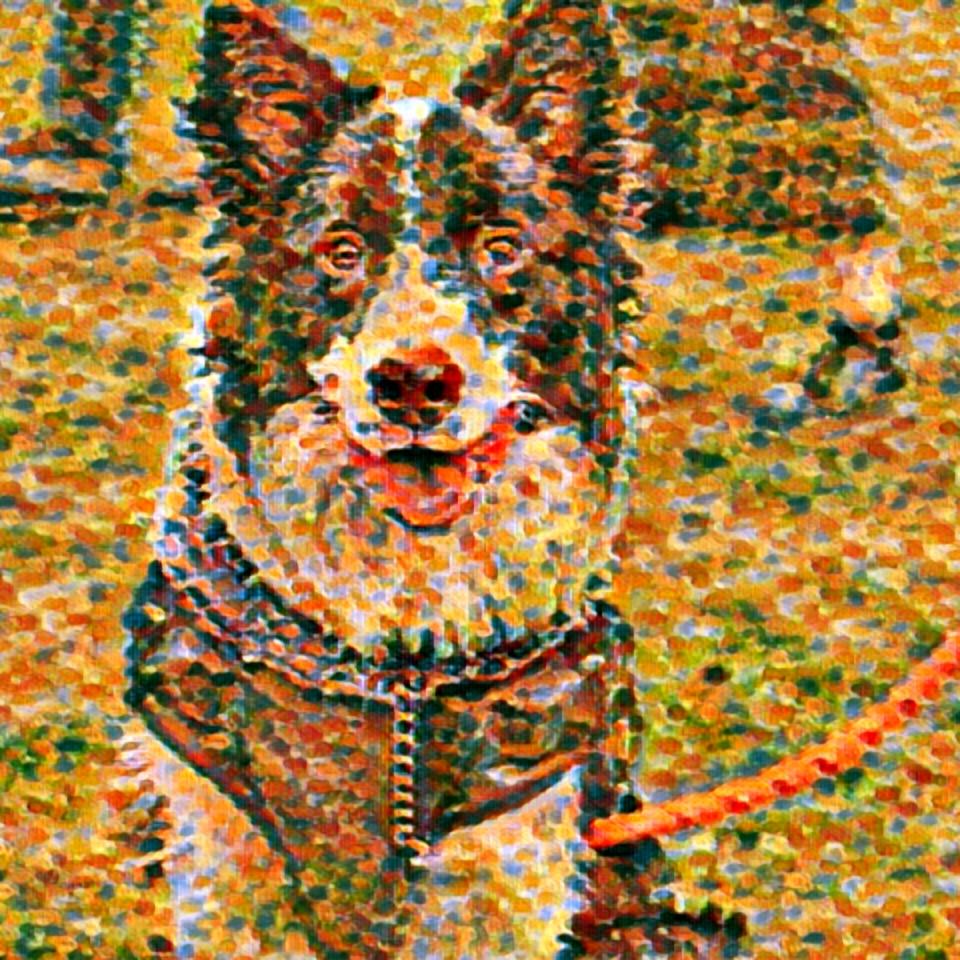} 
        & \hspace*{-4mm} \includegraphics[width=.08\linewidth,height=!]{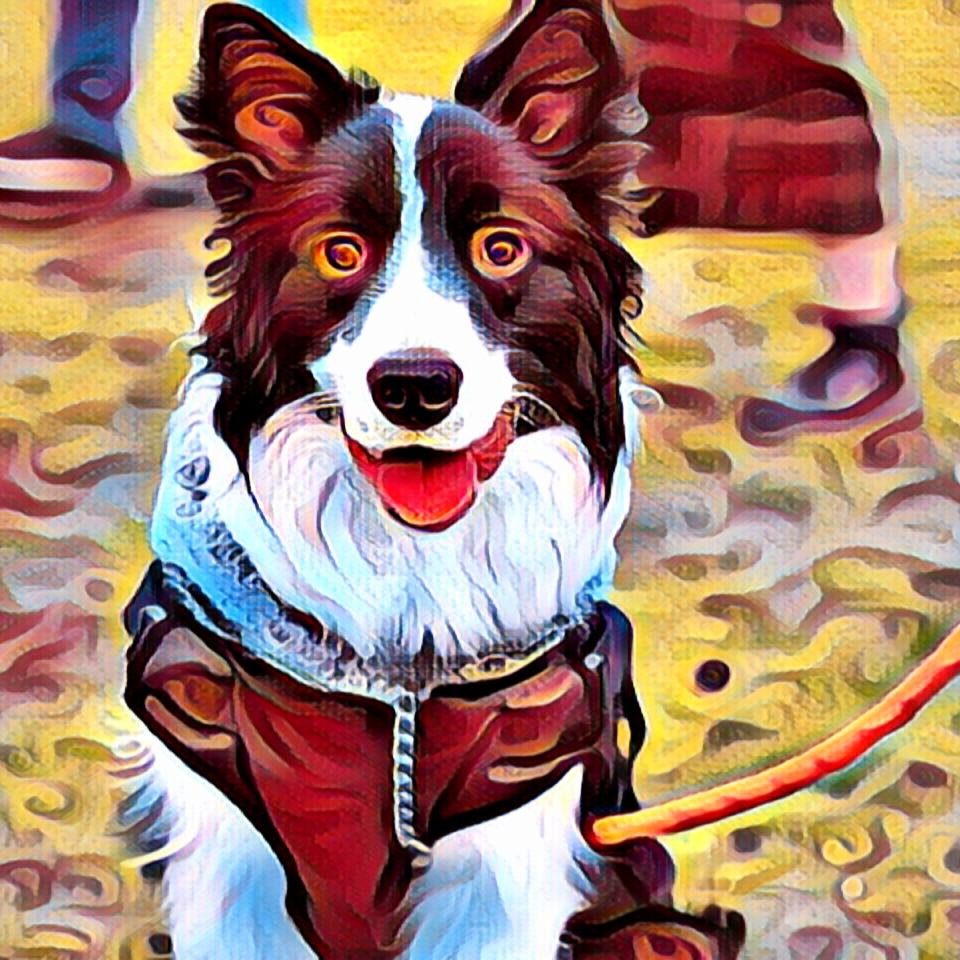} 
        & \hspace*{-4mm} \includegraphics[width=.08\linewidth,height=!]{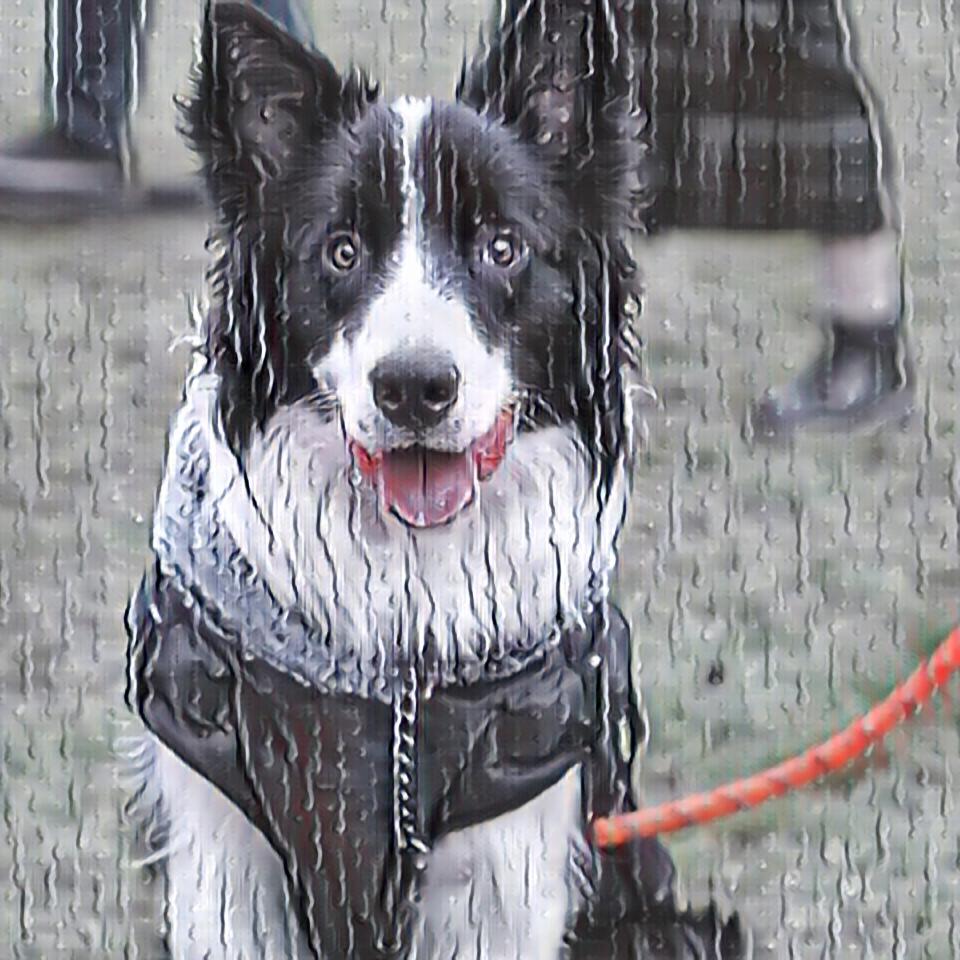} 
        & \hspace*{-4mm} \includegraphics[width=.08\linewidth,height=!]{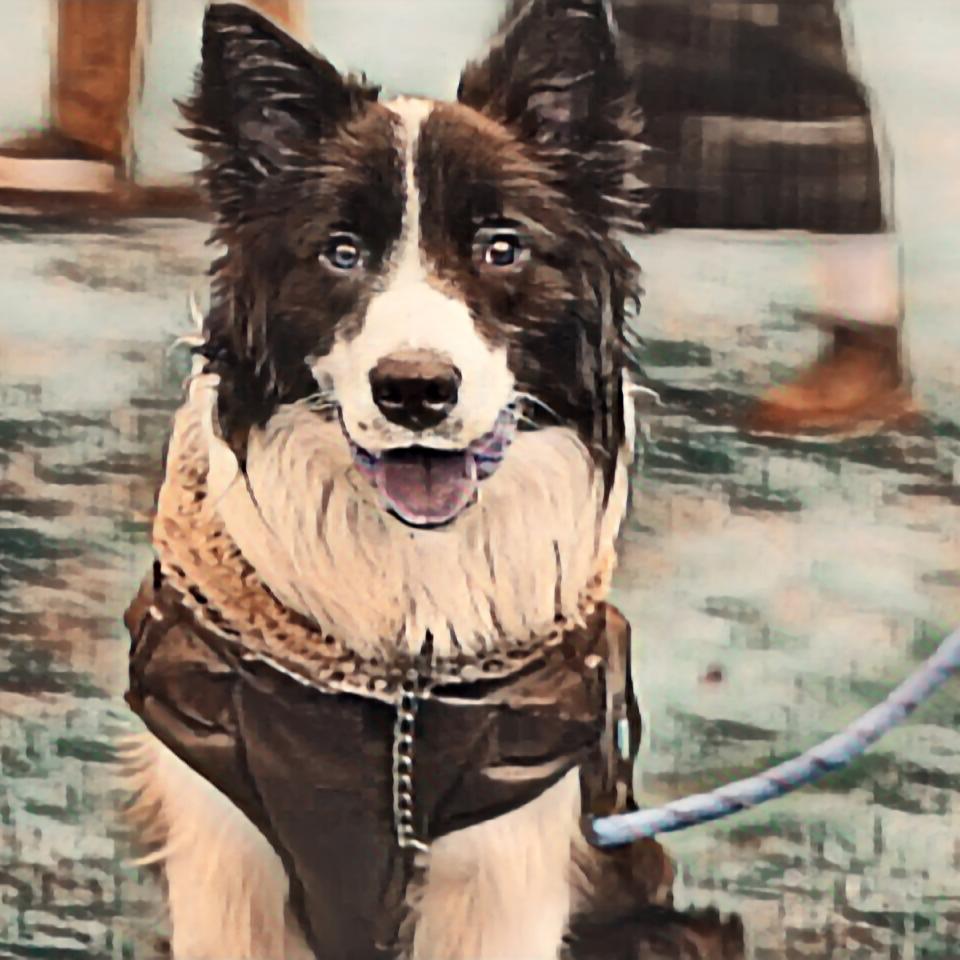} 
        & \hspace*{-4mm} \includegraphics[width=.08\linewidth,height=!]{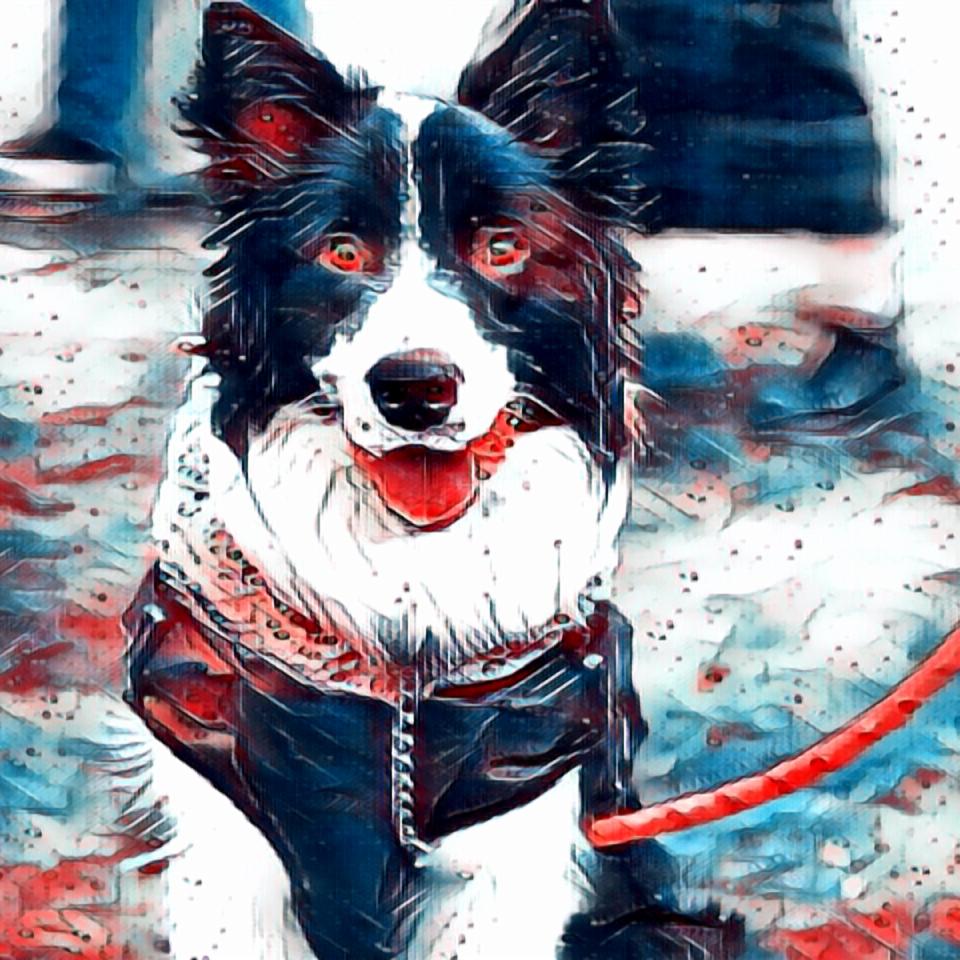} 
        & \hspace*{-4mm} \includegraphics[width=.08\linewidth,height=!]{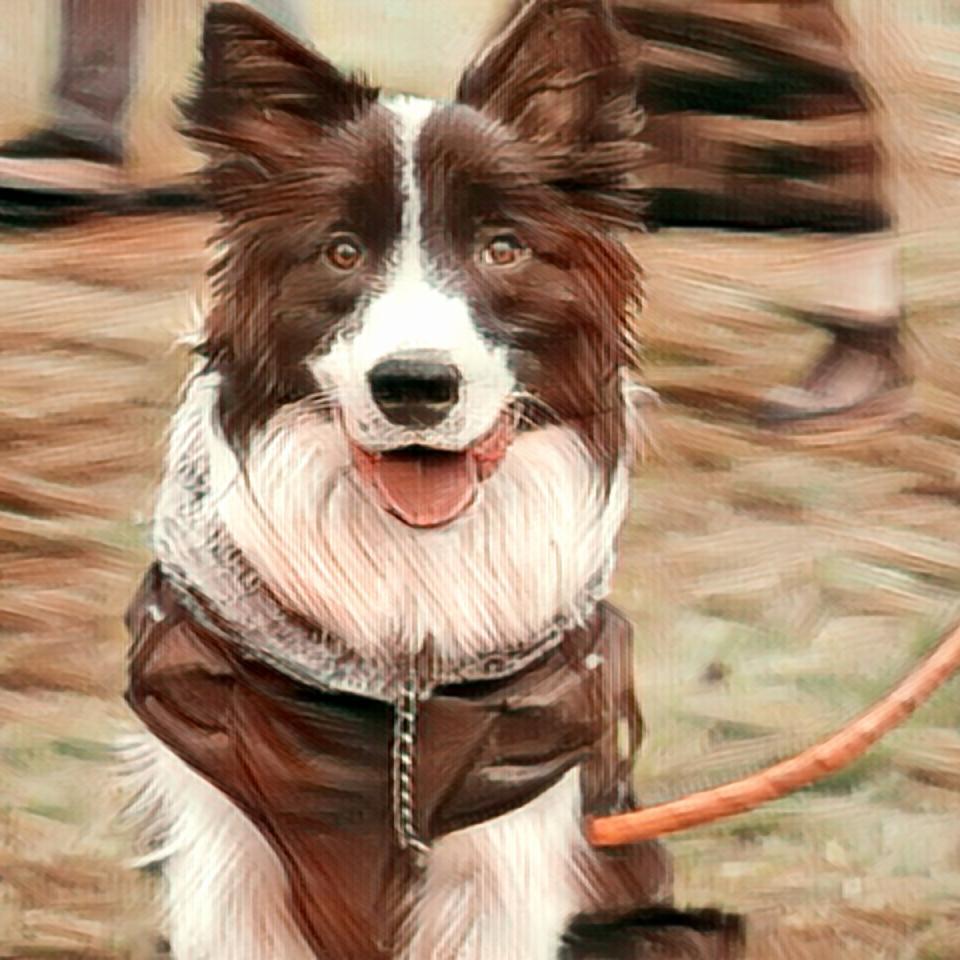} 
        & \hspace*{-4mm} \includegraphics[width=.08\linewidth,height=!]{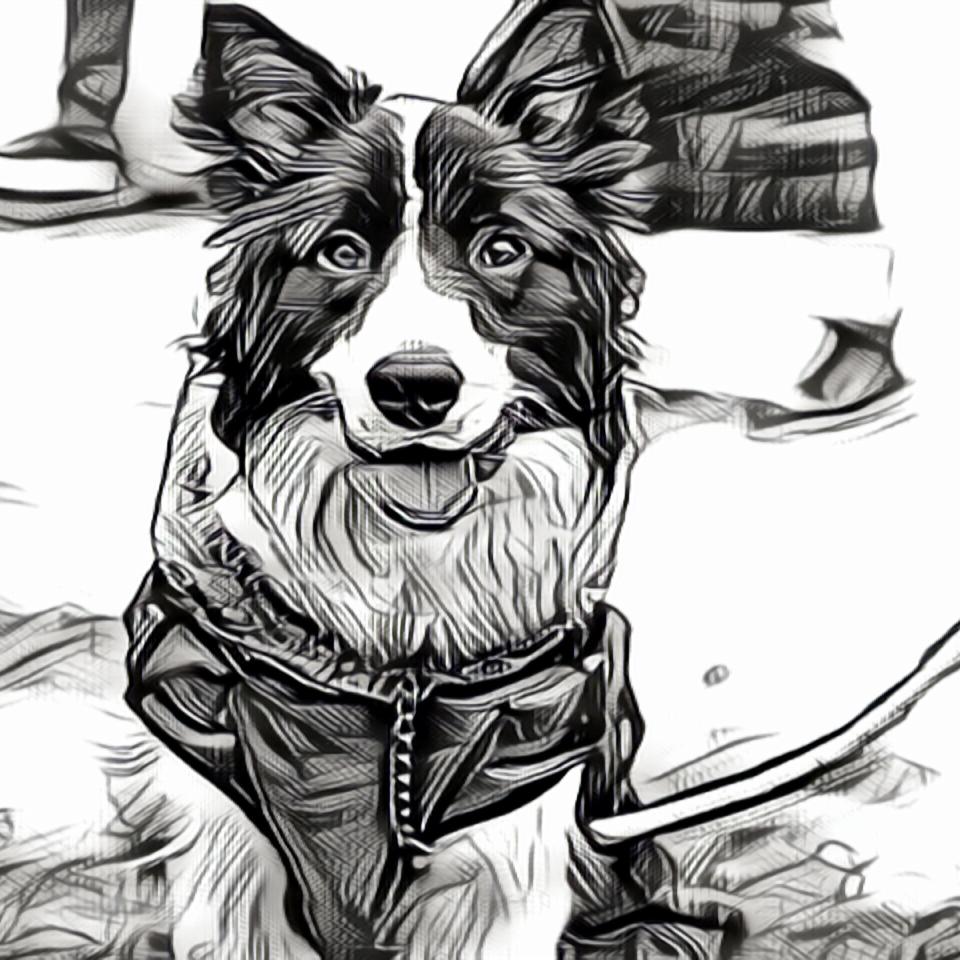} 
        & \hspace*{-4mm} \includegraphics[width=.08\linewidth,height=!]{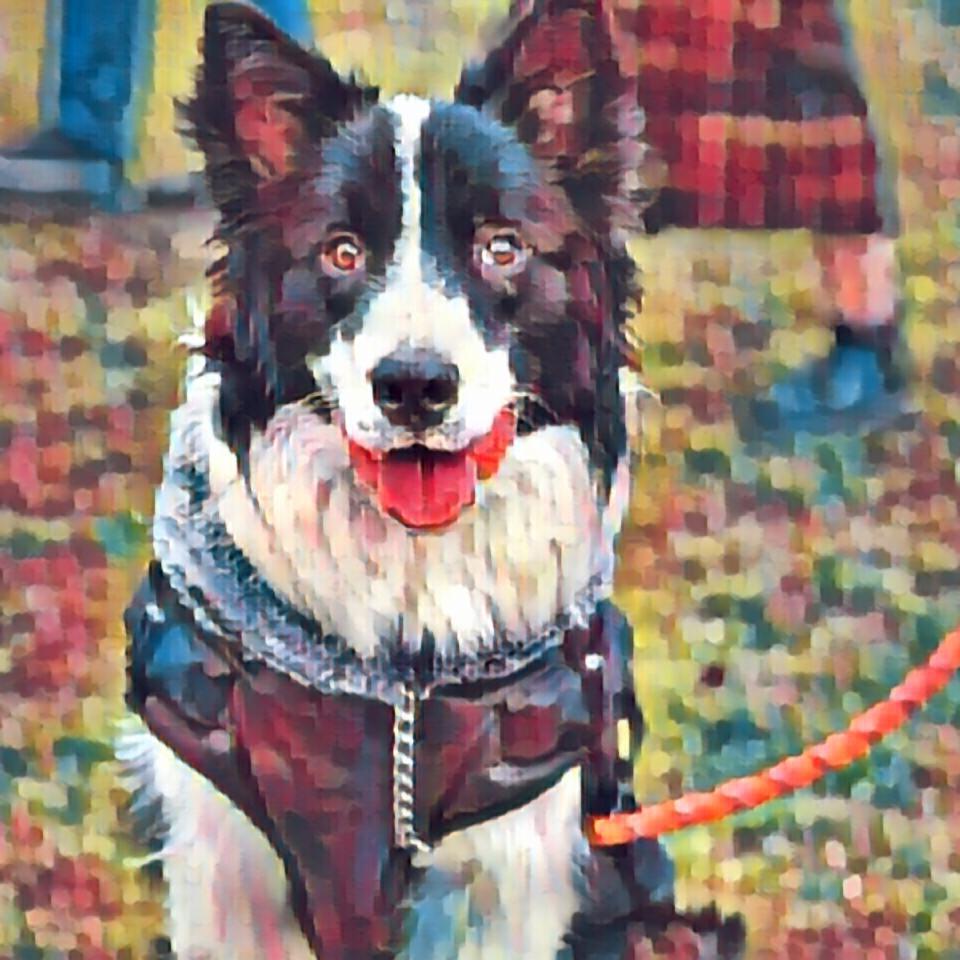} 
        & \hspace*{-4mm} \includegraphics[width=.08\linewidth,height=!]{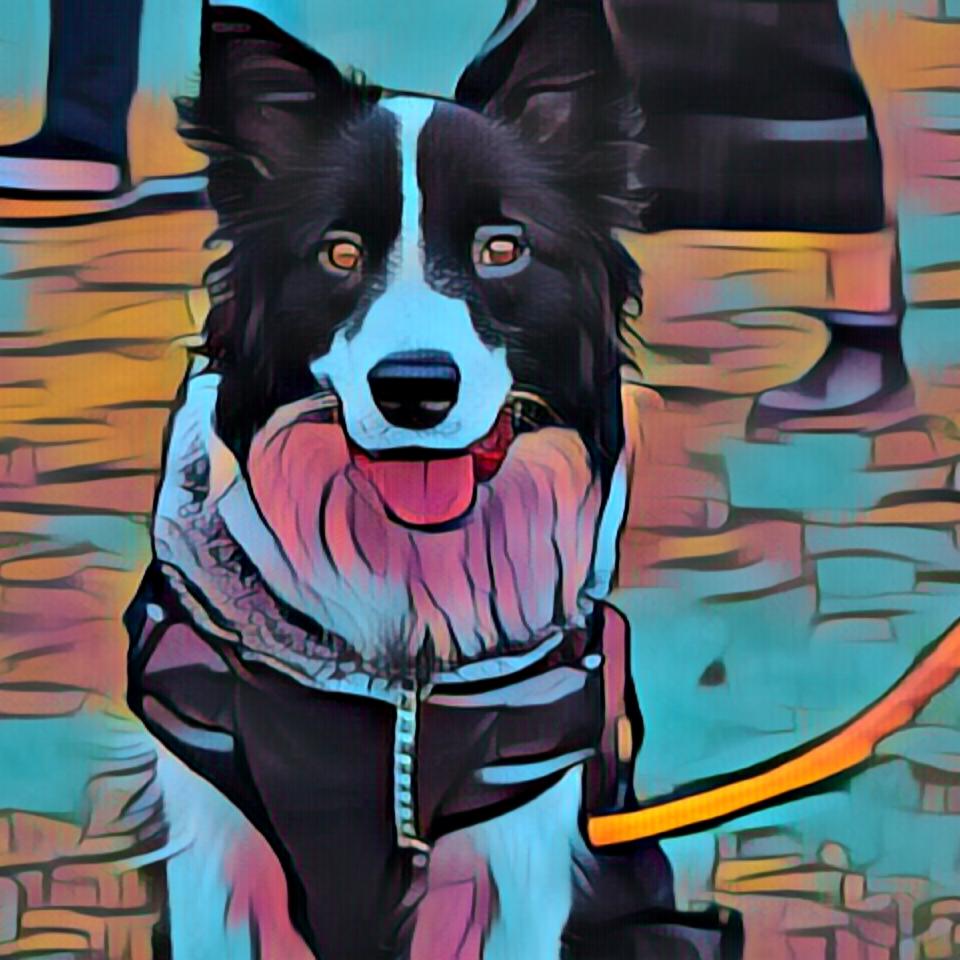} 
        \\
        \hspace*{-4.1mm} (41)
        & \hspace*{-4mm} (42)
        & \hspace*{-4mm} (43)
        & \hspace*{-4mm} (44)
        & \hspace*{-4mm} (45)
        & \hspace*{-4mm} (46)
        & \hspace*{-4mm} (47)
        & \hspace*{-4mm} (48)
        & \hspace*{-4mm} (49)
        & \hspace*{-4mm} (50)
        \\
        \hspace*{-4.1mm} \includegraphics[width=.08\linewidth,height=!]{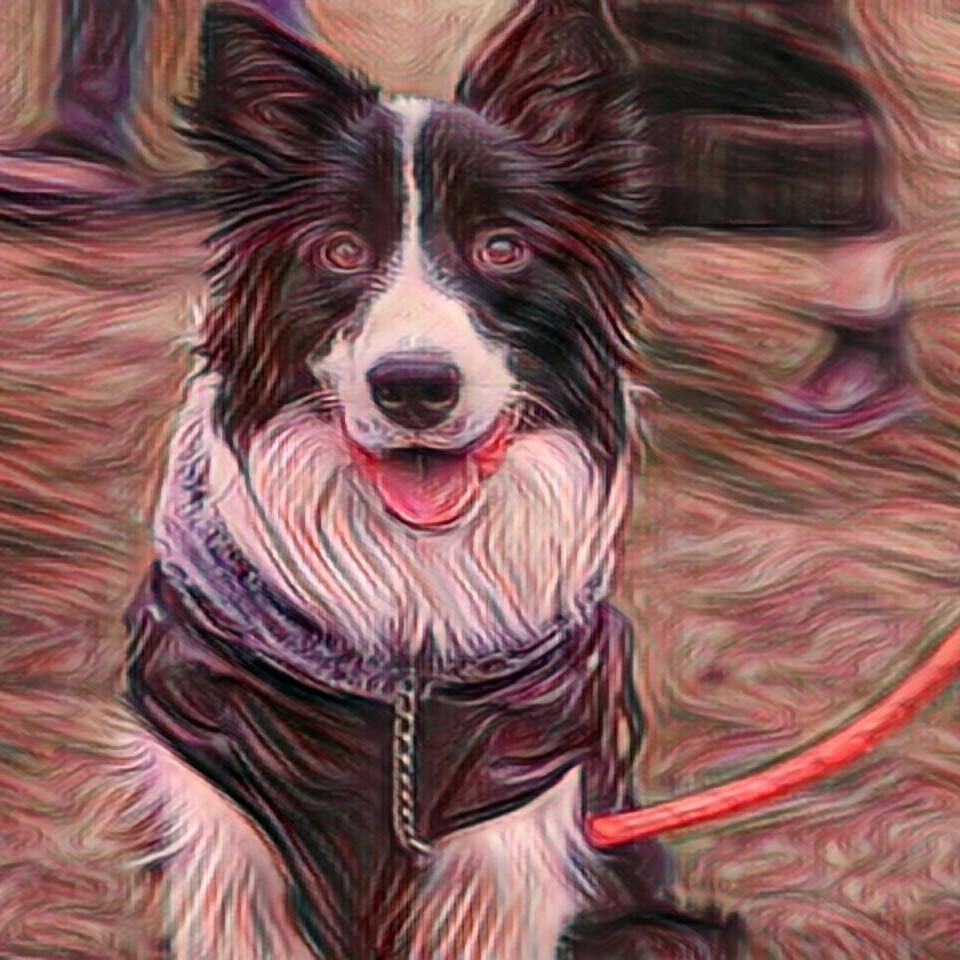} & \hspace*{-4mm} \includegraphics[width=.08\linewidth,height=!]{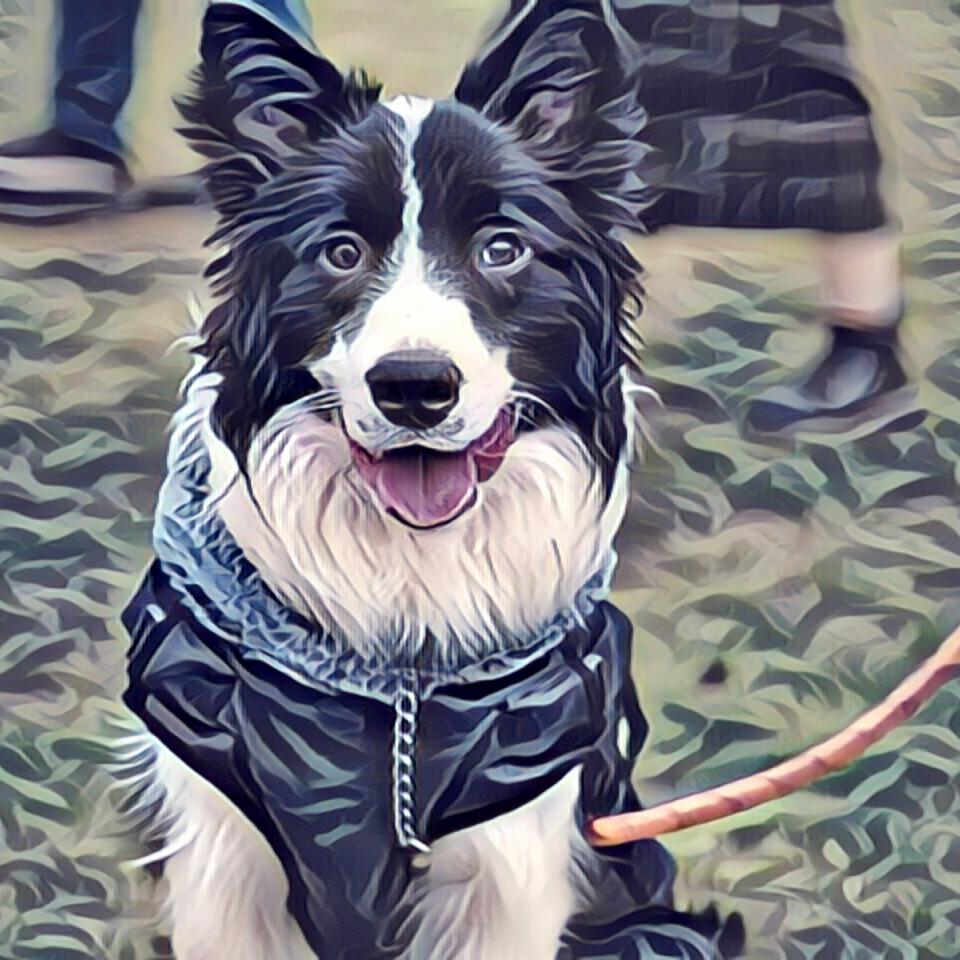} 
        & \hspace*{-4mm} \includegraphics[width=.08\linewidth,height=!]{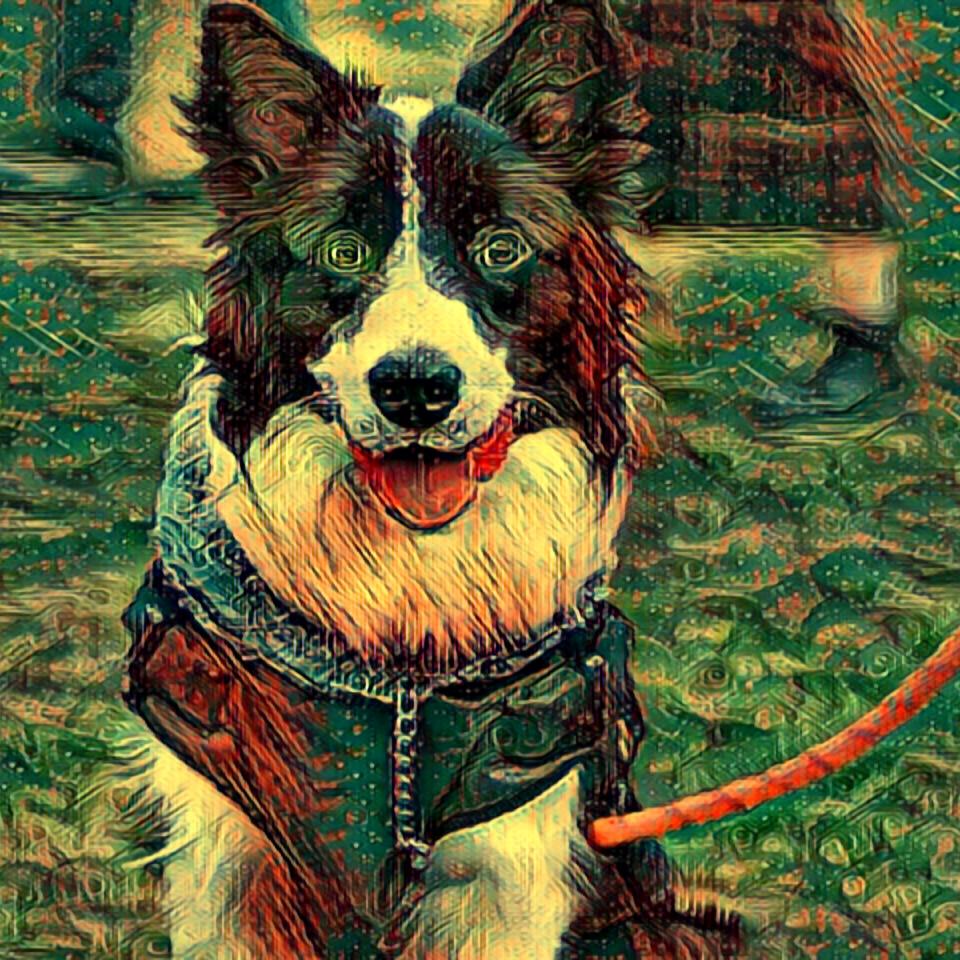} 
        & \hspace*{-4mm} \includegraphics[width=.08\linewidth,height=!]{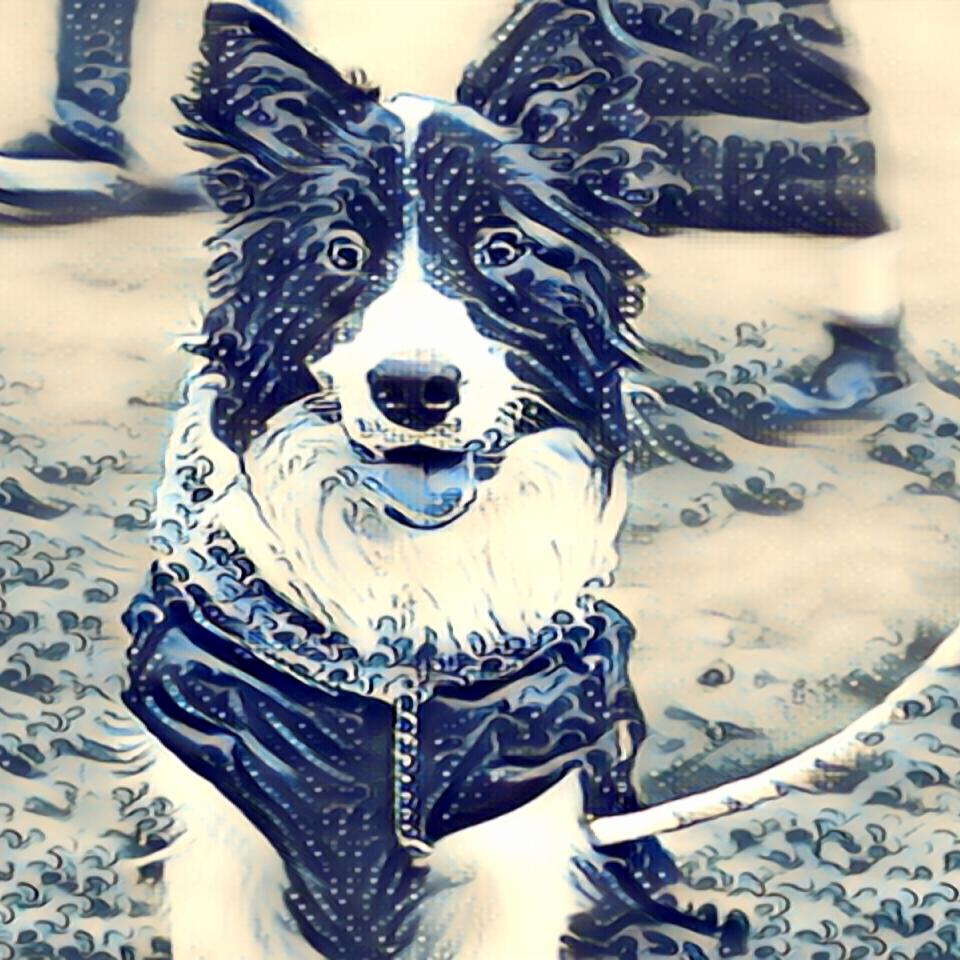} 
        & \hspace*{-4mm} \includegraphics[width=.08\linewidth,height=!]{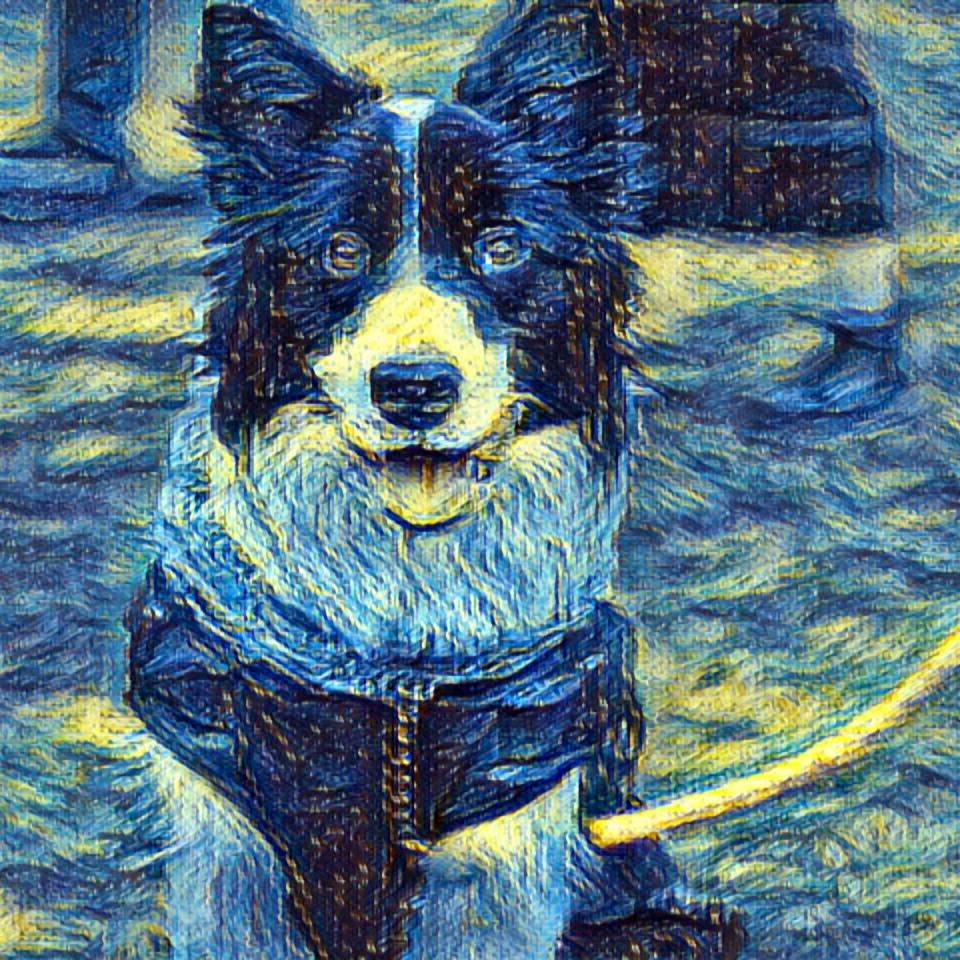} 
        & \hspace*{-4mm} \includegraphics[width=.08\linewidth,height=!]{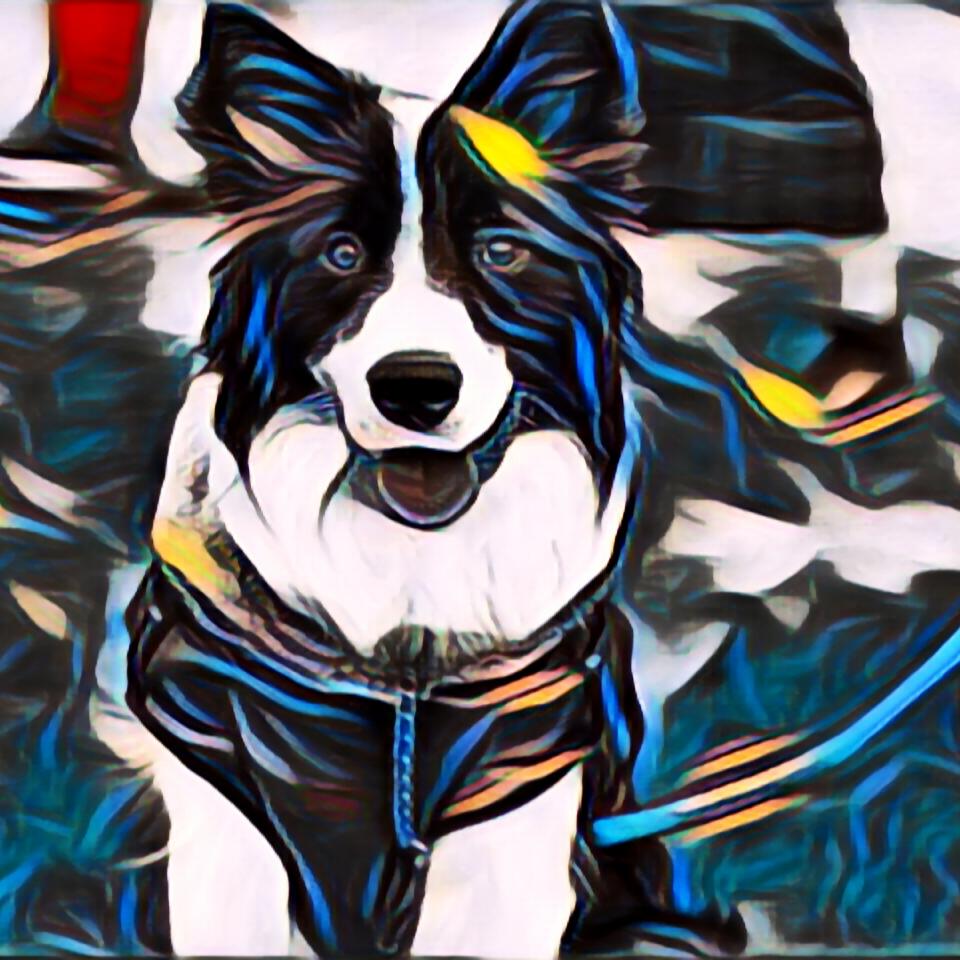} 
        & \hspace*{-4mm} \includegraphics[width=.08\linewidth,height=!]{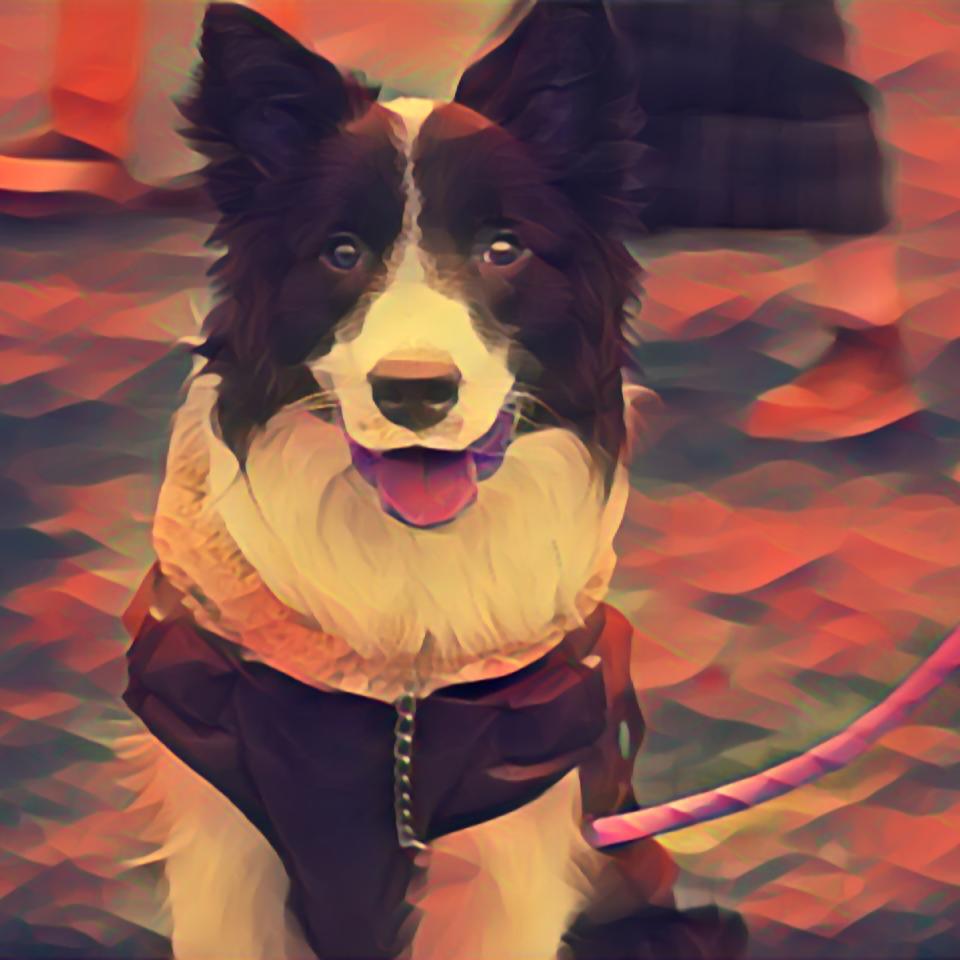} 
        & \hspace*{-4mm} \includegraphics[width=.08\linewidth,height=!]{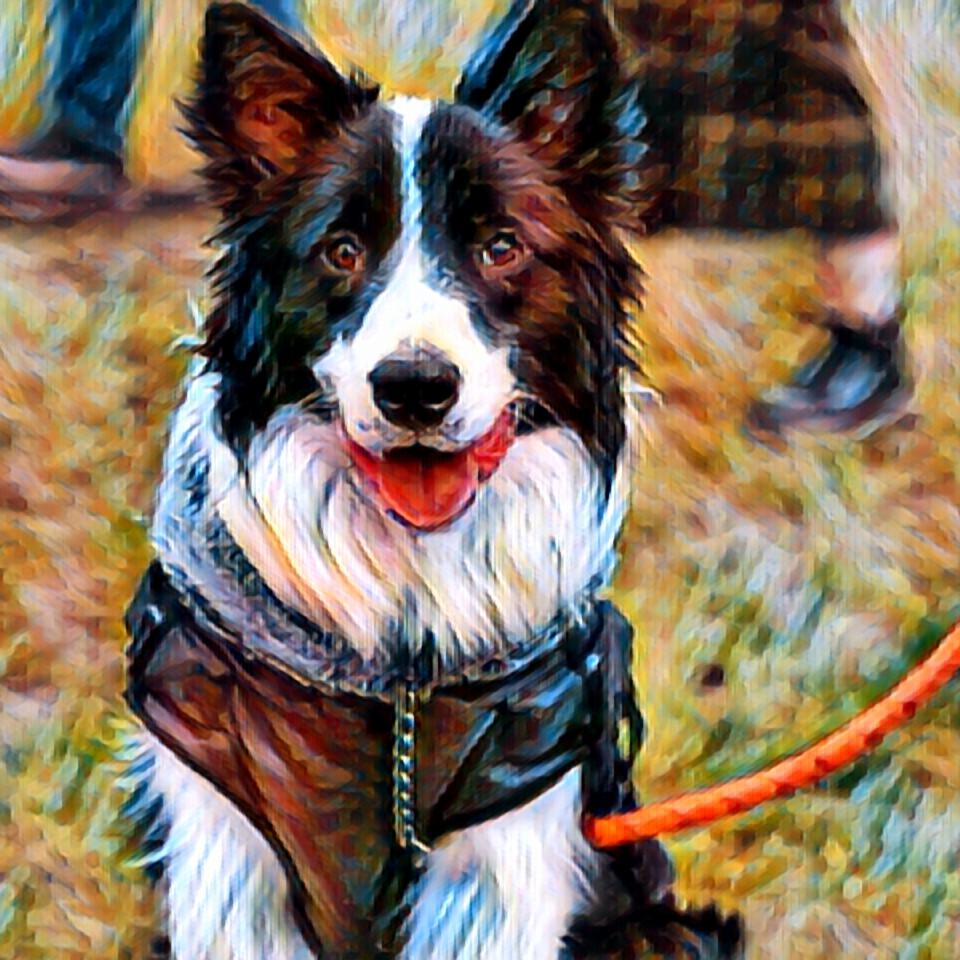} 
        & \hspace*{-4mm} \includegraphics[width=.08\linewidth,height=!]{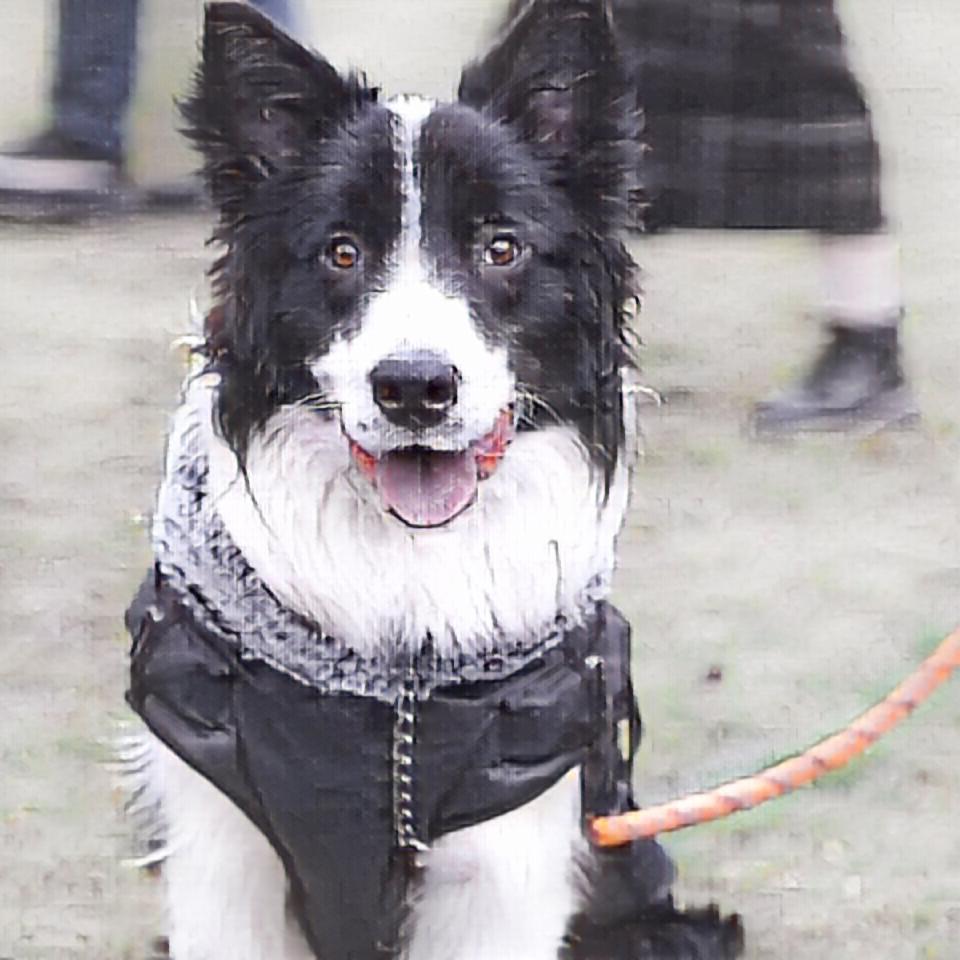} 
        & \hspace*{-4mm} \includegraphics[width=.08\linewidth,height=!]{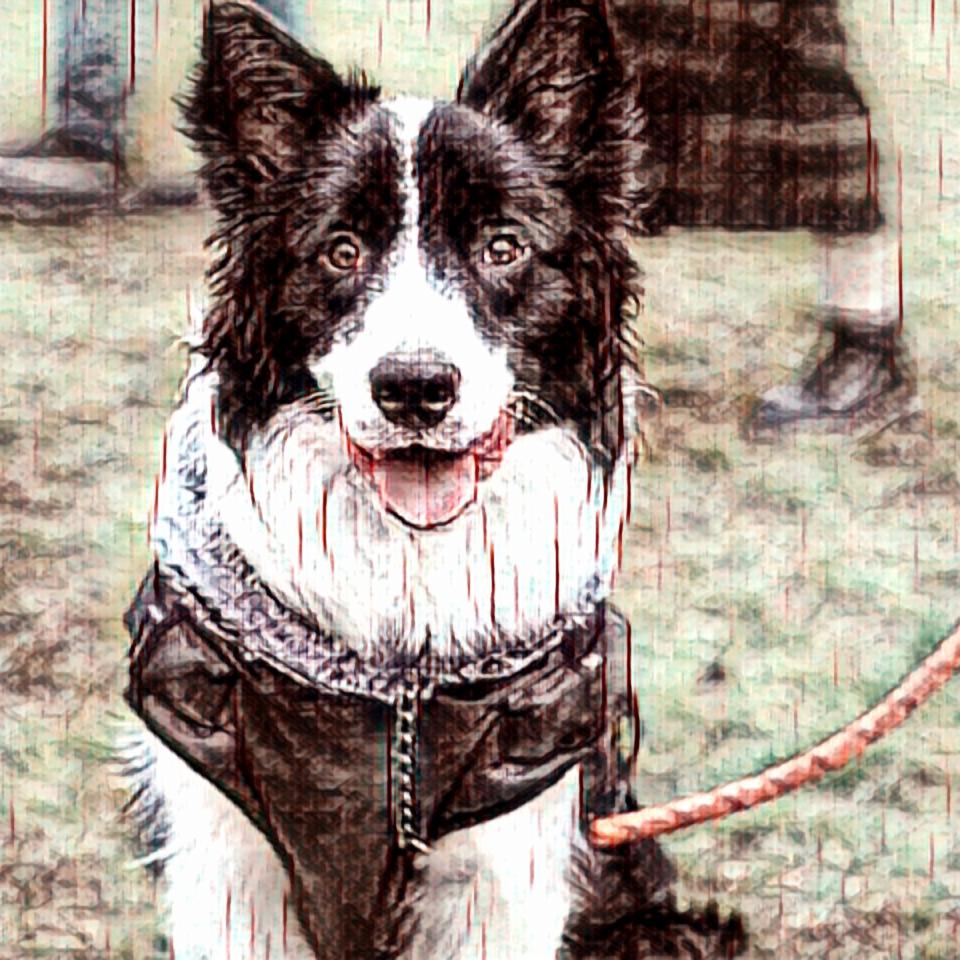} 
        \\
        \hspace*{-4.1mm} (51)
        & \hspace*{-4mm} (52)
        & \hspace*{-4mm} (53)
        & \hspace*{-4mm} (54)
        & \hspace*{-4mm} (55)
        & \hspace*{-4mm} (56)
        & \hspace*{-4mm} (57)
        & \hspace*{-4mm} (58)
        & \hspace*{-4mm} (59)
        & \hspace*{-4mm} (60)
        \\

    \end{tabular}
    \caption{An illustration of the images in each style in {\ours} used in, which are stylized from the same seed image from the `Dogs' object class. The seed image is presented in Fig.\,\ref{fig: dataset_object_details} (6). Images are cropped and down-scaled for illustration purpose. The name of the styles are: (1) Abstractionism; (2) Artist Sketch; (3) Blossom Season; (4) Blue Blooming; (5) Bricks; (6) Byzantine; (7) Cartoon; (8) Cold Warm; (9) Color Fantasy; (10) Comic Etch; (11) Crayon; (12) Crypto Punks; (13) Cubism; (14) Dadaism; (15) Dapple; (16) Defoliation; (17) Dreamweave; (18) Early Autumn; (19) Expressionism; (20) Fauvism; (21) Foliage Patchwork; (22) French; (23) Glowing Sunset; (24) Gorgeous Love; (25) Greenfield; (26) Impasto; (27) Impressionism; (28) Ink Art; (29) Joy; (30) Liquid Dreams; (31) Palette Knife; (32) Magic Cube; (33) Meta Physics; (34) Meteor Shower; (35) Monet; (36) Mosaic; (37) Neon Lines; (38) On Fire; (39) Pastel; (40) Pencil Drawing; (41) Picasso; (42) Pointillism; (43) Pop Art; (44) Rainwash; (45) Realistic Watercolor; (46) Red Blue Ink; (47) Rust; (48) Sketch; (49) Sponge Dabbed; (50) Structuralism; (51) Superstring; (52) Surrealism; (53) Techno; (54) Ukiyoe; (55) Van Gogh; (56) Vibrant Flow; (57) Warm Love; (58) Warm Smear; (59) Watercolor; (60) Winter.}
    \label{fig: dataset_style_details}
\end{figure}

\begin{figure}
    \centering
    \begin{tabular}{cccccccccc}
    \hspace*{-4.1mm} \includegraphics[width=.08\linewidth,height=!]{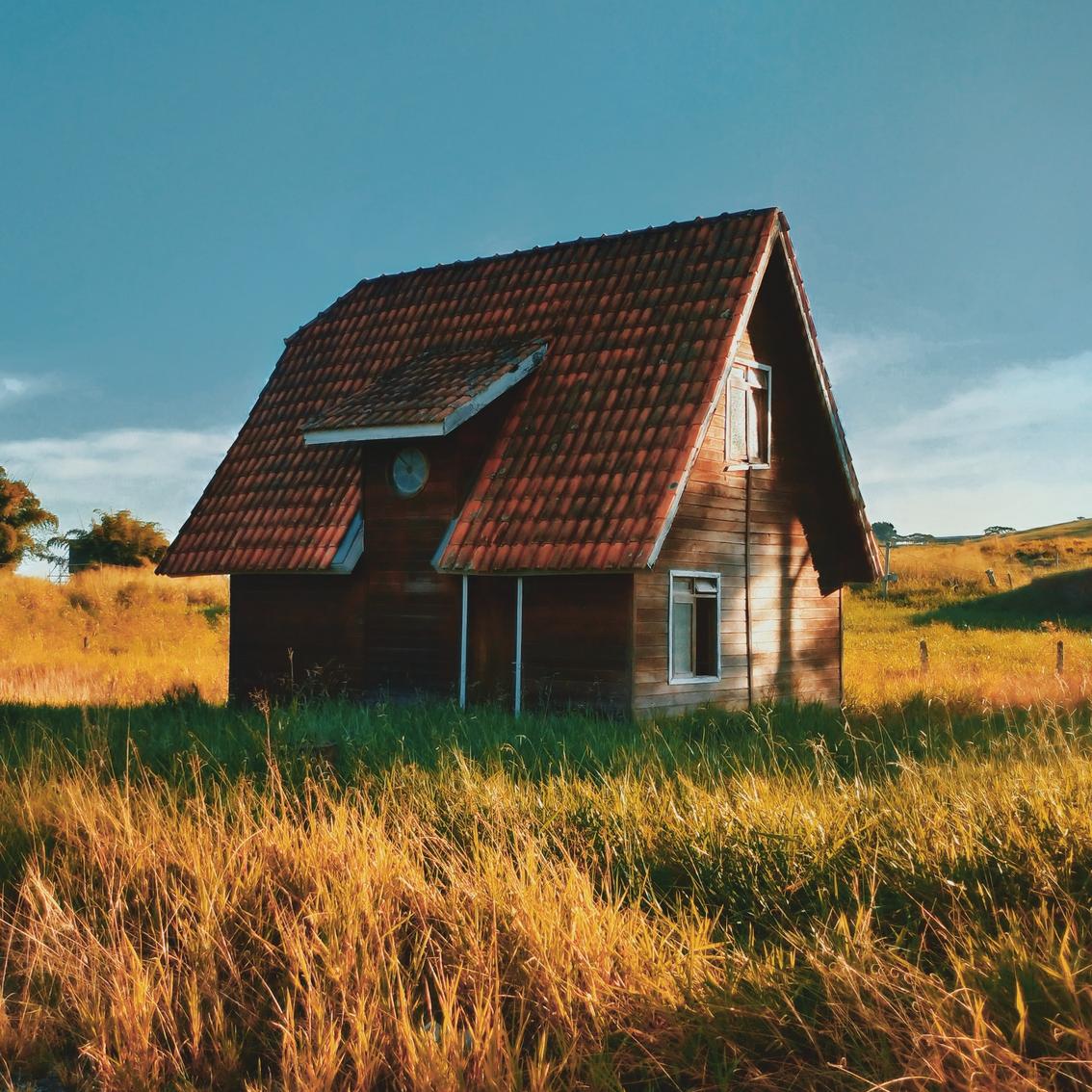} & \hspace*{-4mm} \includegraphics[width=.08\linewidth,height=!]{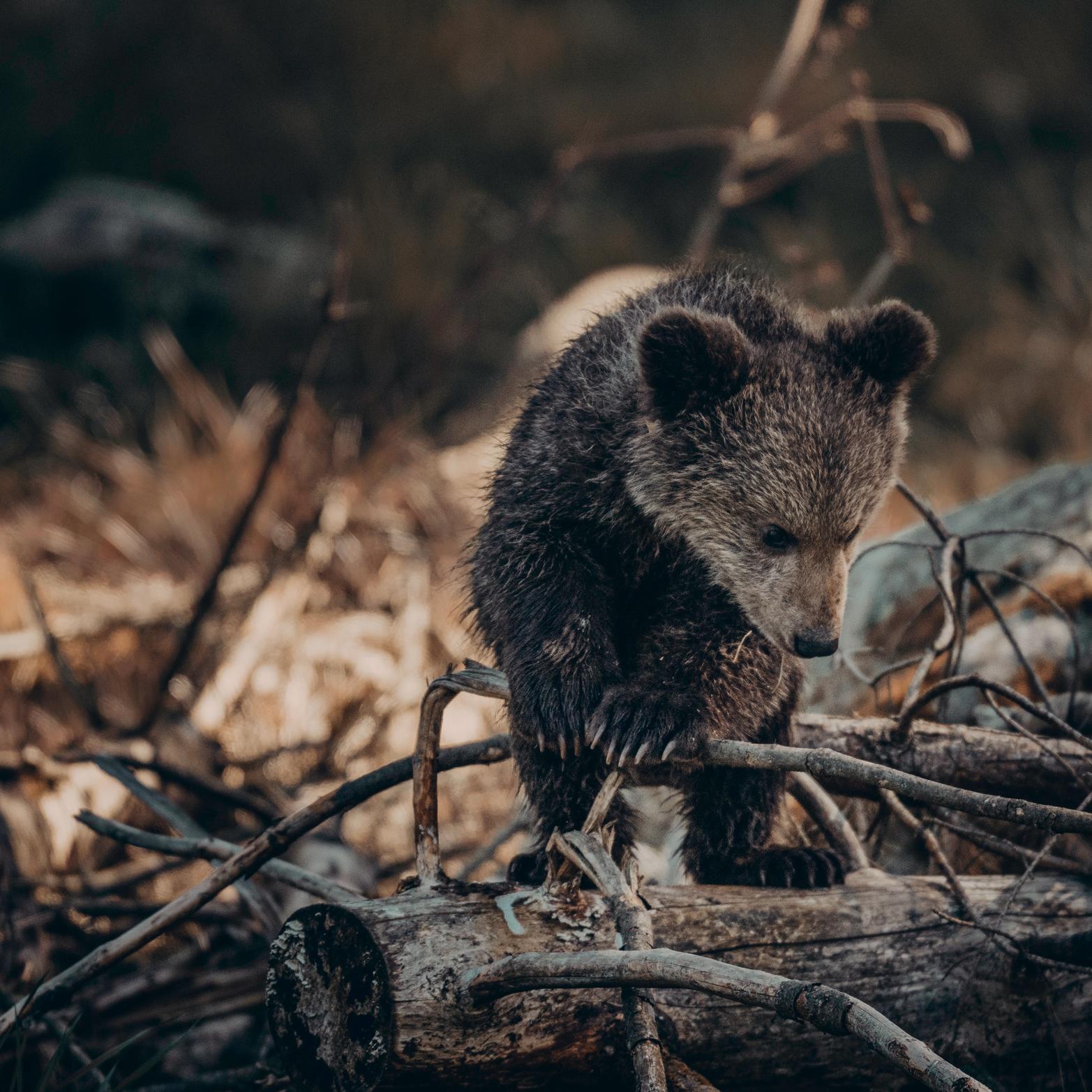} 
& \hspace*{-4mm} \includegraphics[width=.08\linewidth,height=!]{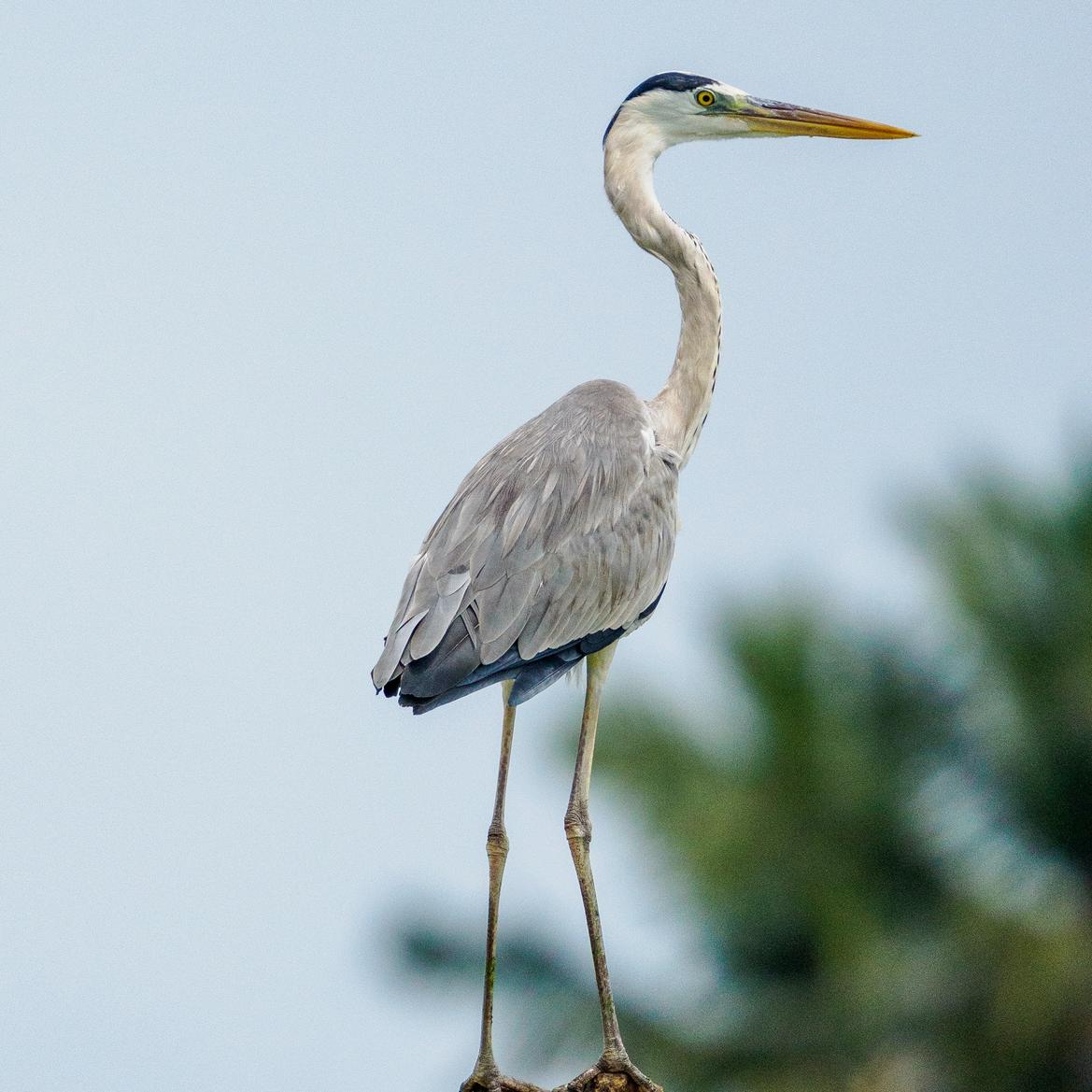} 
& \hspace*{-4mm} \includegraphics[width=.08\linewidth,height=!]{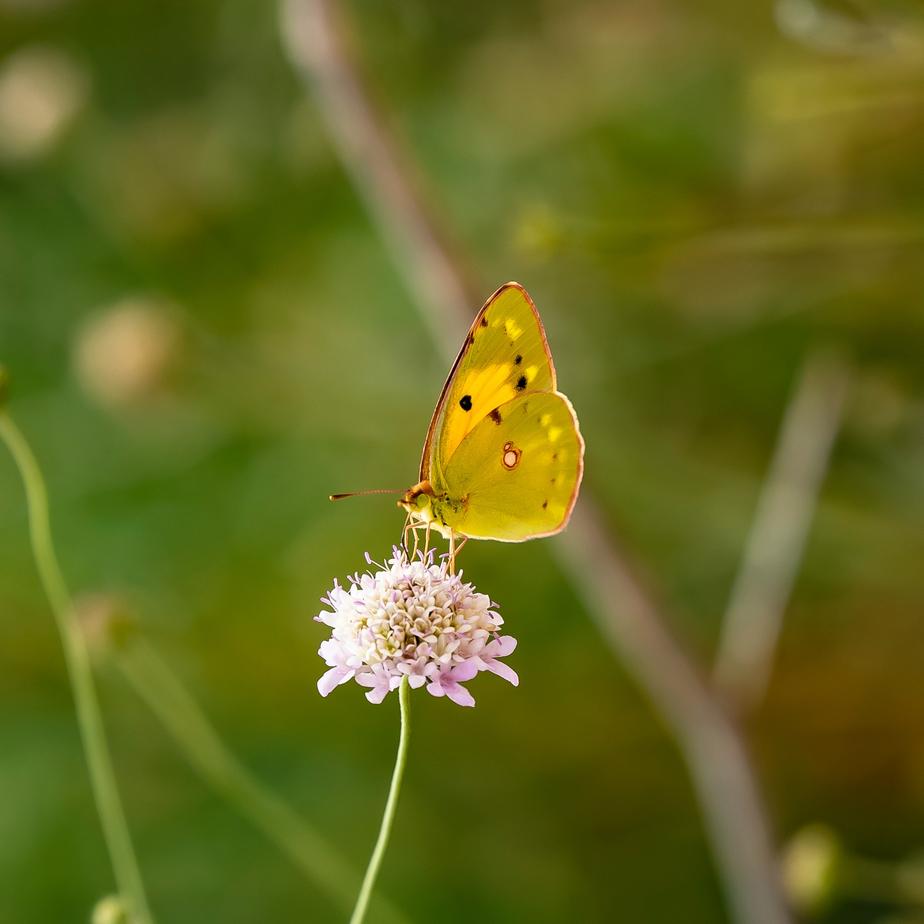} 
& \hspace*{-4mm} \includegraphics[width=.08\linewidth,height=!]{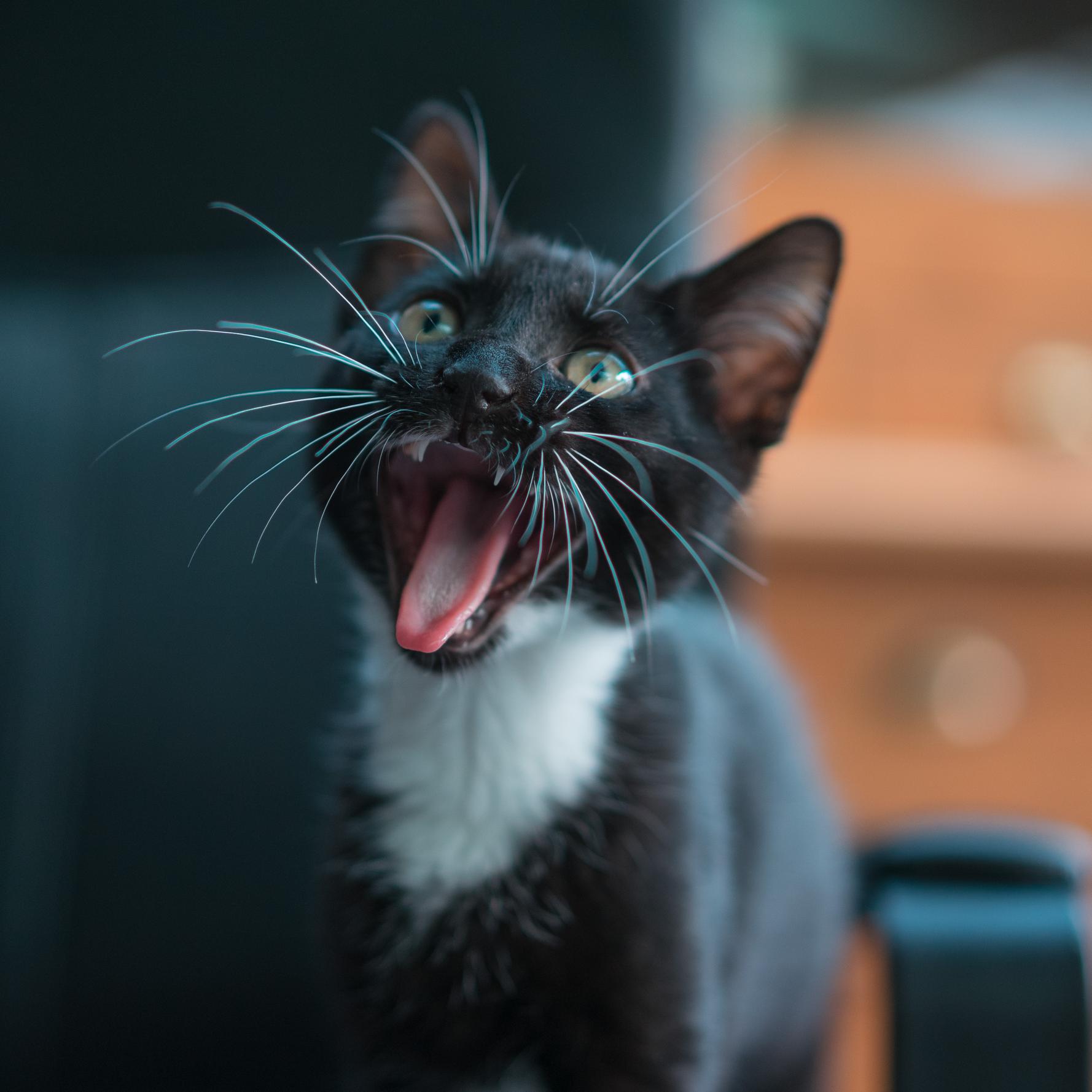} 
& \hspace*{-4mm} \includegraphics[width=.08\linewidth,height=!]{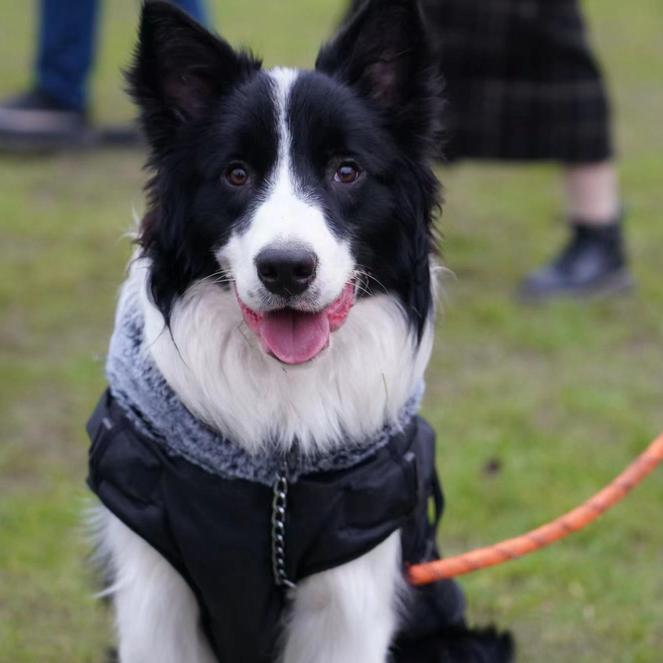} 
& \hspace*{-4mm} \includegraphics[width=.08\linewidth,height=!]{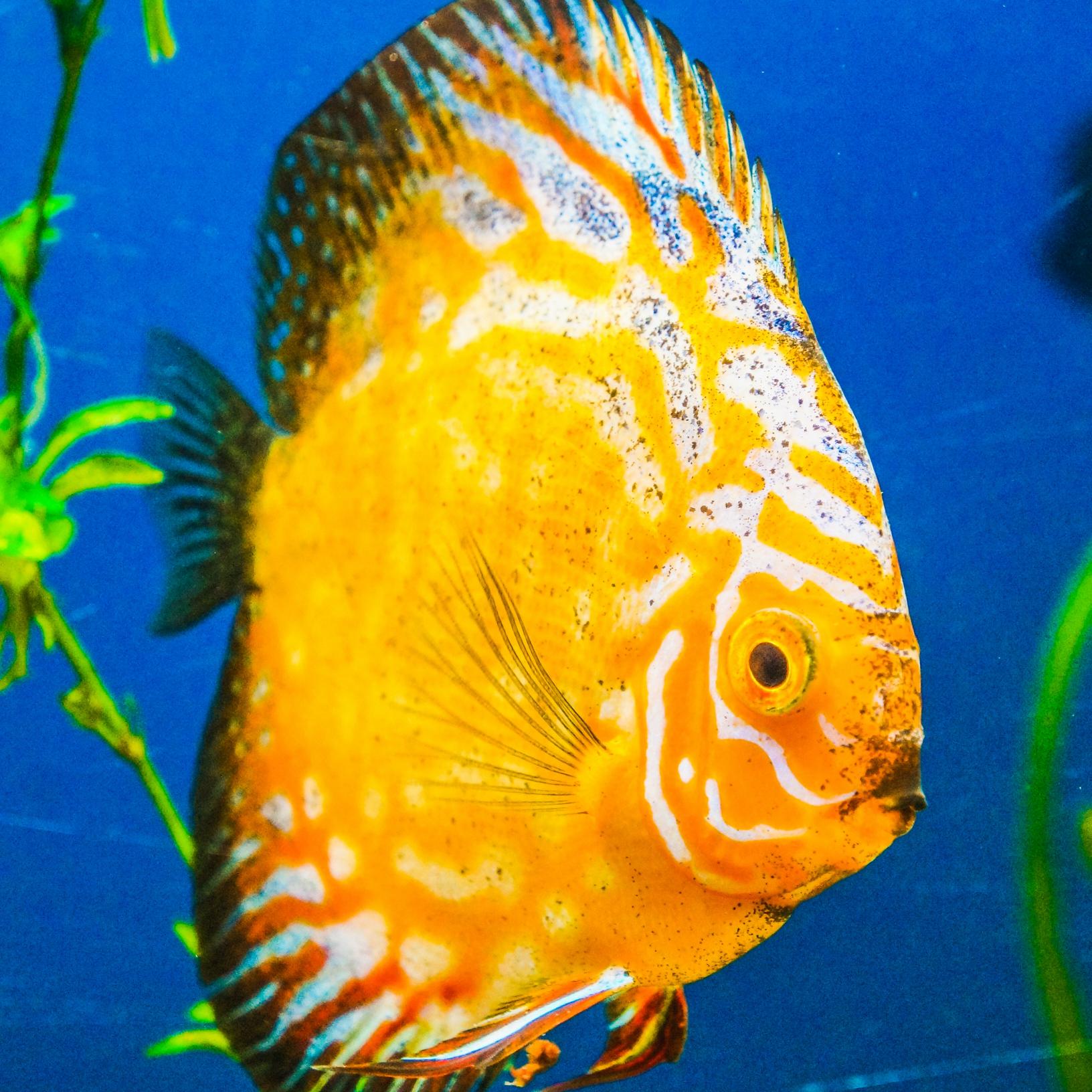} 
& \hspace*{-4mm} \includegraphics[width=.08\linewidth,height=!]{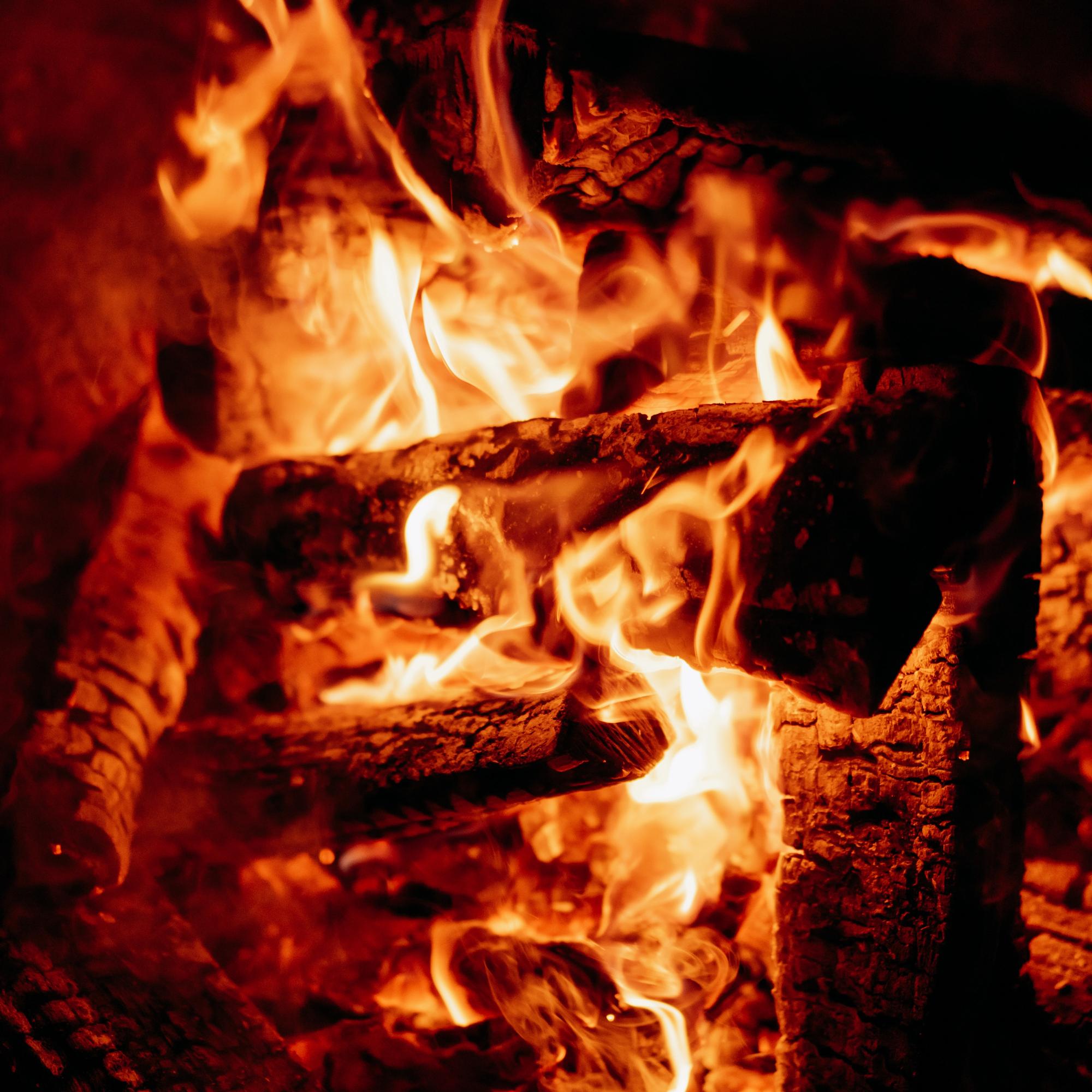} 
& \hspace*{-4mm} \includegraphics[width=.08\linewidth,height=!]{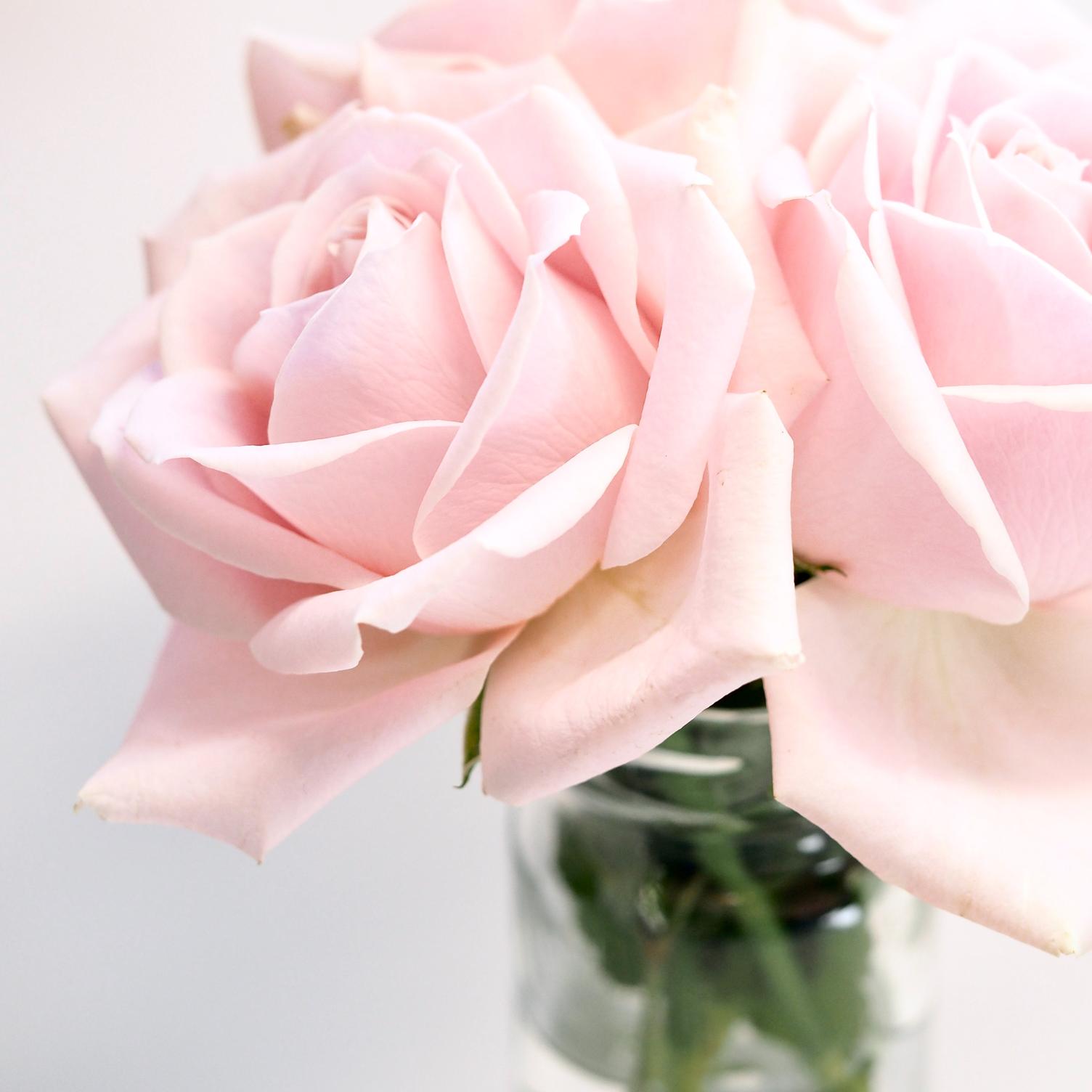} 
& \hspace*{-4mm} \includegraphics[width=.08\linewidth,height=!]{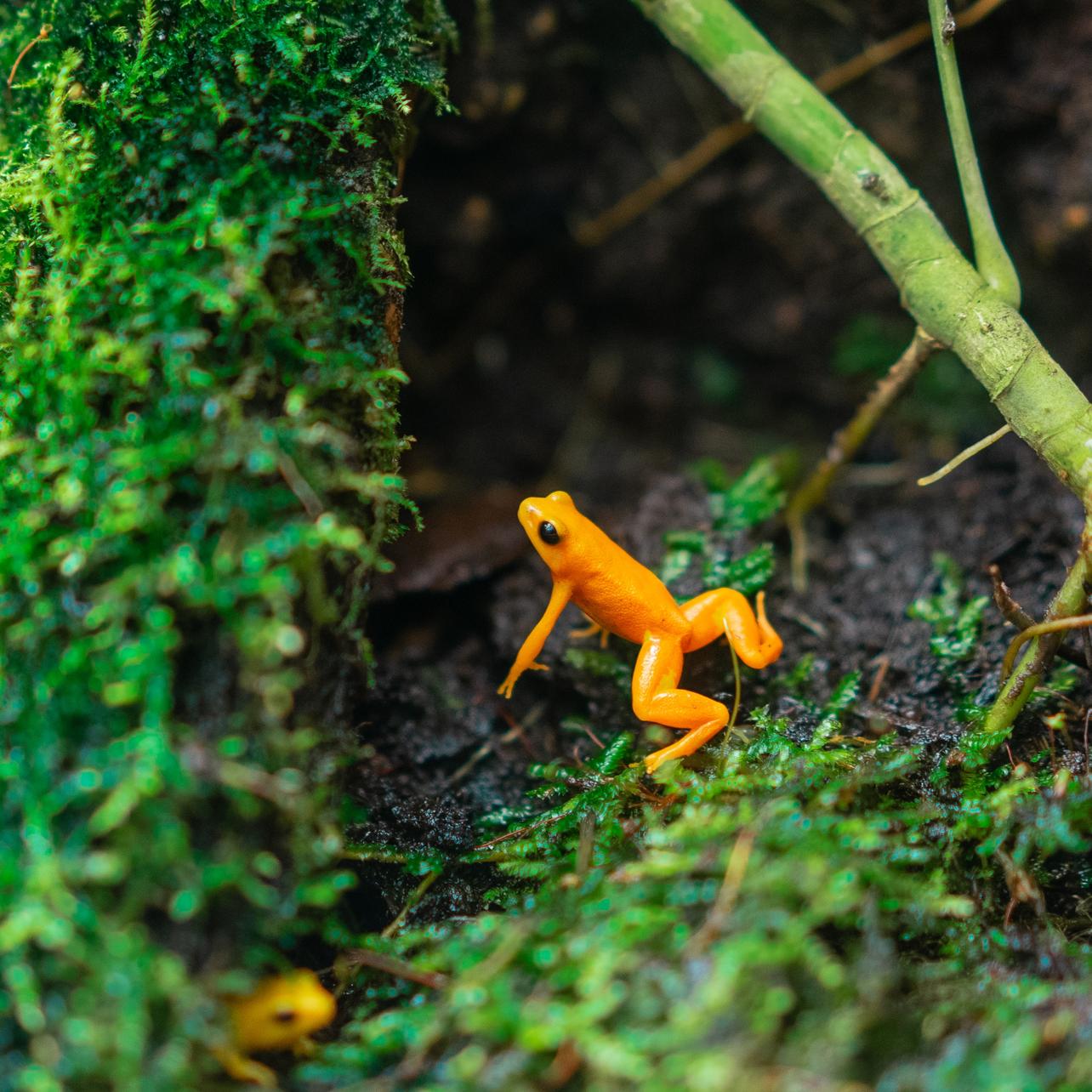} 
\\
\hspace*{-4.1mm} (1)
& \hspace*{-4mm} (2)
& \hspace*{-4mm} (3)
& \hspace*{-4mm} (4)
& \hspace*{-4mm} (5)
& \hspace*{-4mm} (6)
& \hspace*{-4mm} (7)
& \hspace*{-4mm} (8)
& \hspace*{-4mm} (9)
& \hspace*{-4mm} (10)
\\
\hspace*{-4.1mm} \includegraphics[width=.08\linewidth,height=!]{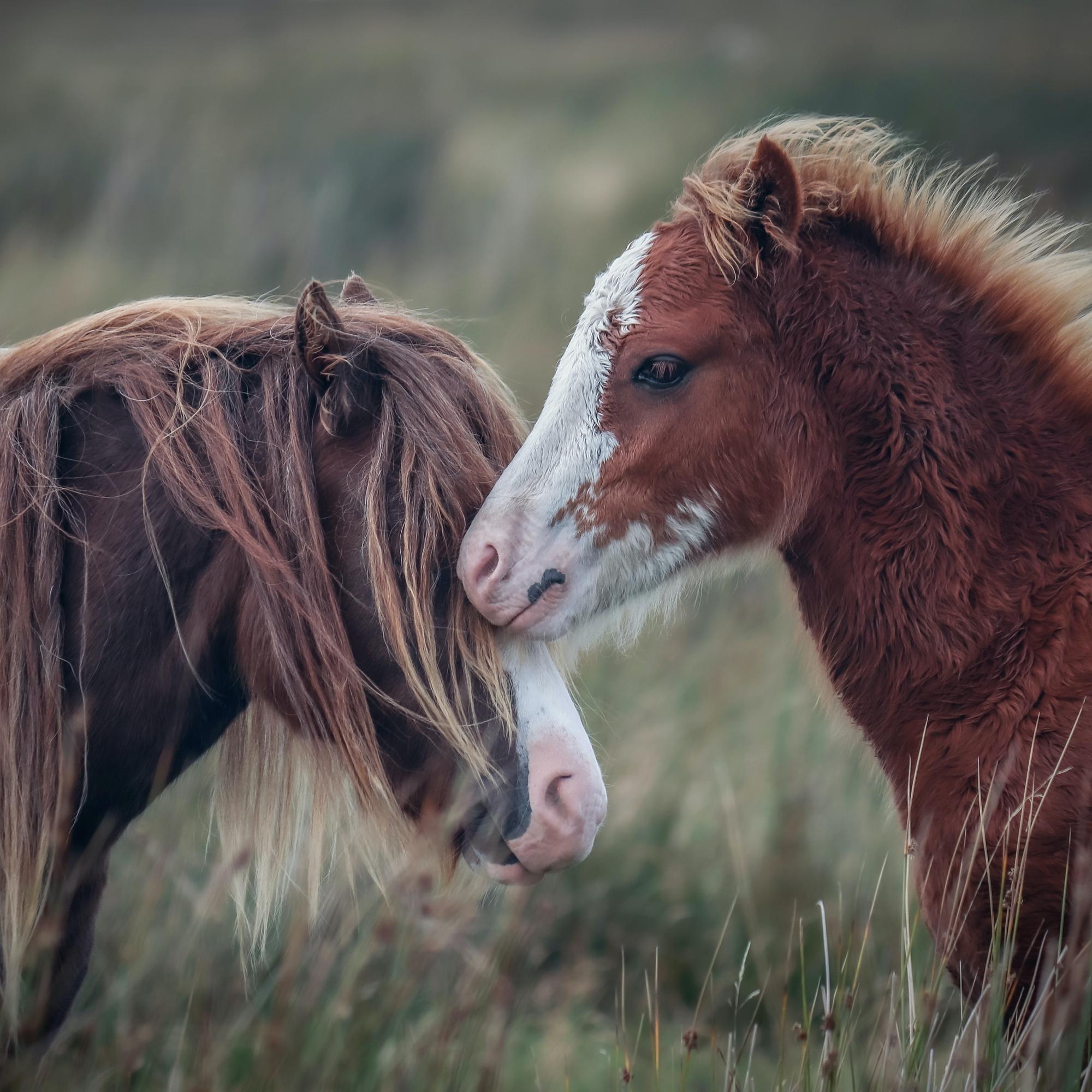} & \hspace*{-4mm} \includegraphics[width=.08\linewidth,height=!]{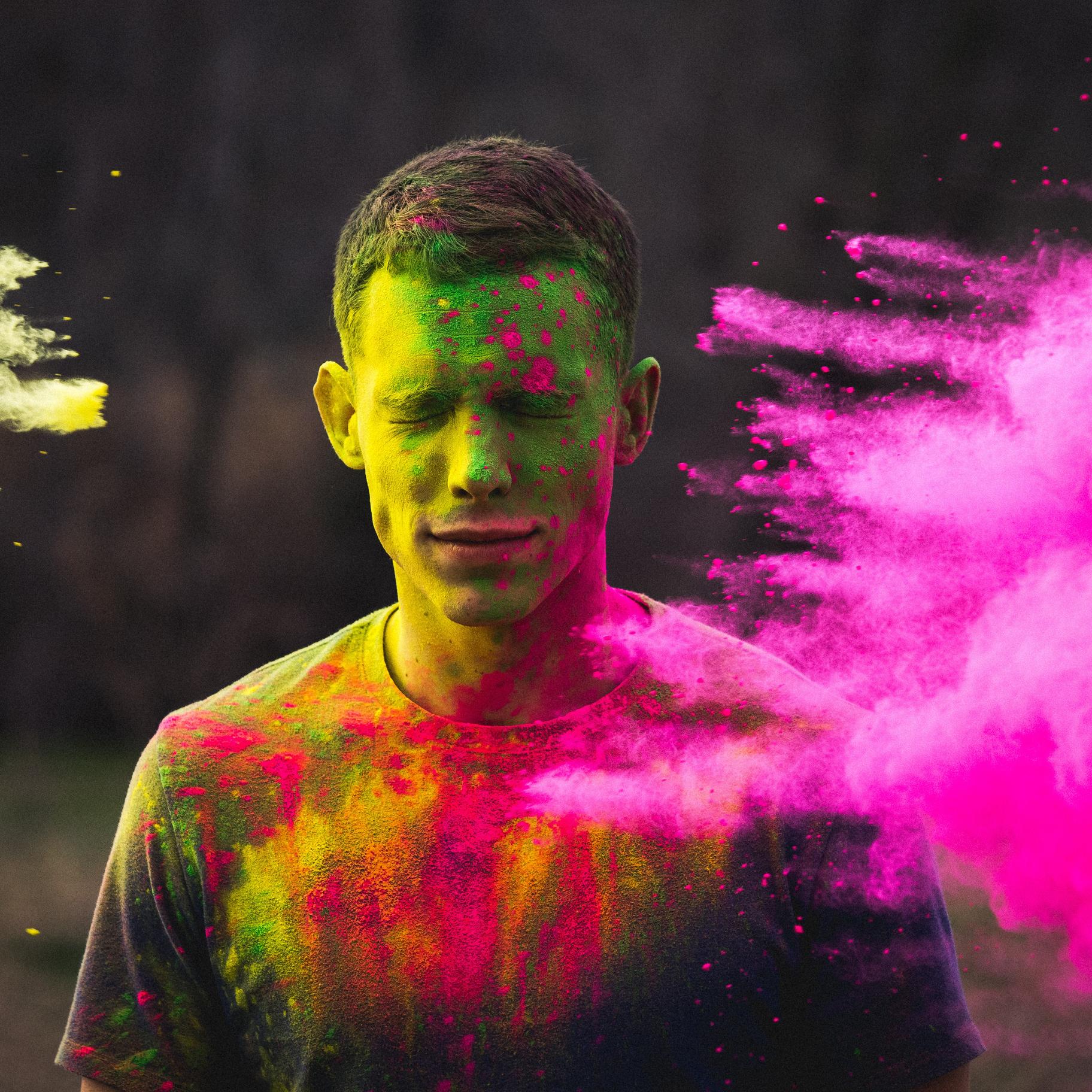} 
& \hspace*{-4mm} \includegraphics[width=.08\linewidth,height=!]{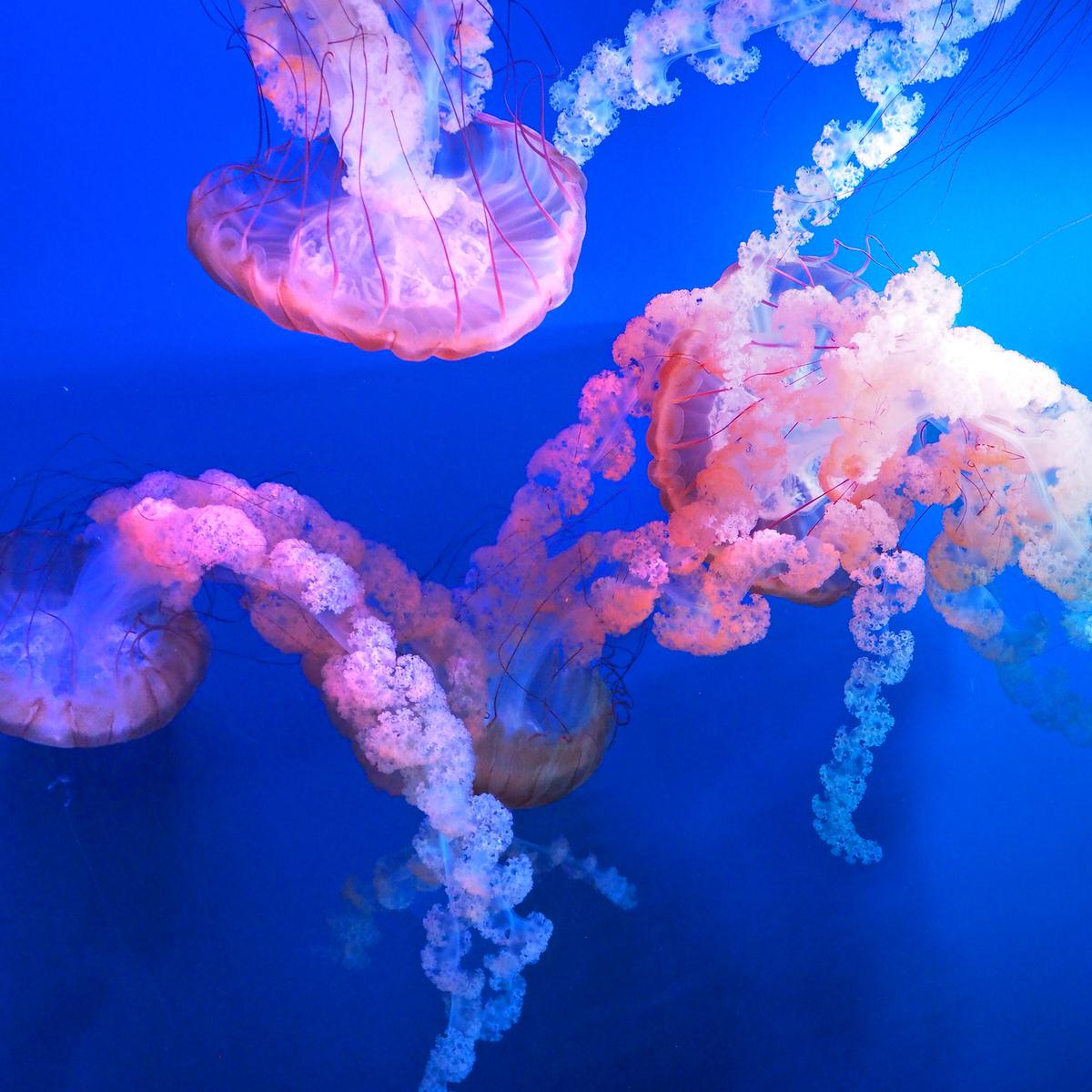} 
& \hspace*{-4mm} \includegraphics[width=.08\linewidth,height=!]{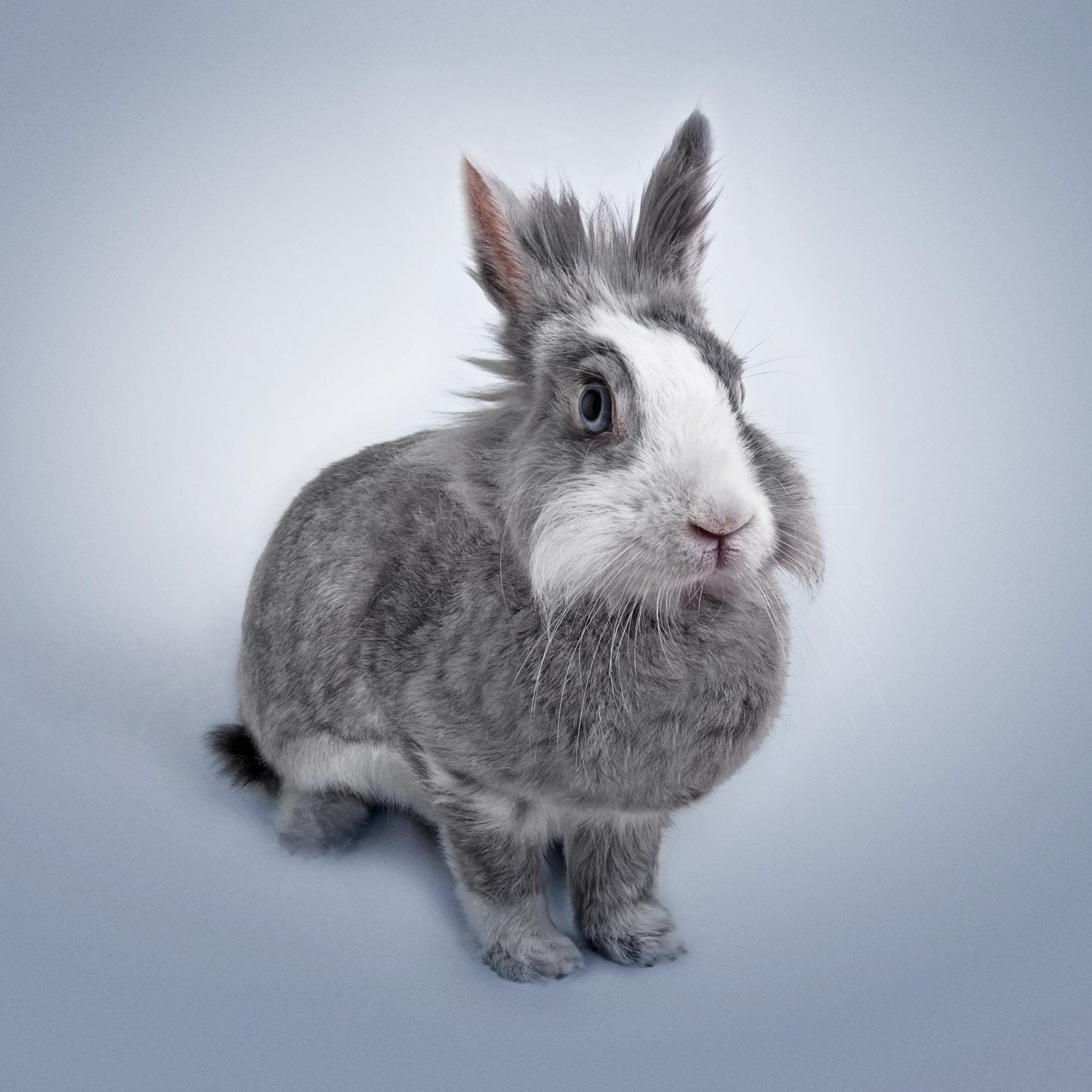} 
& \hspace*{-4mm} \includegraphics[width=.08\linewidth,height=!]{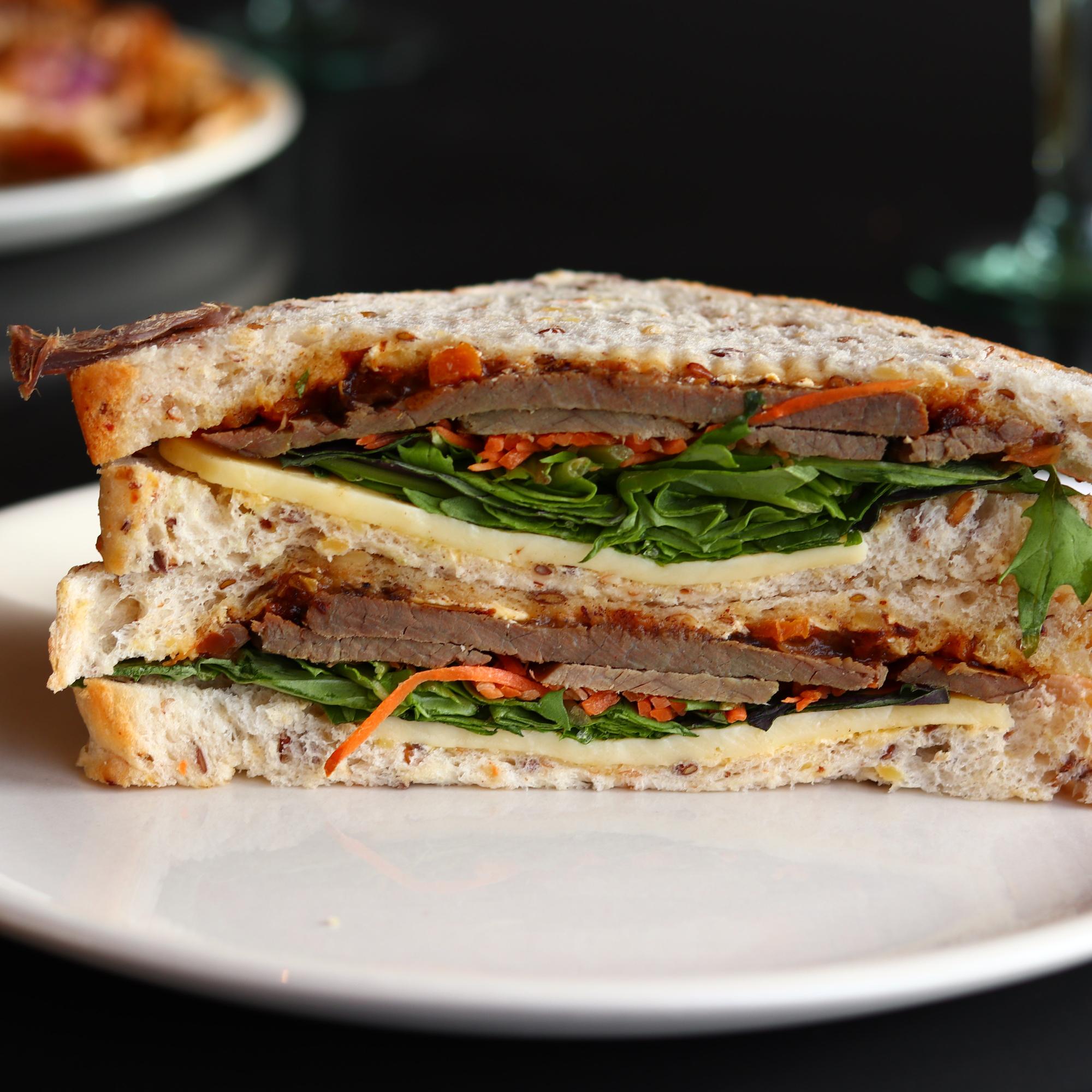} 
& \hspace*{-4mm} \includegraphics[width=.08\linewidth,height=!]{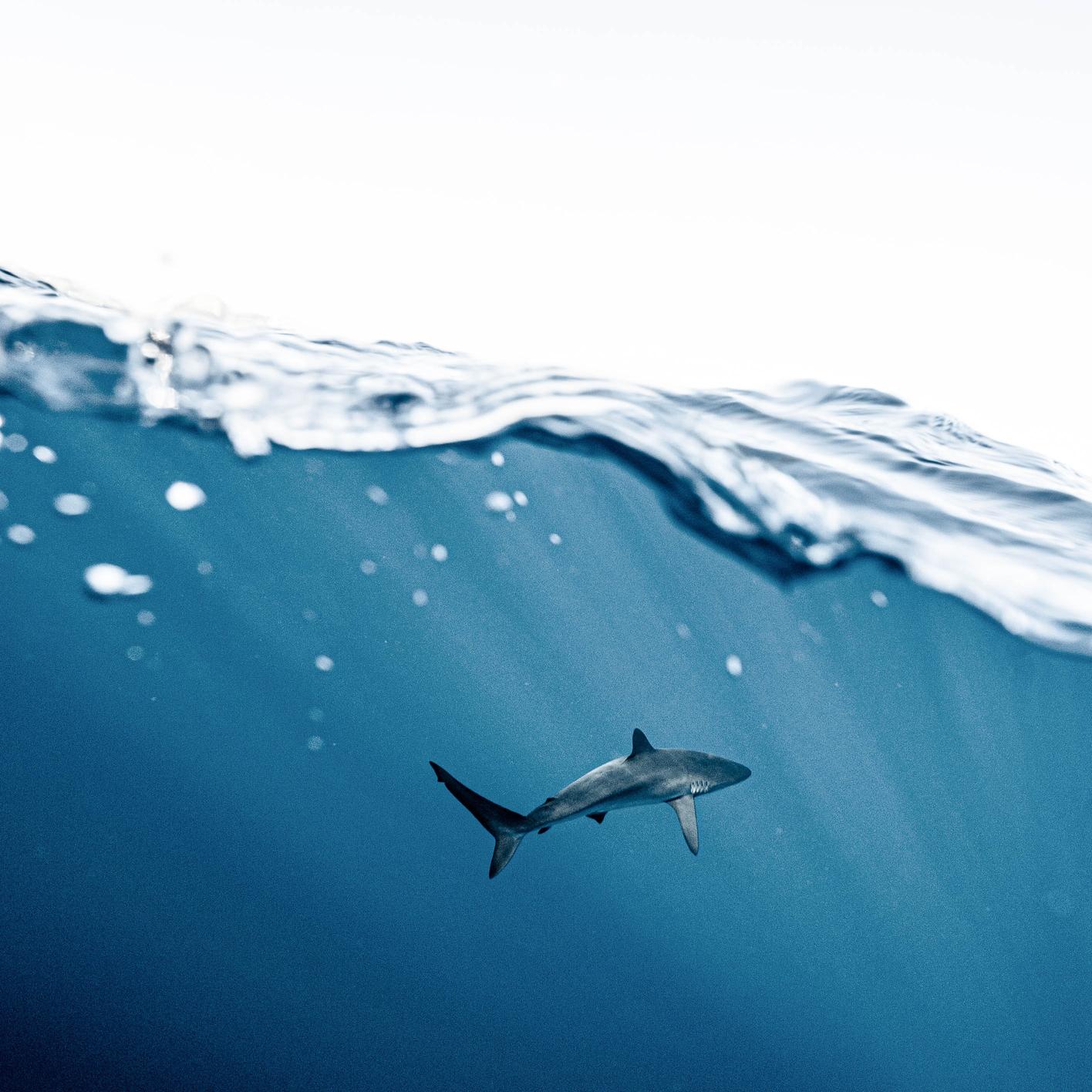} 
& \hspace*{-4mm} \includegraphics[width=.08\linewidth,height=!]{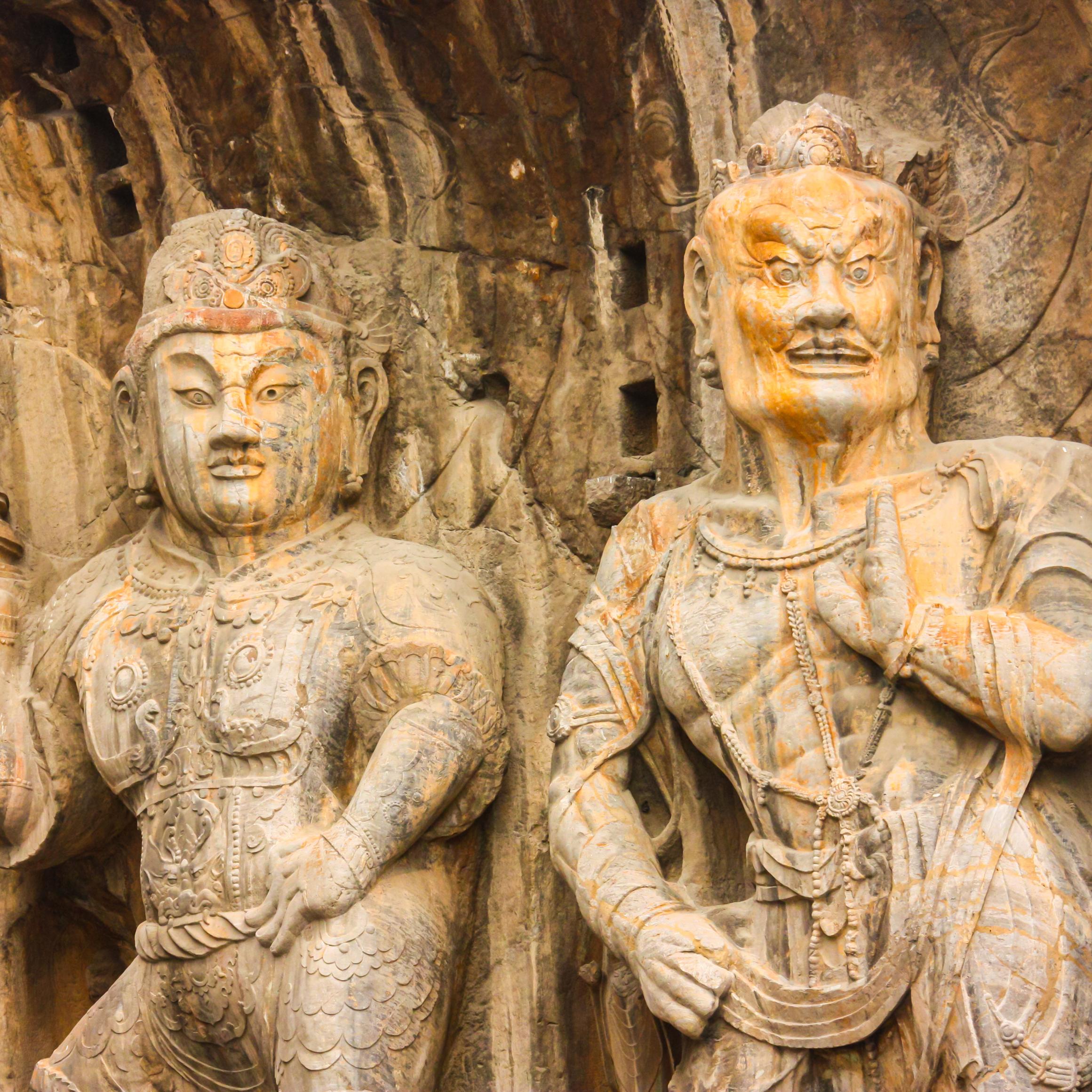} 
& \hspace*{-4mm} \includegraphics[width=.08\linewidth,height=!]{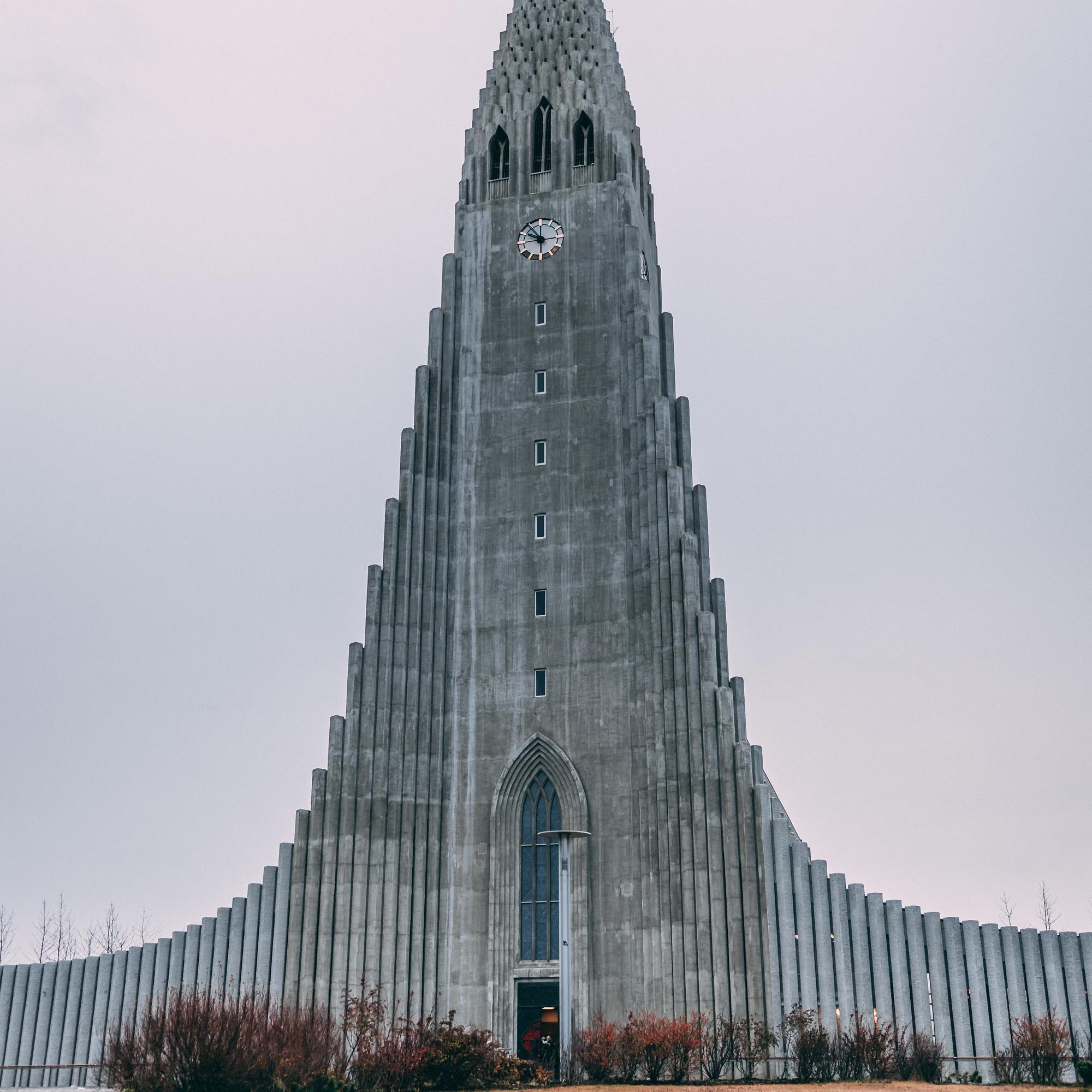} 
& \hspace*{-4mm} \includegraphics[width=.08\linewidth,height=!]{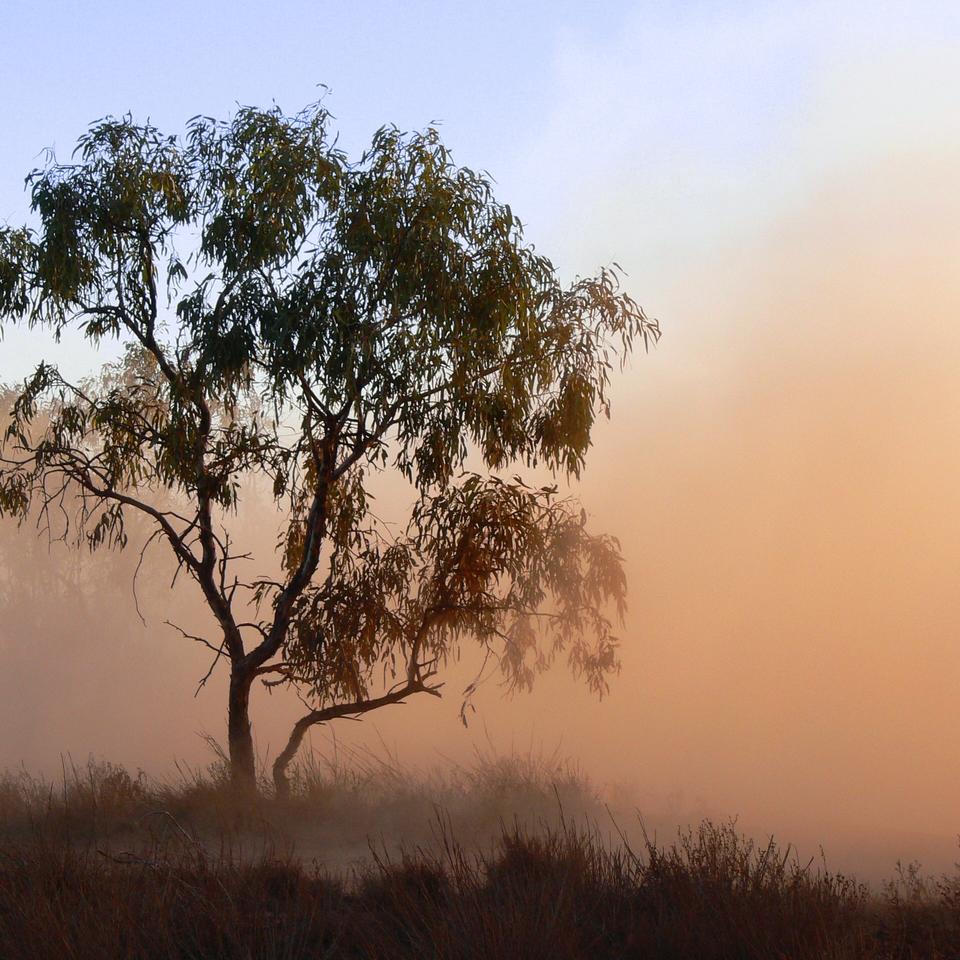} 
& \hspace*{-4mm} \includegraphics[width=.08\linewidth,height=!]{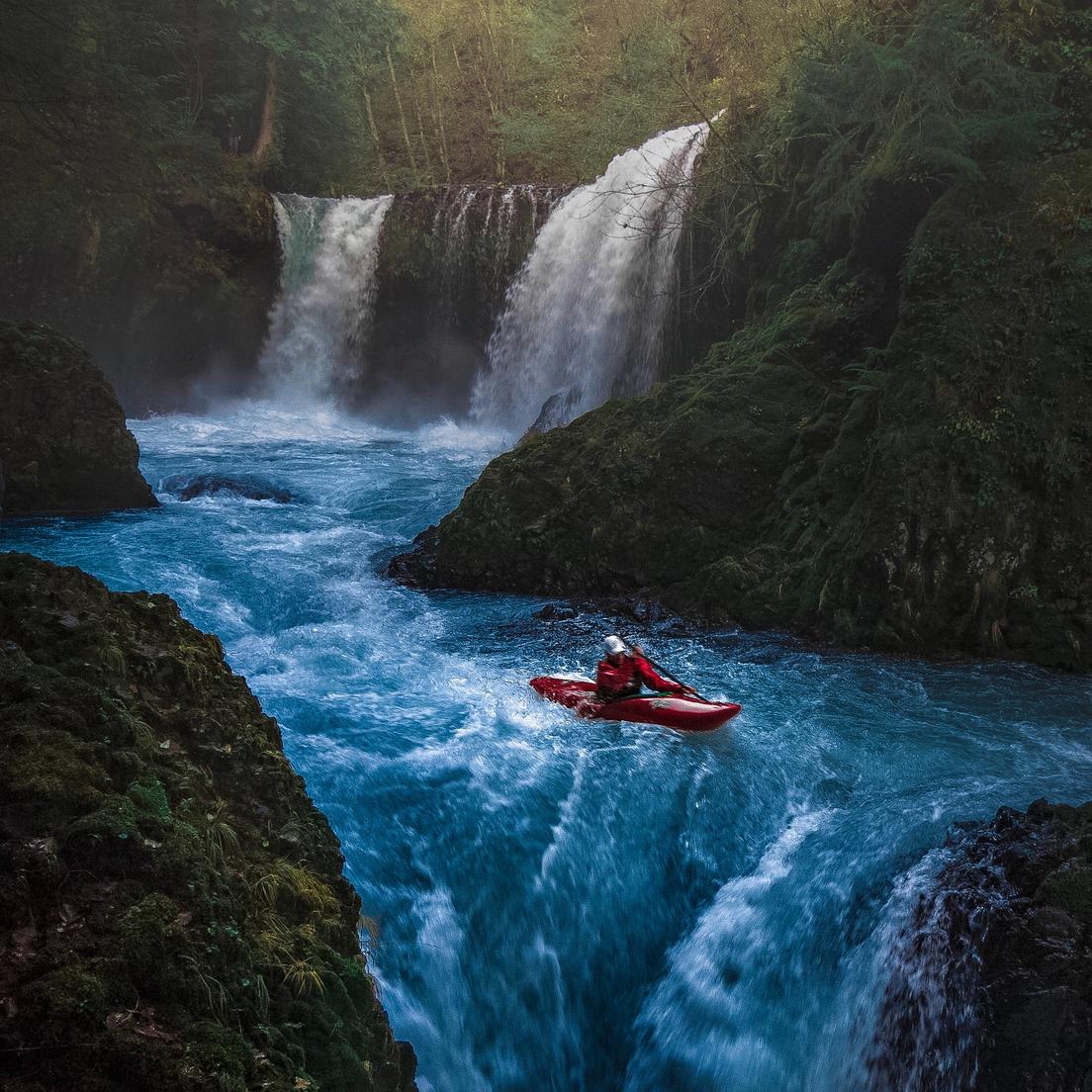} 
\\
\hspace*{-4.1mm} (11)
& \hspace*{-4mm} (12)
& \hspace*{-4mm} (13)
& \hspace*{-4mm} (14)
& \hspace*{-4mm} (15)
& \hspace*{-4mm} (16)
& \hspace*{-4mm} (17)
& \hspace*{-4mm} (18)
& \hspace*{-4mm} (19)
& \hspace*{-4mm} (20)
\\
    \end{tabular}
    \caption{An illustration of the seed images in each object class in {\ours}. Images are cropped and down-scaled for illustration purpose. The name of the object classes are: (1) Architecture; (2) Bear; (3) Bird; (4) Butterfly; (5) Cat; (6) Dog; (7) Fish; (8) Flame; (9) Flowers; (10) Frog; (11) Horse; (12) Human; (13) Jellyfish; (14) Rabbits; (15) Sandwich; (16) Sea; (17) Statue; (18) Tower; (19) Tree; (20) Waterfalls.}
    \label{fig: dataset_object_details}
\end{figure}

We first provide an illustration of the styles and object classes included in {\ours}. Specifically, we show the styles in Fig.\,\ref{fig: dataset_style_details} and object classes in Fig.\,\ref{fig: dataset_object_details} with the style and object names disclosed in the captions.

\newpage
\subsection{Visualization of Style Unlearning}

\begin{figure}
    \centering
    \includegraphics[width=\linewidth]{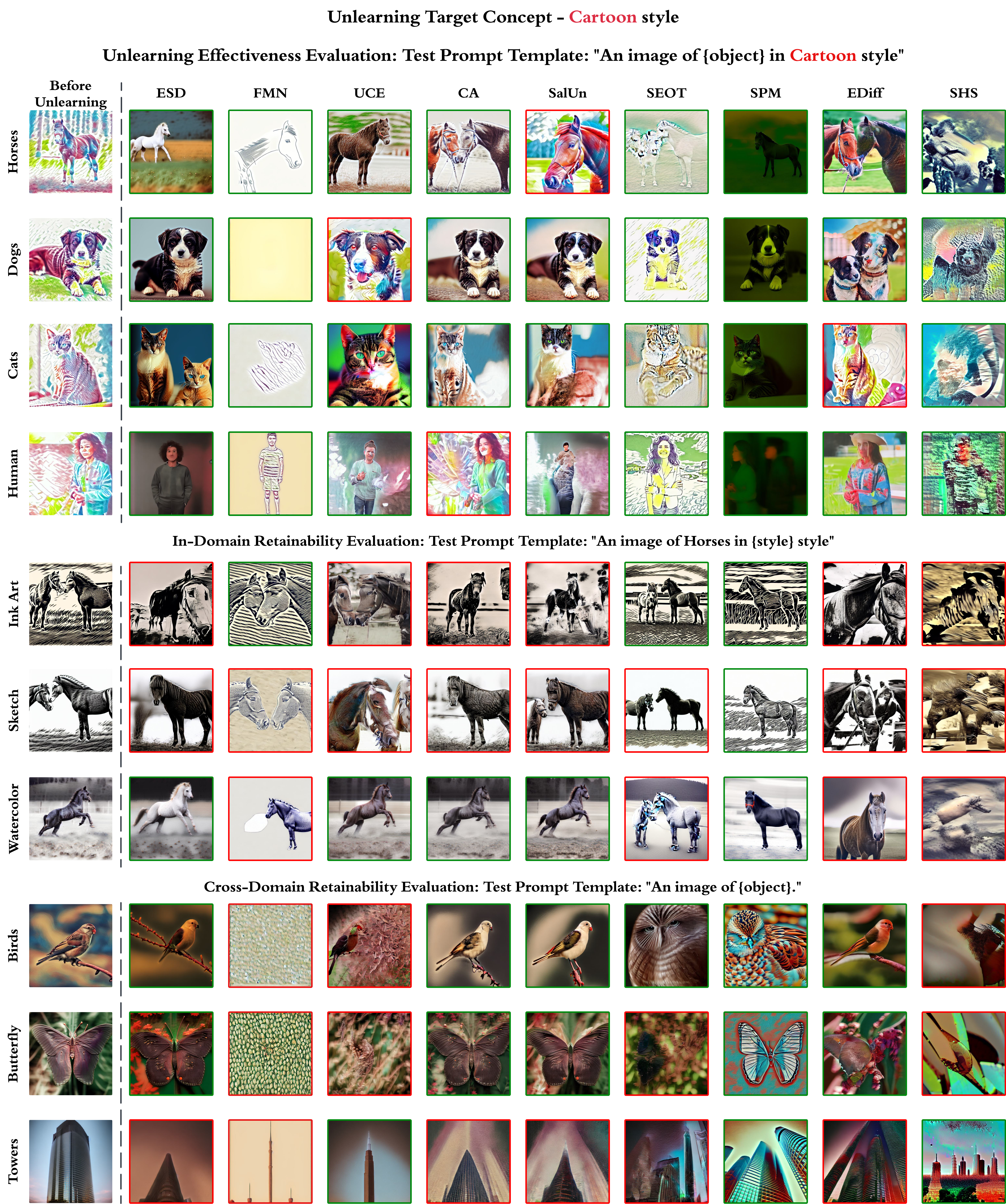}
    \caption{visualization of the unlearning performance of different methods on the task of style unlearning. Three text prompt templates are used to evaluate the unlearning effectiveness, in-domain retainability, and cross-domain retainability of each method. Images with \textcolor{green}{green} frame denote desirable results, while the ones with \textcolor{red}{red} frame denote unlearning or retaining failures.}
    \label{fig: visual_style_unlearning}
\end{figure}

In Fig.\,\ref{fig: visual_style_unlearning}, we provide abundant generation examples of all the 9 methods benchmarked in this work in a case study of unlearning the `Cartoon' style. Both the successful and failure cases are demonstrated in the context of unlearning effectiveness, in-domain retainability, and cross-domain retainability. 

\subsection{Visualization of the Unlearning Performance in the Presence of Adversarial Prompts} 

\begin{figure}
    \centering
    \includegraphics[width=\linewidth]{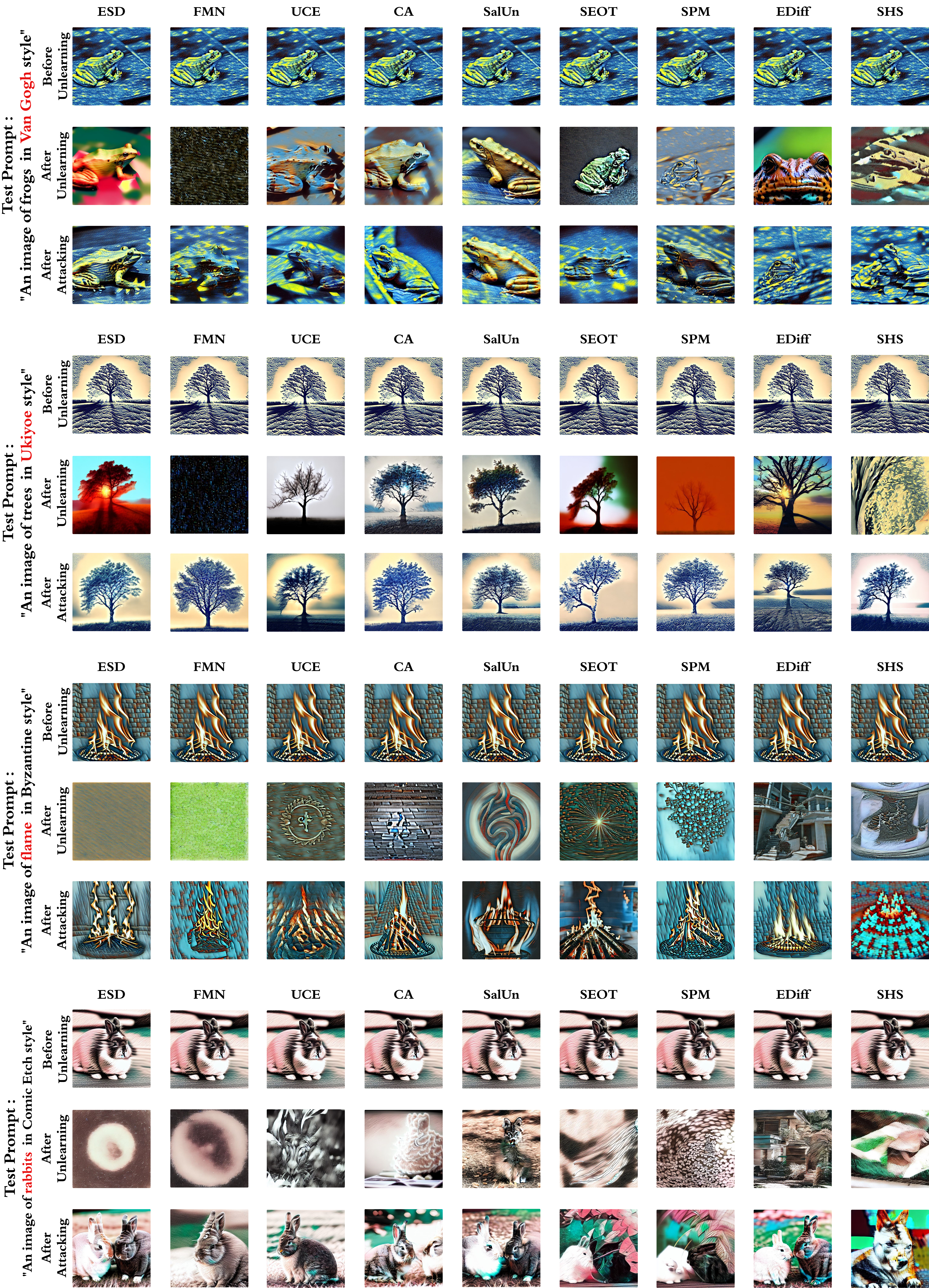}
    \caption{Visualization of the images generated by the unlearned DMs using different unlearning methods in the absence or presence of adversarial prompts. }
    \label{fig: visual_adv_prompt}
\end{figure}

In Fig.\,\ref{fig: visual_adv_prompt}, we provide visualizations for the effect of adversarial prompts. As revealed in Fig.\,\ref{fig: adv_prompt}, all the DM unlearning methods experience a significant drop in unlearning effectiveness when attacked by the adversarial prompt, enabled by UnlearnDiffAtk\,\cite{zhang2023generate}. We provide the image generation in four unlearning cases (two for style unlearning and two for object unlearning), and show the images of the unlearning target successfully generated in the presence of adversarial prompts.

\clearpage
\newpage
\input{sections/broader_impact}

\input{sections/impact_statement}

%% file: sections/broader_impact.tex
\section{Broader Use Cases of {\ours}}
\label{app: broader}

Although {\ours} is originally designed for benchmarking MU methods, we would like to demonstrate its broader use cases of benchmarking more generative modeling tasks, thanks to its good properties discussed in Sec.\,\ref{sec: dataset_proposal}. In this section, we start with a case study on the task of style transfer, which is a much more well-studied topic than MU, but surprisingly also faces great challenges in building up a comprehensive and precise evaluation framework. In the next, we will first dissect the key challenges of the current style transfer evaluation framework, and then demonstrate how {\ours} efficiently resolves these challenges and further proposes a comprehensive and automated benchmark. Through extensive experiments, we draw demonstrate new insights from these results and illuminate the challenges of the future research directions. In the end, we will discuss the possibility of using {\ours} to benchmark more generative modeling tasks.

\subsection{Benchmarking Style Transfer using {\ours}}

\begin{wrapfigure}{r}{.4\linewidth}
\vspace*{-2em}
\centering
\includegraphics[width=\linewidth]{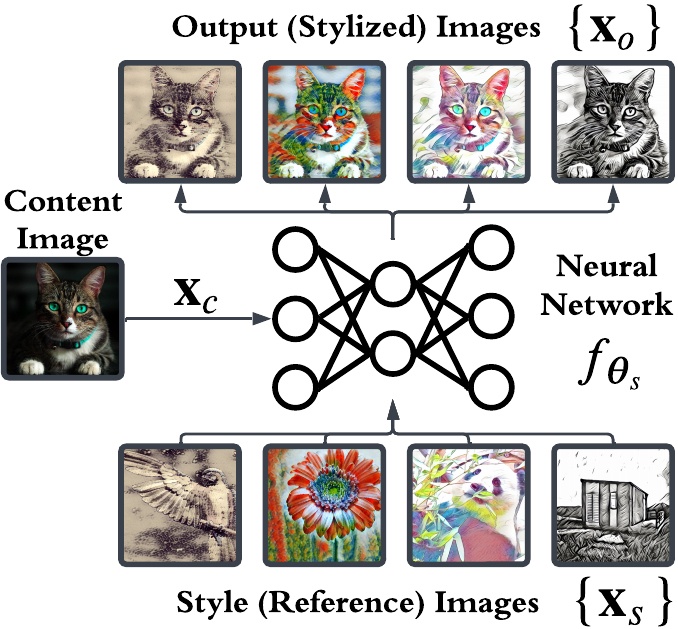}
\caption{An illustration of the task of style transfer.}
\label{fig: style_transfer_illustration.}
\vspace*{-1em}
\end{wrapfigure}

\paragraph{Style transfer.} Style transfer is a long-standing topic and focuses on transferring the artistic style from one \textbf{s}tyle image  (also known as the \textit{reference} image) $\bx_s$ to a target \textbf{c}ontent image $\bx_c$. The most recent methods typically employ a neural network, denoted by $\btheta_s$, to extract style features and perform stylization in a single inference step, expressed as $\hat{\bx}_\mathrm{o}=f_{\btheta_s}(\bx_s, \bx_c)$. Current state-of-the-art (SOTA) style transfer techniques exhibit remarkable generalization capabilities, successfully transferring styles not encountered during training and not requiring any further back-propagations. 
Figure\,\ref{fig: problems_illustration} illustrates the pipeline of this task. Most existing literature {\cite{liu2021adaattn,huang2017arbitrary,li2019learning,chen2016fast,park2019arbitrary,an2021artflow}} utilize \textsc{WikiArt} {\cite{phillips2011wiki}} to provide different styles for training the stylization network $\btheta_s$. During the evaluation, the validation set of \textsc{WikiArt} will serve as the style (reference) images, together with the content images from the COCO dataset \cite{lin2014microsoft}, to form a test bed for style transfer and style learning methods. However, such a evaluation scheme has some inherent limitations, which will be detailed below.

\paragraph{Issues and challenges in the evaluating methods for style transfer}
\label{sec: preliminary_style_transfer}
Although the task of style transfer has been widely studied, its evaluations are still based on very limited quantitative or even sorely qualitative assessments, potentially leading to incomplete and inaccurate assessments {\cite{cai2023image}}. 
Upon examining the evaluation pipelines of over $10$ state-of-the-art (SOTA) style transfer methods, three significant challenges are identified within the current widely accepted evaluation frameworks.
\noindent $\bullet$ \textbf{Challenge I (C1): The lack of the ground truth images for style similarity evaluation.}
Unlike other vision tasks like classification {\cite{singh2020image}}, detection {\cite{zou2023object}}, and segmentation {\cite{minaee2021image}}, one of the key shortcomings of the current style transfer evaluation lies in the lack of the ground truth images $\bx_g$ for the given reference style image $\bx_s$ and the content image $\bx_c$. 
Consequently, existing evaluation metrics, such as the style loss $\ell_\mathrm{style}$ \cite{gatys2015texture, gatys2015neural}, has to be calculated with the reference style image $\bx_s$ as the ground truth $\bx_g$, namely $\ell_\mathrm{style}(\bx_s, \hat{\bx}_o)$, rather than directly using the ground truth $\ell_\mathrm{style}(\bx_o, {\bx}_g)$. 
Obviously, such a indirect evaluation may lead to inaccurate results due to the different contents held in $\bx_s$ and $\bx_o$. Existing work has demonstrated that such evaluation metrics can lead to very different conclusion from that of the user study \cite{an2021artflow}. Therefore, the creation of a \textit{supervised} dataset with ground truth stylized images for every content image under each style is a timely remedy.
\noindent $\bullet$ \textbf{Challenge II (C2): The lack of algorithm stability evaluation against varied reference images $\bx_s$.} 
The evaluation of algorithm stability in style transfer has often been neglected. This assessment requires consistent performance of a method on a content image $\bx_c$ across different style reference images ${\bx_s}$ representing the same target artistic style. Ideally, an algorithm should maintain uniform quality across various references within a style, avoiding significant performance variations. Current challenges in such evaluations stem from the lack of reference sets that exhibit \textit{high stylistic consistency} within each style. Figure \ref{fig: wikiart_examples} showcases examples from the widely-used \textsc{WikiArt} dataset. Despite sharing the same artistic label, images within a row show considerable divergence in visual appearance and style. Consequently, using these images as style references can lead to stable algorithms producing stylistically varied outputs, leading to misleading assessment results. Therefore, creating a dataset with high stylistic uniformity within each style category and clear differentiation between styles is crucial for accurate measurement of algorithm stability.


\begin{figure*}[thb]
\centerline{
\begin{tabular}{ccc}
\hspace*{0mm}\includegraphics[width=.22\textwidth,height=!]{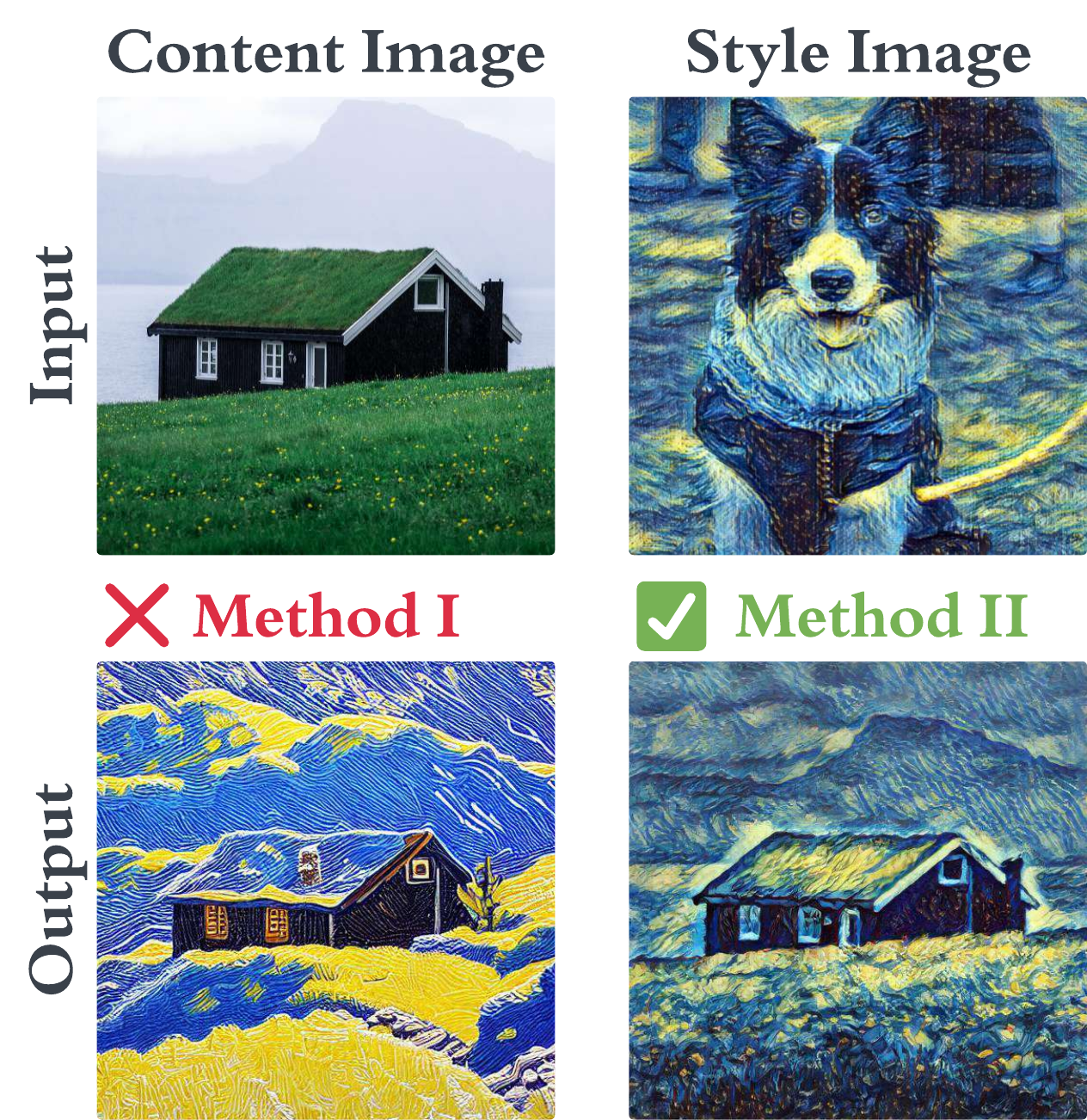}  
& \hspace*{0mm}\includegraphics[width=.22\textwidth,height=!]{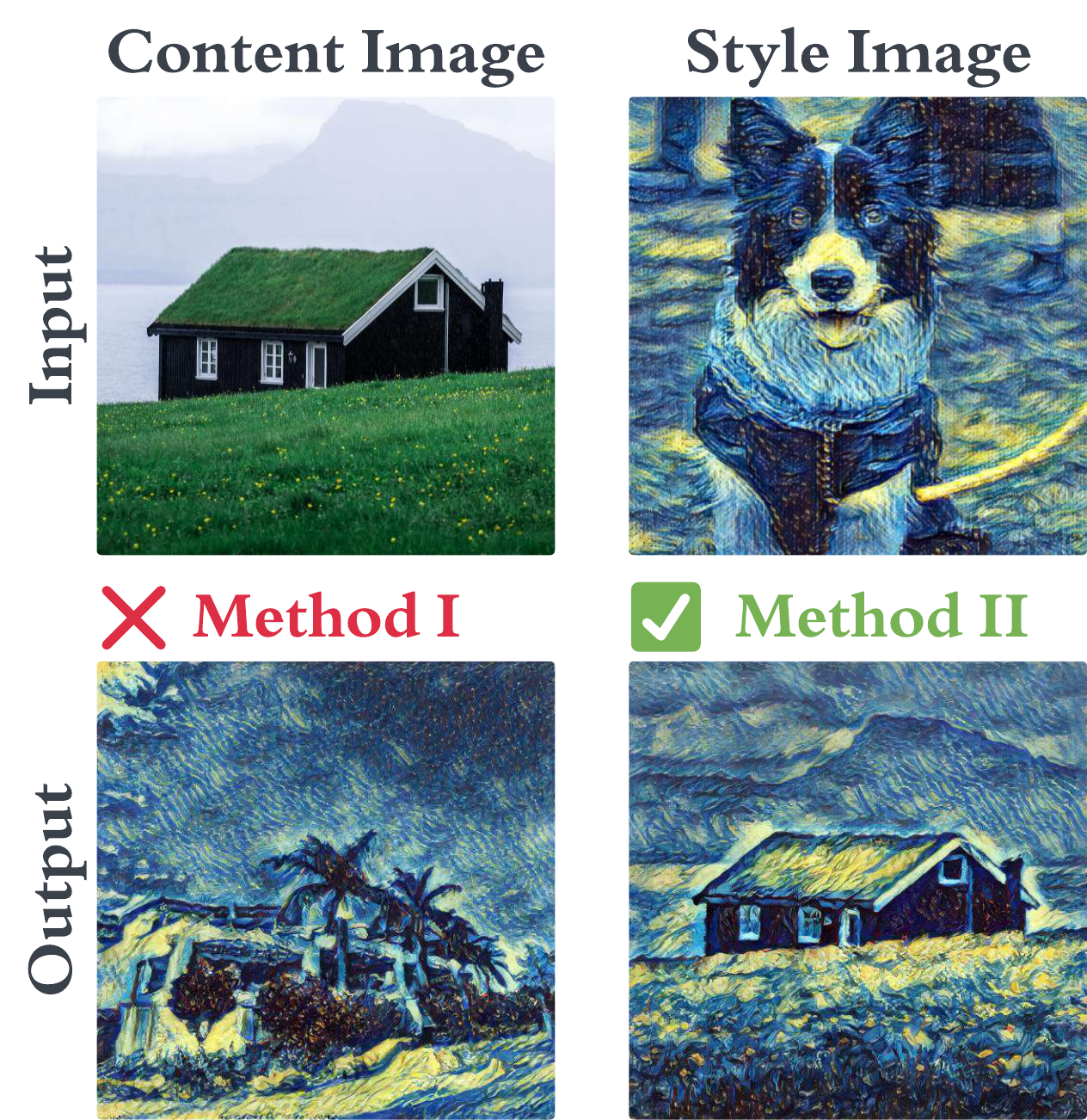}
& \hspace*{0mm}\includegraphics[width=.44\textwidth,height=!]{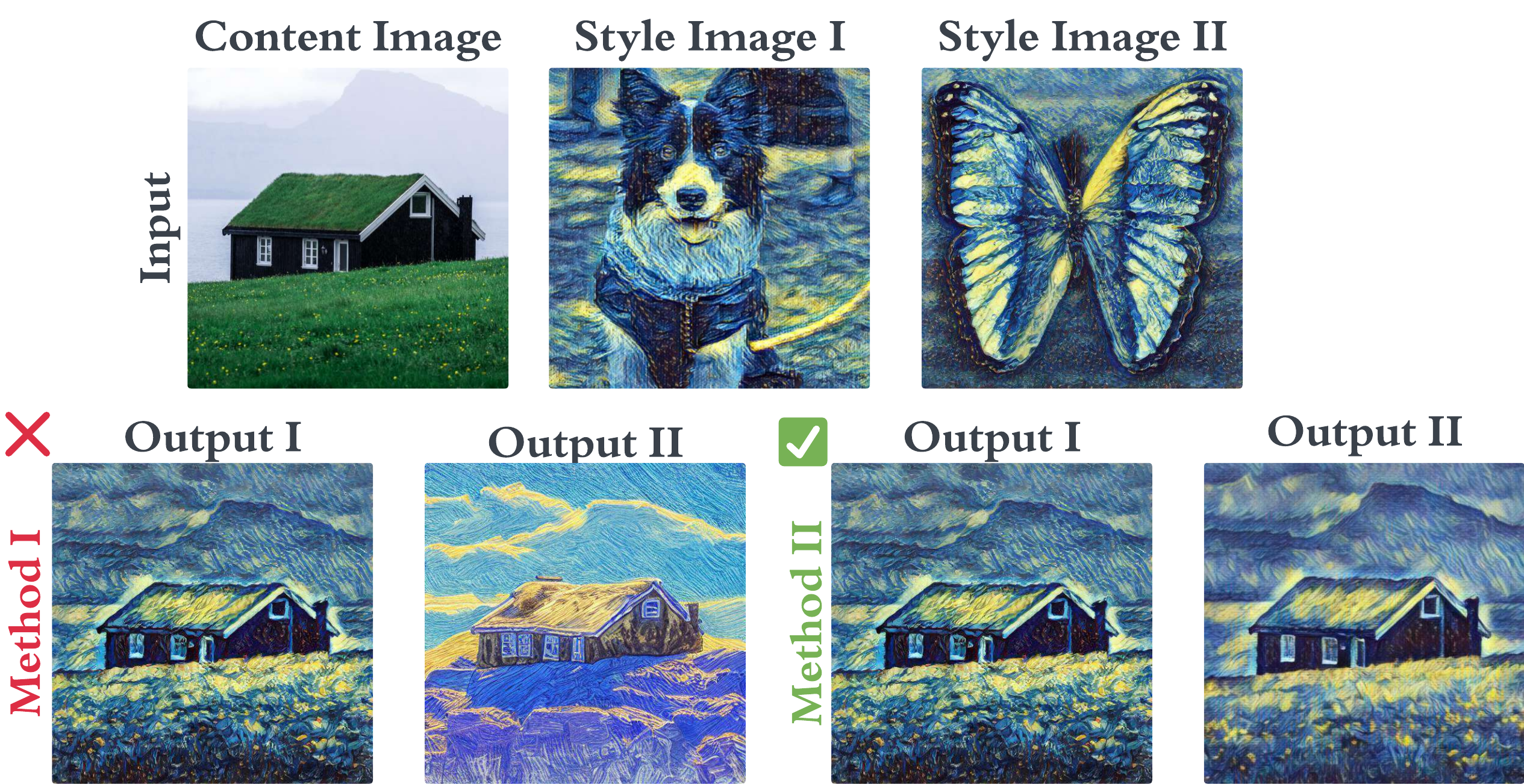} \\
(a) Style Authenticity
& \hspace*{-2mm} (b) Content Integrity
& \hspace*{0mm} (c) Algorithm Stability
\end{tabular}}
\caption{An illustration of comprehensive performance metrics for style transfer tasks. (a) Style authenticity measures the style similarity between the generated image and the style image. (b) Content integrity measures the content preservation between the generated image and the content image. (c) Algorithm stability reflects the sensitivity of the algorithm to different style images. For all the metrics in the illustration, ``Method II'' is always better than ``Method I''.}
\label{fig: style_evaluation}
\vspace*{-1.5em}
\end{figure*}

\paragraph{Building up a comprehensive evaluation pipeline with {\ours} for style transfer.} To revolutionize the evaluation framework with {\ours}, it is crucial to first understand the components that form a comprehensive evaluation pipeline, ensuring a thorough and impartial assessment of performance. In the realm of style transfer, three pivotal metrics stand paramount, as depicted in \textbf{Figure\,\ref{fig: style_evaluation}}:

\ding{182} Style Authenticity: Assesses the extent to which the style of the generated image aligns with that of the provided reference style image(s).
\ding{183} Content Integrity: Measures how well the content features are preserved post style transfer.
\ding{184} Algorithm Stability: Evaluates the algorithm's robustness against variations in content subjects, target styles, or style reference image selections, while keeping other parameters constant.

In addressing the challenges (\textbf{C1}-\textbf{C2}), {\ours} proves to be inherently advantageous. \textbf{First}, the style-specific supervision embedded in {\ours} enables the provision of ground truth for quantitative evaluations of stylized images. \textbf{Second}, the extensive image collection within the same style category in {\ours} allows for the assessment of algorithm stability by applying style transfer methods to varied reference images. To encompass the aforementioned aspects of style transfer performance, we propose the following quantitative evaluation metrics:

\ding{172} Style Loss: Utilizes a feature map-based style loss \cite{deng2022stytr2} to quantify the stylistic \textit{dissimilarity} between image pairs, effectively representing the inverse of style authenticity.
\ding{173} Content Loss: Employs a VGG-based, feature-map content loss \cite{deng2022stytr2, zhang2022domain} to measure the visual dissimilarity between the reference and generated images, essentially mirroring content integrity.
\ding{174} Averaged Standard Deviation (STD): Computes the average STD of style and content loss \textit{w.r.t.} the same reference image, reflecting algorithm stability.

Furthermore, similar to the MU task, we also consider efficiency metrics for each method, including:

\ding{175} Average Time Consumption: Measures the time required for performing style transfer.
\ding{176} Peak GPU Memory Consumption: Records the maximum GPU memory usage during the style transfer process.
\ding{177} Model Storage Memory Consumption: Assesses the memory requirement for storing the style transfer model.


With the introduction of evaluation metrics (\ding{172}-\ding{177}), we establish a comprehensive evaluation pipeline for style transfer. The process is as follows:

For each style transfer method, style transfer is executed within each object class. Specifically, for every style, images indexed from $1$ to $18$ serve as style reference images, while seed images corresponding to indices $19$ and $20$ are used as content images. Style and content loss are computed for each pair of style reference and content images, with the stylized images derived from the used seed content images serving as the ground truth for style loss. This results in a total of $60 \times 20 \times 18\times 2$ experimental trials for each method. For a specific style and content image pair, $18$ experiments are performed to calculate the Standard Deviation (STD) values for both style and content loss. These STD values are then averaged over all content images (amounting to $60 \times 20\times 2$ cases). The findings from these comprehensive evaluations are presented in Table \ref{tab: style_transfer_overview}.

Following this evaluation pipeline, we scrutinized $9$ prominent style transfer methods, including \textsc{SaNet} \cite{park2019arbitrary}, \textsc{Mcc} \cite{deng2021arbitrary}, \textsc{Mast} \cite{huo2021manifold}, \textsc{ArtFlow} with its two variants (\textsc{AF-AdaIn} and \textsc{AF-WCT}) \cite{an2021artflow}, \textsc{IE-Contrast} \cite{chen2021artistic}, \textsc{Cast} \cite{zhang2022domain}, \textsc{StyTr2} \cite{deng2022stytr2}, and \textsc{BLIP} \cite{li2023blip}.

\input{tables/style_transfer_main}

\paragraph{Experiment results analysis.} 
Tab.\,\ref{tab: style_transfer_overview} provides a systematic evaluation of the performance of various methods tested. From the analysis, we can derive several crucial insights:

\textbf{First}, it is evident that no single method excels across all evaluation metrics. Notably, \textsc{Mcc} demonstrates superior performance in maintaining stylistic authenticity, as indicated by a low style loss. Conversely, \textsc{StyTr2} stands out in preserving content integrity, reflected by its minimal content loss.

\textbf{Second}, the assessment of standard deviation is indispensable for a comprehensive evaluation. The method with the optimal performance does not necessarily exhibit the greatest stability. This is particularly apparent in the style loss evaluation, where \textsc{Mcc}, despite achieving the best result in terms of style loss, exhibits the least stability, denoted by the highest standard deviation.



\subsection{Other Possible Applications of {\ours}}

In the preceding section, we demonstrated the application of {\ours} in refining evaluation metrics and frameworks for style transfer. Beyond this, we recognize the potential of {\ours} in diverse domains. Here, we delve into two illustrative examples:

\textbf{Bias mitigation.} Bias mitigation in DMs, which are now gaining popularity, can also benefit from {\ours}. Its hierarchical and balanced architecture enables the deliberate introduction of artificial biases by selectively omitting data from specific groups. For instance, by predominantly excluding images from styles other than the `Van Gogh Style' within the `Dogs' class, DMs finetuned on this dataset will inherently exhibit a tendency to generate images of dogs in the Van Gogh style, particularly when the style is not explicitly specified in the prompt. This approach not only allows for the manipulation and quantification of biases but also paves the way for {\ours} to become a standardized benchmark for bias mitigation, similar to the role of MU for DMs.

\textbf{Vision in-context learning (V-ICL)}. V-ICL \cite{wang2023context, wang2023images, bai2023sequential, wang2024instantid} is another domain where {\ours} can be effectively applied. The field of V-ICL is in urgent need of robust, comprehensive methods for the fair assessment of existing models. In this context, the image pairs from {\ours} are ideally suited for evaluating various tasks such as style transfer, image inpainting, and image segmentation, offering a rich resource for nuanced and quantitative analyses.

%% file: tables/style_transfer_main.tex
\begin{table*}[htb]
\centering
\caption{Performance overview of different style transfer methods evaluated with {\ours} dataset. The performance are assessed from the perspectives of stylistic authenticity (style loss), content integrity (content loss), algorithm stability (standard deviations from different dimensions), and efficiency. For all the metrics, \textit{smaller} values are always preferred for better performance. The best performance per each metric is highlighted in \textbf{bold}. The standard deviations are first calculated with respect to different styles, object classes, or tested content images and then averaged in order to depict the algorithm stability from different perspectives.}
\resizebox{\textwidth}{!}{%
\begin{tabular}{c|c|ccc|c|ccc|ccc}
\toprule[1pt]
\midrule
\multirow{3}{*}{Method} & \multicolumn{4}{c|}{Style Loss} & \multicolumn{4}{c|}{Content Loss} & \multicolumn{3}{c}{Efficiency} \\

& \multirow{2}{*}{Mean} & \multicolumn{3}{c|}{STD (Averaged over)} & \multirow{2}{*}{Mean} & \multicolumn{3}{c|}{STD (Averaged over)} & \multirow{2}{*}{\begin{tabular}[c]{@{}c@{}}Time\\ (s/image)\end{tabular}} & \multirow{2}{*}{\begin{tabular}[c]{@{}c@{}}Memory\\ (GB)\end{tabular}} & \multirow{2}{*}{\begin{tabular}[c]{@{}c@{}}Storage\\ (GB)\end{tabular}} \\

&  & Style & Object & Content Image &  & Style & Object & Content Image &  &  &  \\
\midrule
\textsc{SaNet}          & 23.48 & \textbf{2.73} & \textbf{2.87} & 1.87 & 0.85 & 0.12 & 0.17 & 0.09 & 0.29 & \textbf{2.3} & 0.11 \\
\textsc{Mcc}            & \textbf{17.92} & 4.59 & 4.82 & 2.14 & 0.96 & 0.14 & 0.21 & 0.07 & 0.38 & 5.4 & 0.10 \\
\textsc{Mast}           & 24.10 & 2.87 & 3.16 & 1.74 & 1.42 & 0.33 & 0.34 & 0.18 & 2.86 & 4.8 & 0.16 \\
\textsc{AF-AdaIn}       & 20.78 & 2.96 & 3.13 & 1.65 & 1.09 & 0.11 & 0.19 & 0.05 & 0.53 & 6.3 & 0.08 \\
\textsc{AF-WCT}         & 20.22 & 2.94 & 3.19 & 1.75 & 1.02 & 0.12 & 0.18 & 0.05 & 0.53 & 6.3 & \textbf{0.08} \\
\textsc{IE-Contrast}    & 21.27 & 3.01 & 3.32 & 2.05 & 1.08 & 0.29 & 0.31 & 0.15 & \textbf{0.05} & 3.8 & 0.11 \\
\textsc{Cast}           & 24.01 & 2.78 & 2.90 & \textbf{1.35} & 1.38 & 0.32 & 0.40 & 0.16 & 0.32 & 6.7 & 0.19 \\
\textsc{StyTr2}         & 19.75 & 3.04 & 3.30 & 1.91 & \textbf{0.62} & \textbf{0.10} & \textbf{0.12} & \textbf{0.04} & 0.58 & 3.9 & 0.21 \\
\textsc{BLIP}           & 25.43 & 2.90 & 3.06 & 2.03 & 1.61 & 0.30 & 0.34 & 0.16 & 8.87 & 7.2 & 7.23 \\
 \midrule
 \bottomrule[1pt]
\end{tabular}%
}
\label{tab: style_transfer_overview}
\end{table*}

%% file: sections/impact_statement.tex
\section{Impact Statement}

This work helps improve the assessment and further promotes the advancement of MU (machine unlearning) methods for DMs (diffusion models), which are known to be effective in relieving or mitigating the various negative societal influences brought by the prevalent usage of DMs, which include but are not limited to the following aspects.

\noindent $\bullet$ \textbf{Avoiding Copyright Issues.} There is an urgent need for the generative model providers to scrub the influence of certain data on an already-trained model. In January 2023, a notable lawsuit targeted two leading AI art generators, Stable Diffusion \cite{rombach2022high} and Midjourney \cite{midjourney}, for alleged copyright infringement. Concurrently, incidents with the recently released Midjourney V6 \cite{midjourney} also highlighted a visual plagiarism issue on famous film scenes. These instances illuminate the broad copyright challenges inherent in the way of training data collection method of those foundation generative models' training datasets. MU methods can be used as an effective method to remove the influence of the private data and avoid unnecessary retraining.

\noindent $\bullet$ \textbf{Mitigating biases and stereotypes.} Generative AI systems are known to have tendencies towards bias, stereotypes, and reductionism, when it comes to gender, race and national identities \cite{turk2023ai}. For example, a recent study on the images generated with Midjourney revealed, that images associated with higher-paying job titles featured people with lighter skin tones, and that results for most professional roles were male-dominated \cite{vinker2023concept}. MU is known to be effective in eliminating biases rooted in the training data. Moreover, {\ours} offers a flexible framework to benchmark MU techniques against bias removal, allowing for the creation and quantitative control of biases across different object classes for comprehensive bias removal studies.

%% file: main.bbl
\begin{thebibliography}{96}
\providecommand{\natexlab}[1]{#1}
\providecommand{\url}[1]{#1}
\csname url@samestyle\endcsname
\providecommand{\newblock}{\relax}
\providecommand{\bibinfo}[2]{#2}
\providecommand{\BIBentrySTDinterwordspacing}{\spaceskip=0pt\relax}
\providecommand{\BIBentryALTinterwordstretchfactor}{4}
\providecommand{\BIBentryALTinterwordspacing}{\spaceskip=\fontdimen2\font plus
\BIBentryALTinterwordstretchfactor\fontdimen3\font minus \fontdimen4\font\relax}
\providecommand{\BIBforeignlanguage}[2]{{%
\expandafter\ifx\csname l@#1\endcsname\relax
\typeout{** WARNING: IEEEtranN.bst: No hyphenation pattern has been}%
\typeout{** loaded for the language `#1'. Using the pattern for}%
\typeout{** the default language instead.}%
\else
\language=\csname l@#1\endcsname
\fi
#2}}
\providecommand{\BIBdecl}{\relax}
\BIBdecl

\bibitem[Rombach et~al.(2022)Rombach, Blattmann, Lorenz, Esser, and Ommer]{rombach2022high}
R.~Rombach, A.~Blattmann, D.~Lorenz, P.~Esser, and B.~Ommer, ``High-resolution image synthesis with latent diffusion models,'' in \emph{Proceedings of the IEEE/CVF conference on computer vision and pattern recognition}, 2022, pp. 10\,684--10\,695.

\bibitem[Midjourney et al., 2023()]{midjourney}
``{Midjourney},'' \url{https://www.midjourney.com/}, 2023, accessed: 2024-01-19.

\bibitem[Kang et~al.(2023)Kang, Zhu, Zhang, Park, Shechtman, Paris, and Park]{kang2023scaling}
M.~Kang, J.-Y. Zhu, R.~Zhang, J.~Park, E.~Shechtman, S.~Paris, and T.~Park, ``Scaling up gans for text-to-image synthesis,'' in \emph{Proceedings of the IEEE/CVF Conference on Computer Vision and Pattern Recognition}, 2023, pp. 10\,124--10\,134.

\bibitem[Tao et~al.(2023)Tao, Bao, Tang, and Xu]{tao2023galip}
M.~Tao, B.-K. Bao, H.~Tang, and C.~Xu, ``Galip: Generative adversarial clips for text-to-image synthesis,'' in \emph{Proceedings of the IEEE/CVF Conference on Computer Vision and Pattern Recognition}, 2023, pp. 14\,214--14\,223.

\bibitem[Tao et~al.(2022)Tao, Tang, Wu, Jing, Bao, and Xu]{tao2022df}
M.~Tao, H.~Tang, F.~Wu, X.-Y. Jing, B.-K. Bao, and C.~Xu, ``Df-gan: A simple and effective baseline for text-to-image synthesis,'' in \emph{Proceedings of the IEEE/CVF Conference on Computer Vision and Pattern Recognition}, 2022, pp. 16\,515--16\,525.

\bibitem[Li et~al.(2019{\natexlab{a}})Li, Qi, Lukasiewicz, and Torr]{li2019controllable}
B.~Li, X.~Qi, T.~Lukasiewicz, and P.~Torr, ``Controllable text-to-image generation,'' \emph{Advances in Neural Information Processing Systems}, vol.~32, 2019.

\bibitem[Reed et~al.(2016)Reed, Akata, Yan, Logeswaran, Schiele, and Lee]{reed2016generative}
S.~Reed, Z.~Akata, X.~Yan, L.~Logeswaran, B.~Schiele, and H.~Lee, ``Generative adversarial text to image synthesis,'' in \emph{International conference on machine learning}.\hskip 1em plus 0.5em minus 0.4em\relax PMLR, 2016, pp. 1060--1069.

\bibitem[Zhou et~al.(2022)Zhou, Yang, Loy, and Liu]{zhou2022learning}
K.~Zhou, J.~Yang, C.~C. Loy, and Z.~Liu, ``Learning to prompt for vision-language models,'' \emph{International Journal of Computer Vision}, vol. 130, no.~9, pp. 2337--2348, 2022.

\bibitem[Kawar et~al.(2023)Kawar, Zada, Lang, Tov, Chang, Dekel, Mosseri, and Irani]{Kawar_2023_CVPR}
B.~Kawar, S.~Zada, O.~Lang, O.~Tov, H.~Chang, T.~Dekel, I.~Mosseri, and M.~Irani, ``Imagic: Text-based real image editing with diffusion models,'' in \emph{Proceedings of the IEEE/CVF Conference on Computer Vision and Pattern Recognition (CVPR)}, June 2023, pp. 6007--6017.

\bibitem[Zhang et~al.(2022{\natexlab{a}})Zhang, Cai, Pan, Hong, Guo, Yang, and Liu]{zhang2022motiondiffuse}
M.~Zhang, Z.~Cai, L.~Pan, F.~Hong, X.~Guo, L.~Yang, and Z.~Liu, ``Motiondiffuse: Text-driven human motion generation with diffusion model,'' \emph{arXiv preprint arXiv:2208.15001}, 2022.

\bibitem[Fotor et al., 2022()]{fotor}
``Fotor: Online photo editor for everyone.'' \url{https://www.fotor.com/}, 2023.

\bibitem[LabLab AI et al., 2023()]{lablab2023forensic}
``Forensic sketch artist at lablab's openai event,'' 2023.

\bibitem[Schuhmann et~al.(2022)Schuhmann, Beaumont, Vencu, Gordon, Wightman, Cherti, Coombes, Katta, Mullis, Wortsman, et~al.]{schuhmann2022laion}
C.~Schuhmann, R.~Beaumont, R.~Vencu, C.~Gordon, R.~Wightman, M.~Cherti, T.~Coombes, A.~Katta, C.~Mullis, M.~Wortsman \emph{et~al.}, ``Laion-5b: An open large-scale dataset for training next generation image-text models,'' \emph{Advances in Neural Information Processing Systems}, vol.~35, pp. 25\,278--25\,294, 2022.

\bibitem[Schramowski et~al.(2023)Schramowski, Brack, Deiseroth, and Kersting]{schramowski2023safe}
P.~Schramowski, M.~Brack, B.~Deiseroth, and K.~Kersting, ``Safe latent diffusion: Mitigating inappropriate degeneration in diffusion models,'' in \emph{Proceedings of the IEEE/CVF Conference on Computer Vision and Pattern Recognition}, 2023, pp. 22\,522--22\,531.

\bibitem[Krietzberg(2023)]{author2023midjourney}
I.~Krietzberg, ``Users of midjourney text-to-image site claim issues with new update,'' \emph{The Street}, 2023.

\bibitem[Vinker et~al.(2023)Vinker, Voynov, Cohen-Or, and Shamir]{vinker2023concept}
Y.~Vinker, A.~Voynov, D.~Cohen-Or, and A.~Shamir, ``Concept decomposition for visual exploration and inspiration,'' \emph{ACM Transactions on Graphics (TOG)}, vol.~42, no.~6, pp. 1--13, 2023.

\bibitem[Turk(2023)]{turk2023ai}
\BIBentryALTinterwordspacing
V.~Turk, ``How ai reduces the world to stereotypes,'' \emph{Rest of World}, 2023. [Online]. Available: \url{https://restofworld.org/2023/ai-image-stereotypes/?utm_source=pocket-newtab-en-us}
\BIBentrySTDinterwordspacing

\bibitem[Tsai et~al.(2023)Tsai, Hsu, Xie, Lin, Chen, Li, Chen, Yu, and Huang]{tsai2023ring}
Y.-L. Tsai, C.-Y. Hsu, C.~Xie, C.-H. Lin, J.-Y. Chen, B.~Li, P.-Y. Chen, C.-M. Yu, and C.-Y. Huang, ``Ring-a-bell! how reliable are concept removal methods for diffusion models?'' \emph{arXiv preprint arXiv:2310.10012}, 2023.

\bibitem[Dong et~al.(2024)Dong, Guo, Wang, Wang, Zhang, and Liu]{dong2024towards}
P.~Dong, S.~Guo, J.~Wang, B.~Wang, J.~Zhang, and Z.~Liu, ``Towards test-time refusals via concept negation,'' \emph{Advances in Neural Information Processing Systems}, vol.~36, 2024.

\bibitem[Hong et~al.(2024)Hong, Lee, and Woo]{hong2024all}
S.~Hong, J.~Lee, and S.~S. Woo, ``All but one: Surgical concept erasing with model preservation in text-to-image diffusion models,'' in \emph{Proceedings of the AAAI Conference on Artificial Intelligence}, vol.~38, no.~19, 2024, pp. 21\,143--21\,151.

\bibitem[Zhao et~al.(2024)Zhao, Zhang, Zheng, Kong, and Yin]{zhao2024separable}
M.~Zhao, L.~Zhang, T.~Zheng, Y.~Kong, and B.~Yin, ``Separable multi-concept erasure from diffusion models,'' \emph{arXiv preprint arXiv:2402.05947}, 2024.

\bibitem[Huang et~al.(2023)Huang, Chang, Tsai, Lai, and Wang]{huang2023receler}
C.-P. Huang, K.-P. Chang, C.-T. Tsai, Y.-H. Lai, and Y.-C.~F. Wang, ``Receler: Reliable concept erasing of text-to-image diffusion models via lightweight erasers,'' \emph{arXiv preprint arXiv:2311.17717}, 2023.

\bibitem[Gandikota et~al.(2023{\natexlab{a}})Gandikota, Materzynska, Fiotto-Kaufman, and Bau]{gandikota2023erasing}
R.~Gandikota, J.~Materzynska, J.~Fiotto-Kaufman, and D.~Bau, ``Erasing concepts from diffusion models,'' \emph{arXiv preprint arXiv:2303.07345}, 2023.

\bibitem[Gandikota et~al.(2023{\natexlab{b}})Gandikota, Orgad, Belinkov, Materzy\'nska, and Bau]{gandikota2023unified}
R.~Gandikota, H.~Orgad, Y.~Belinkov, J.~Materzy\'nska, and D.~Bau, ``Unified concept editing in diffusion models,'' \emph{arXiv preprint arXiv:2308.14761}, 2023.

\bibitem[Kumari et~al.(2023)Kumari, Zhang, Wang, Shechtman, Zhang, and Zhu]{kumari2023ablating}
N.~Kumari, B.~Zhang, S.-Y. Wang, E.~Shechtman, R.~Zhang, and J.-Y. Zhu, ``Ablating concepts in text-to-image diffusion models,'' \emph{arXiv preprint arXiv:2303.13516}, 2023.

\bibitem[Lyu et~al.(2023)Lyu, Yang, Hong, Chen, Jin, He, Xue, Han, and Ding]{lyu2023one}
M.~Lyu, Y.~Yang, H.~Hong, H.~Chen, X.~Jin, Y.~He, H.~Xue, J.~Han, and G.~Ding, ``One-dimensional adapter to rule them all: Concepts, diffusion models and erasing applications,'' \emph{arXiv preprint arXiv:2312.16145}, 2023.

\bibitem[Fan et~al.(2023)Fan, Liu, Zhang, Wei, Wong, and Liu]{fan2023salun}
C.~Fan, J.~Liu, Y.~Zhang, D.~Wei, E.~Wong, and S.~Liu, ``Salun: Empowering machine unlearning via gradient-based weight saliency in both image classification and generation,'' \emph{arXiv preprint arXiv:2310.12508}, 2023.

\bibitem[Zhang et~al.(2023{\natexlab{a}})Zhang, Wang, Xu, Wang, and Shi]{zhang2023forget}
E.~Zhang, K.~Wang, X.~Xu, Z.~Wang, and H.~Shi, ``Forget-me-not: Learning to forget in text-to-image diffusion models,'' \emph{arXiv preprint arXiv:2303.17591}, 2023.

\bibitem[Heng and Soh(2023)]{heng2023selective}
A.~Heng and H.~Soh, ``Selective amnesia: A continual learning approach to forgetting in deep generative models,'' \emph{arXiv preprint arXiv:2305.10120}, 2023.

\bibitem[Li et~al.(2024{\natexlab{a}})Li, van~de Weijer, Hu, Khan, Hou, Wang, and Yang]{li2024get}
S.~Li, J.~van~de Weijer, T.~Hu, F.~S. Khan, Q.~Hou, Y.~Wang, and J.~Yang, ``Get what you want, not what you don't: Image content suppression for text-to-image diffusion models,'' \emph{arXiv preprint arXiv:2402.05375}, 2024.

\bibitem[Wu et~al.(2024)Wu, Le, Hayat, and Harandi]{wu2024erasediff}
J.~Wu, T.~Le, M.~Hayat, and M.~Harandi, ``Erasediff: Erasing data influence in diffusion models,'' \emph{arXiv preprint arXiv:2401.05779}, 2024.

\bibitem[Wu and Harandi(2024)]{wu2024scissorhands}
J.~Wu and M.~Harandi, ``Scissorhands: Scrub data influence via connection sensitivity in networks,'' \emph{arXiv preprint arXiv:2401.06187}, 2024.

\bibitem[Zhang et~al.(2024)Zhang, Chen, Jia, Zhang, Fan, Liu, Hong, Ding, and Liu]{zhang2024defensive}
Y.~Zhang, X.~Chen, J.~Jia, Y.~Zhang, C.~Fan, J.~Liu, M.~Hong, K.~Ding, and S.~Liu, ``Defensive unlearning with adversarial training for robust concept erasure in diffusion models,'' \emph{arXiv preprint arXiv:2405.15234}, 2024.

\bibitem[Cao and Yang(2015)]{cao2015towards}
Y.~Cao and J.~Yang, ``Towards making systems forget with machine unlearning,'' in \emph{2015 IEEE symposium on security and privacy}.\hskip 1em plus 0.5em minus 0.4em\relax IEEE, 2015, pp. 463--480.

\bibitem[Golatkar et~al.(2020)Golatkar, Achille, and Soatto]{golatkar2020eternal}
A.~Golatkar, A.~Achille, and S.~Soatto, ``Eternal sunshine of the spotless net: Selective forgetting in deep networks,'' in \emph{Proceedings of the IEEE/CVF Conference on Computer Vision and Pattern Recognition}, 2020, pp. 9304--9312.

\bibitem[Becker and Liebig(2022)]{becker2022evaluating}
A.~Becker and T.~Liebig, ``Evaluating machine unlearning via epistemic uncertainty,'' \emph{arXiv preprint arXiv:2208.10836}, 2022.

\bibitem[Thudi et~al.(2022)Thudi, Deza, Chandrasekaran, and Papernot]{thudi2022unrolling}
A.~Thudi, G.~Deza, V.~Chandrasekaran, and N.~Papernot, ``Unrolling sgd: Understanding factors influencing machine unlearning,'' in \emph{2022 IEEE 7th European Symposium on Security and Privacy (EuroS\&P)}.\hskip 1em plus 0.5em minus 0.4em\relax IEEE, 2022, pp. 303--319.

\bibitem[Jia et~al.(2023)Jia, Liu, Ram, Yao, Liu, Liu, Sharma, and Liu]{jia2023model}
J.~Jia, J.~Liu, P.~Ram, Y.~Yao, G.~Liu, Y.~Liu, P.~Sharma, and S.~Liu, ``Model sparsity can simplify machine unlearning,'' in \emph{Annual Conference on Neural Information Processing Systems}, 2023.

\bibitem[Chen et~al.(2023{\natexlab{a}})Chen, Gao, Liu, Peng, and Wang]{chen2023boundary}
M.~Chen, W.~Gao, G.~Liu, K.~Peng, and C.~Wang, ``Boundary unlearning: Rapid forgetting of deep networks via shifting the decision boundary,'' in \emph{Proceedings of the IEEE/CVF Conference on Computer Vision and Pattern Recognition}, 2023, pp. 7766--7775.

\bibitem[Warnecke et~al.(2021)Warnecke, Pirch, Wressnegger, and Rieck]{warnecke2021machine}
A.~Warnecke, L.~Pirch, C.~Wressnegger, and K.~Rieck, ``Machine unlearning of features and labels,'' \emph{arXiv preprint arXiv:2108.11577}, 2021.

\bibitem[Fan et~al.(2024)Fan, Liu, Hero, and Liu]{fan2024challenging}
C.~Fan, J.~Liu, A.~Hero, and S.~Liu, ``Challenging forgets: Unveiling the worst-case forget sets in machine unlearning,'' \emph{arXiv preprint arXiv:2403.07362}, 2024.

\bibitem[Ginart et~al.(2019)Ginart, Guan, Valiant, and Zou]{ginart2019making}
A.~Ginart, M.~Guan, G.~Valiant, and J.~Y. Zou, ``Making ai forget you: Data deletion in machine learning,'' \emph{Advances in neural information processing systems}, vol.~32, 2019.

\bibitem[Neel et~al.(2021)Neel, Roth, and Sharifi-Malvajerdi]{neel2021descent}
S.~Neel, A.~Roth, and S.~Sharifi-Malvajerdi, ``Descent-to-delete: Gradient-based methods for machine unlearning,'' in \emph{Algorithmic Learning Theory}.\hskip 1em plus 0.5em minus 0.4em\relax PMLR, 2021, pp. 931--962.

\bibitem[Ullah et~al.(2021)Ullah, Mai, Rao, Rossi, and Arora]{ullah2021machine}
E.~Ullah, T.~Mai, A.~Rao, R.~A. Rossi, and R.~Arora, ``Machine unlearning via algorithmic stability,'' in \emph{Conference on Learning Theory}.\hskip 1em plus 0.5em minus 0.4em\relax PMLR, 2021, pp. 4126--4142.

\bibitem[Sekhari et~al.(2021)Sekhari, Acharya, Kamath, and Suresh]{sekhari2021remember}
A.~Sekhari, J.~Acharya, G.~Kamath, and A.~T. Suresh, ``Remember what you want to forget: Algorithms for machine unlearning,'' \emph{Advances in Neural Information Processing Systems}, vol.~34, pp. 18\,075--18\,086, 2021.

\bibitem[Bourtoule et~al.(2021)Bourtoule, Chandrasekaran, Choquette-Choo, Jia, Travers, Zhang, Lie, and Papernot]{bourtoule2021machine}
L.~Bourtoule, V.~Chandrasekaran, C.~A. Choquette-Choo, H.~Jia, A.~Travers, B.~Zhang, D.~Lie, and N.~Papernot, ``Machine unlearning,'' in \emph{2021 IEEE Symposium on Security and Privacy (SP)}.\hskip 1em plus 0.5em minus 0.4em\relax IEEE, 2021, pp. 141--159.

\bibitem[Nguyen et~al.(2022)Nguyen, Huynh, Nguyen, Liew, Yin, and Nguyen]{nguyen2022survey}
T.~T. Nguyen, T.~T. Huynh, P.~L. Nguyen, A.~W.-C. Liew, H.~Yin, and Q.~V.~H. Nguyen, ``A survey of machine unlearning,'' \emph{arXiv preprint arXiv:2209.02299}, 2022.

\bibitem[Izzo et~al.(2021)Izzo, Smart, Chaudhuri, and Zou]{izzo2021approximate}
Z.~Izzo, M.~A. Smart, K.~Chaudhuri, and J.~Zou, ``Approximate data deletion from machine learning models,'' in \emph{International Conference on Artificial Intelligence and Statistics}.\hskip 1em plus 0.5em minus 0.4em\relax PMLR, 2021, pp. 2008--2016.

\bibitem[Graves et~al.(2021)Graves, Nagisetty, and Ganesh]{graves2021amnesiac}
L.~Graves, V.~Nagisetty, and V.~Ganesh, ``Amnesiac machine learning,'' in \emph{Proceedings of the AAAI Conference on Artificial Intelligence}, vol.~35, no.~13, 2021, pp. 11\,516--11\,524.

\bibitem[Guo et~al.(2019)Guo, Goldstein, Hannun, and Van Der~Maaten]{guo2019certified}
C.~Guo, T.~Goldstein, A.~Hannun, and L.~Van Der~Maaten, ``Certified data removal from machine learning models,'' \emph{arXiv preprint arXiv:1911.03030}, 2019.

\bibitem[Xu et~al.(2023)Xu, Zhu, Zhang, Zhou, and Yu]{xu2023machine}
H.~Xu, T.~Zhu, L.~Zhang, W.~Zhou, and P.~S. Yu, ``Machine unlearning: A survey,'' \emph{ACM Computing Surveys}, vol.~56, no.~1, pp. 1--36, 2023.

\bibitem[Jia et~al.(2024)Jia, Zhang, Zhang, Liu, Runwal, Diffenderfer, Kailkhura, and Liu]{jia2024soul}
J.~Jia, Y.~Zhang, Y.~Zhang, J.~Liu, B.~Runwal, J.~Diffenderfer, B.~Kailkhura, and S.~Liu, ``Soul: Unlocking the power of second-order optimization for llm unlearning,'' \emph{arXiv preprint arXiv:2404.18239}, 2024.

\bibitem[Hoofnagle et~al.(2019)Hoofnagle, van~der Sloot, and Borgesius]{hoofnagle2019european}
C.~J. Hoofnagle, B.~van~der Sloot, and F.~Z. Borgesius, ``The european union general data protection regulation: what it is and what it means,'' \emph{Information \& Communications Technology Law}, vol.~28, no.~1, pp. 65--98, 2019.

\bibitem[Liu et~al.(2022)Liu, Fan, Chen, Liu, Ma, Wang, and Ma]{liu2022backdoor}
Y.~Liu, M.~Fan, C.~Chen, X.~Liu, Z.~Ma, L.~Wang, and J.~Ma, ``Backdoor defense with machine unlearning,'' \emph{arXiv preprint arXiv:2201.09538}, 2022.

\bibitem[Chen et~al.(2023{\natexlab{b}})Chen, Yang, Xiong, Bai, Hu, Hao, Feng, Zhou, Wu, and Liu]{chen2023fast}
R.~Chen, J.~Yang, H.~Xiong, J.~Bai, T.~Hu, J.~Hao, Y.~Feng, J.~T. Zhou, J.~Wu, and Z.~Liu, ``Fast model debias with machine unlearning,'' \emph{arXiv preprint arXiv:2310.12560}, 2023.

\bibitem[Oesterling et~al.(2023)Oesterling, Ma, Calmon, and Lakkaraju]{oesterling2023fair}
A.~Oesterling, J.~Ma, F.~P. Calmon, and H.~Lakkaraju, ``Fair machine unlearning: Data removal while mitigating disparities,'' \emph{arXiv preprint arXiv:2307.14754}, 2023.

\bibitem[Halimi et~al.(2022)Halimi, Kadhe, Rawat, and Baracaldo]{halimi2022federated}
A.~Halimi, S.~Kadhe, A.~Rawat, and N.~Baracaldo, ``Federated unlearning: How to efficiently erase a client in fl?'' \emph{arXiv preprint arXiv:2207.05521}, 2022.

\bibitem[Jin et~al.(2023)Jin, Chen, Zhang, and Li]{jin2023forgettable}
R.~Jin, M.~Chen, Q.~Zhang, and X.~Li, ``Forgettable federated linear learning with certified data removal,'' \emph{arXiv preprint arXiv:2306.02216}, 2023.

\bibitem[Zhang et~al.(2023{\natexlab{b}})Zhang, Jia, Chen, Chen, Zhang, Liu, Ding, and Liu]{zhang2023generate}
Y.~Zhang, J.~Jia, X.~Chen, A.~Chen, Y.~Zhang, J.~Liu, K.~Ding, and S.~Liu, ``To generate or not? safety-driven unlearned diffusion models are still easy to generate unsafe images... for now,'' \emph{arXiv preprint arXiv:2310.11868}, 2023.

\bibitem[Chin et~al.(2023)Chin, Jiang, Huang, Chen, and Chiu]{chin2023prompting4debugging}
Z.-Y. Chin, C.-M. Jiang, C.-C. Huang, P.-Y. Chen, and W.-C. Chiu, ``Prompting4debugging: Red-teaming text-to-image diffusion models by finding problematic prompts,'' \emph{arXiv preprint arXiv:2309.06135}, 2023.

\bibitem[Krizhevsky and Hinton(2009)]{Krizhevsky2009learning}
A.~Krizhevsky and G.~Hinton, ``Learning multiple layers of features from tiny images,'' \emph{Master's thesis, Department of Computer Science, University of Toronto}, 2009.

\bibitem[Deng et~al.(2009)Deng, Dong, Socher, Li, Li, and Fei-Fei]{deng2009imagenet}
J.~Deng, W.~Dong, R.~Socher, L.-J. Li, K.~Li, and L.~Fei-Fei, ``Imagenet: A large-scale hierarchical image database,'' in \emph{Computer Vision and Pattern Recognition, 2009. CVPR 2009. IEEE Conference on}.\hskip 1em plus 0.5em minus 0.4em\relax IEEE, 2009, pp. 248--255.

\bibitem[Choi and Na(2023)]{choi2023towards}
D.~Choi and D.~Na, ``Towards machine unlearning benchmarks: Forgetting the personal identities in facial recognition systems,'' \emph{arXiv preprint arXiv:2311.02240}, 2023.

\bibitem[Maini et~al.(2024)Maini, Feng, Schwarzschild, Lipton, and Kolter]{maini2024tofu}
P.~Maini, Z.~Feng, A.~Schwarzschild, Z.~C. Lipton, and J.~Z. Kolter, ``Tofu: A task of fictitious unlearning for llms,'' \emph{arXiv preprint arXiv:2401.06121}, 2024.

\bibitem[Li et~al.(2024{\natexlab{b}})Li, Pan, Gopal, Yue, Berrios, Gatti, Li, Dombrowski, Goel, Phan, et~al.]{li2024wmdp}
N.~Li, A.~Pan, A.~Gopal, S.~Yue, D.~Berrios, A.~Gatti, J.~D. Li, A.-K. Dombrowski, S.~Goel, L.~Phan \emph{et~al.}, ``The wmdp benchmark: Measuring and reducing malicious use with unlearning,'' \emph{arXiv preprint arXiv:2403.03218}, 2024.

\bibitem[Phillips and Mackintosh(2011)]{phillips2011wiki}
F.~Phillips and B.~Mackintosh, ``Wiki art gallery, inc.: A case for critical thinking,'' \emph{Issues in Accounting Education}, vol.~26, no.~3, pp. 593--608, 2011.

\bibitem[Moayeri et~al.(2024)Moayeri, Basu, Balasubramanian, Kattakinda, Chengini, Brauneis, and Feizi]{moayeri2024rethinking}
M.~Moayeri, S.~Basu, S.~Balasubramanian, P.~Kattakinda, A.~Chengini, R.~Brauneis, and S.~Feizi, ``Rethinking artistic copyright infringements in the era of text-to-image generative models,'' \emph{arXiv preprint arXiv:2404.08030}, 2024.

\bibitem[Dosovitskiy et~al.(2020)Dosovitskiy, Beyer, Kolesnikov, Weissenborn, Zhai, Unterthiner, Dehghani, Minderer, Heigold, Gelly, et~al.]{dosovitskiy2020image}
A.~Dosovitskiy, L.~Beyer, A.~Kolesnikov, D.~Weissenborn, X.~Zhai, T.~Unterthiner, M.~Dehghani, M.~Minderer, G.~Heigold, S.~Gelly \emph{et~al.}, ``An image is worth 16x16 words: Transformers for image recognition at scale,'' \emph{arXiv preprint arXiv:2010.11929}, 2020.

\bibitem[Pexels et al., 2023()]{pexels}
``Pexels: The best free stock photos, royalty free images and videos shared by creators.'' \url{https://www.pexels.com/}, 2023.

\bibitem[Jang et~al.(2022)Jang, Yoon, Yang, Cha, Lee, Logeswaran, and Seo]{jang2022knowledge}
J.~Jang, D.~Yoon, S.~Yang, S.~Cha, M.~Lee, L.~Logeswaran, and M.~Seo, ``Knowledge unlearning for mitigating privacy risks in language models,'' \emph{arXiv preprint arXiv:2210.01504}, 2022.

\bibitem[Chen and Yang(2023)]{chen2023unlearn}
J.~Chen and D.~Yang, ``Unlearn what you want to forget: Efficient unlearning for llms,'' \emph{arXiv preprint arXiv:2310.20150}, 2023.

\bibitem[Gebru et~al.(2021)Gebru, Morgenstern, Vecchione, Vaughan, Wallach, Iii, and Crawford]{gebru2021datasheets}
T.~Gebru, J.~Morgenstern, B.~Vecchione, J.~W. Vaughan, H.~Wallach, H.~D. Iii, and K.~Crawford, ``Datasheets for datasets,'' \emph{Communications of the ACM}, vol.~64, no.~12, pp. 86--92, 2021.

\bibitem[Nichol(2016)]{nichol2016painter}
K.~Nichol, ``Painter by numbers,'' WikiArt, 2016, accessed: [insert date of access].

\bibitem[Liu et~al.(2021)Liu, Lin, He, Li, Wang, Li, Sun, Li, and Ding]{liu2021adaattn}
S.~Liu, T.~Lin, D.~He, F.~Li, M.~Wang, X.~Li, Z.~Sun, Q.~Li, and E.~Ding, ``Adaattn: Revisit attention mechanism in arbitrary neural style transfer,'' in \emph{Proceedings of the IEEE/CVF international conference on computer vision}, 2021, pp. 6649--6658.

\bibitem[Huang and Belongie(2017)]{huang2017arbitrary}
X.~Huang and S.~Belongie, ``Arbitrary style transfer in real-time with adaptive instance normalization,'' in \emph{Proceedings of the IEEE international conference on computer vision}, 2017, pp. 1501--1510.

\bibitem[Li et~al.(2019{\natexlab{b}})Li, Liu, Kautz, and Yang]{li2019learning}
X.~Li, S.~Liu, J.~Kautz, and M.-H. Yang, ``Learning linear transformations for fast image and video style transfer,'' in \emph{Proceedings of the IEEE/CVF Conference on Computer Vision and Pattern Recognition}, 2019, pp. 3809--3817.

\bibitem[Chen and Schmidt(2016)]{chen2016fast}
T.~Q. Chen and M.~Schmidt, ``Fast patch-based style transfer of arbitrary style,'' \emph{arXiv preprint arXiv:1612.04337}, 2016.

\bibitem[Park and Lee(2019)]{park2019arbitrary}
D.~Y. Park and K.~H. Lee, ``Arbitrary style transfer with style-attentional networks,'' in \emph{proceedings of the IEEE/CVF conference on computer vision and pattern recognition}, 2019, pp. 5880--5888.

\bibitem[An et~al.(2021)An, Huang, Song, Dou, Liu, and Luo]{an2021artflow}
J.~An, S.~Huang, Y.~Song, D.~Dou, W.~Liu, and J.~Luo, ``Artflow: Unbiased image style transfer via reversible neural flows,'' in \emph{Proceedings of the IEEE/CVF Conference on Computer Vision and Pattern Recognition}, 2021, pp. 862--871.

\bibitem[Lin et~al.(2014)Lin, Maire, Belongie, Hays, Perona, Ramanan, Doll{\'a}r, and Zitnick]{lin2014microsoft}
T.-Y. Lin, M.~Maire, S.~Belongie, J.~Hays, P.~Perona, D.~Ramanan, P.~Doll{\'a}r, and C.~L. Zitnick, ``Microsoft coco: Common objects in context,'' in \emph{European conference on computer vision}.\hskip 1em plus 0.5em minus 0.4em\relax Springer, 2014, pp. 740--755.

\bibitem[Cai et~al.(2023)Cai, Ma, Wang, and Li]{cai2023image}
Q.~Cai, M.~Ma, C.~Wang, and H.~Li, ``Image neural style transfer: A review,'' \emph{Computers and Electrical Engineering}, vol. 108, p. 108723, 2023.

\bibitem[Singh and Singh(2020)]{singh2020image}
A.~Singh and P.~Singh, ``Image classification: a survey,'' \emph{Journal of Informatics Electrical and Electronics Engineering (JIEEE)}, vol.~1, no.~2, pp. 1--9, 2020.

\bibitem[Zou et~al.(2023)Zou, Chen, Shi, Guo, and Ye]{zou2023object}
Z.~Zou, K.~Chen, Z.~Shi, Y.~Guo, and J.~Ye, ``Object detection in 20 years: A survey,'' \emph{Proceedings of the IEEE}, 2023.

\bibitem[Minaee et~al.(2021)Minaee, Boykov, Porikli, Plaza, Kehtarnavaz, and Terzopoulos]{minaee2021image}
S.~Minaee, Y.~Boykov, F.~Porikli, A.~Plaza, N.~Kehtarnavaz, and D.~Terzopoulos, ``Image segmentation using deep learning: A survey,'' \emph{IEEE transactions on pattern analysis and machine intelligence}, vol.~44, no.~7, pp. 3523--3542, 2021.

\bibitem[Gatys et~al.(2015{\natexlab{a}})Gatys, Ecker, and Bethge]{gatys2015texture}
L.~Gatys, A.~S. Ecker, and M.~Bethge, ``Texture synthesis using convolutional neural networks,'' \emph{Advances in neural information processing systems}, vol.~28, 2015.

\bibitem[Gatys et~al.(2015{\natexlab{b}})Gatys, Ecker, and Bethge]{gatys2015neural}
L.~A. Gatys, A.~S. Ecker, and M.~Bethge, ``A neural algorithm of artistic style,'' \emph{arXiv preprint arXiv:1508.06576}, 2015.

\bibitem[Deng et~al.(2022)Deng, Tang, Dong, Ma, Pan, Wang, and Xu]{deng2022stytr2}
Y.~Deng, F.~Tang, W.~Dong, C.~Ma, X.~Pan, L.~Wang, and C.~Xu, ``Stytr2: Image style transfer with transformers,'' in \emph{Proceedings of the IEEE/CVF conference on computer vision and pattern recognition}, 2022, pp. 11\,326--11\,336.

\bibitem[Zhang et~al.(2022{\natexlab{b}})Zhang, Tang, Dong, Huang, Ma, Lee, and Xu]{zhang2022domain}
Y.~Zhang, F.~Tang, W.~Dong, H.~Huang, C.~Ma, T.-Y. Lee, and C.~Xu, ``Domain enhanced arbitrary image style transfer via contrastive learning,'' in \emph{ACM SIGGRAPH 2022 Conference Proceedings}, 2022, pp. 1--8.

\bibitem[Deng et~al.(2021)Deng, Tang, Dong, Huang, Ma, and Xu]{deng2021arbitrary}
Y.~Deng, F.~Tang, W.~Dong, H.~Huang, C.~Ma, and C.~Xu, ``Arbitrary video style transfer via multi-channel correlation,'' in \emph{Proceedings of the AAAI Conference on Artificial Intelligence}, vol.~35, 2021, pp. 1210--1217.

\bibitem[Huo et~al.(2021)Huo, Jin, Li, Wu, Lai, Shi, and Gao]{huo2021manifold}
J.~Huo, S.~Jin, W.~Li, J.~Wu, Y.-K. Lai, Y.~Shi, and Y.~Gao, ``Manifold alignment for semantically aligned style transfer,'' in \emph{Proceedings of the IEEE/CVF International Conference on Computer Vision}, 2021, pp. 14\,861--14\,869.

\bibitem[Chen et~al.(2021)Chen, Wang, Zhang, Zuo, Li, Xing, Lu, et~al.]{chen2021artistic}
H.~Chen, Z.~Wang, H.~Zhang, Z.~Zuo, A.~Li, W.~Xing, D.~Lu \emph{et~al.}, ``Artistic style transfer with internal-external learning and contrastive learning,'' \emph{Advances in Neural Information Processing Systems}, vol.~34, pp. 26\,561--26\,573, 2021.

\bibitem[Li et~al.(2023)Li, Li, and Hoi]{li2023blip}
D.~Li, J.~Li, and S.~C. Hoi, ``Blip-diffusion: Pre-trained subject representation for controllable text-to-image generation and editing,'' \emph{arXiv preprint arXiv:2305.14720}, 2023.

\bibitem[Wang et~al.(2023{\natexlab{a}})Wang, Jiang, Lu, Shen, He, Chen, Wang, and Zhou]{wang2023context}
Z.~Wang, Y.~Jiang, Y.~Lu, Y.~Shen, P.~He, W.~Chen, Z.~Wang, and M.~Zhou, ``In-context learning unlocked for diffusion models,'' \emph{arXiv preprint arXiv:2305.01115}, 2023.

\bibitem[Wang et~al.(2023{\natexlab{b}})Wang, Wang, Cao, Shen, and Huang]{wang2023images}
X.~Wang, W.~Wang, Y.~Cao, C.~Shen, and T.~Huang, ``Images speak in images: A generalist painter for in-context visual learning,'' in \emph{Proceedings of the IEEE/CVF Conference on Computer Vision and Pattern Recognition}, 2023, pp. 6830--6839.

\bibitem[Bai et~al.(2023)Bai, Geng, Mangalam, Bar, Yuille, Darrell, Malik, and Efros]{bai2023sequential}
Y.~Bai, X.~Geng, K.~Mangalam, A.~Bar, A.~Yuille, T.~Darrell, J.~Malik, and A.~A. Efros, ``Sequential modeling enables scalable learning for large vision models,'' \emph{arXiv preprint arXiv:2312.00785}, 2023.

\bibitem[Wang et~al.(2024)Wang, Bai, Wang, Qin, and Chen]{wang2024instantid}
Q.~Wang, X.~Bai, H.~Wang, Z.~Qin, and A.~Chen, ``Instantid: Zero-shot identity-preserving generation in seconds,'' \emph{arXiv preprint arXiv:2401.07519}, 2024.

\end{thebibliography}
